\documentclass{article}  % For LaTeX2e
\usepackage{hyperref}
\usepackage{iclr2017_conference,times}
\pdfoutput=1

\title{Recurrent Environment Simulators}

\author{Silvia Chiappa, S{\'e}bastien Racaniere, Daan Wierstra \& Shakir Mohamed \\
DeepMind, London, UK\\
\texttt{\{csilvia, sracaniere, wierstra, shakir\}@google.com} \\
}

\usepackage[T1]{fontenc}    % use 8-bit T1 fonts
\usepackage{microtype}      % microtypography

\usepackage{enumitem}
\usepackage{algorithm2e}
\usepackage{booktabs}       % professional-quality tables
\usepackage{amsmath,amsfonts,subfigure,tikz}
\usetikzlibrary{shapes}

\tikzstyle{obs}=[fill=blue!20,thick]  % observed node
\tikzstyle{ocont}=[ellipse,draw=blue!50,thick,minimum size=6mm,>=stealth]  % continuous  node
\tikzstyle{dgraph}=[->, line width=1.5pt]

\newcommand{\figref}[1]{Fig. \ref{#1}}
\newcommand{\secref}[1]{Sec. \ref{#1}}
\newcommand{\appref}[1]{Appendix \ref{#1}}
\definecolor{myred}{cmyk}{0,0.9,0.9,0.1}
\definecolor{myblue}{cmyk}{0,0.3,0.9,0.1}

\newcommand{\myblue}{\color{myblue}}
\newcommand{\PDT}{PDT}

\newcommand{\OD}{observation-dependent} 
\newcommand{\PD}{prediction-dependent} 
\newcommand{\myvec}[1]{\mathbf{#1}}
\newcommand{\va}{\myvec{a}}
\newcommand{\vc}{\myvec{c}}
\newcommand{\vf}{\myvec{f}}
\newcommand{\vh}{\myvec{h}}
\newcommand{\vi}{\myvec{i}}
\newcommand{\vo}{\myvec{o}}
\newcommand{\vs}{\myvec{s}}
\newcommand{\vv}{\myvec{v}}
\newcommand{\vx}{\myvec{x}}
\newcommand{\vy}{\myvec{y}}
\newcommand{\vz}{\myvec{z}}
\newcommand{\vW}{\myvec{W}}
\newcommand{\ha}{\vv}

\newcommand{\ie}{\emph{i.e.}}

\iclrfinalcopy

\begin{document}

\maketitle  

\begin{abstract}
Models that can simulate how environments change in response to actions can be used by agents to plan and act efficiently.
We improve on previous environment simulators from high-dimensional pixel observations 
by introducing recurrent neural networks that are able to make temporally and spatially coherent predictions for hundreds of time-steps into the future. 
We present an in-depth analysis of the factors affecting performance, providing the most extensive 
attempt to advance the understanding of the properties of these models. 
We address the issue of computationally inefficiency with a model that does not need to generate a high-dimensional image at each time-step.
We show that our approach can be used to improve exploration and is adaptable to many diverse environments, namely 10 Atari games, a 3D car racing environment, and complex 3D mazes. 
\end{abstract}

\section{Introduction}
\vspace{-2mm}
In order to plan and act effectively, agent-based systems require an ability to anticipate the consequences of their actions within an environment, often for an extended period into the future.
Agents can be equipped with this ability by having access to models that can simulate how the environments changes in response to their actions. 
The need for environment simulation is widespread: in psychology, model-based predictive abilities form sensorimotor contingencies that are seen as essential for perception \citep{oregan01sensorimotor}; 
in neuroscience, environment simulation forms part of deliberative planning systems used by the brain \citep{niv09reinforcement}; and in reinforcement learning, 
the ability to imagine the future evolution of an environment is needed to form predictive state representations \citep{littman06predictive} and for Monte Carlo planning 
\citep{sutton98reinforcement}.

Simulating an environment requires models of temporal sequences that must possess a number of properties to be useful: the models should make predictions 
that are accurate, temporally and spatially coherent over long time periods; and allow for flexibility in the policies and action sequences that are used.  
In addition, these models should be general-purpose and scalable, and able to learn from high-dimensional perceptual inputs and from diverse and realistic environments. 
A model that achieves these desiderata can empower agent-based systems with a vast array of abilities, 
including counterfactual reasoning \citep{pearl09causality}, intuitive physical reasoning \citep{mccloskey83intuitive}, model-based exploration, episodic control 
\citep{lengyel07hippocampal}, intrinsic motivation \citep{oudeyer07intrinsic}, and hierarchical control.

Deep neural networks have recently enabled significant advances in simulating complex environments, allowing for models that consider high-dimensional visual inputs across a wide variety of domains \citep{wahlstrom15pixels, watter15embed, sun2015learning, patraucean15spatio}.
The model of \citet{oh15action} represents the state-of-the-art in this area, demonstrating high long-term accuracy in deterministic and discrete-action environments.

Despite these advances, there are still several challenges and open questions. 
Firstly, the properties of these simulators in terms of generalisation and sensitivity to the choices of model structure and training are poorly understood.
Secondly, accurate prediction for long time periods into the future remains difficult to achieve. 
Finally, these models are computationally inefficient, since they require the prediction of a high-dimensional image each time an action is executed, 
which is unnecessary in situations where the agent is interested only in the final prediction after taking several actions.

In this paper we advance the state-of-the-art in environment modelling. 
We build on the work of \citet{oh15action}, and develop alternative architectures and training schemes that significantly improve performance, 
and provide in-depth analysis to advance our understanding of the properties of these models. 
We also introduce a simulator that does not need to predict visual inputs after every action, reducing the computational burden in the use of the model.
We test our simulators on three diverse and challenging families of environments, 
namely Atari 2600 games, a first-person game where an agent moves in randomly generated 3D mazes, and a 3D car racing environment;
and show that they can be used for model-based exploration. 

\section{Recurrent Environment Simulators} 
\vspace{-2mm}
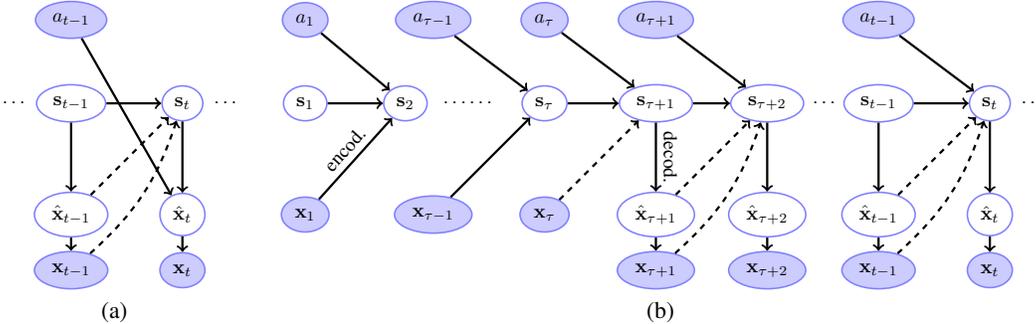
\begin{figure}[t]
\vskip-0.3cm
\begin{center}
\hskip-0.2cm
\subfigure[]{\scalebox{0.74}{
\begin{tikzpicture}[dgraph]
\node[ocont,obs] (atm) at (1,1.5) {$a_{t-1}$};
\node[ocont] (htm) at (1,0) {$\vs_{t-1}$};
\node[ocont] (ht) at (3,0) {$\vs_t$};
\node[] at (0,0) {$\cdots$};
\node[] at (3.8,0) {$\cdots$};
\node[ocont] (hxtm) at (1,-2) {$\hat \vx_{t-1}$};
\node[ocont] (hxt) at (3,-2) {$\hat \vx_t$};
\node[ocont,obs] (xtm) at (1,-3) {$\vx_{t-1}$};
\node[ocont,obs] (xt) at (3,-3) {$\vx_{t}$};
\draw[line width=1.15pt,dashed](xtm)to [bend left=-15](ht);
\draw[line width=1.15pt,dashed](hxtm)--(ht);
\draw[line width=1.15pt](htm)--(ht);
\draw[line width=1.15pt](htm)--(hxtm);\draw[line width=1.15pt](ht)--(hxt);
\draw[line width=1.15pt](hxtm)--(xtm);\draw[line width=1.15pt](hxt)--(xt);
\draw[line width=1.15pt](atm)--(hxt);color=myred,
\end{tikzpicture}}}
\hskip0.3cm
\subfigure[]{\scalebox{0.74}{
\begin{tikzpicture}[dgraph]
\node[ocont,obs] (a1) at (-0.3,1.5) {$a_1$};
\node[ocont,obs] (a9) at (2,1.5) {$a_{\tau-1}$};
\node[ocont,obs] (a10) at (4,1.5) {$a_{\tau}$};
\node[ocont,obs] (a11) at (6,1.5) {$a_{\tau+1}$};
\node[ocont,obs] (atm) at (10,1.5) {$a_{t-1}$};
\node[ocont] (h1) at (-0.3,0) {$\vs_1$};
\node[ocont] (h2) at (1.5,0) {$\vs_2$};
\node[ocont] (h10) at (4,0) {$\vs_{\tau}$};
\node[ocont] (h11) at (6,0) {$\vs_{\tau+1}$};
\node[ocont] (h12) at (8,0) {$\vs_{\tau+2}$};
\node[ocont] (htm) at (10,0) {$\vs_{t-1}$};
\node[ocont] (ht) at (12,0) {$\vs_t$};
\node[] at (2.65,0) {$\cdots\cdots$};
\node[] at (9.05,0) {$\cdots$};
\node[] at (12.8,0) {$\cdots$};
\node[ocont] (hx11) at (6,-2) {$\hat \vx_{\tau+1}$};
\node[ocont] (hx12) at (8,-2) {$\hat \vx_{\tau+2}$};
\node[ocont] (hxtm) at (10,-2) {$\hat \vx_{t-1}$};
\node[ocont] (hxt) at (12,-2) {$\hat \vx_t$};
\node[ocont,obs] (x1) at (-0.3,-2) {$\vx_1$};
\node[ocont,obs] (x9) at (2,-2) {$\vx_{\tau-1}$};
\node[ocont,obs] (x10) at (4,-2) {$\vx_{\tau}$};
\node[ocont,obs] (x11) at (6,-3) {$\vx_{\tau+1}$};
\node[ocont,obs] (x12) at (8,-3) {$\vx_{\tau+2}$};
\node[ocont,obs] (xtm) at (10,-3) {$\vx_{t-1}$};
\node[ocont,obs] (xt) at (12,-3) {$\vx_{t}$};
\draw[line width=1.15pt](x1)--node[sloped,above]{encod.}++(h2);\draw[line width=1.15pt](x9)--(h10);\draw[line width=1.15pt,dashed](x10)--(h11);\draw[line width=1.15pt,dashed](x11)to [bend left=-15](h12);\draw[line width=1.15pt,dashed](xtm)to [bend left=-15](ht);
\draw[line width=1.15pt](h1)--(h2);\draw[line width=1.15pt](h10)--(h11);\draw[line width=1.15pt](h11)--(h12);\draw[line width=1.15pt](htm)--(ht);
\draw[line width=1.15pt](a1)--(h2);\draw[line width=1.15pt](a9)--(h10);\draw[line width=1.15pt](a10)--(h11);\draw[line width=1.15pt](a11)--(h12);\draw[line width=1.15pt](atm)--(ht);
\draw[line width=1.15pt](h11)--node[sloped,above]{decod.}++(hx11);\draw[line width=1.15pt](h12)--(hx12);\draw[line width=1.15pt](htm)--(hxtm);\draw[line width=1.15pt](ht)--(hxt);
\draw[line width=1.15pt,dashed](hx11)--(h12);\draw[line width=1.15pt,dashed](hxtm)--(ht);
\draw[line width=1.15pt](hx11)--(x11);\draw[line width=1.15pt](hx12)--(x12);\draw[line width=1.15pt](hxtm)--(xtm);\draw[line width=1.15pt](hxt)--(xt);
\end{tikzpicture}}}
\end{center}
\caption{Graphical model representing (a) the recurrent structure used in \citet{oh15action} and (b) our recurrent structure. Filled and empty nodes indicate observed and hidden variables respectively.}
\label{fig:Model}
\end{figure}
An environment simulator is a model that, given a sequence of actions $a_1,\ldots,a_{\tau-1}\equiv a_{1:\tau-1}$ and corresponding observations $\vx_{1:\tau}$ of the environment, 
is able to predict the effect of subsequent actions $a_{\tau:\tau+\tau'-1}$, such as forming predictions $\hat \vx_{\tau+1:\tau+\tau'}$ or state representations $\vs_{\tau+1:\tau+\tau'}$ of the environment.

Our starting point is the recurrent simulator of \citet{oh15action}, which is the state-of-the-art in simulating deterministic environments with visual observations (frames) and discrete actions. This simulator is a recurrent neural network with the following backbone structure:
\begin{equation*}
\vs_t = f(\vs_{t-1}, \mathcal{C}(\mathbb{I}(\hat \vx_{t-1}, \vx_{t-1})))\,, \quad \hat{\vx}_t = \mathcal{D}(\vs_{t}, a_{t-1})\,. 
\end{equation*}
In this equation, $\vs_t$ is a hidden state representation of the environment, and $f$ a non-linear deterministic state transition function.
The symbol $\mathbb{I}$ indicates the selection of the predicted frame $\hat{\vx}_{t-1}$  or real frame $\vx_{t-1}$, producing two types of state transition 
called \textbf{prediction-dependent transition} and \textbf{observation-dependent transition} respectively.
$\mathcal{C}$ is an encoding function consisting of a series of convolutions, and  
$\mathcal{D}$ is a decoding function that combines the state $\vs_t$ with the action $a_{t-1}$ through a multiplicative interaction,
and then transforms it using a series of full convolutions to form the predicted frame $\hat{\vx}_{t}$. 

The model is trained to minimise the mean squared error between the observed time-series $\vx_{\tau+1:\tau+\tau'}$, 
corresponding to the evolution of the environment, and its prediction. 
In a probabilistic framework, this corresponds to maximising the log-likelihood in the graphical model depicted in \figref{fig:Model}(a). 
In this graph, the link from $\hat \vx_t$ to $\vx_t$ represents stochastic dependence, as $\vx_t$ is formed by adding to $\hat \vx_t$ a Gaussian noise term with zero mean and unit variance,
whilst all remaining links represent deterministic dependences.
The dashed lines indicate that only one of the two links is active, depending on whether the state transition is \PD~or \OD. 

The model is trained using stochastic gradient descent, in which each mini-batch consists of a set of segments of length $\tau+T$ randomly sub-sampled from $\vx_{1:\tau+\tau'}$.
For each segment in the mini-batch, the model uses the first $\tau$ observations to evolve the state and forms predictions of the last $T$ observations only.
\textit{Training} comprises three phases differing in the use of \PD~or \OD~transitions (after the first $\tau$ transitions) and in the value of the \textit{prediction length} $T$. In the first phase, the model uses \OD~transitions and predicts for $T=10$ time-steps.
In the second and third phases, the model uses \PD~transitions and predicts for $T=3$ and $T=5$ time-steps respectively. 
During \textit{evaluation} or \textit{usage}, the model can only use \PD~transitions. 

\subsection*{Action-Dependent State Transition}
\vspace{-2mm}
A strong feature of the model of \citet{oh15action} described above is that the actions influence the state transitions only \textit{indirectly} through the predictions or the observations.
Allowing the actions to condition the state transitions directly could potentially enable the model to incorporate action information more effectively. 
We therefore propose the following backbone structure:
\begin{equation*}
\vs_{t} = f(\vs_{t-1},a_{t-1},\mathcal{C}(\mathbb{I}(\hat \vx_{t-1}, \vx_{t-1})))\,, \quad \hat{\vx}_t = \mathcal{D}(\vs_{t})\,.
\end{equation*} 
In the graphical model representation, this corresponds to replacing the link from $a_{t-1}$ to $\hat \vx_t$ with a link from $a_{t-1}$ to $\vs_t$ as in \figref{fig:Model}(b). 

\subsection*{Short-Term versus Long-Term Accuracy}
\vspace{-2mm}
The last two phases in the training scheme of \citet{oh15action} described above are used to address 
the issue of poor accuracy that recurrent neural networks trained using only \OD~transitions display 
when asked to predict several time-steps ahead. However, the paper does not analyse nor discuss alternative training schemes.

In principle, the highest accuracy should be obtained by training the model as closely as possible to the way it will be used, 
and therefore by using a number of \PD~transitions which is as close as possible to the number of time-steps the model will be asked to predict for.
However, \PD~transitions increase the complexity of the objective function such that alternative schemes are most often used \citep{talvitie14model,bengio15scheduled,oh15action}.
Current training approaches are guided by the belief that using the observation $\vx_{t-1}$, rather than the prediction $\hat\vx_{t-1}$, to form the state $\vs_{t}$ has the effect of reducing 
the propagation of the errors made in the predictions, which are higher at earlier stages of the training, enabling the model to \textit{correct} itself from the mistakes made up to time-step $t-1$.
For example, \cite{bengio15scheduled} introduce a scheduled sampling approach where at each time-step the type of state transition is sampled from a Bernoulli distribution,  
with parameter annealed from an initial value corresponding to using only \OD~transitions to a final value corresponding to using only \PD~transitions, according to a schedule selected 
by validation. 

Our analysis of different training schemes on Atari, which considered the interplay among \textit{warm-up} length $\tau$, prediction length $T$, and number of \PD~transitions,
suggests that, rather than as having a corrective effect, \OD~transitions should be seen as restricting the time interval in which the model considers its predictive abilities, 
and therefore focuses resources.
Indeed we found that, the higher the number of \textit{consecutive} \PD~transitions, 
the more the model is encouraged to focus on learning the global dynamics of the environment, which results in higher long-term accuracy.
The highest long-term accuracy is always obtained by a training scheme that uses only \PD~transitions even at the early stages of the training.
Focussing on learning the global dynamics comes at the price of shifting model resources away from learning the precise details of the frames, leading to a decrease in short-term accuracy. 
Therefore, for complex games for which reasonable long-term accuracy cannot be obtained, training schemes that mix \PD~and \OD~transitions are preferable.
It follows from this analysis that percentage of consecutive \PD~transitions, rather than just percentage of such transitions, should be considered when designing training schemes.

From this viewpoint, the poor results obtained in \cite{bengio15scheduled} when using only \PD~transitions can be explained by the difference in the type of the tasks considered.  
Indeed, unlike our case in which the model is tolerant to some degree of error such as blurriness in earlier predictions, the discrete problems considered in \cite{bengio15scheduled} are such that one prediction error at earlier time-steps can severely affect predictions at later time-steps, 
so that the model needs to be highly accurate short-term in order to perform reasonably longer-term. Also, \cite{bengio15scheduled} treated the prediction used to form $\vs_t$ as a fixed quantity, rather than as a function of $\vs_{t-1}$, 
and therefore did not perform exact maximum likelihood.

\subsection*{Prediction-Independent State Transition}
\vspace{-2mm}
In addition to potentially enabling the model to incorporate action information more effectively, allowing the actions to directly influence the state dynamics has another crucial advantage:
it allows to consider the case of a state transition that does not depend on the frame, \ie~of the form $\vs_t = f(\vs_{t-1},a_{t-1})$, 
corresponding to removing the dashed links from $\hat \vx_{t-1}$ and from $\vx_{t-1}$ to $\vs_t$ in \figref{fig:Model}(b).
We shall call such a model \textit{prediction-independent simulator}, referring to its ability to evolve the state without using the prediction during usage.
Prediction-independent state transitions for high-dimensional observation problems have also been considered in \cite{srivastava15unsupervised}.

A prediction-independent simulator can dramatically increase computational efficiency in situations is which the agent is interested in the effect of a sequence of actions rather than of a single action.
Indeed, such a model does not need to project from the lower dimensional state space into the higher dimensional observation space through the set of convolutions, and vice versa, at each time-step. 

\section{Prediction-Dependent Simulators}
\vspace{-2mm}
We analyse simulators with state transition of the form $\vs_{t} = f(\vs_{t-1},a_{t-1},\mathcal{C}(\mathbb{I}(\hat \vx_{t-1}, \vx_{t-1})))$ 
on three families of environments with different characteristics and challenges, 
namely Atari 2600 games from the arcade learning environment \citep{bellemare13arcade}, 
a first-person game where an agent moves in randomly generated 3D mazes \citep{beattie16deepmind}, and a 3D car racing environment called TORCS \citep{wymann13torcs}. 
We use two evaluation protocols. In the first one, the model is asked to predict for 100 or 200 time-steps into the future using actions from the test data.
In the second one, a human uses the model as an interactive simulator. 
The first protocol enables us to determine how the model performs within the action policy of the training data, 
whilst the second protocol enables us to explore how the model generalises to other action policies.

As state transition, we used the following \textit{action-conditioned} 
long short-term memory (LSTM) \citep{hochreiter97long}: 
\begin{align}
\textrm{Encoding: } & \vz_{t-1}=
{\cal C}(\mathbb{I}(\hat \vx_{t-1}, \vx_{t-1})) \,, \label{eq:state_enc} \\
\textrm{Action fusion: } & \ha_{t} = \vW^h \vh_{t-1}\otimes \vW^a \va_{t-1}\,, \label{eq:alstm_action}\\
\textrm{Gate update: } & \vi_t = {\sigma}(\vW^{iv}\ha_{t} +\vW^{iz}\vz_{t-1})\,, \hskip0.1cm \vf_t = {\sigma}(\vW^{fv}\ha_t +\vW^{fz}\vz_{t-1})\,, \nonumber\\ 
& \vo_t = {\sigma}(\vW^{ov}\ha_t + \vW^{oz}\vz_{t-1}) \label{eq:alstm_gate}\,,\\
\textrm{Cell update: } & \vc_t = \vf_t\otimes \vc_{t-1} + \vi_t\otimes \textrm{tanh}(\vW^{cv}\ha_{t}+\vW^{cz}\vz_{t-1})\,, \label{eq:alstm_cell}\\
\textrm{State update: } & \vh_t = \vo_t\otimes \textrm{tanh}(\vc_t)\,, \label{eq:alstm_state}
\end{align}
where $\otimes$ denotes the Hadamard product, $\sigma$ the logistic sigmoid function, $\va_{t-1}$ is a one-hot vector representation of $a_{t-1}$, 
and $\vW$ are parameter matrices. In Eqs. \eqref{eq:alstm_action}--\eqref{eq:alstm_state},  $\vh_t$ and $\vc_t$ are the LSTM state and cell forming the model state $\vs_t = (\vh_t,\vc_t)$; 
and $\vi_t, \vf_t$, and $\vo_t$ are the input, forget, and output gates respectively (for simplicity, we omit the biases in their updates).  
The vectors $\vh_t$ and $\ha_t$ had dimension 1024 and 2048 respectively.  
Details about the encoding and decoding functions $\mathcal{C}$ and $\mathcal{D}$ for the three families of environments 
can be found in Appendix \ref{sec:AppAtari}, \ref{sec:AppTorcs} and \ref{sec:AppMazes}. 
We used a warm-up phase of length $\tau = 10$ and we did not backpropagate the gradient to this phase.

\subsection{Atari} 
We considered the 10 Atari games Freeway, Ms Pacman, Qbert, Seaquest, Space Invaders, Bowling, Breakout, Fishing Derby, Pong, and Riverraid. 
Of these, the first five were analysed in \citet{oh15action} and are used for comparison. The remaining five were chosen to better test the ability of the model in environments with other challenging characteristics,
such as scrolling backgrounds (Riverraid), small/thin objects that are key aspects of the game (lines in Fishing Derby, ball in Pong and 
Breakout), and sparse-reward games that require very long-term predictions (Bowling).
We used training and test datasets consisting of five and one million 210$\times$160 RGB images respectively, with 
actions chosen from a trained DQN agent \citep{mnih15dqn} according to an $\epsilon=0.2$-greedy policy. Such a large number of training frames ensured that our simulators did not strongly overfit to the training data
(see training and test lines in Figs. \ref{fig:predErrFBTypeBF} and \ref{fig:predErrFBTypePS}, and the discussion in \appref{sec:AppAtari}).

\subsubsection*{Short-Term versus Long-Term Accuracy}
Below we summarise our results on the interplay among warm-up length $\tau$, prediction length $T$, and number of \PD~transitions 
-- the full analysis is given in \appref{sec:AppPDT}. 

The warm-up and prediction lengths $\tau$ and $T$ regulate degree of accuracy in two different ways. 
1) The value of $\tau+T$ determines how far into the past the model can access information -- this is the case irrespectively of the type of transition used, although when using \PD~transitions 
information about the last $T$ time-steps of the environment would need to be inferred. Accessing information far back into the past can be necessary
even when the model is used to perform one-step ahead prediction only.
2) The higher the value of $T$ and the number of \PD~transitions, the more the corresponding objective function encourages long-term accuracy. 
This is achieved by guiding the one-step ahead prediction error in such a way that further predictions will not be strongly affected, and by teaching the 
model to make use of information from the far past. The more precise the model is in performing one-step ahead prediction, the less noise guidance 
should be required. Therefore, models with very accurate convolutional and transition structures should need less encouragement.

\paragraph{Increasing the percentage of consecutive \PD~transitions increases long-term accuracy, often at the expense of short-term accuracy.\label{sec:PDT}}
We found that using only \OD~transitions leads to poor performance in most games. 
Increasing the number of consecutive \PD~transitions produces an increase in long-term accuracy, 
but also a decrease in short-term accuracy usually corresponding to reduction in sharpness. 
For games that are too complex, although the lowest long-term prediction error is still achieved with using only \PD~transitions, 
reasonable long-term accuracy cannot be obtained, and training schemes that mix \PD~and \OD~transitions are therefore preferable.

To illustrate these results, we compare the following training schemes for prediction length $T=15$:
\begin{description}[leftmargin=*]
\setlength{\itemsep}{-6pt}  
\setlength{\parskip}{-4pt}
\setlength{\parsep}{-4pt}
\item[$\bullet$ 0\% \PDT:] Only \OD~transitions.\\
\item[$\bullet$ 33\% \PDT:] Observation and \PD~transitions for the first 10 and last 5 time-steps respectively.\\ %[-5pt]
\item[$\bullet$ 0\%-20\%-33\% \PDT:] Only \OD~transitions in the first 10,000 parameter updates; \OD~transitions for the first 12 time-steps and \PD~transitions for the last 3 time-steps for the subsequent 100,000 parameters updates;  
\OD~transitions for the first 10 time-steps and \PD~transitions for the last 5 time-steps for the remaining parameter updates (adaptation of the training scheme of \citet{oh15action} to $T=15$).\\ % [-5pt]
\item[$\bullet$ 46\% \PDT~Alt.:] Alternate between \OD~and \PD~transitions from a time-step to the next.\\ %[-5pt]
\item[$\bullet$ 46\% \PDT:] Observation and \PD~transitions for the first 8 and last 7 time-steps respectively.\\ %[-5pt]
\item[$\bullet$ 67\% \PDT:]  Observation and \PD~transitions for the first 5 and last 10 time-steps respectively.\\ %[-5pt]
\item[$\bullet$ 0\%-100\% \PDT:] Only \OD~transitions in the first 1000 parameter updates; only \PD~transitions in the subsequent parameter updates.\\ %[-5pt]
\item[$\bullet$ 100\% \PDT:] Only \PD~transitions.%[-5pt]
\end{description}
For completeness, we also consider a training scheme as in \citet{oh15action}, which consists of three phases with $T=10, T=3, T=5$, and 500,000, 250,000, 750,000 parameter updates respectively.
In the first phase $\vs_t$ is formed by using the observed frame $\vx_{t-1}$, whilst in the two subsequent phases $\vs_t$ is formed by using the predicted frame $\hat \vx_{t-1}$.

\begin{figure}[t]
%\hskip-0.2cm
%\centering
\subfigure[]{\scalebox{0.79}{\includegraphics[]{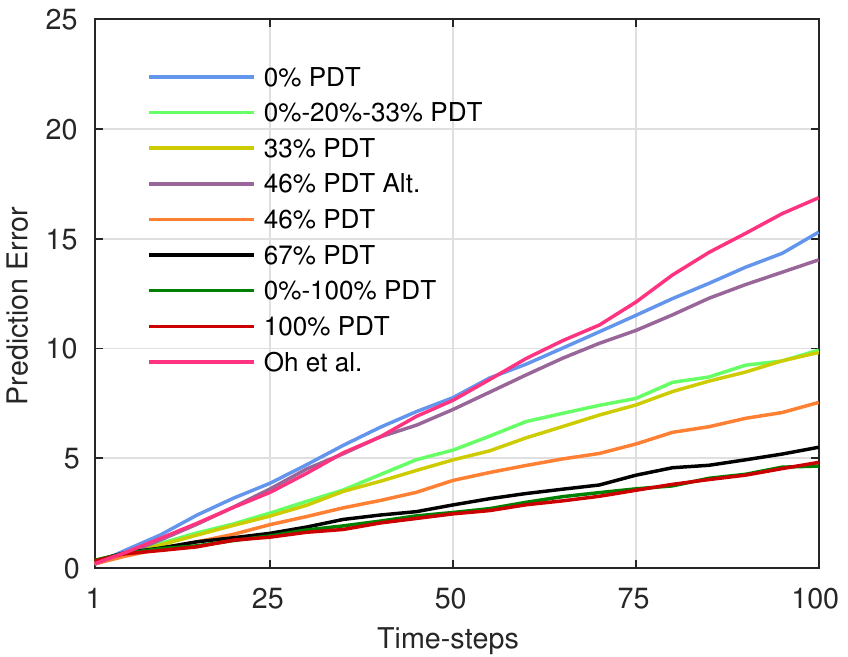}}}
%\hskip0.1cm
\subfigure[]{\scalebox{0.79}{\includegraphics[]{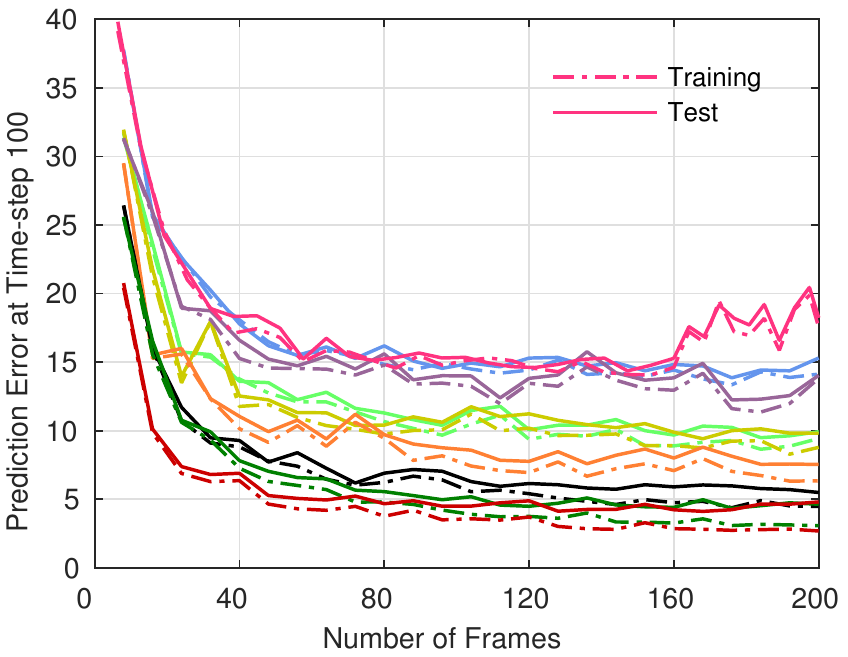}}}
\subfigure[]{\scalebox{0.79}{\includegraphics[]{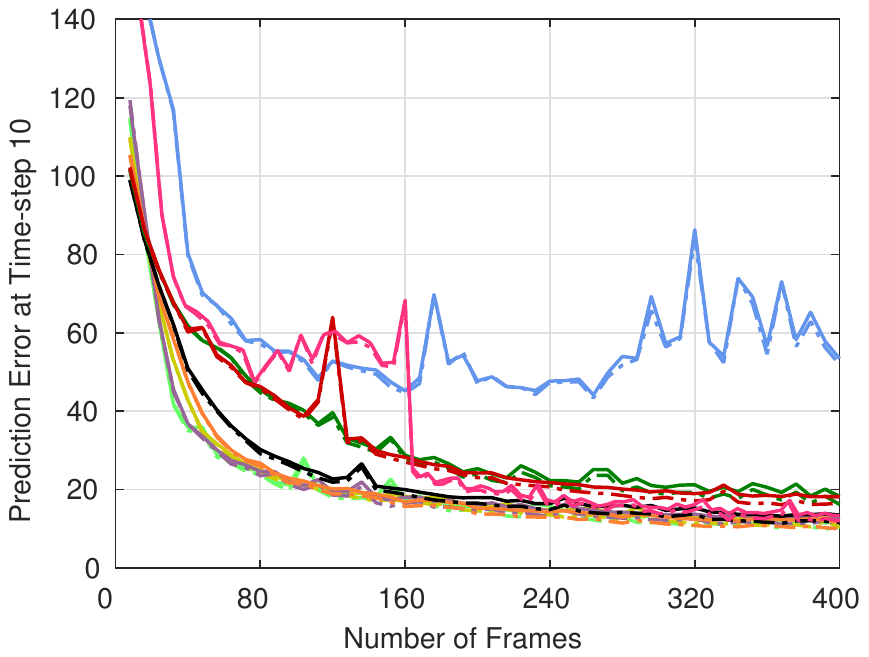}}
%\hskip0.1cm
\scalebox{0.79}{\includegraphics[]{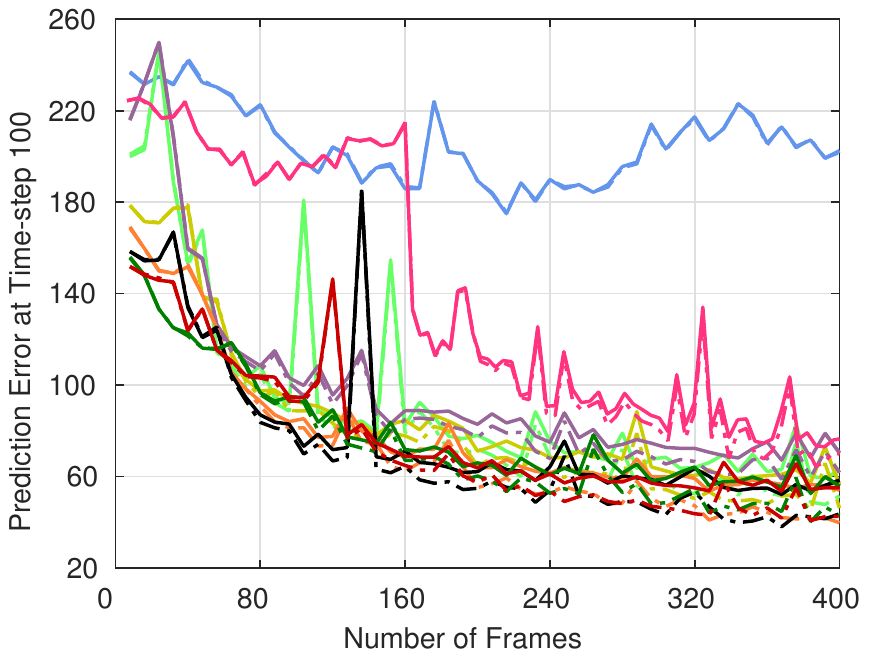}}}\\
\caption{Prediction error averaged over 10,000 sequences on (a)-(b) Bowling and (c) Fishing Derby for different training schemes. The same color and line code is used in all figures.
(a): Prediction error vs time-steps after the model has seen 200 million frames. 
(b)-(c): Prediction error vs number of frames seen by the model at time-steps 10 and 100. }
\label{fig:predErrFBTypeBF}
% Figure obtained with predErrFBTypePaper.m
\end{figure}
\begin{figure}[t]
%\hskip-0.2cm
%\centering
\subfigure[]{\scalebox{0.79}{\includegraphics[]{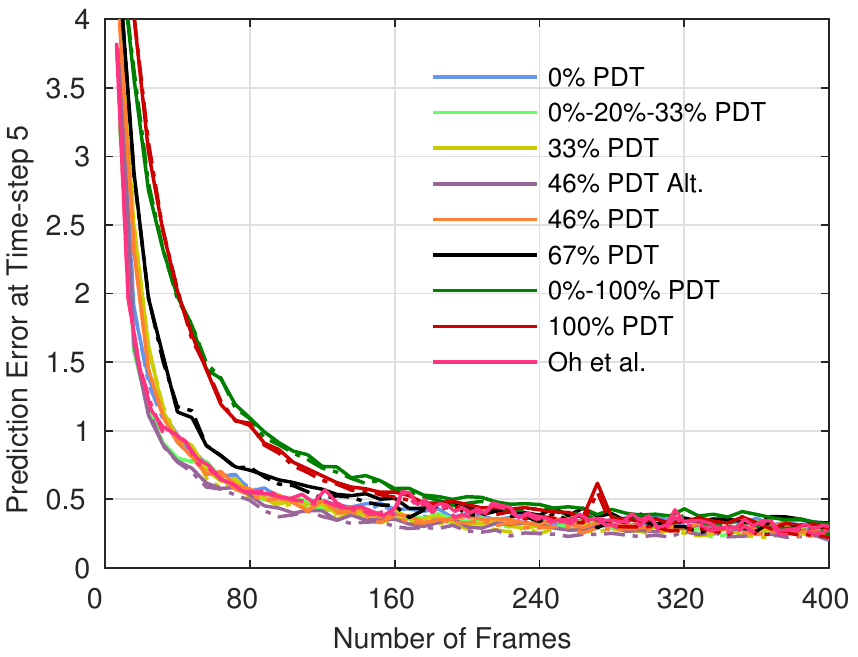}}
%\hskip0.1cm
\scalebox{0.79}{\includegraphics[]{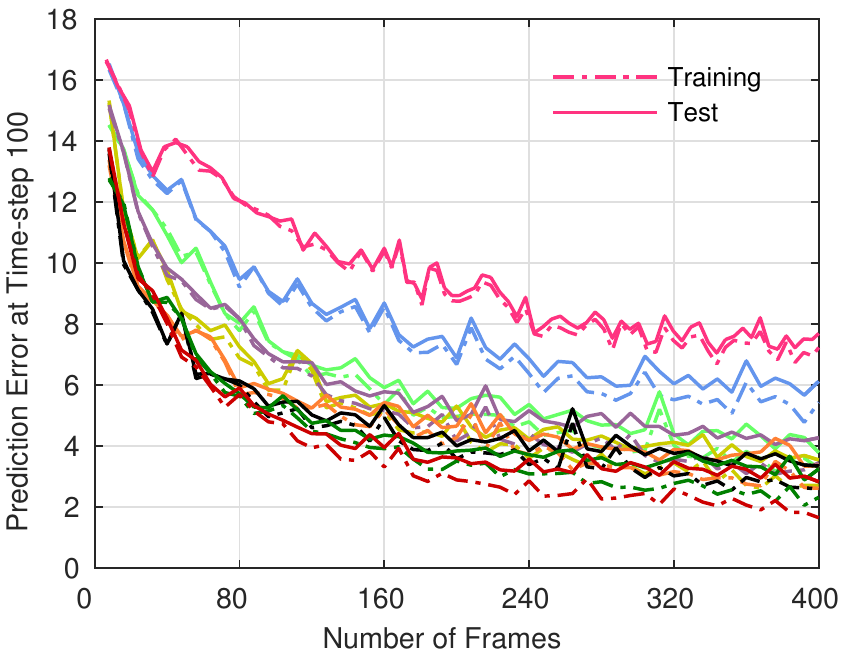}}}\\
\subfigure[]{\scalebox{0.79}{\includegraphics[]{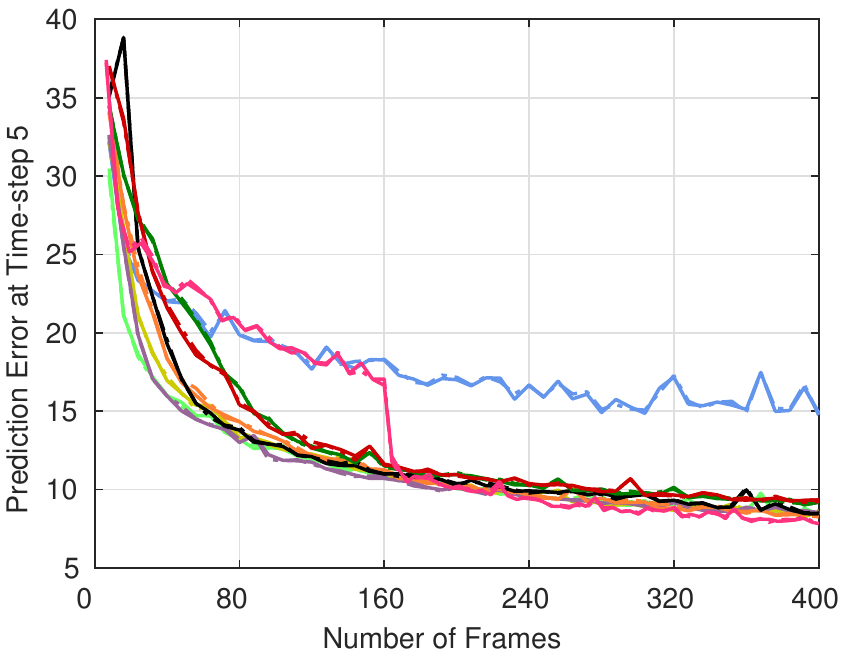}}
%\hskip0.1cm
\scalebox{0.79}{\includegraphics[]{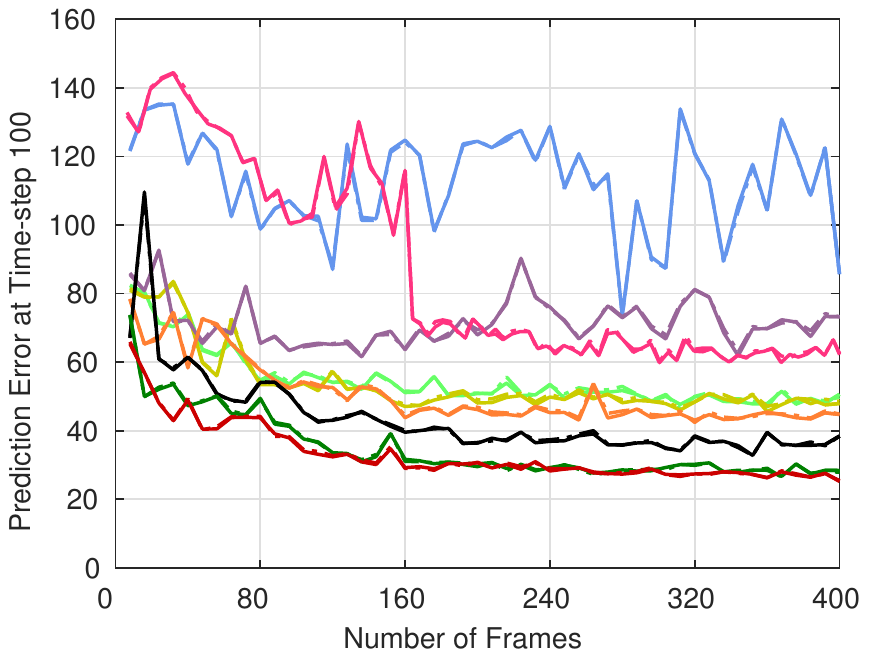}}}
\caption{Prediction error on (a) Pong and (b) Seaquest for different training schemes. }
\label{fig:predErrFBTypePS}
% Figure obtained with predErrFBTypePaper.m
\end{figure}

\begin{figure}[t]
%\hskip-0.2cm
%\centering
\subfigure[]{\scalebox{0.79}{\includegraphics[]{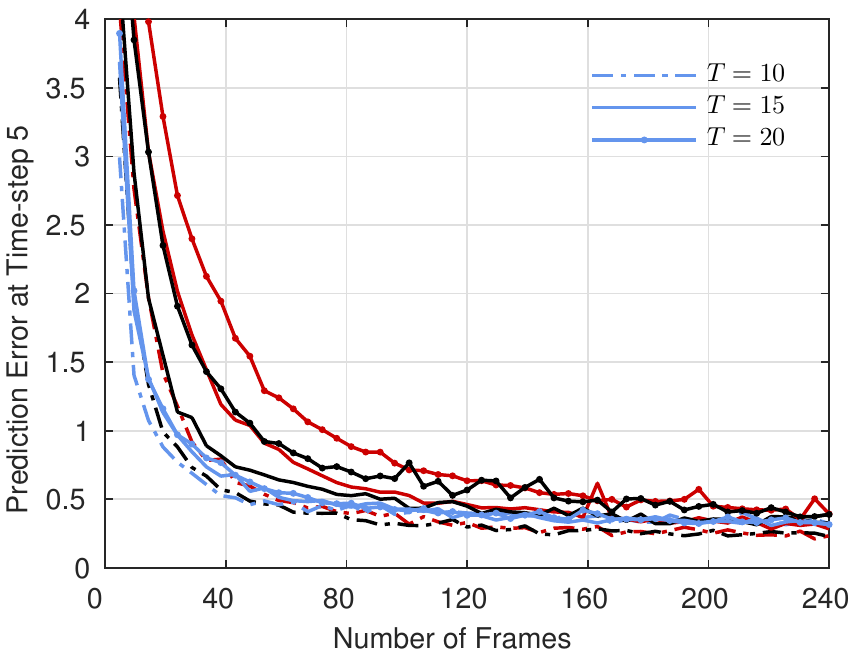}}
%\hskip0.1cm
\scalebox{0.79}{\includegraphics[]{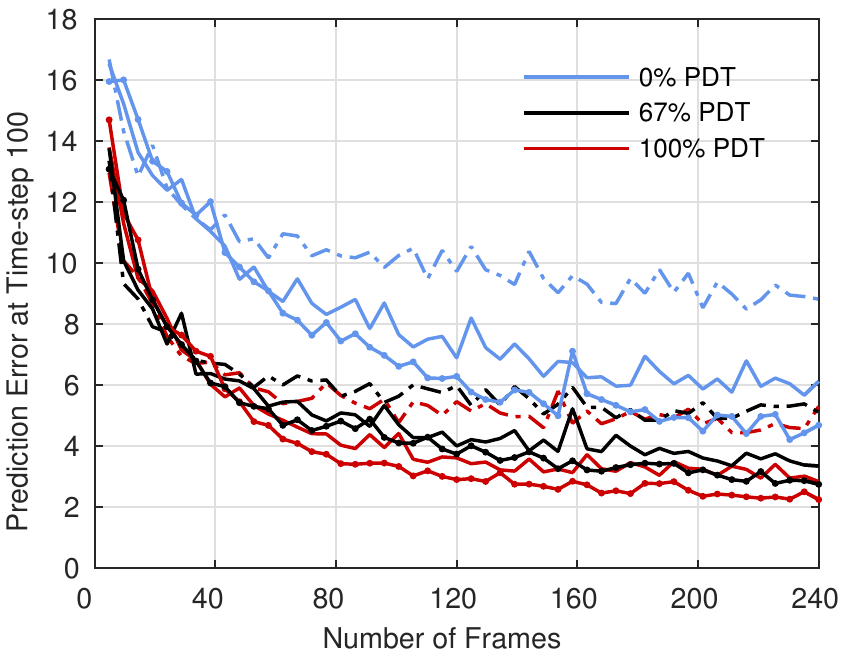}}}\\
\subfigure[]{\scalebox{0.79}{\includegraphics[]{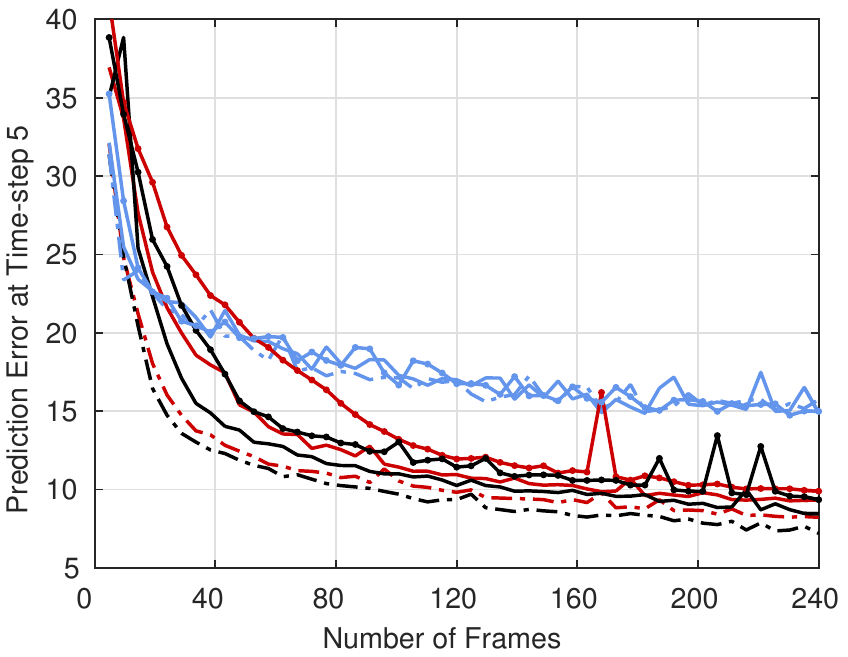}}
%\hskip0.1cm
\scalebox{0.79}{\includegraphics[]{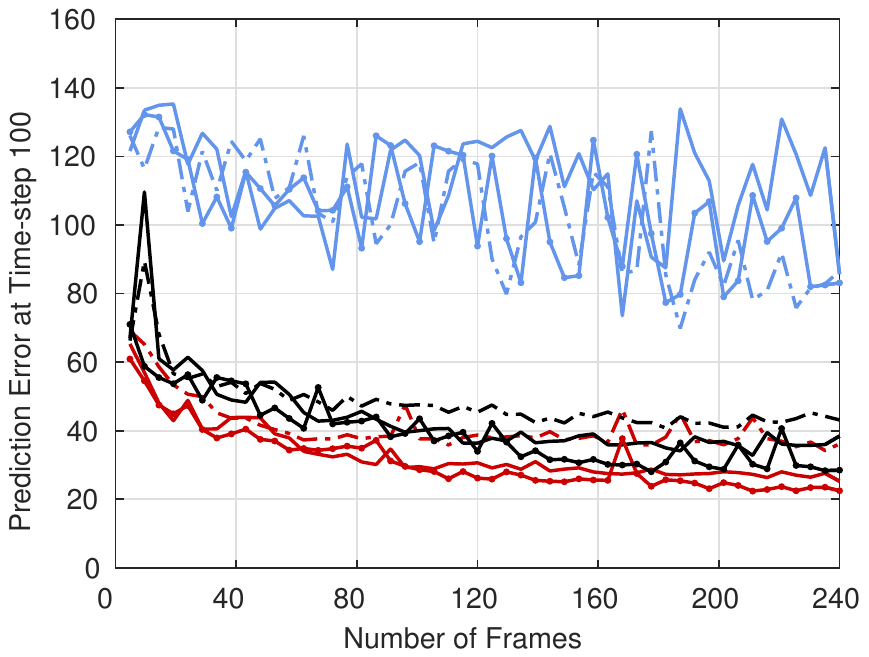}}}
\caption{Prediction error vs number of frames seen by the model (excluding warm-up frames) for (a) Pong and (b) Seaquest, using prediction lengths $T=10, 15$, and 20, 
and training schemes 0\%\PDT, 67\%\PDT, and 100\%\PDT.}
\label{fig:predErrSeqLengthPS}
% Figure obtained with predErrSeqLengthPaper.m
\end{figure}

\begin{figure}[t]
%\hskip-0.2cm
%\centering
\subfigure[]{\scalebox{0.79}{\includegraphics[]{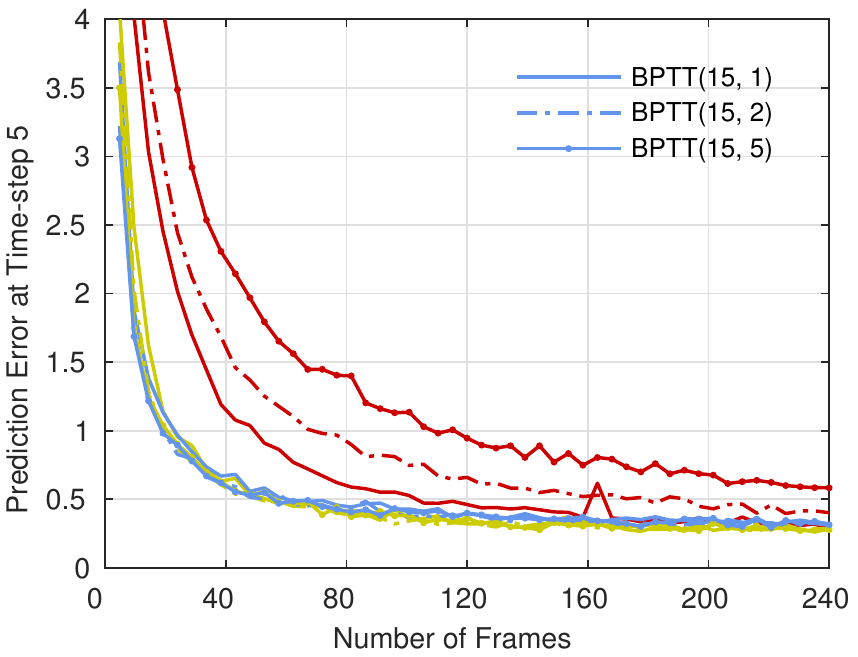}}
%\hskip0.1cm
\scalebox{0.79}{\includegraphics[]{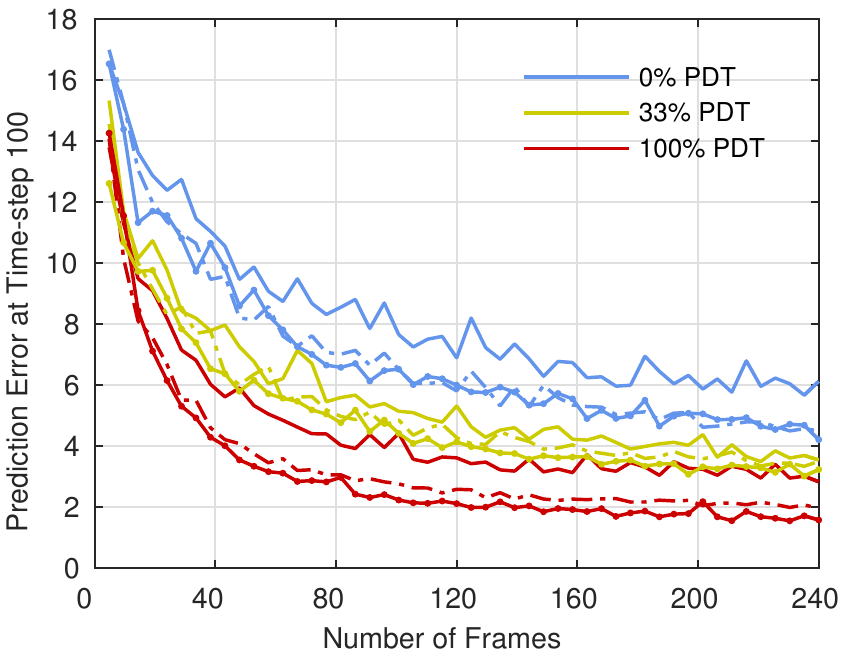}}}\\
\subfigure[]{\scalebox{0.79}{\includegraphics[]{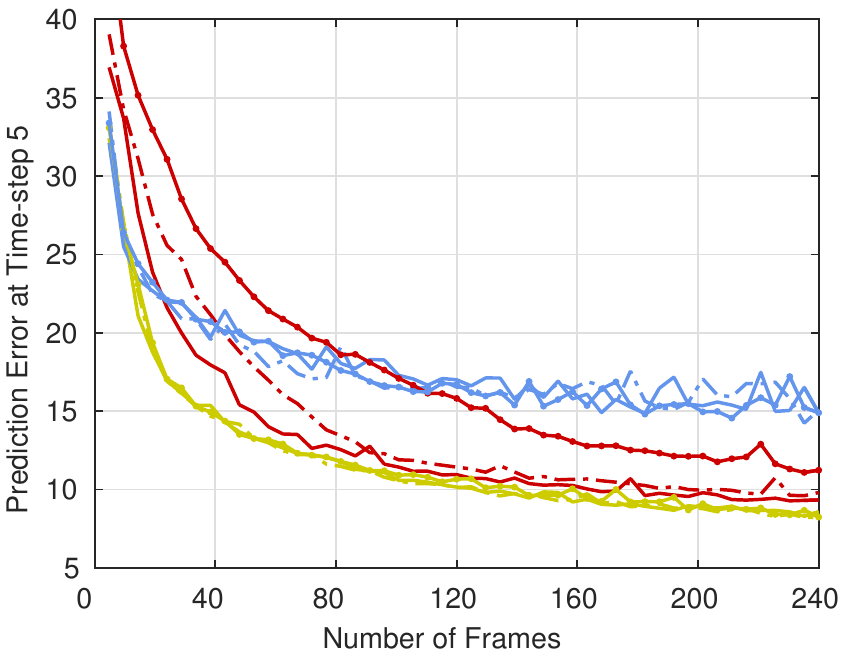}}
%\hskip0.1cm
\scalebox{0.79}{\includegraphics[]{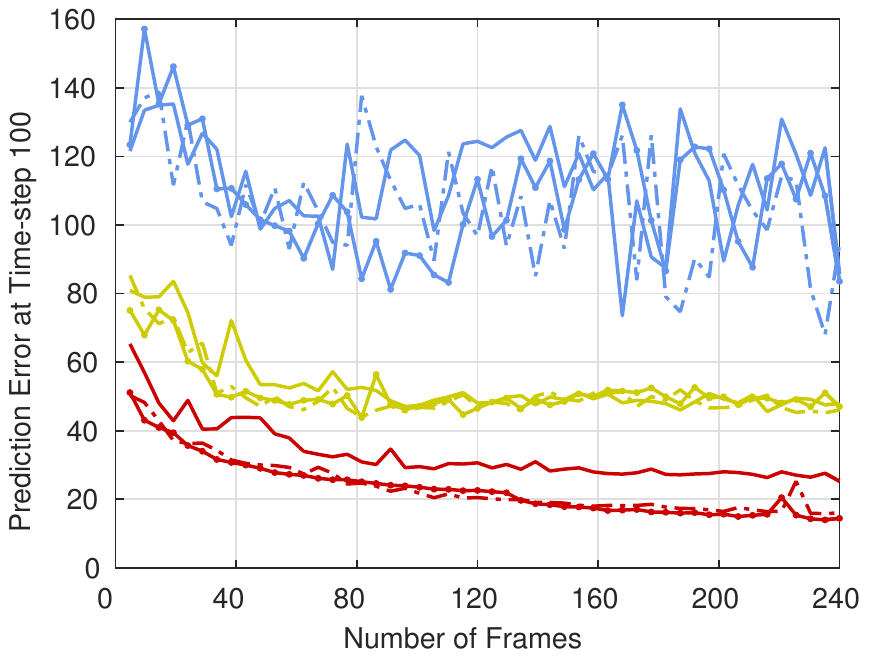}}}
\caption{Prediction error vs number of frames seen by the model (excluding warm-up frames) for (a) Pong and (b) Seaquest, using BPTT(15, 1), BPTT(15, 2), and BTT(15, 5), 
and training schemes 0\%\PDT, 33\%\PDT, and 100\%\PDT.}
\label{fig:predErrSeqNumPS}
% Figure obtained with predErrSeqNumPaper.m
\end{figure}

\begin{figure}[t]
%\hskip-0.2cm
%\centering
\scalebox{0.79}{\includegraphics[]{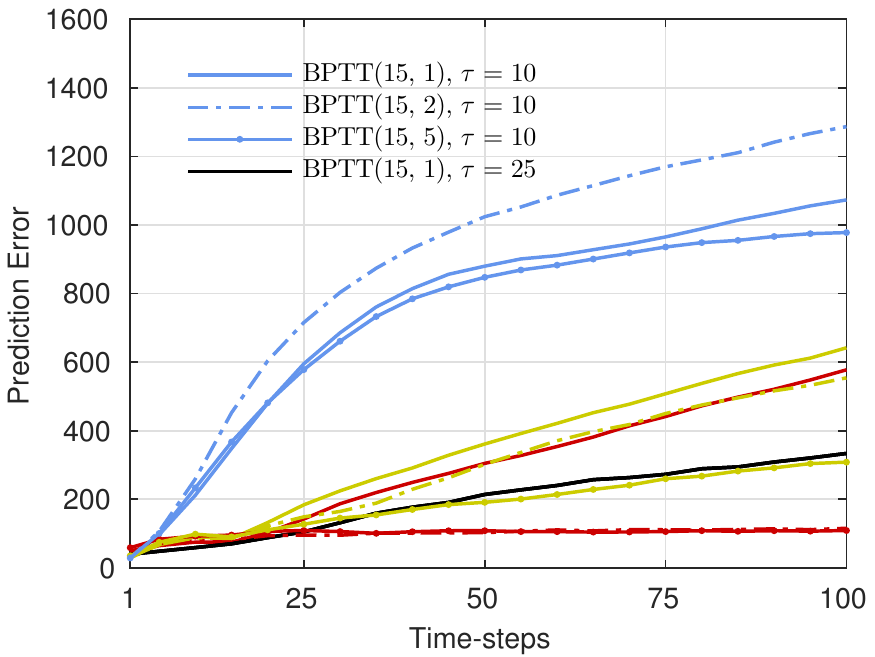}}
%\hskip0.1cm
\scalebox{0.79}{\includegraphics[]{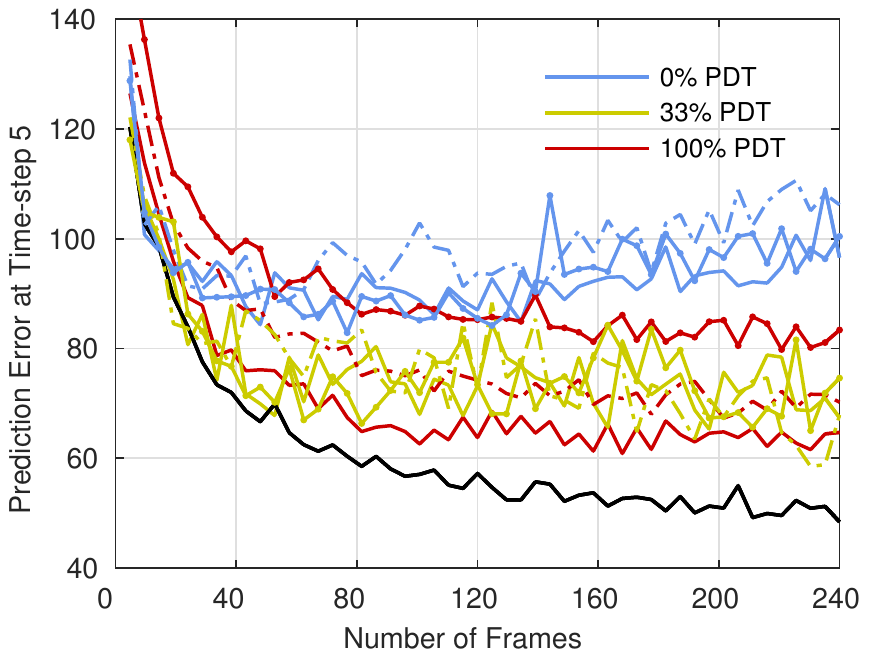}}\\
\scalebox{0.79}{\includegraphics[]{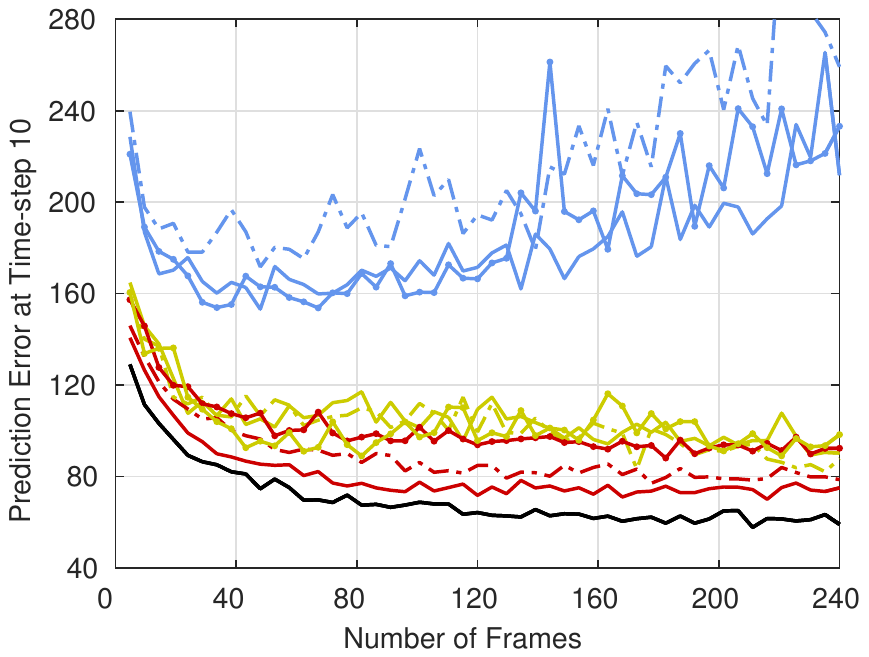}}
%\hskip0.1cm
\scalebox{0.79}{\includegraphics[]{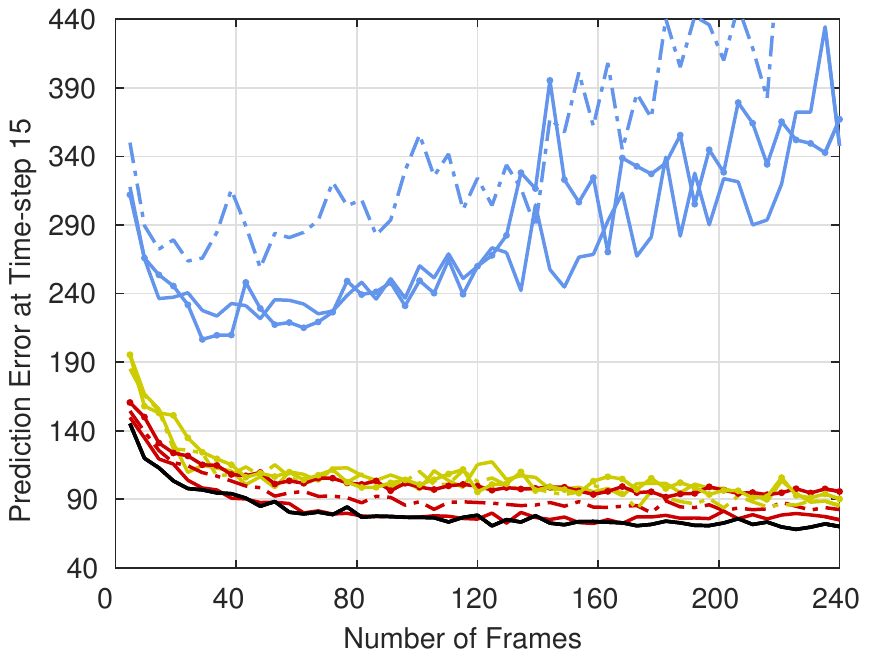}}
\caption{Prediction error vs number of frames seen by the model for Riverraid, using BPTT(15, 1), BPTT(15, 2), and BTT(15, 5), and training schemes 0\%\PDT, 33\%\PDT, and 100\%\PDT. 
The black line is obtained with the 100\%\PDT~training scheme.}
\label{fig:predErrSeqNumR}
% Figure obtained with predErrSeqNumPaperRiverraid.m
\end{figure}

In Figs. \ref{fig:predErrFBTypeBF} and \ref{fig:predErrFBTypePS} we show the prediction error averaged over 10,000 sequences\footnote{We define the prediction error as 
$\frac{1}{3*10,000}\sum_{n=1}^{10,000}\parallel \vx^{n}_t -\hat \vx^{n}_t\parallel^2$.} for the games of Bowling\footnote{In this game, the player is given two chances to roll a ball down an alley in an attempt 
to knock down as many of the ten pins as possible, after which the score is updated and the knocked pins are relocated. 
Knocking down every pin on the first shot is a strike, while knocking 
every pin down in both shots is a spare. The player's score is determined by the number of pins knocked down, as well as the number of strikes and spares acquired.}, Fishing Derby, Pong and Seaquest. 
More specifically, \figref{fig:predErrFBTypeBF}(a) shows the error for predicting up to 100 time-steps ahead after the model has seen 200 million frames (corresponding to half million parameter updates using mini-batches 
of 16 sequences), using actions and warm-up frames from the test data,
whilst Figs. \ref{fig:predErrFBTypeBF}(b)-(c) and \ref{fig:predErrFBTypePS} show the error at time-steps 5, 10 and 100 versus number of frames seen by the model.

These figures clearly show that long-term accuracy generally improves with increasing number of consecutive \PD~transitions. 
When using alternating (46\% \PDT~Alt.), rather than consecutive (46\% \PDT), \PD~transitions 
long-term accuracy is worse, as we are effectively asking the model to predict at most two time-steps ahead.
We can also see that using more \PD~transitions produces lower short-term accuracy and/or slower short-term convergence.
Finally, the figures show that using a training phase with only \OD~transitions that is too long, as in \citet{oh15action}, can be detrimental:
the models reaches at best a performance similar to the 46\% \PDT~Alt. training scheme (the sudden drop in prediction error corresponds to transitioning to the second training phase), but is most often worse.

By looking at the predicted frames we could notice that, in games containing balls and paddles, 
using only \OD~transitions gives rise to errors in reproducing the dynamics of these objects. Such errors decrease with increasing \PD~transitions. 
In other games, using only \OD~transitions causes the model to fail in representing moving objects, except for the agent in most cases. 
Training schemes containing more \PD~transitions encourage the model to focus more on learning the dynamics of the moving objects 
and less on details that would only increase short-term accuracy, giving rise to more globally accurate but less sharp predictions. 
Finally, in games that are too complex, the strong emphasis on long-term accuracy produces predictions that are overall not sufficiently good.

More specifically, from the videos available at\footnote{Highlighted names like these are direct links to folders containing videos. 
Each video consists of 5 randomly selected 200 time-steps ahead predictions separated by black frames (the generated frames are shown on the left,
whilst the real frames are shown on the right -- the same convention will be used throughout the paper). 
Shown are 15 frames per seconds. Videos associated with the material discussed in this and following sections can also be found at \url{https://sites.google.com/site/resvideos1729}.} 
{\myblue \href{https://drive.google.com/drive/folders/0B_L2b7VHvBW2SEllTmlEX1l0RGc?usp=sharing}{PDTvsODT}}, 
%{\myblue \href{https://drive.google.com/file/d/0B_L2b7VHvBW2WlN4UUQ5TkNqbDg/view?usp=sharing}{Breakout-\ODT}}, 
%{\myblue\href{https://drive.google.com/a/google.com/file/d/0B_R02INZWFe9cWQzVVBfMVJmdzQ/view?usp=sharing}{FDerby-\ODT}},
%{\myblue \href{https://drive.google.com/file/d/0B_L2b7VHvBW2cF9QeFB6VGpnUGc/view?usp=sharing}{MSPacman-\ODT}}, 
%{\myblue\href{https://drive.google.com/file/d/0B_L2b7VHvBW2aXAycGFUM21STDg/view?usp=sharing}{Qbert-\ODT}}, 
%{\myblue\href {https://drive.google.com/a/google.com/file/d/0B_R02INZWFe9RnNMWDZKYi1PTjA/view?usp=sharing}{Riverraid-\ODT}}, 
%{\myblue\href{https://drive.google.com/a/google.com/file/d/0B_R02INZWFe9WGtWNzZjWU9UM2s/view?usp=sharing}{Seaquest-\ODT}}, 
%{\myblue\href{https://drive.google.com/file/d/0B_L2b7VHvBW2eGdjSGh6U2JWSzA/view?usp=sharing}{SInvaders-\ODT}}, 
we can see that using only \OD~transitions has a detrimental effect on long-term accuracy for Fishing Derby, Ms Pacman, Qbert, Riverraid, Seaquest and Space Invaders. 
The most salient features of the videos are: consistent inaccuracy in predicting the paddle and ball in Breakout;
reset to a new life after a few time-steps in Ms Pacman;
prediction of background only after a few time-steps in Qbert;
no generation of new objects or background in Riverraid; 
quick disappearance of existing fish and no appearance of new fish from the sides of the frame in Seaquest.
For Bowling, Freeway, and Pong, long-term accuracy is generally good, but the movement of the ball is not always correctly predicted in Bowling and Pong
and the chicken sometimes disappears in Freeway.
On the other hand, using only \PD~transitions results in good long-term accuracy for Bowling, Fishing Derby, Freeway, Pong, Riverraid, and Seaquest: 
%{\myblue\href{https://drive.google.com/file/d/0B_L2b7VHvBW2Q0RXU0c5Rkd5ZWs/view?usp=sharing}{Bowling-\PDT}}, 
%{\myblue\href{https://drive.google.com/file/d/0B_L2b7VHvBW2eUdSeG9IZFBNenM/view?usp=sharing}{FDerby-\PDT}},
%{\myblue\href{https://drive.google.com/a/google.com/file/d/0B_R02INZWFe9MWcxLW40S2xDdEU/view?usp=sharing}{Freeway-\PDT}},
%{\myblue\href{https://drive.google.com/file/d/0B_L2b7VHvBW2NTYtWDVtX3hWVzQ/view?usp=sharing}{Seaquest-\PDT}},
the model accurately represents the paddle and ball dynamics in Bowling and Pong;  the chicken hardly disappears in Freeway,
and new objects and background are created and most often correctly positioned in Riverraid and Seaquest.

The trading-off of long for short-term accuracy when using more \PD~transitions is particularly evident in the videos of Seaquest: the higher the number 
of such transitions, the better the model learns the dynamics of the game, with new fish appearing in the right location more often.
However, this comes at the price of reduced sharpness, mostly in representing the fish. 
%{\myblue\href{https://drive.google.com/drive/folders/0B_R02INZWFe9Y0dwTFJSZ3ZTZGM?usp=sharing}{Breakout}}
%{\myblue\href{https://drive.google.com/file/d/0B_L2b7VHvBW2R1F6eFBNLVZ1dXc/view?usp=sharing}{Seaquest-0\%-20\%-33\%Pred.Frames}} 

This trade-off causes problems in Breakout, Ms Pacman, Qbert, and Space Invaders, so that schemes that also use \OD~transitions are preferable for these games.
For example, in Breakout, the model fails at representing the ball, making the predictions not sufficiently good. 
Notice that the prediction error (see \figref{fig:predErrFBTBowling-Breakout}) is misleading in terms of desired performance,
as the 100\%\PDT~training scheme performs as well as other mixing schemes for long-term accuracy -- this highlights 
the difficulties in evaluating the performance of these models. 
%{\myblue\href{https://drive.google.com/file/d/0B_L2b7VHvBW2NElRNHgtZmVhU3c/view?usp=sharing}{Breakout-66\%Pred.Frames}}, 
%{\myblue\href{https://drive.google.com/file/d/0B_L2b7VHvBW2dHVreDgyYjhIQ1U/view?usp=sharing}{Ms Pacman-66\%Pred.Frames}}, 
%{\myblue\href{https://drive.google.com/file/d/0B_L2b7VHvBW2WTJ3MDJtRDY4U1E/view?usp=sharing}{Qbert-66\%Pred.Frames}}, 
%{\myblue\href{https://drive.google.com/file/d/0B_L2b7VHvBW2bUxkUkJNSjhkNTQ/view?usp=sharing}{SInvaders-66\%Pred.Frames}}).
%{\myblue\href{https://drive.google.com/a/google.com/file/d/0B_R02INZWFe9V0xaWnVHMUNuX1k/view?usp=sharing}{Oh}}

\paragraph{Increasing the prediction length $T$ increases long-term accuracy when using \PD~transitions.} 
In \figref{fig:predErrSeqLengthPS}, we show the effect of using different prediction lengths $T\leq 20$ on the training schemes 0\%\PDT, 67\%\PDT, and 100\%\PDT~for Pong and Seaquest.
In Pong, with the 0\%\PDT~training scheme, using higher $T$ improves long-term accuracy: this is a game for which this scheme gives reasonable accuracy and the model is able to benefit from longer history. 
This is however not the case for Seaquest (or other games as shown in \appref{sec:AppPDT}). 
On the other hand, with the 100\%\PDT~training scheme, using higher $T$ improves long-term accuracy in most games (the difference is more pronounced between 
$T=10$ and $T=15$ than between $T=15$ and $T=20$), but decreases short-term accuracy. 
Similarly to above, reduced short-term accuracy corresponds to reduced sharpness: from the videos available at 
{\myblue \href{https://drive.google.com/drive/folders/0B_L2b7VHvBW2WjQ3YTZwOE9qMVU?usp=sharing}{$T\leq 20$}} 
we can see, for example, that the moving caught fish in Fishing Derby, the fish in Seaquest, and the ball in Pong are less sharp for higher $T$.

\paragraph{Truncated backpropagation still enables increase in long-term accuracy.}
Due to memory constraints, we could only backpropagate gradients over sequences of length up to 20. 
To use $T>20$, we split the prediction sequence into subsequences and performed parameter updates separately 
for each subsequence. For example, to use $T=30$ we split the prediction sequence into two successive subsequences of length 15,
performed parameter updates over the first subsequence, initialised the state of the second subsequence with the final state from the first subsequence, 
and then performed parameter updates over the second subsequence. 
This approach corresponds to a form of truncated backpropagation through time \citep{williams95gradient} --
the extreme of this strategy (with $T$ equal to the length of the whole training sequence) was used by \citet{zarembe14recurrent}.

In \figref{fig:predErrSeqNumPS}, we show the effect of using 2 and 5 subsequences of length $15$ (indicated by BPTT(15, 2) and BTT(15, 5))
on the training schemes 0\%\PDT, 33\%\PDT, and 100\%\PDT~for Pong and Seaquest. 
We can see that the 0\%\PDT~and 33\%\PDT~training schemes display no difference in accuracy for different values of $T$. 
On the other hand, with the 100\%\PDT~training scheme, using more than one subsequence improves long-term accuracy (the difference is more pronounced between 
$T=15$ and $T=30$ than between $T=30$ and $T=75$), but decreases short-term accuracy (the difference is small at convergence between $T=15$ and $T=30$, but big between $T=30$ and $T=75$). 
The decrease in accuracy with 5 subsequences is drastic in some games.

For Riverraid, using more than one subsequence with the 33\%\PDT~and 100\%\PDT~training schemes improves long-term accuracy dramatically, as shown in \figref{fig:predErrSeqNumR}, 
as it enables correct prediction after a jet loss.
Interestingly, for the 100\%\PDT~training scheme, using $\tau=25$ with prediction length $T=15$ (black line) does not give the same amount of gain as when using BPTT(15, 2), 
even if history length $\tau+T$ is the same. 
This would seem to suggest that some improvement in BPTT(15, 2) is due to encouraging longer-term accuracy, indicating that this can be achieved even when not fully backpropagating the gradient.
%(as shown in {\myblue\href{https://drive.google.com/a/google.com/file/d/0B_R02INZWFe9dW5qbWFaaDZPVVU/view?usp=sharing}{Riverraid-\PDT}}). 

From the videos available at {\myblue \href{https://drive.google.com/drive/folders/0B_L2b7VHvBW2bVRvLTUwZmEtM1U?usp=sharing}{$T> 20$}}, 
we can see that with $T=75$ the predictions in some of the Fishing Derby videos are faded,
whilst in Pong the model can suddenly switch from one dynamics to another for the ball and the opponent's paddle.

In conclusion, using higher $T$ through truncated backpropagation can improve performance. However, in schemes that use many \PD~transitions, a high value of $T$ can lead to poor predictions.

\subsubsection*{Evaluation through Human Play}
Whilst we cannot expect our simulators to generalise to structured sequences of actions never chosen by the DQN and that are not present in the training data, such as moving the agent up and down the alley in Bowling, 
it is reasonable to expect some degree of generalisation in the action-wise simple environments of Breakout, Freeway and Pong.

We tested these three games by having humans using the models as interactive simulators. We generally found that models trained using only 
\PD~transitions were more fragile to states of the environment not experienced during training, 
such that the humans were able to play these games for longer with simulators trained with mixing training schemes.
This seems to indicate that models with higher long-term test accuracy are 
at higher risk of overfitting to the training policy. 

\begin{figure}[t]
\hskip-0.05cm
\centering
\subfigure[]{\scalebox{1}{\includegraphics[width=6.8cm,height=6.4cm]{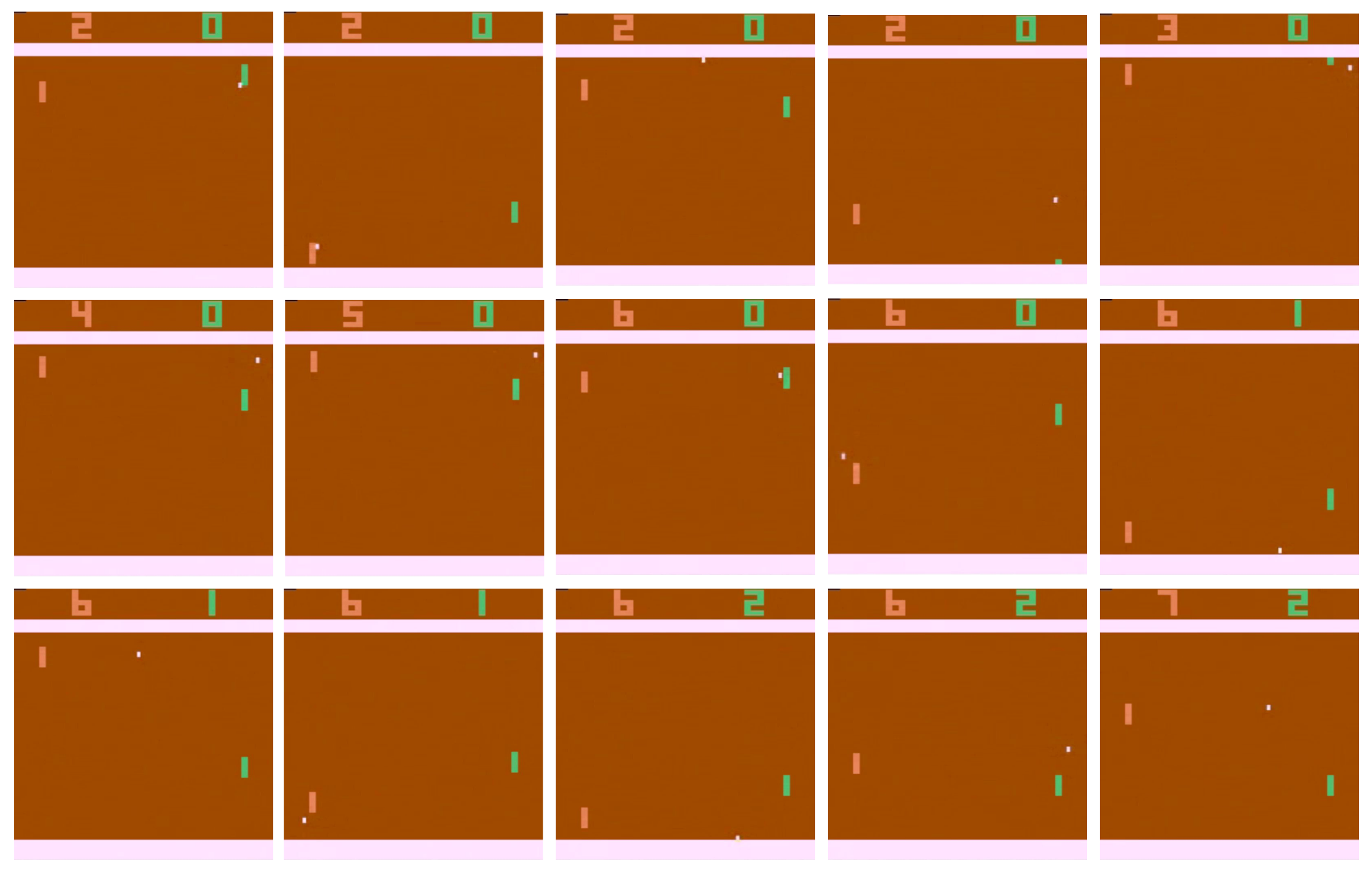}}}
\hskip0.3cm
\subfigure[]{\scalebox{1}{\includegraphics[width=6.8cm,height=6.4cm]{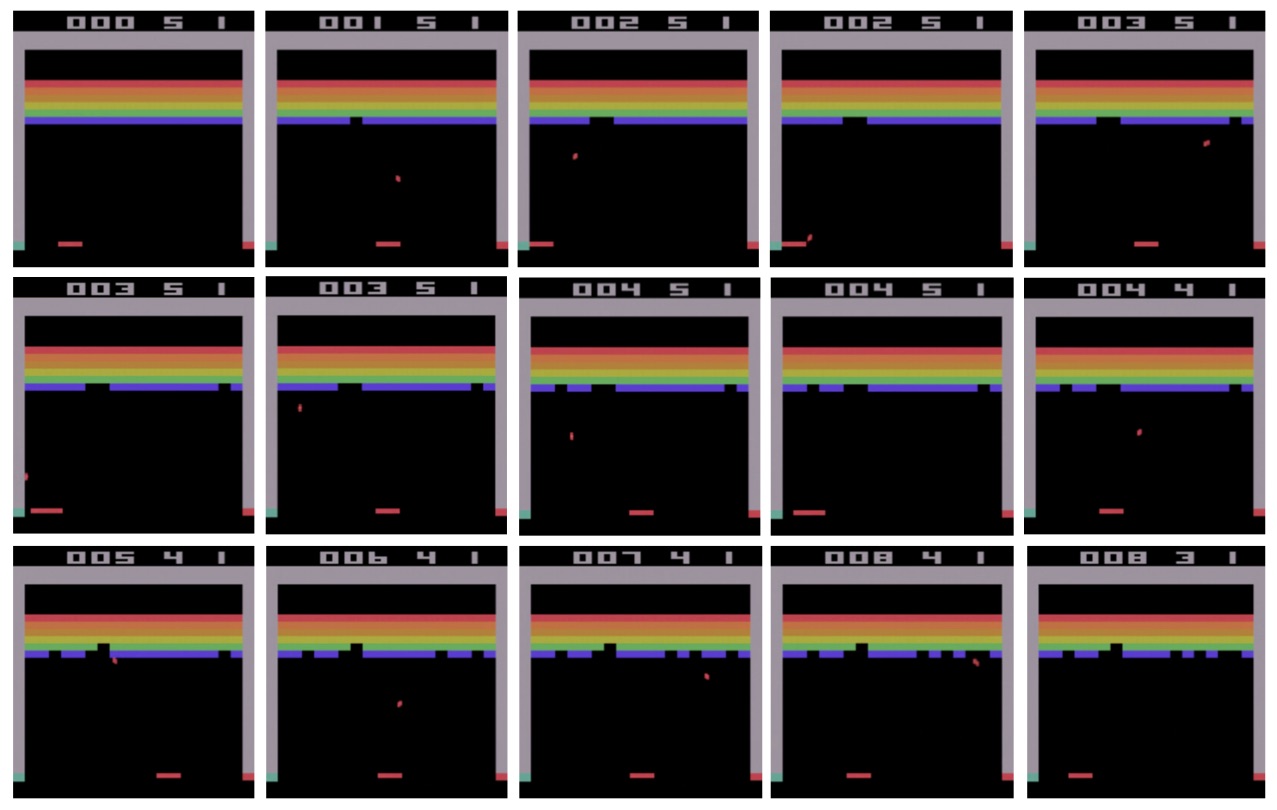}}}
\caption{Salient frames extracted from (a) 500 frames of Pong and (b) 350
frames of Breakout generated using our simulator with actions taken by a human player (larger versions can be found in Figs. \ref{fig:pong-large} and \ref{fig:breakout-large}).}
\label{fig:pong-breakout}
\end{figure}

%Below we discuss generalization to human policies in Breakout and Pong and leave the discussion of Freeway to the Appendix.
In \figref{fig:pong-breakout}(a), we show some salient frames from a game of Pong played by a human for 500 time-steps (the corresponding video is available at {\myblue \href{https://drive.google.com/open?id=0B_L2b7VHvBW2VGJwOVVueV8yTWM}{Pong-HPlay}}). 
The game starts with score (2,0), after which the opponent scores five times, whilst the human player scores twice. 
As we can see, the scoring is updated correctly and the game dynamics is accurate. 
In \figref{fig:pong-breakout}(b), we show some salient fames from a game of Breakout played by a human for 350 time-steps (the corresponding video is available at {\myblue \href{https://drive.google.com/file/d/0B_L2b7VHvBW2QVY0a2xnR0FXR1k/view?usp=sharing}{Breakout-HPlay}}).
As for Pong,  the scoring is updated correctly and the game dynamics is accurate. 
These images demonstrate some degree of generalisation of the model to a human style of play. 

\subsubsection*{Evaluation of State Transitions Structures}
In Appendix \ref{sec:AppStructures} and \ref{sec:AppAction} we present an extensive evaluation of different action-dependent state transitions, 
including convolutional transformations for the action fusion, and gate and cell updates, and different ways of incorporating action information. 
We also present a comparison between action-dependent and action-independent state transitions. 

Some action-dependent state transitions give better performance than the baseline (Eqs. \eqref{eq:state_enc}--\eqref{eq:alstm_state}) in some games. 
For example, we found that increasing the state dimension from 1024 to the dimension of the convolved frame, namely 2816, might be preferable. 
Interestingly, this is not due to an increase in the number of parameters, as the same gain is obtained using convolutions for the gate and cell updates.
These results seem to suggest that high-dimensional sparse state transition structures could be a promising direction for further improvement. 
Regarding different ways of incorporation action information, we found that using local incorporation 
such as augmenting the frame with action information and indirect action influence gives worse performance that direct and global action influence, but that there are several ways of incorporating
action information directly and globally that give similar performance. 

\subsection{3D Environments}
Both TORCS and the 3D maze environments highlight the need to learn dynamics that are temporally and spatially coherent: TORCS exposes the need to learn fast moving dynamics and consistency under motion, 
whilst 3D mazes are partially-observed and therefore require the simulator to build an internal representation of its surrounding using memory, as well learn basic physics, such as rotation, momentum, and the solid properties of walls. 

\textbf{TORCS.} The data was generated using an artificial agent controlling a fast car without opponents (more details are given in \appref{sec:AppTorcs}). 

When using actions from the test set (see \figref{fig:torcstest} and the corresponding video at {\myblue \href{https://drive.google.com/open?id=0B_L2b7VHvBW2NEQ1djNjU25tWUE}{TORCS}}), 
the simulator was able to produce accurate predictions for up to several hundreds time-steps. 
As the car moved around the racing track, the simulator was able to predict the appearance of new features in the background (towers, sitting areas, lamp posts, etc.), 
as well as model the jerky motion of the car caused by our choices of random actions. Finally, the instruments (speedometer and rpm) were correctly displayed. 

\begin{figure}[t]
%\vskip-0.2cm
\scalebox{0.66}{\includegraphics[]{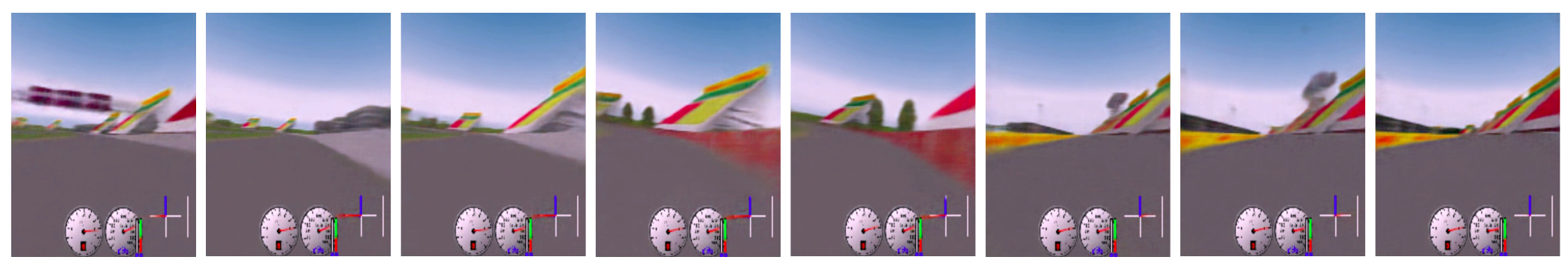}}
%vspace{-5mm}
\caption{Salient frames highlighting coherence extracted from 700 frames of TORCS generated using our simulator with actions taken by a human player.}
\label{fig:Torcs} %1:28, 2:01, 2:29, 2:54, 3:02, 4:02, 4:011
%\vspace{-1mm}
\end{figure}

The simulator was good enough to be used interactively for several hundred frames, using actions provided by a human.
This showed that the model had learnt well how to deal with the car hitting the wall on the right side of the track. 
Some salient frames from the game are shown in \figref{fig:Torcs} (the corresponding video can be seen at {\myblue \href{https://drive.google.com/open?id=0B_L2b7VHvBW2UjMwWVRoM3lTeFU}{TORCS-HPlay}}).

\textbf{3D Mazes.} We used an environment that consists of randomly generated 3D mazes, containing textured surfaces with occasional paintings on the walls: the mazes were all of the same size, but differed in the layout of rooms and corridors, 
and in the locations of paintings (see \figref{fig:expBaselineVsSim}(b) for an example of layout). More details are given in \appref{sec:AppMazes}. 

\begin{figure}[t]
\hskip-0.2cm
%\centering
\begin{tabular}{c c c c c c c}
 \includegraphics[width=69pt]{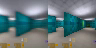} &
 \includegraphics[width=69pt]{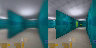} &
 \includegraphics[width=69pt]{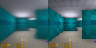} &
 \includegraphics[width=69pt]{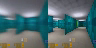} &
 \includegraphics[width=69pt]{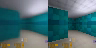}
\end{tabular}
\caption{Predicted (left) and real (right) frames at time-steps $1$, $25$, $66$, $158$ and $200$ using actions from the test data.}
\label{fig:labyrinth_emulator_vs_simulator}
\vspace{-3mm}
\end{figure}

When using actions from the test set, the simulator was able to very reasonably predict frames even after 200 steps. In \figref{fig:labyrinth_emulator_vs_simulator} we compare predicted frames to the real frames at several time-steps 
(the corresponding video can be seen at {\myblue \href{https://drive.google.com/open?id=0B_L2b7VHvBW2UWl5YUtSMXdUbnc}{3DMazes}}). 
We can see that the wall layout is better predicted when walls are closer to the agent, and that corridors 
and far away-walls are not as long as they should be. The lighting on the ceiling is correct on all the frames shown. 

\begin{figure}[t]
%\vskip-0.2cm
\centering
\begin{tabular}{c c c c c c c c c}
 \includegraphics[width=32pt, height=32pt]{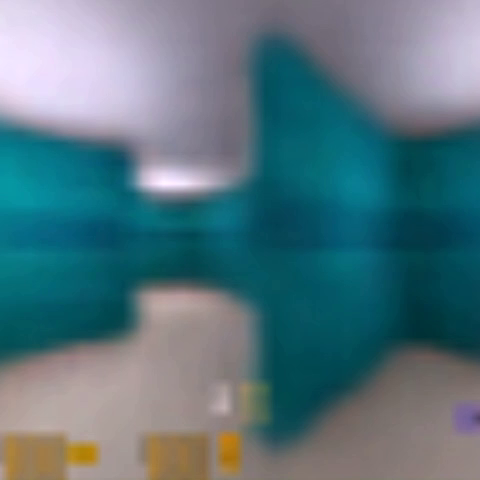} &
 \includegraphics[width=32pt, height=32pt]{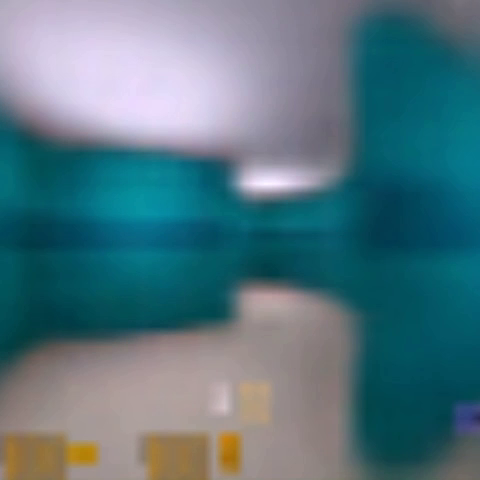} &
 \includegraphics[width=32pt, height=32pt]{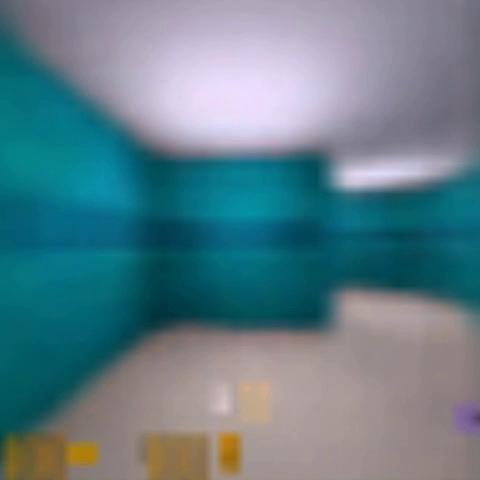} &
 \includegraphics[width=32pt, height=32pt]{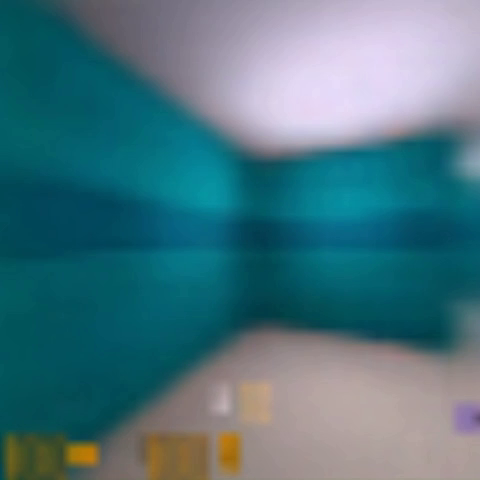} &
 \includegraphics[width=32pt, height=32pt]{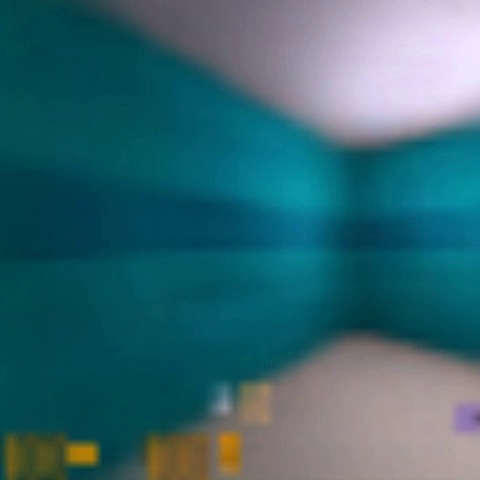} &
 \includegraphics[width=32pt, height=32pt]{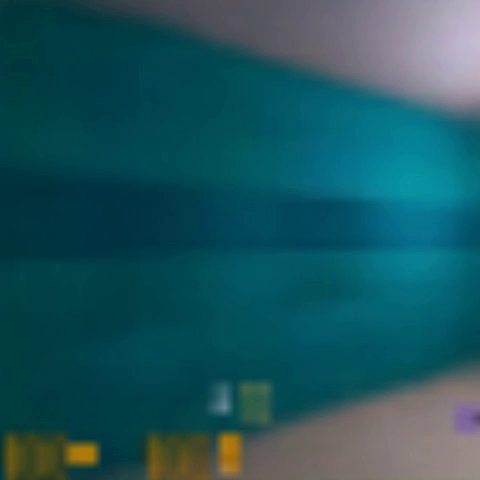} &
 \includegraphics[width=32pt, height=32pt]{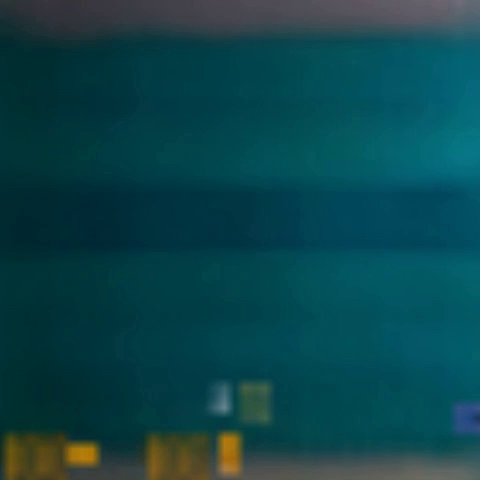} &
 \includegraphics[width=32pt, height=32pt]{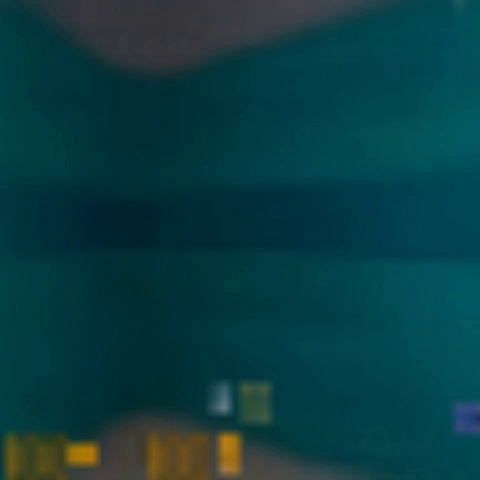} &
 \includegraphics[width=32pt, height=32pt]{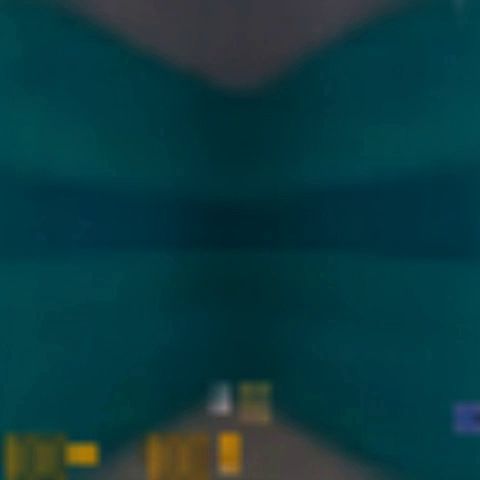} \\
 \includegraphics[width=32pt, height=32pt]{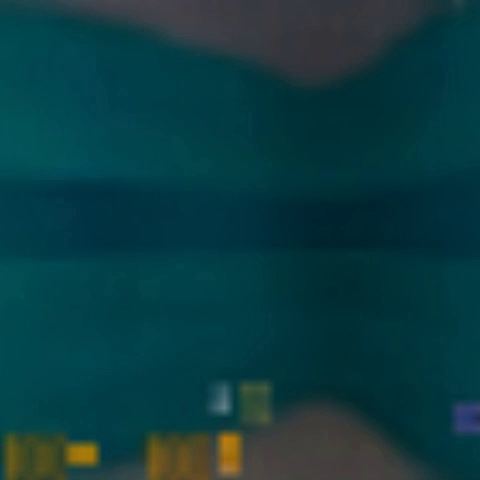} &
 \includegraphics[width=32pt, height=32pt]{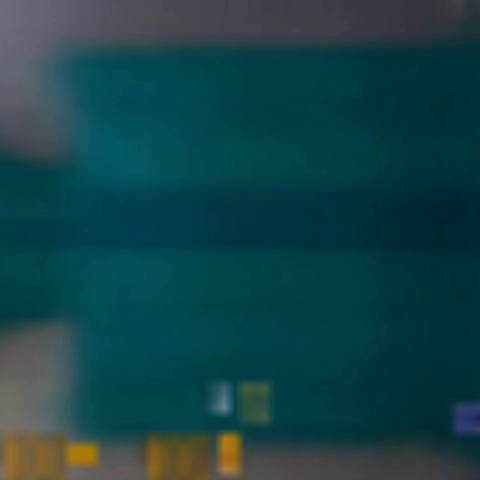} &
 \includegraphics[width=32pt, height=32pt]{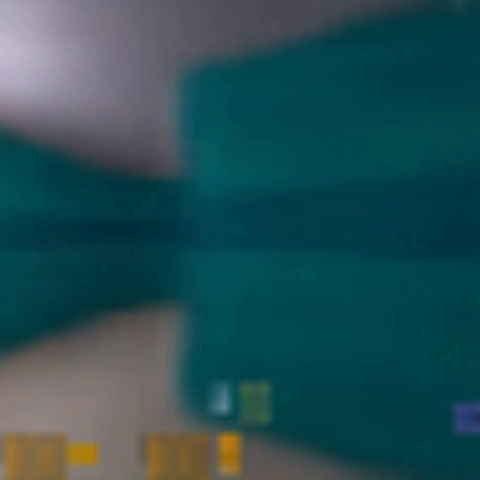} &
 \includegraphics[width=32pt, height=32pt]{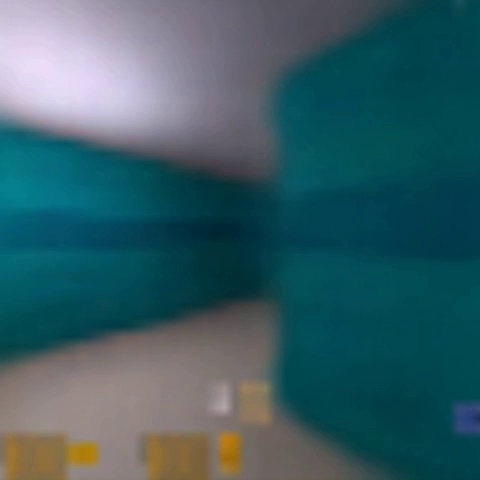} &
 \includegraphics[width=32pt, height=32pt]{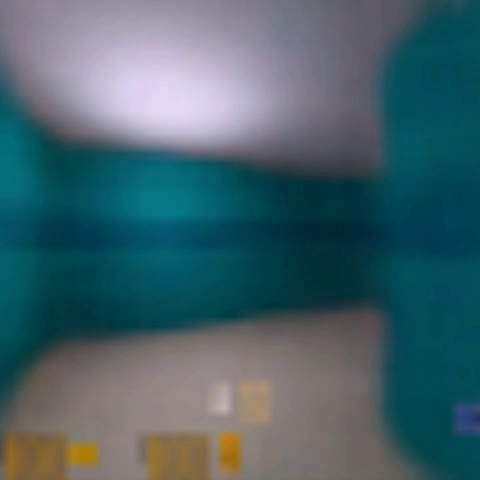} &
 \includegraphics[width=32pt, height=32pt]{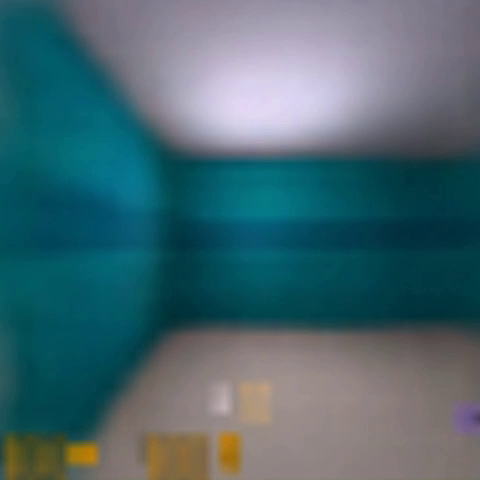} &
 \includegraphics[width=32pt, height=32pt]{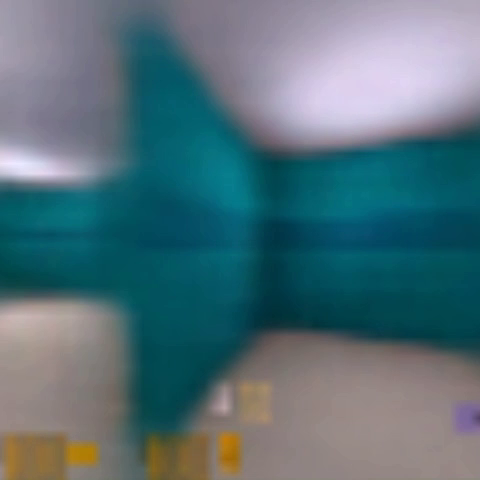} &
 \includegraphics[width=32pt, height=32pt]{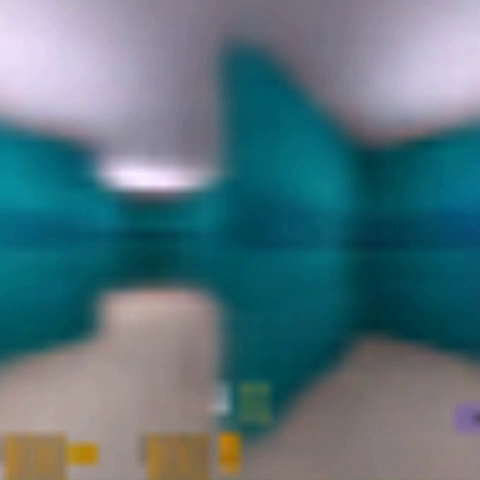} &
 \includegraphics[width=32pt, height=32pt]{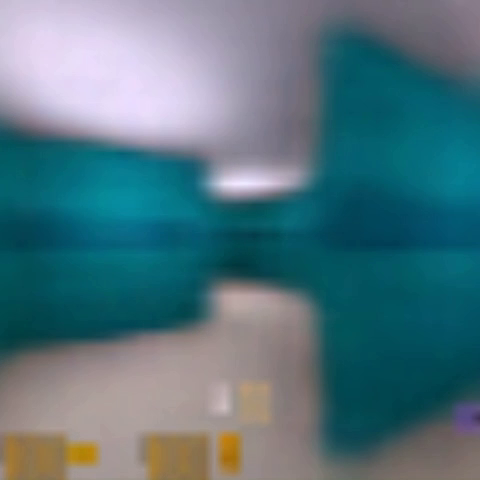}
\end{tabular}
\caption{Salient frames highlighting wall-layout memory after $360^\circ$ spin generated using our simulator with actions taken by a human player.}
\label{fig:lab360spin}
%\vspace{-3mm}
\end{figure}

When using the simulator interactively with actions provided by a human, we could test that the simulator had learnt consistent aspects of the maze: 
when walking into walls, the model maintained their position and layout (in one case we were able to walk through a painting on the wall -- paintings are rare in the dataset and hence it is not unreasonable that they would not be maintained when stress testing the model in this way). 
When taking $360^\circ$ spins, the wall configurations were the same as previously generated and not regenerated afresh, 
and shown in \figref{fig:lab360spin} (see also {\myblue \href{https://drive.google.com/open?id=0B_L2b7VHvBW2TEU3R0xMdC1GVEU}{3DMazes-HPLay}}). 
The coherence of the maze was good for nearby walls, but not at the end of long-corridors. 
%\vspace{-2mm}
\subsection{Model-based Exploration\label{sect:MBexploration}}
%\vspace{-2mm}
The search for exploration strategies better than $\epsilon$-greedy is an active area of research. Various solutions have been proposed, such as density based or optimistic exploration \citep{auer02finite}. 
\cite{oh15action} considered a memory-based approach that steers the agent towards previously unobserved frames. In this section, we test our simulators using a similar approach, but select a group of actions rather than a single action at a time. 
Furthermore, rather than a fixed 2D environment, we consider the more challenging 3D mazes environment. 
This also enables us this present a qualitative analysis, as we can exactly measure and plot the proportion of the maze visited over time. Our aim is to be quantitatively and qualitatively better than random exploration 
(using dithering of $0.7$, as this lead to the best possible random agent). 

We used a 3D maze simulator to predict the outcome of sequences of actions, chosen with a hard-coded policy. Our algorithm (see below) did $N$ Monte-Carlo simulations with randomly selected sequences of actions of fixed length $d$. 
At each time-step $t$, we stored the last 10 observed frames in an episodic memory buffer and compared predicted frames to those in memory.

\begin{figure}[t]
%\vskip-0.2cm
\centering
\subfigure[]{\scalebox{1}{
\includegraphics[width=128pt, height=90pt]{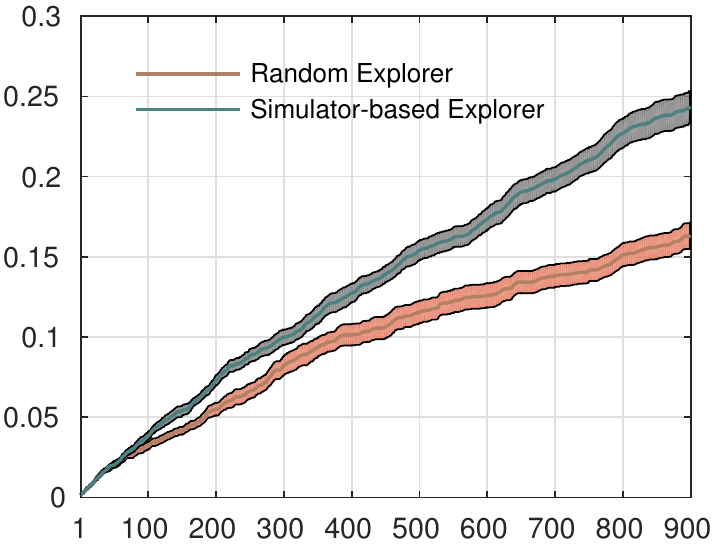}}}
\hskip0.2cm
\subfigure[]{\scalebox{1}{
\includegraphics[width=124pt]{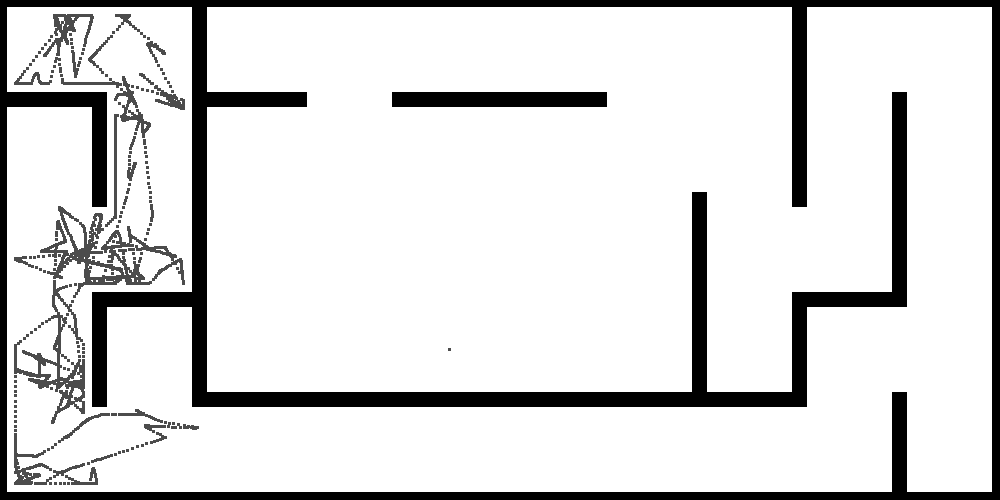} 
\hskip0.1cm
\includegraphics[width=124pt]{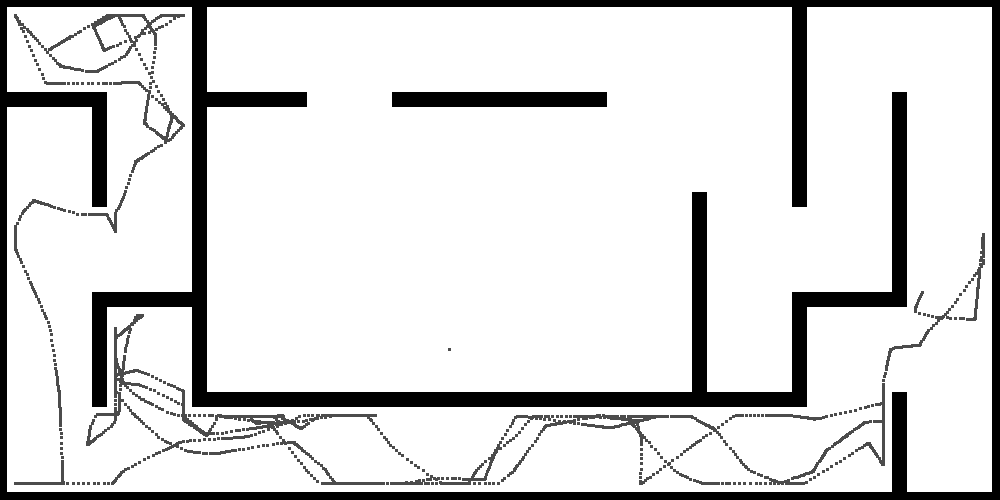}}}
%\vspace{-4mm}
\caption{(a) Average ratio over 10 mazes (shaded is the 68\% confidence interval) of area visited by the random agent and an agent using our model. 
(b) Typical example of paths followed by (left) the random agent and (right) our agent (see the Appendix for more examples).}
\label{fig:expBaselineVsSim}
%\vspace{-5mm}
\end{figure}

\begin{minipage}[]{0.5\textwidth}
\begin{algorithm}[H]
\SetAlgoLined
\For{t = 1, episodeLength, d}{
\For{n = 1, N}{
    Choose random actions $A^n = a_{t:t+d-1}$\;
    Predict $\hat \vx^n_{t+1:t+d}$\;
}
Follow actions in $A^{n_0}$ where $n_0 = {\rm argmax}_n{\rm min}_{j = 0, 10} \parallel\hat \vx^n_{t+d} - \vx^{}_{t - j}\parallel$ } 
\end{algorithm}
\end{minipage}
\begin{minipage}[]{0.5\textwidth}
Our method (see \figref{fig:expBaselineVsSim}(a)) covered $50$\% more of the maze area after $900$ time-steps than random exploration.
These results were obtained with 100 Monte-Carlo simulations and sequences of 6 actions (more details are given in \appref{sec:AppExploration}). 
Comparing typical paths chosen by the random explorer and by our explorer (see \figref{fig:expBaselineVsSim}(b)), we see the our explorer has much smoother trajectories. 
\end{minipage}
This is a good local exploration strategy that leads to faster movement through corridors. To transform this into a good global exploration strategy, our explorer would have to be augmented with a better memory in order to avoid going down the same corridor twice. 
These sorts of smooth local exploration strategies could also be useful in navigation problems.

\section{Prediction-Independent Simulators\label{sect:jumpy}} 
A prediction-independent simulator has state transitions of the form  $\vs_t = f(\vs_{t-1},a_{t-1})$, which therefore do not require the high-dimensional predictions. 
In the Atari environment, for example, this avoids having to project from the state space of dimension 1024 into the observation space of dimension 100,800 (210$\times$160$\times$3)  
through the decoding function ${\cal D}$, and vice versa through the encoding function ${\cal C}$ -- in the used structure this enables saving around 200 million flops at each time-step.

For the state transition, we found that a working structure was to use Eqs. \eqref{eq:state_enc}--\eqref{eq:alstm_state} with $\vz_t=\vh_t$ and with different parameters for the warm-up and prediction phases.
As for the prediction-dependent simulator, we used a warm-up phase of length $\tau = 10$, but we did backpropagate the gradient back to time-step five in order to learn the encoding function ${\cal C}$. 
%  -- Height (210-8)/2+1=102; (102+2-6)/2+1=50; (50+2-6)/2+1=24; (24-4)/2+1=11
%  -- Width (160+2-8)/2+1=78; (78+2-6)/2+1=38; (38+2-6)/2+1=18; (18-4)/2+1=8
% 2x3x102x(2x8x8)x64+2x64x50x(2x6x6)x32+2x32x24x(2x6x6)x32+2x32x11x(2x4x4)x32
% 2*3*102*(2*8*8)*64 + 2*64*50*(2*6*6)*32 + 2*32*24*(2*6*6)*32 + 2*3*11*(2*4*4)*32= 23,365,632

Our analysis on Atari (see \appref{sec:AppPISimulators}) suggests that the prediction-independent simulator is much more sensitive to changes in the state transition structure and in the training scheme than the prediction-dependent simulator.
We found that using prediction length $T=15$ gave much worse long-term accuracy than with the prediction-dependent simulator. This problem could be alleviated with the use of prediction length $T=30$ through truncated backpropagation.

\begin{figure}[t]
%\hskip-0.3cm
\centering
\subfigure[]{
\scalebox{0.79}{\includegraphics[]{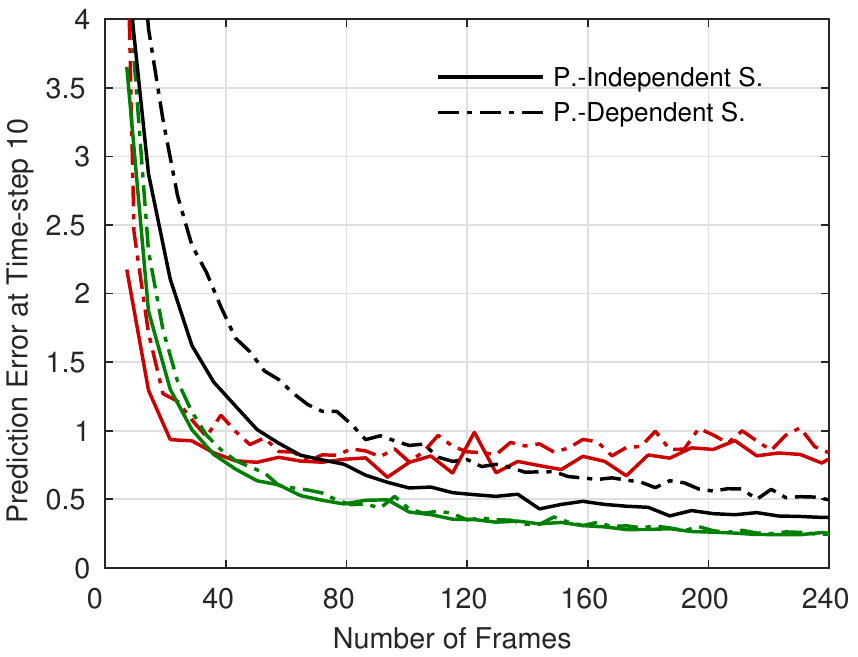}}
\scalebox{0.79}{\includegraphics[]{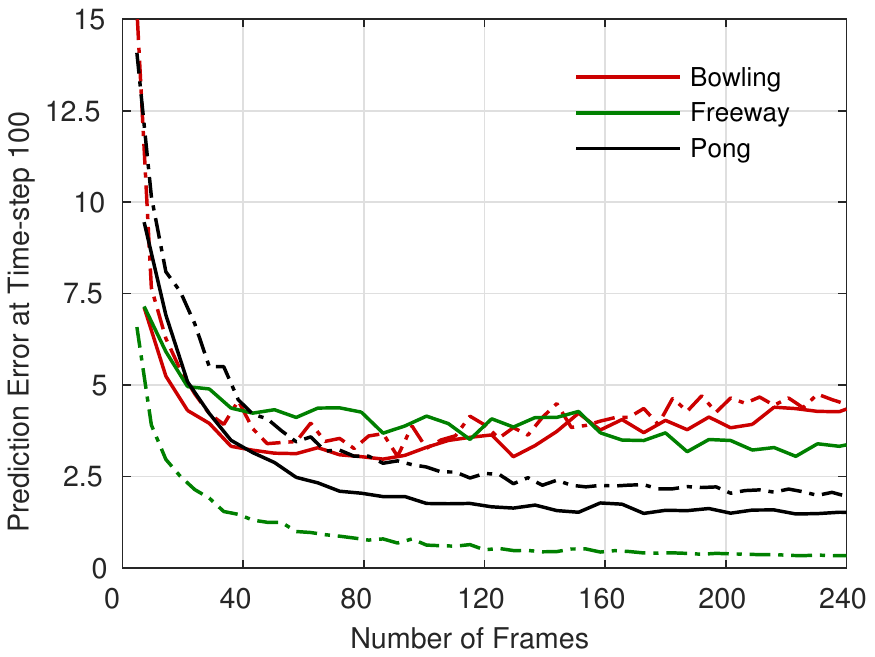}}}
\subfigure[]{
\scalebox{0.79}{\includegraphics[]{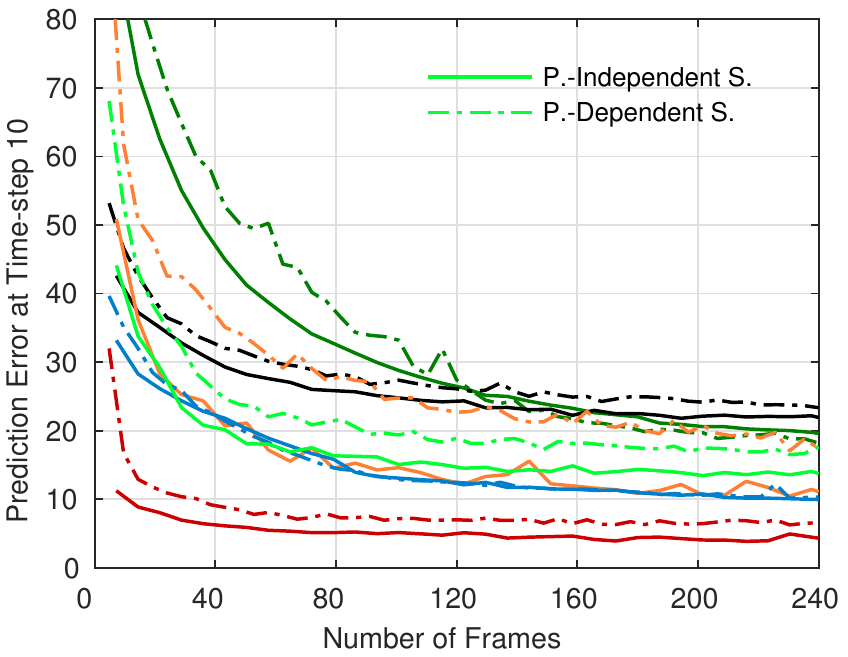}}
\scalebox{0.79}{\includegraphics[]{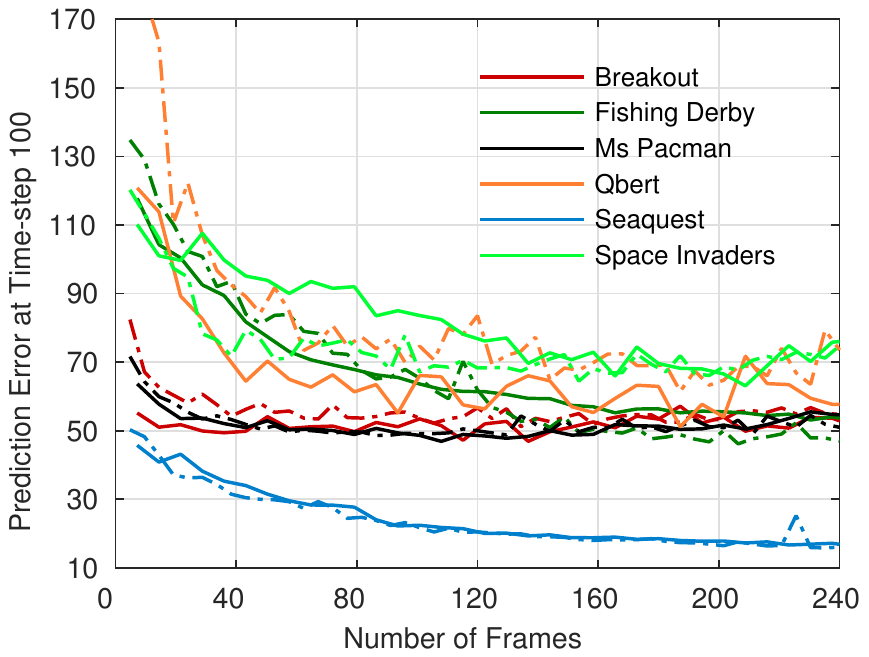}}}
%\vspace{-2mm}
\caption{Prediction error vs number of frames seen by the model (excluding warm-up frames) for the prediction-dependent and prediction-independent simulators using BPTT(15, 2) for (a) Bowling, Freeway, Pong and (b) Breakout, Fishing Derby, Ms Pacman, Qbert, Seaquest, Space Invaders (the prediction-dependent simulator is trained with the 0\%-100\%\PDT~training scheme).}
\label{fig:predErr}
\end{figure}

\figref{fig:predErr} shows a comparison of the prediction-dependent and prediction-independent simulators using $T=30$ through two subsequences of length 15 (we indicate this as BPTT(15, 2), even though in the prediction-independent simulator we did backpropagate 
the gradient to the warm-up phase).

When looking at the videos available at {\myblue  \href{https://drive.google.com/drive/folders/0B_L2b7VHvBW2c1l0NGZjZ191QUU?usp=sharing}{PI-Simulators}}, 
we can notice that the prediction-independent simulator tends to give 
worse type of long-term prediction. In Fishing Derby for example, in the long-term the model tends to create fish of smaller dimension in addition to the fish present in the real frames. 
Nevertheless, for some difficult games the prediction-independent simulator achieves better performance than the prediction-dependent simulator. 
More investigation about alternative state transitions and training schemes would need to be performed to obtain the same overall level of accuracy as with the prediction-dependent simulator.
%{\myblue \href{https://drive.google.com/drive/folders/0B_L2b7VHvBW2c1l0NGZjZ191QUU?usp=sharing}{PI-Simulators}}
%{\myblue \href{https://drive.google.com/file/d/0B_L2b7VHvBW2VkctdThfUEJfZEE/view?usp=sharing}{Bowling}}, 
%{\myblue \href{https://drive.google.com/file/d/0B_L2b7VHvBW2aDRWamNtMHZOeGs/view?usp=sharing}{FDerby}}, 
%{\myblue \href{https://drive.google.com/file/d/0B_L2b7VHvBW2d1l4ME5VeDJzamc/view?usp=sharing}{Freeway}}, 
%{\myblue \href{https://drive.google.com/file/d/0B_L2b7VHvBW2bWRPZ2JLTWNORkE/view?usp=sharing}{Pong}}, 
%{\myblue \href{https://drive.google.com/file/d/0B_L2b7VHvBW2NXdmOTVMZWhwRms/view?usp=sharing}{Riverraid}}, 
%{\myblue \href{https://drive.google.com/file/d/0B_L2b7VHvBW2bWc1QS1DV0txblU/view?usp=sharing}{Seaquest}}. 

\section{Discussion}
In this paper we have introduced an approach to simulate action-conditional dynamics and demonstrated that is highly adaptable to different environments,
ranging from Atari games to 3D car racing environments and mazes. We showed state-of-the-art results on Atari, and demonstrated the feasibility 
of live human play in all three task families. The system is able to capture complex and long-term interactions, and displays a sense of spatial and temporal coherence that has, 
to our knowledge, not been demonstrated on high-dimensional time-series data such as these.

We have presented an in-deep analysis on the effect of different training approaches on short and long-term prediction capabilities, and showed that moving towards schemes in which 
the simulator relies less on past observations to form future predictions has the effect on focussing model resources on learning the global dynamics of the environment, 
leading to dramatic improvements in the long-term predictions. However, this requires a distribution of resources that impacts short-term performance, which can be harmful to the overall performance of the model for some games. 
This trade-off is also causing the model to be less robust to states of the environment not seen during training.
To alleviate this problem would require the design of more sophisticated model architectures than the ones considered here. 
Whilst it is also expected that more ad-hoc architectures would be less sensitive to different training approaches, we believe that 
guiding the noise as well as teaching the model to make use of past information through the objective function would still be beneficial for improving long-term prediction.

Complex environments have compositional structure, such as independently moving objects and other phenomena that only rarely interact. 
In order for our simulators to better capture this compositional structure, we may need to develop specialised functional forms and memory 
stores that are better suited to dealing with independent representations and their interlinked interactions and relationships. More homogeneous 
deep network architectures such as the one presented here are clearly not optimal for these domains, as can be seen in Atari environments such as Ms Pacman 
where the system has trouble keeping track of multiple independently moving ghosts. 
Whilst the LSTM memory and our training scheme have proven to capture long-term dependencies, alternative memory structures are required in order, for example, 
to learn spatial coherence at a more global level than the one displayed by our model in the 3D mazes in oder to do navigation.

In the case of action-conditional dynamics, the policy-induced data distribution does not cover the state space and might in fact be nonstationary over an agent lifetime. 
This can cause some regions of the state space to be oversampled, whereas the regions we might actually care about the most -- those just \emph{around} the agent policy state distribution -- to be underrepresented. 
In addition, this induces biases in the data that will ultimately not enable the model learn the environment dynamics correctly. 
As verified from the experiments in this paper, both on live human play and model-based exploration, this problem is not yet as pressing as might be expected in some environments.
However, our simulators displayed limitations and faults due to the specificities of the training data, 
such as for example predicting an event based on the recognition of a particular sequence of actions always co-occurring with this event in the training data rather than on the recognition of the real causes.  

Finally, a limitation of our approach is that, however capable it might be, it is a deterministic model designed for deterministic environments. 
Clearly most real world environments involve noisy state transitions, and future work will have to address the extension of the techniques developed in this paper to more generative temporal models.

\subsubsection*{Acknowledgments}
The authors would like to thank David Barber for helping with the graphical model interpretation, Alex Pritzel for preparing the DQN data, Yori Zwols and Frederic Besse for helping with the implementation of the model, and Oriol Vinyals, Yee Whye Teh, Junhyuk Oh, and the anonymous reviewers for useful discussions and feedback on the manuscript.

\begin{small}
\bibliography{RESiNet}
\end{small}

\clearpage
\appendix

\section{Data, Preprocessing and Training Algorithm}
%\vspace{-2mm}
When generating the data, each selected action was repeated for 4 time-steps and only the 4th frame was recorded for the analysis. 
The RGB images were preprocessed by subtracting mean pixel values (calculated separately for each color channel and over an initial set of 2048 frames only) and by dividing each pixel value by 255. 

As stochastic gradient algorithm, we used centered RMSProp \citep{graves13generating} with learning rate\footnote{We found that using a higher learning rate value of 2e-5 would generally increase convergence 
speed but cause major instability issues, suggesting that gradient clipping would need to be used.} 1e-5, epsilon 0.01, momentum 0.9, decay 0.95, and mini-batch size 16.
The model was implemented in \href{http://torch.ch}{Torch}, using the default initialization of the parameters. The state $\vs_1$ was initialized to zero.

\section{Prediction-Dependent Simulators}
\vspace{-2mm}
As baseline for the single-step simulators we used the following state transition:
\begin{align*}
\textrm{Encoding: } & \vz_{t-1} = {\cal C}(\mathbb{I}(\hat \vx_{t-1}, \vx_{t-1}))\,,\\
\textrm{Action fusion: } & \ha_{t} = \vW^h \vh_{t-1}\otimes \vW^a \va_{t-1}\,,\\
\textrm{Gate update: } & \vi_t = {\sigma}(\vW^{iv}\ha_{t} +\vW^{iz}\vz_{t-1})\,, \hskip0.1cm \vf_t = {\sigma}(\vW^{fv}\ha_t +\vW^{fs}\vz_{t-1})\,,\\ 
& \vo_t = {\sigma}(\vW^{ov}\ha_t + \vW^{oz}\vz_{t-1})\,,\\
\textrm{Cell update: } & \vc_t = \vf_t\otimes \vc_{t-1} + \vi_t\otimes \textrm{tanh}(\vW^{cv}\ha_{t}+\vW^{cz}\vz_{t-1})\,,\\
\textrm{State update: } & \vh_t = \vo_t\otimes \textrm{tanh}(\vc_t)\,, 
\end{align*}
with vectors $\vh_{t-1}$ and $\ha_t$ of dimension 1024 and 2048 respectively. 

\subsection{Atari\label{sec:AppAtari}}
\begin{figure}[t]
\centering
\resizebox{\textwidth}{!}{\includegraphics[]{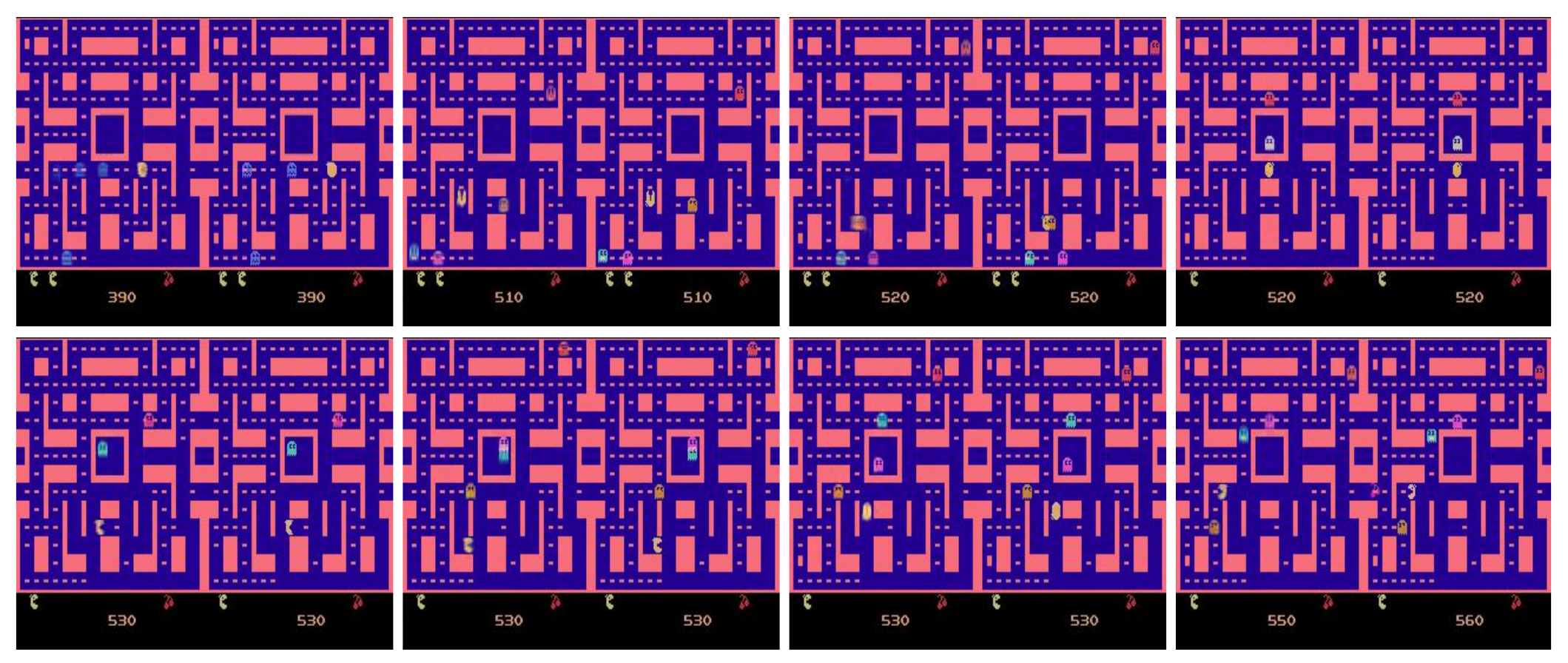}}
\caption{Prediction that demonstrates overfitting of the model when trained on one million frames.}
\label{fig:PacmanOver}
\end{figure}

%\vspace{-2mm}
\begin{minipage}{0.55\textwidth}
We used a trained DQN agent (the scores are given in the table on the right) to generate training and test datasets consisting of 5,000,000 and 1,000,000 
(210$\times$160) RGB images respectively, with actions chosen according to an $\epsilon=0.2$-greedy policy.  
Such a large number of training frames was necessary to prevent our simulators from strongly overfitting to the training data. 
This would be the case with, for example, one million training frames, as shown in \figref{fig:PacmanOver} (the corresponding video can be seen at 
{\myblue \href{https://drive.google.com/open?id=0B_L2b7VHvBW2N1lWSHJKWEdOZ3c}{MSPacman}}). The ghosts are in frightened mode at time-step 1 (first image),  
and have returned to chase mode at time-step 63 (second image). The simulator is able to predict the exact time of return to the chase mode without sufficient history, 
which suggests that the sequence was memorized. 
\end{minipage}
\begin{minipage}{0.45\textwidth}
\centering
\begin{tabular}{lll}
\toprule
%\multicolumn{2}{c}{Part}                   \\
%\cmidrule{1-2}
Game Name     & DQN Score \\%    & Frames\\
\midrule
%Alien 3801.11 200,000,000 
%Amidar 1991.153846 on frame 180000000 
%Assault 5419.037037 on frame 81500000 
%Asterix 16690.909091 on frame 199000000
%Asteroids 1131.151079 on frame 111500000
%Atlantis 232375.000000 on frame 176500000  
%Bank Heist 1064.057971 on frame 161000000
%Battle Zone 33655.737705 on frame 190000000
%Beam Rider 16650.714286 on frame 164000000
%Berzerk 1293.541667 on frame 194000000
Bowling & 51.84\\ %& 12,000,000 \\ 
%Boxing & 87.64 & 34,000,000\\ 
Breakout & 396.25\\ %& 180,500,000\\  
%Centipede 8700.647887 on frame 500000
%Chopper Command 6700.000000 on frame 171500000
%Crazy Climber 127782.758621 on frame 161000000
%Defender 35696.875000 on frame 158000000
%Demon Attack 17 -- 98450.000000 on frame 125000000  
%Double Dunk -4.514286 on frame 195000000
%Enduro & 1357.14 & 18,000,000\\
Fishing Derby & 19.30\\% & 81,000,000\\
Freeway & 33.38\\% & 91,500,000\\
%Frostbite 2318.888889 on frame 4500000 
%Gopher & 13766.67 & 14,000,0000\\
%Gravitar 24 -- 689.613527 on frame 188500000
%Hero 18584.553571 on frame 173000000
%Ice Hockey -2.026316 on frame 151500000
%Jamesbond 625.000000 on frame 193000000
%Kangaroo 14185.714286 on frame 169000000
%Krull 8953.666667 on frame 36500000
%Kung Fu Master 30735.714286 on frame 112500000
%Montezuma Revenge 0.000000 on frame 500000
Ms Pacman & 2963.31\\% & 81,000,000\\
%Name this Game 11090.000000 on frame 115000000
%Phoenix 15608.333333 on frame 20000000
%Pitfall 35 -- 0.000000 on frame 500000 pitfall
Pong & 20.88\\% & 128,500,000\\
%Private Eye 102.173913 on frame 72500000
Qbert & 14,865.43\\% & 188,000,000\\
Riverraid & 13,593.49\\% & 161,500,000\\
%Road Runner & 44348.96 & 131,500,000\\ 
%Robotank 66.611111 on frame 188500000
Seaquest & 17,250.31\\% & 133,000,000 \\
%Skiing -11512.510753 on frame 500000\\
%Solaris 4958.823529 on frame 174500000\\
Space Invaders & 2952.09\\% & 131,000,000\\
%Star Gunner 64885.294118 on frame 194000000\\
%Surround -1.714286 on frame 81500000
%Tennis 0.000000 on frame 5000000
%Time Plot 10036.363636 on frame 182500000
%Tutankham 215.000000 on frame 175000000
%Up n Down 18100.200000 on frame 61500000
%Venture 239.655172 on frame 77000000
%Video Pinball & 509614.00 & 44,000,000\\
%Wizard of Wor 7023.437500 on frame 143000000
%Yars Revenge 26283.428571 on frame 188500000
%Zaxxon 12477.777778 on frame 133000000
\bottomrule
\end{tabular}
%\end{table}
\vspace{+2mm}
\end{minipage}
 % ms_pacman_5FMOVERUP : 0:01, 1:03, 1:21, 1:41, 2:22, 2:56, 3:04, 3:20

The encoding consisted of 4 convolutional layers with 64, 32, 32 and 32 filters, of size $8\times8$, $6\times6$, $6\times6$, and $4\times4$, stride 2, and padding 0, 1, 1, 0 and  1, 1, 1, 0 for the height and width respectively. 
Every layer was followed by a randomized rectified linear function (RReLU) \citep{xu15empirical} with parameters $l=1/8$, $u=1/3$. 
The output tensor of the convolutional layers of dimension $32\times11\times8$ was then flattened into the vector $\vz_t$ of dimension 2816. 
The decoding consisted of one fully-connected layer with 2816 hidden units followed by 4 full convolutional layers with the inverse symmetric structure of the encoding transformation: 
32, 32, 32 and 64 filters, of size $4\times4$, $6\times 6$, $6\times6$, and $8\times8$, stride 2, and padding 0, 1, 1, 0  and  0, 1, 1, 1. Each full convolutional layer (except the last one) was followed by a RReLU. 

\begin{figure}[t]
\hskip-0.1cm
\scalebox{0.66}{\includegraphics[]{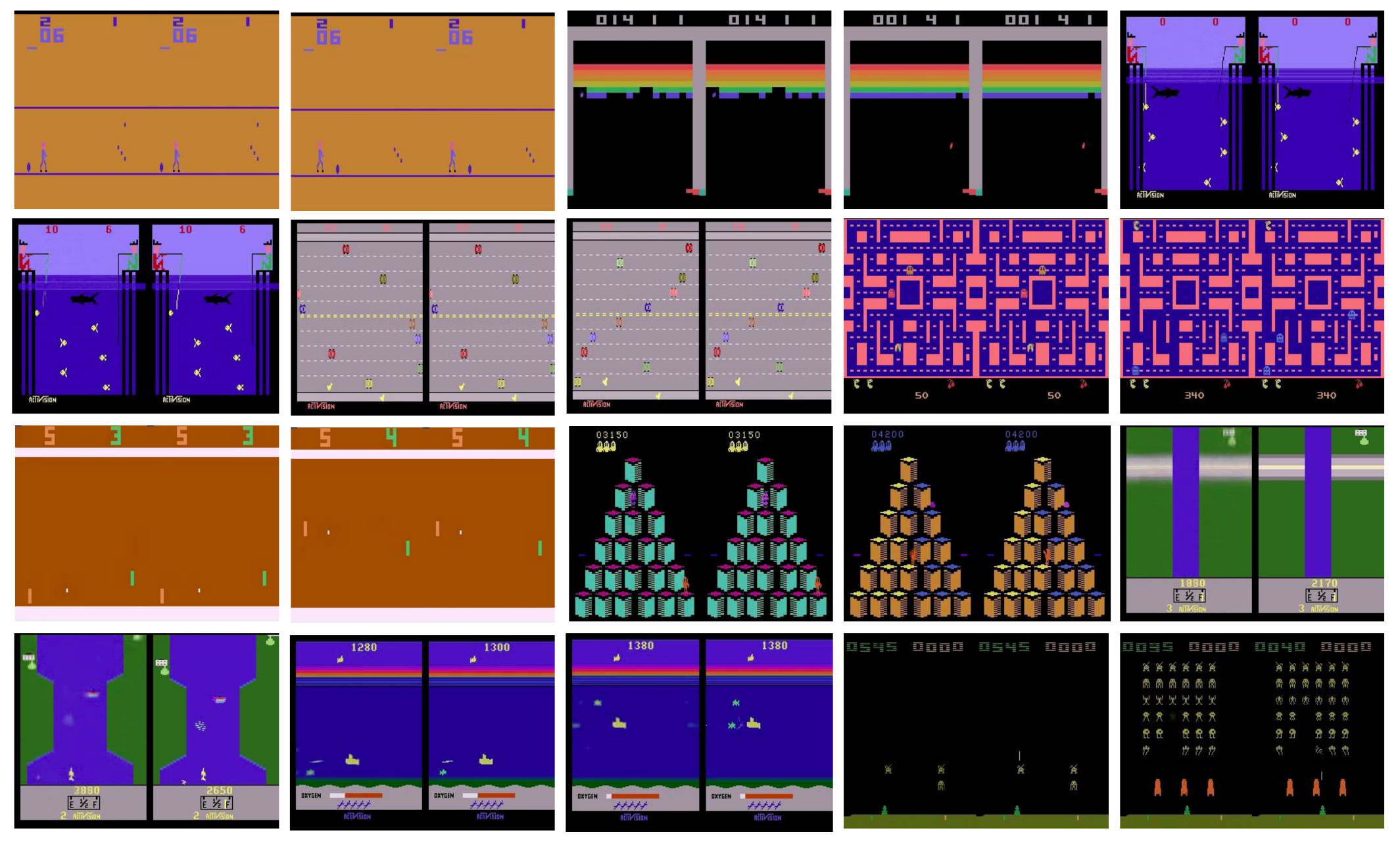}}
\caption{One example of 200 time-step ahead prediction for each of the 10 Atari games. Displayed are predicted (left) and real (right) frames at time-steps 100 and 200.}
\label{fig:AtariAll}
% Breakout ATAbS16nH10mask21-12-20.21-12-20SEnF64RReLUlR2e-5M 13D_47000itN
% Seaquest ATARIbS16nH10mask21-12-20.21-12-20SEnF64RReLUlR2e-5MnH2048 7D14H_23900it
\end{figure}

In \figref{fig:AtariAll}, we show one example of successful prediction at time-steps 100 and 200 for each game. 

\subsubsection{Short-Term Versus Long-Term Accuracy \label{sec:AppPDT}}
\begin{figure}[h!] 
\vskip-0.5cm
\scalebox{0.79}{\includegraphics[]{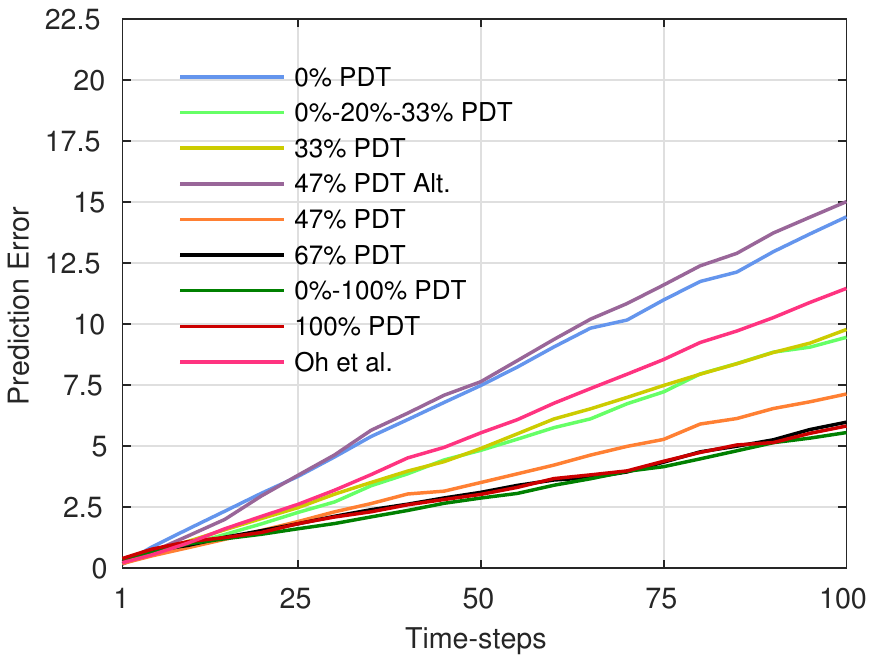}}
%\hskip0.1cm
\scalebox{0.79}{\includegraphics[]{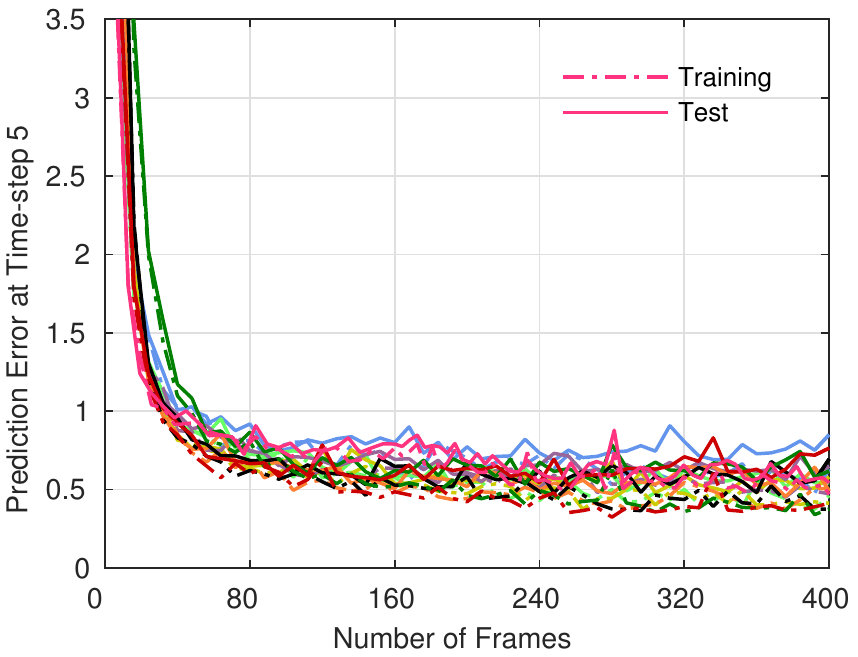}}\\
\subfigure[]{\scalebox{0.79}{\includegraphics[]{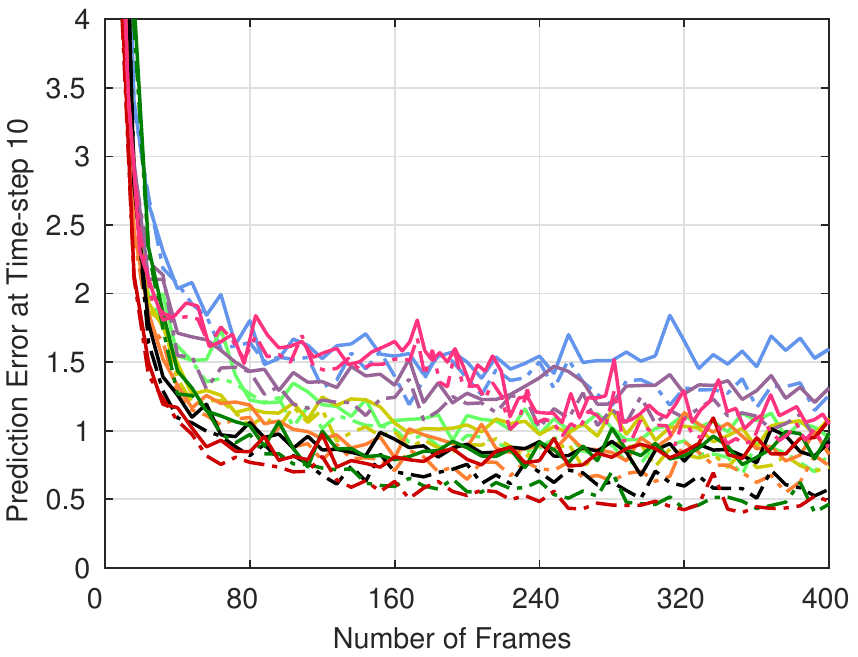}}
%\hskip0.1cm
\scalebox{0.79}{\includegraphics[]{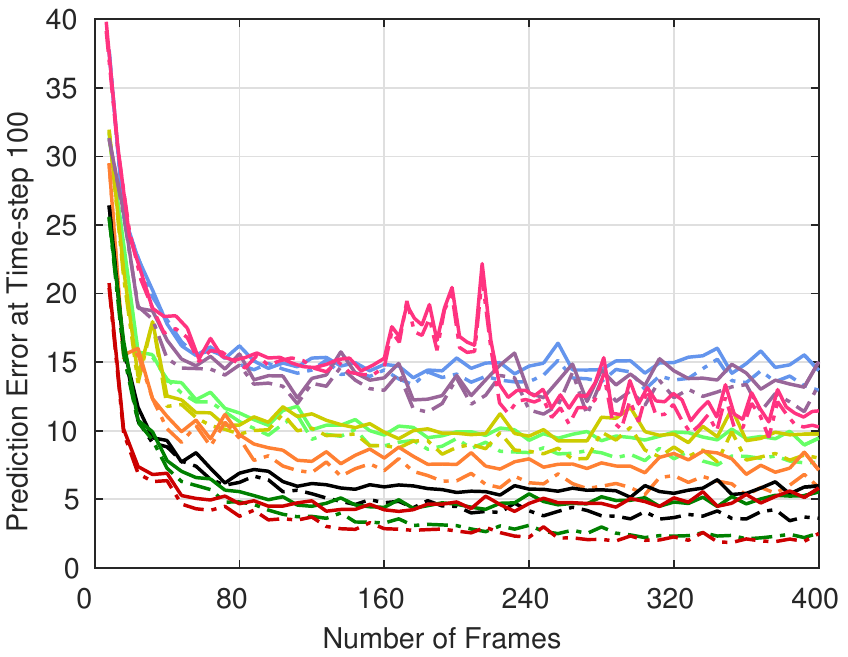}}}
\scalebox{0.79}{\includegraphics[]{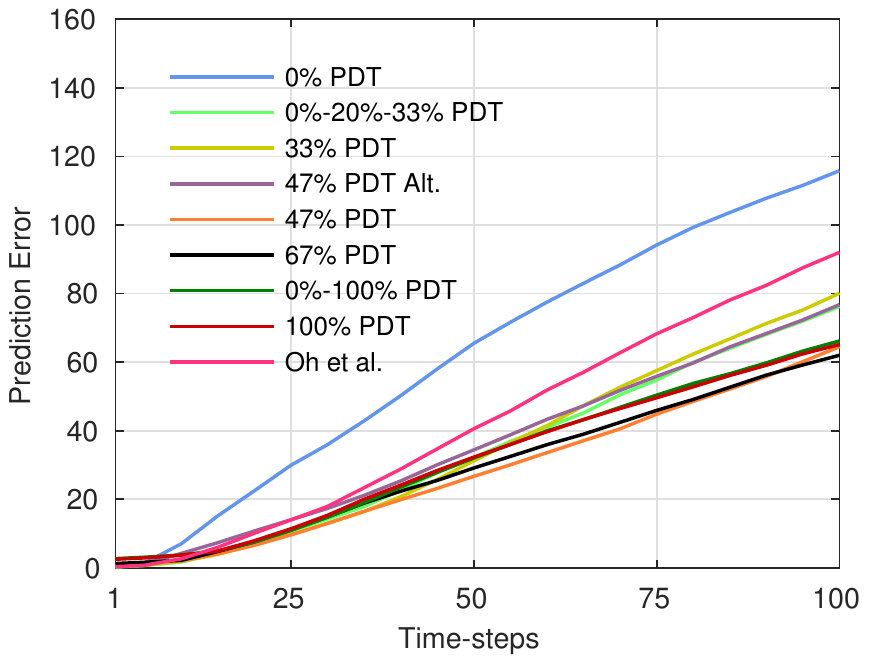}}
%\hskip0.1cm
\scalebox{0.79}{\includegraphics[]{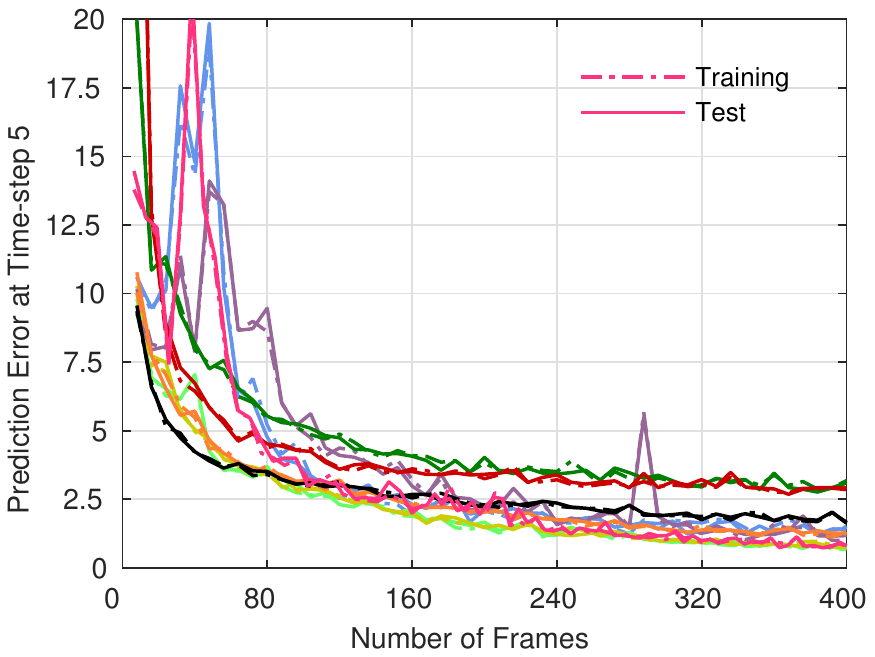}}
\subfigure[]{
\scalebox{0.79}{\includegraphics[]{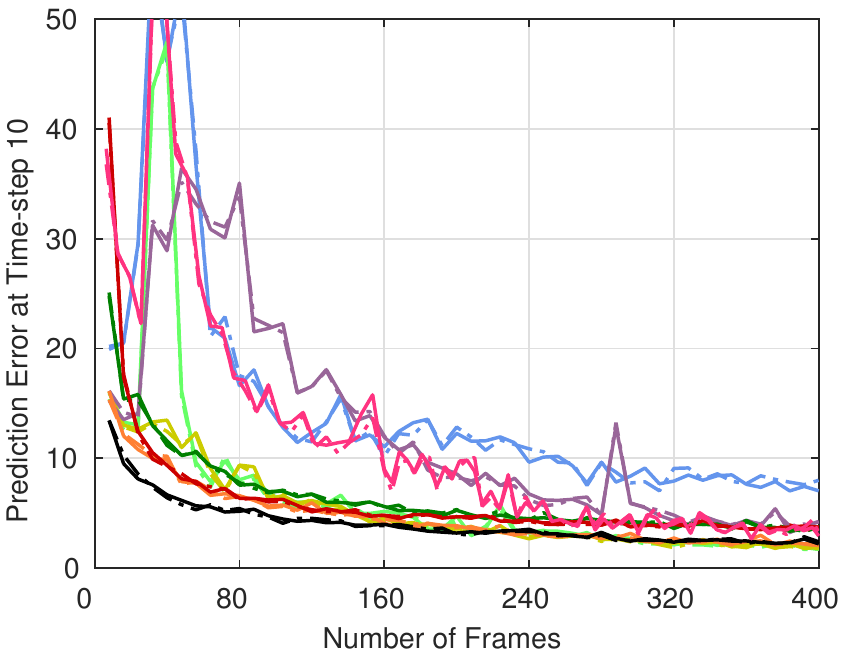}}
%\hskip0.1cm
\scalebox{0.79}{\includegraphics[]{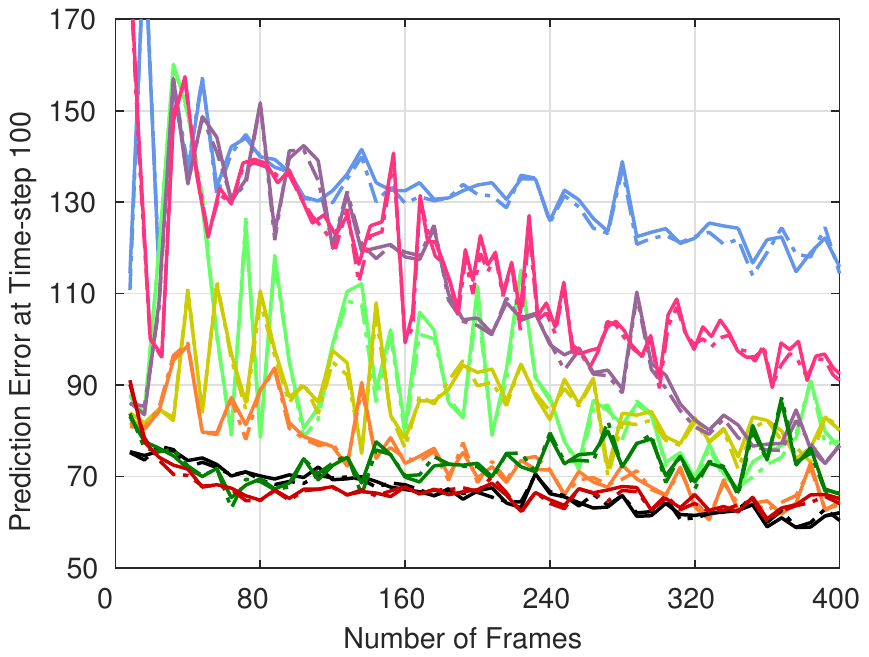}}}
%\vskip-0.3cm
\caption{Prediction error (average over 10,000 sequences) for different training schemes on (a) Bowling and (b) Breakout. Number of frames is in millions.}
\label{fig:predErrFBTBowling-Breakout}
\end{figure}

\begin{figure}[] % Figures obtained with predErrFBTypeAppendix
\vskip-0.5cm
\scalebox{0.79}{\includegraphics[]{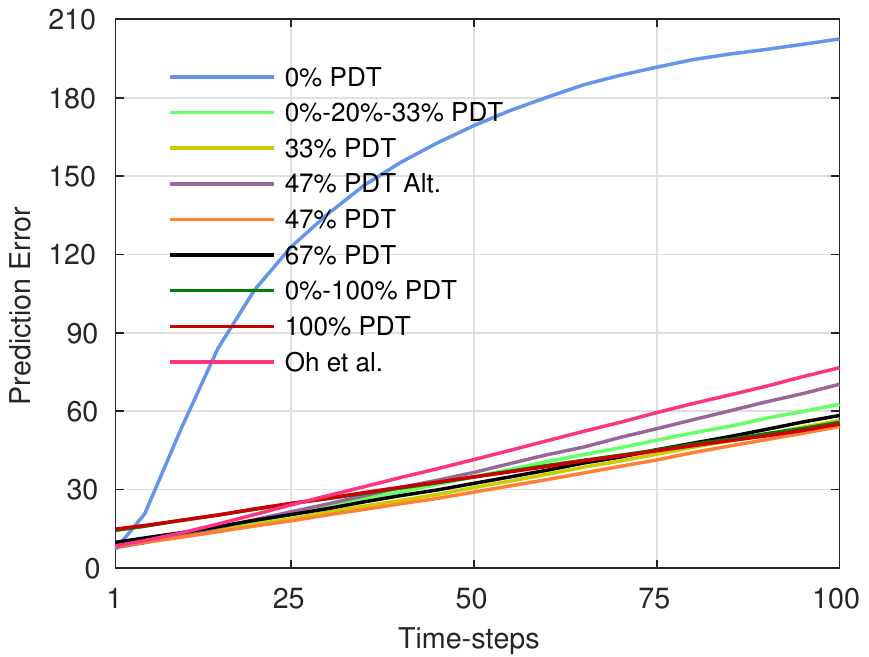}}
%\hskip0.1cm
\scalebox{0.79}{\includegraphics[]{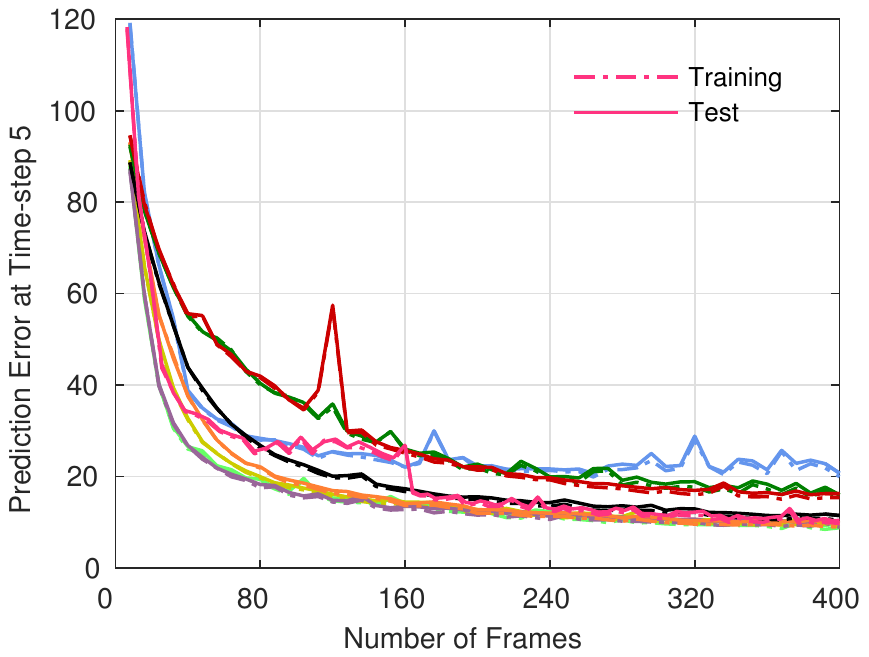}}
\subfigure[]{\scalebox{0.79}{\includegraphics[]{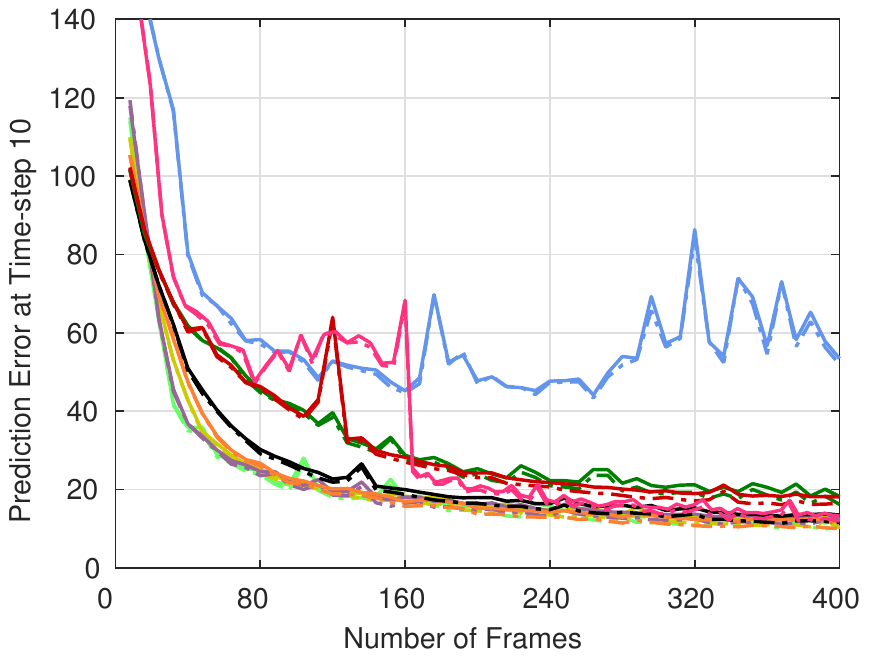}}
%\hskip0.1cm
\scalebox{0.79}{\includegraphics[]{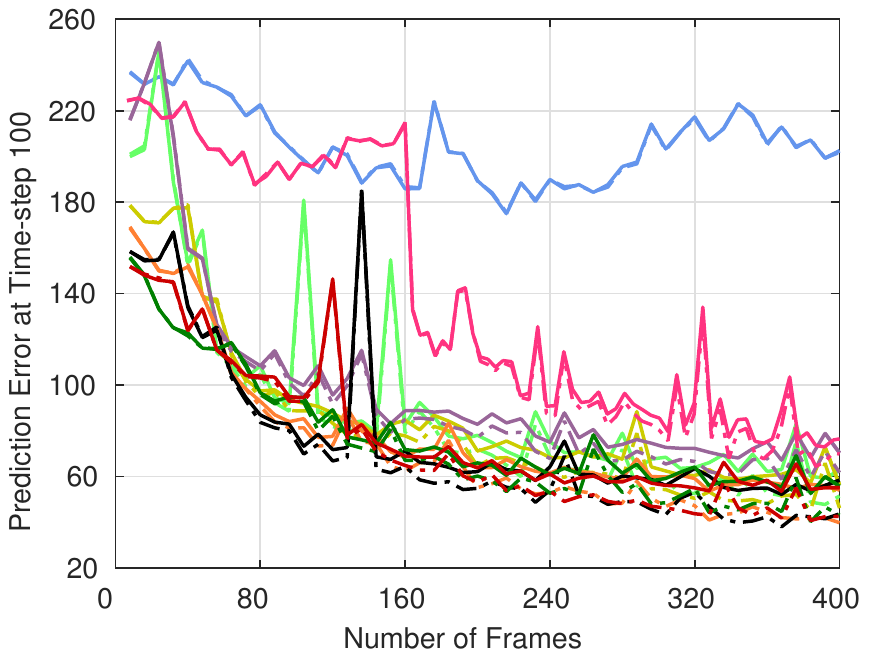}}}
\scalebox{0.79}{\includegraphics[]{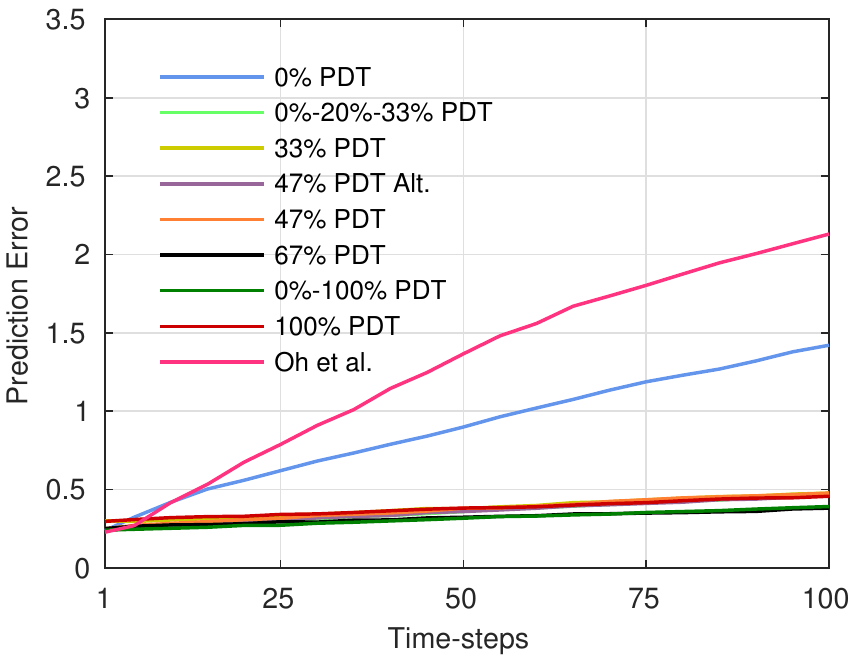}}
%\hskip0.1cm
\scalebox{0.79}{\includegraphics[]{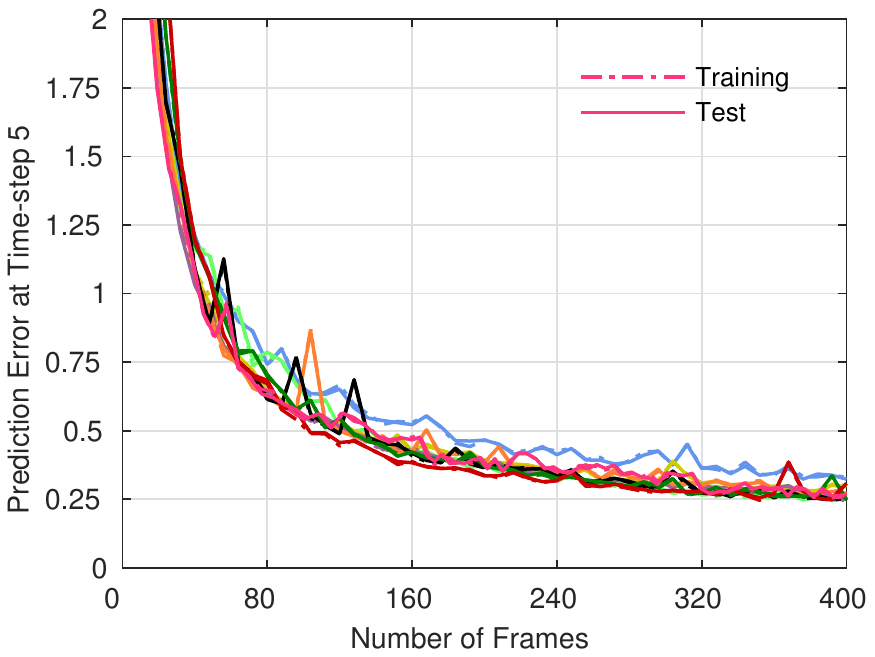}}
\subfigure[]{\scalebox{0.79}{\includegraphics[]{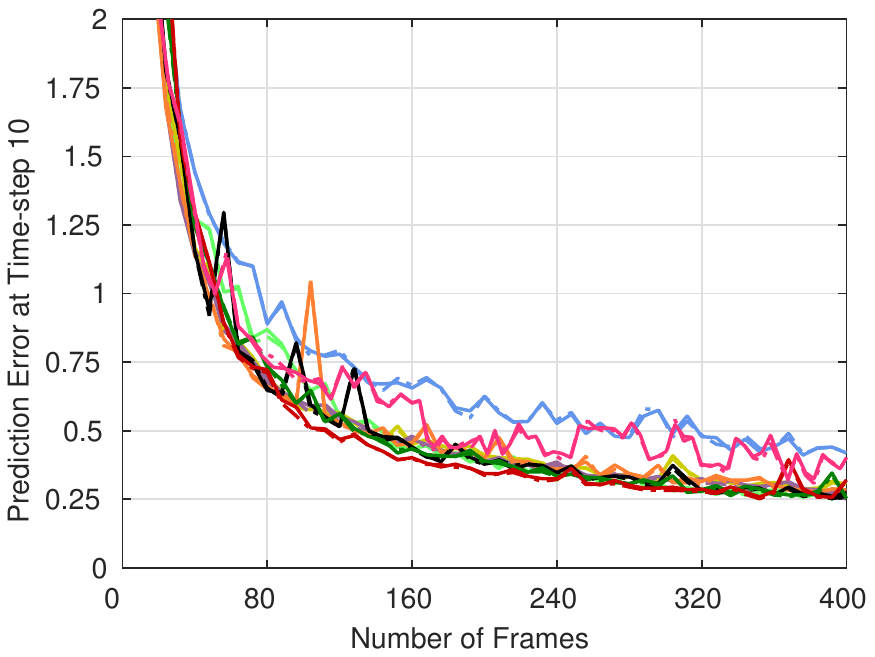}}
\scalebox{0.79}{\includegraphics[]{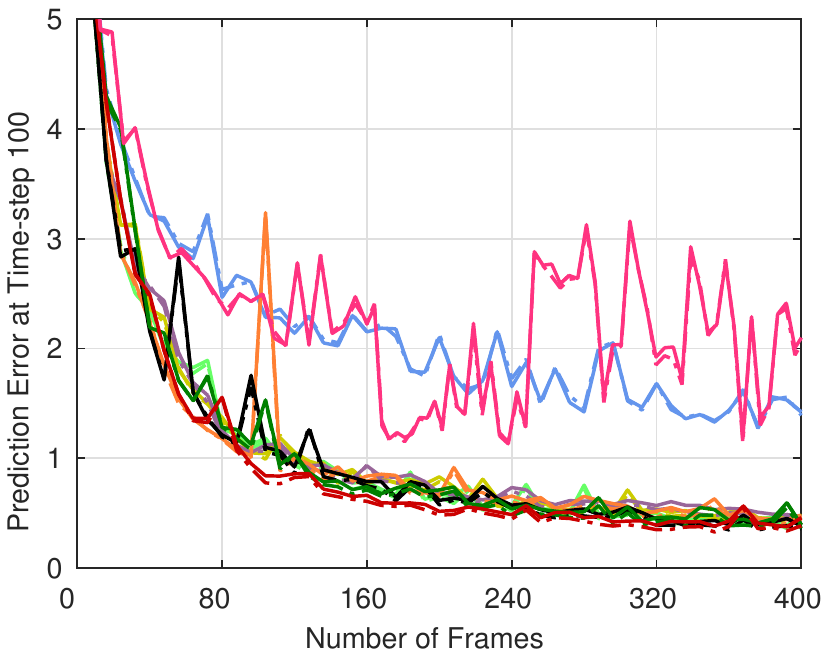}}}
\caption{Prediction error for different training schemes on (a) Fishing Derby and (b) Freeway.}
\label{fig:predErrFBTFishingDerby-Freeway}
\end{figure}

\begin{figure}[] % Figures obtained with predErrFBTypeAppendix
\vskip-0.5cm
\scalebox{0.79}{\includegraphics[]{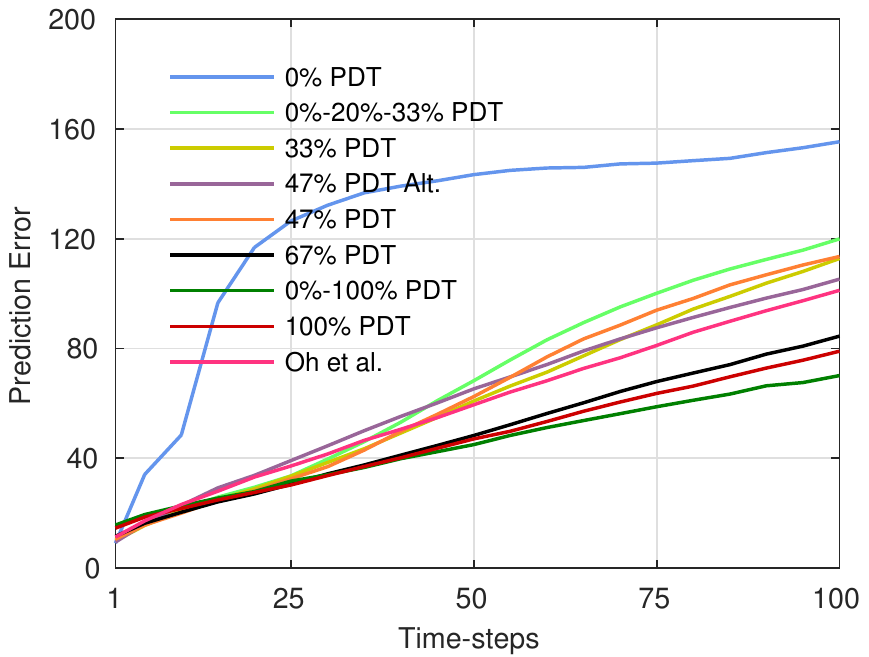}}
%\hskip0.1cm
\scalebox{0.79}{\includegraphics[]{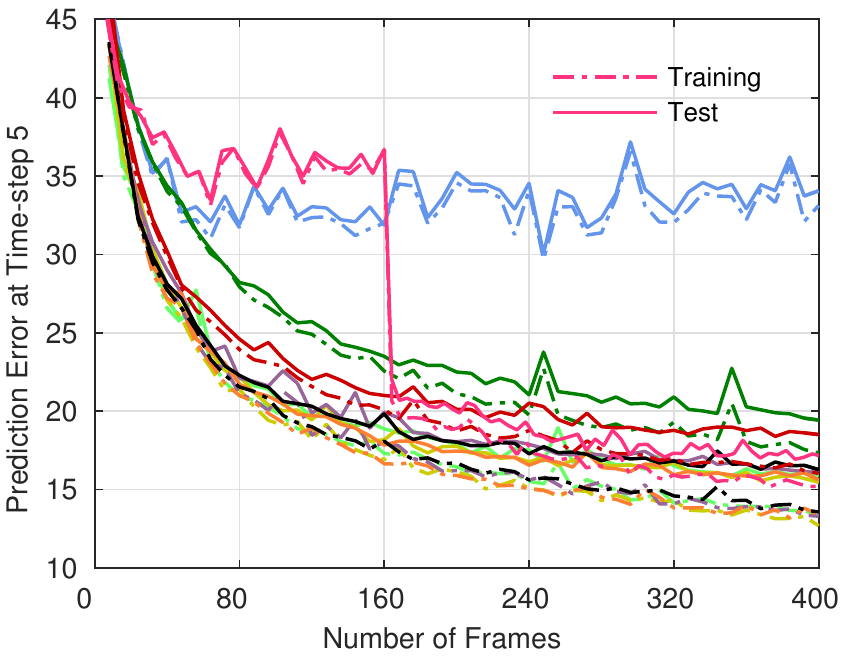}}
\subfigure[]{
\scalebox{0.79}{\includegraphics[]{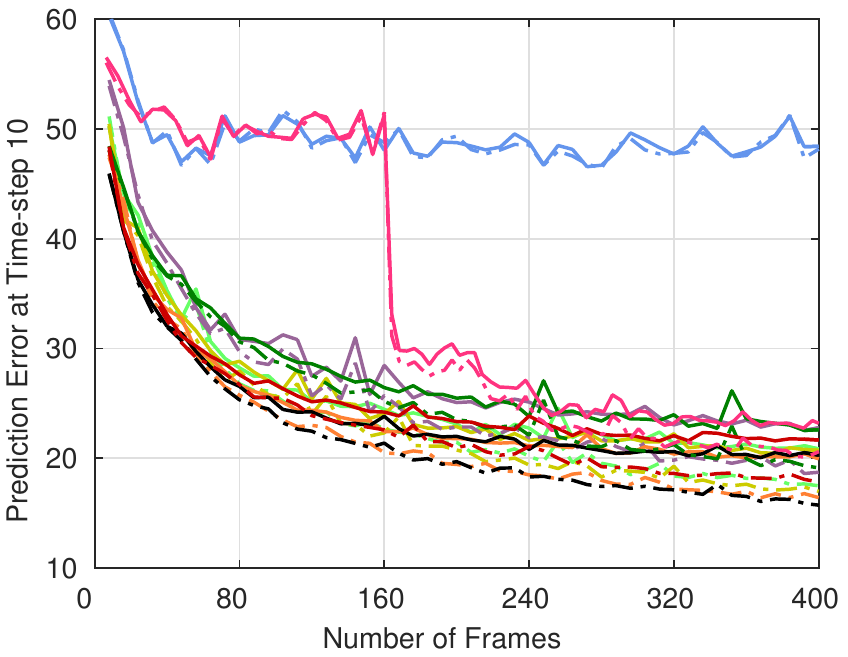}}
%\hskip0.1cm
\scalebox{0.79}{\includegraphics[]{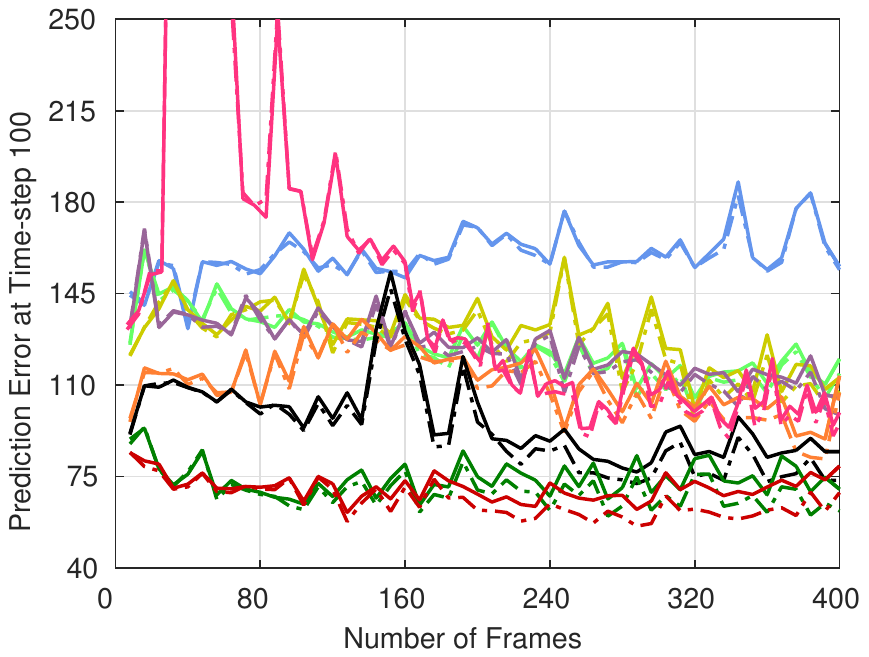}}}
\scalebox{0.79}{\includegraphics[]{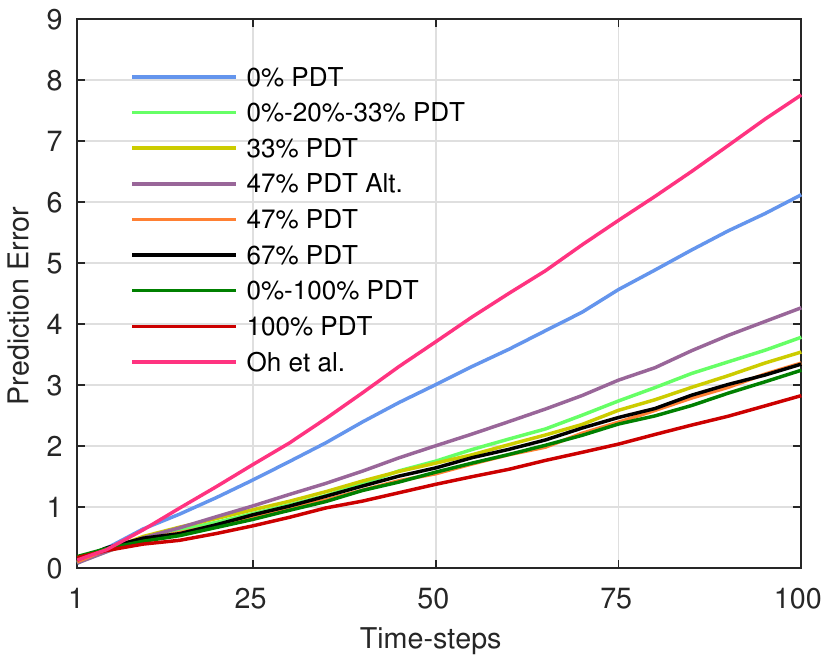}}
%\hskip0.1cm
\scalebox{0.79}{\includegraphics[]{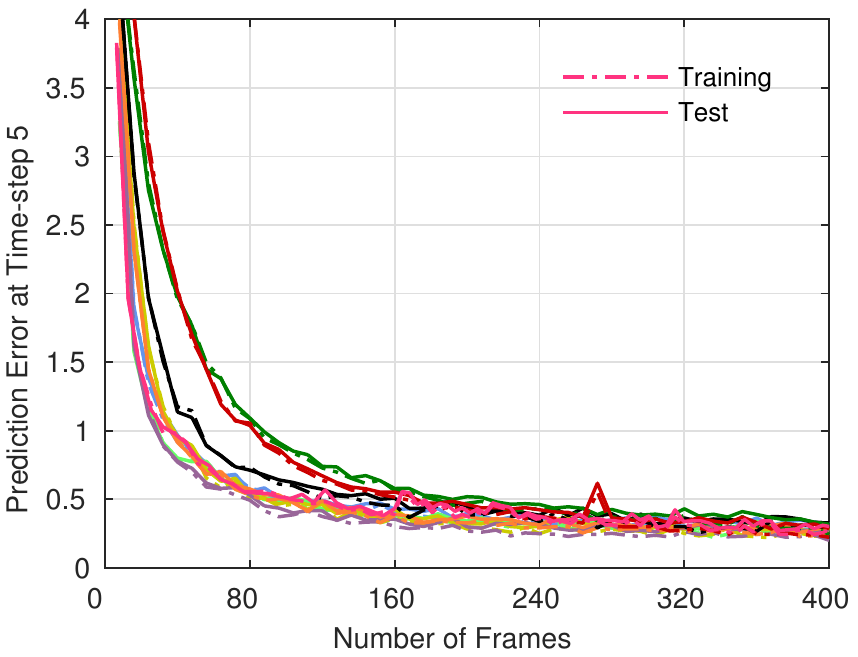}}
\subfigure[]{
\scalebox{0.79}{\includegraphics[]{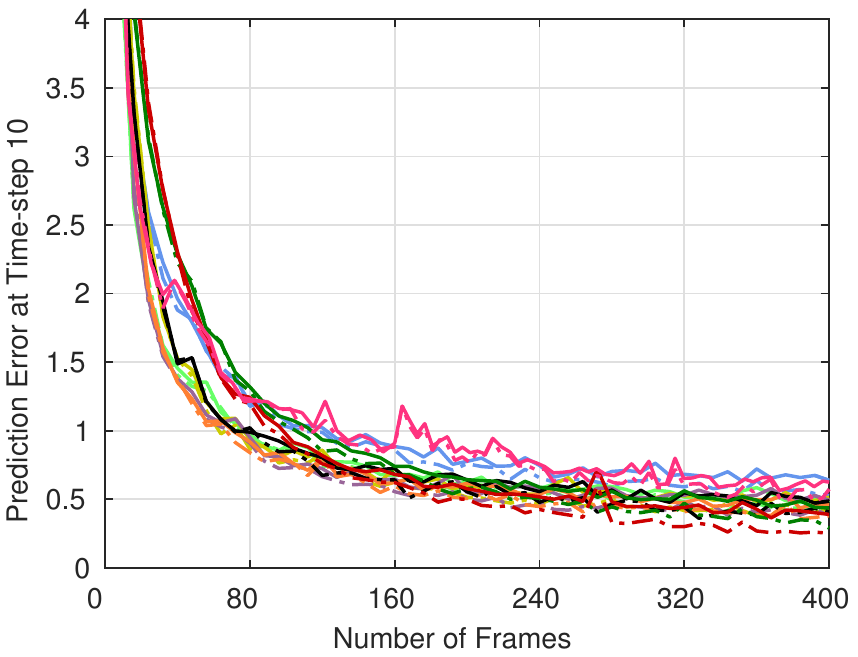}}
%\hskip0.1cm
\scalebox{0.79}{\includegraphics[]{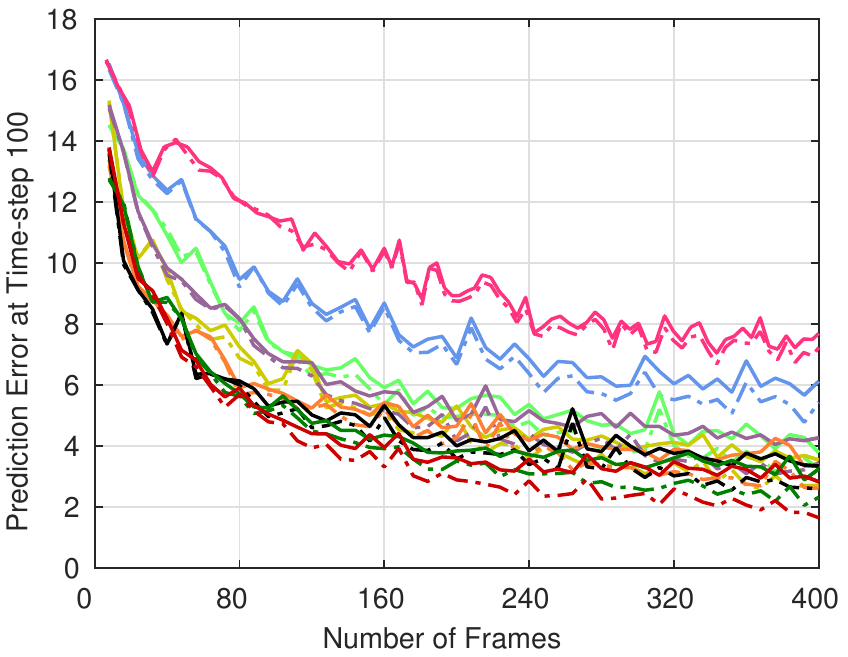}}}
\caption{Prediction error for different training schemes on (a) Ms Pacman and (b) Pong.}
\label{fig:predErrFBTMsPacman-Pong}
\end{figure}

\begin{figure}[] % Figures obtained with predErrFBTypeAppendix
\vskip-0.5cm
\scalebox{0.79}{\includegraphics[]{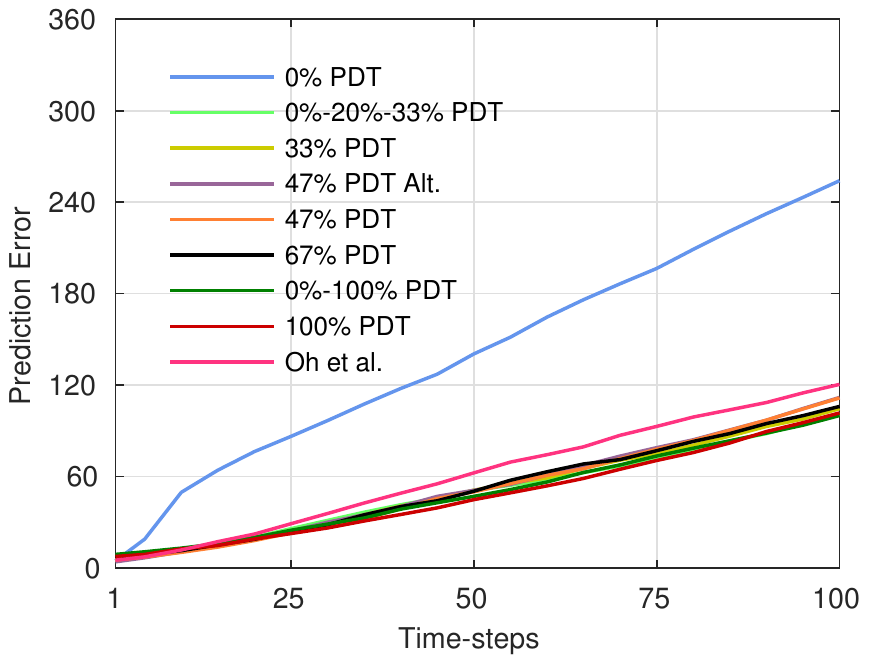}}
%\hskip0.1cm
\scalebox{0.79}{\includegraphics[]{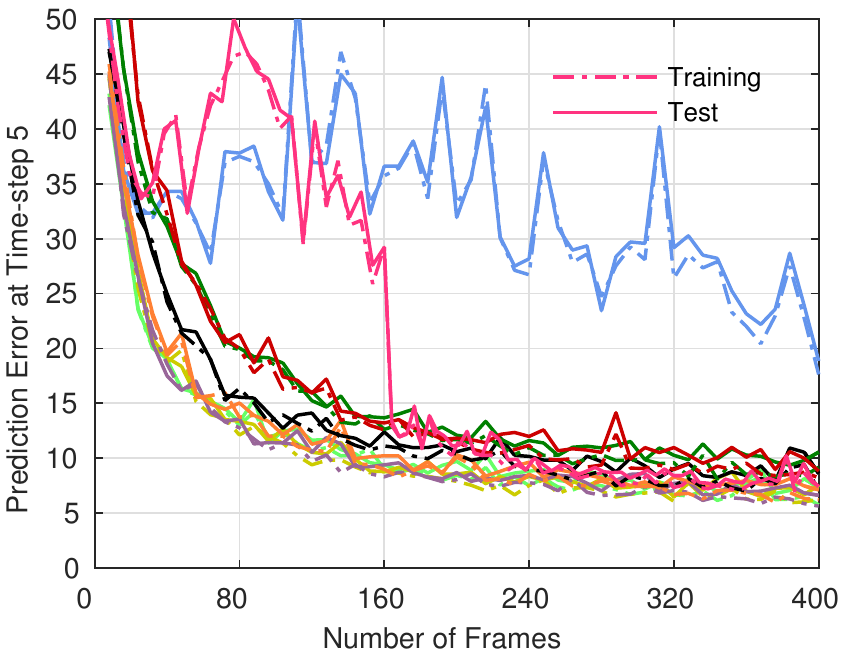}}
\subfigure[]{
\scalebox{0.79}{\includegraphics[]{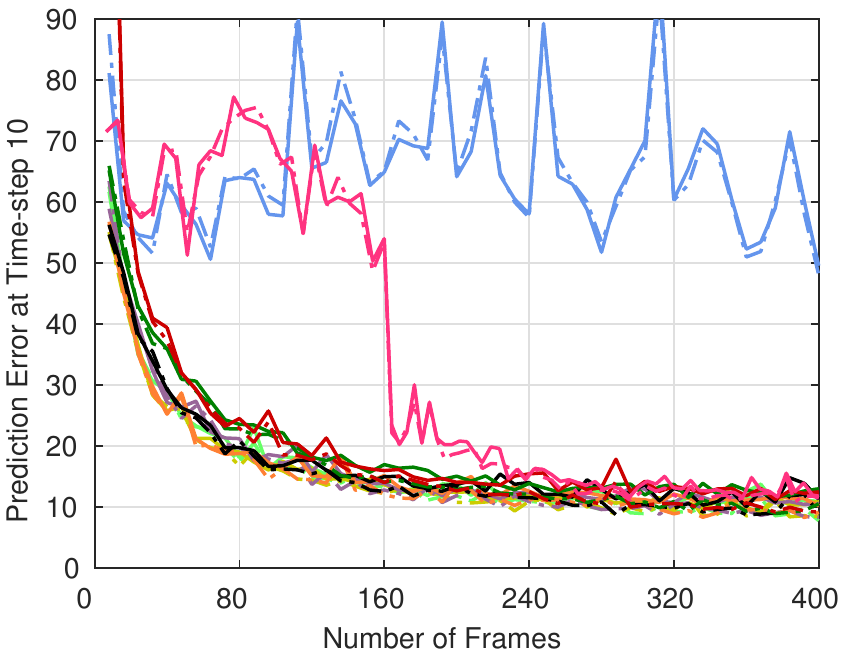}}
%\hskip0.1cm
\scalebox{0.79}{\includegraphics[]{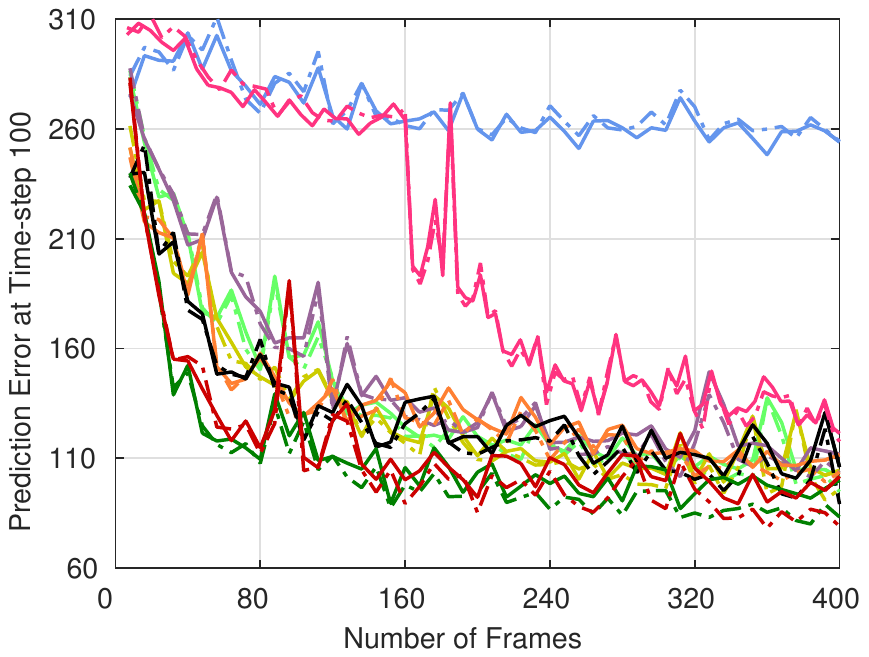}}}
\scalebox{0.79}{\includegraphics[]{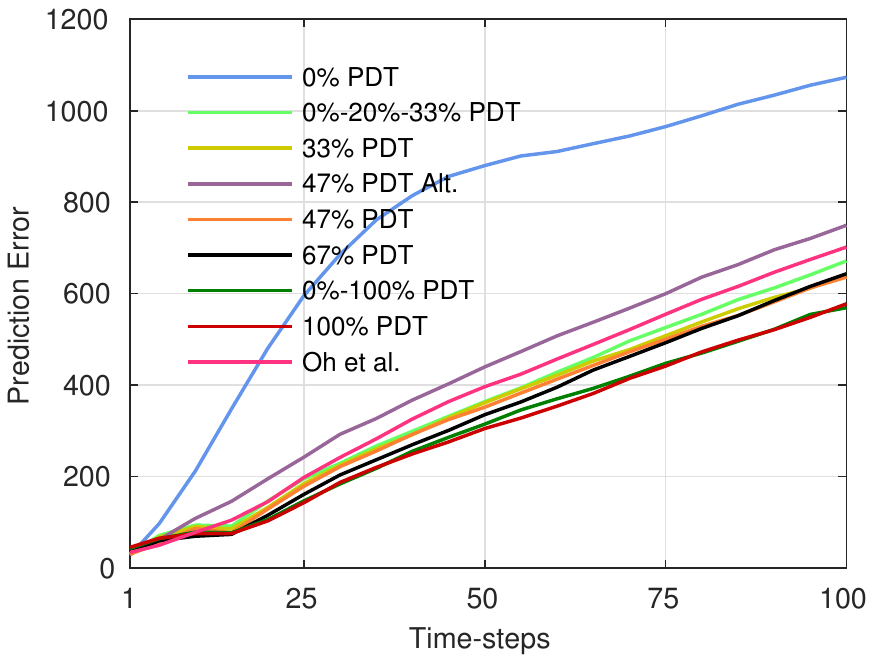}}
%\hskip0.1cm
\scalebox{0.79}{\includegraphics[]{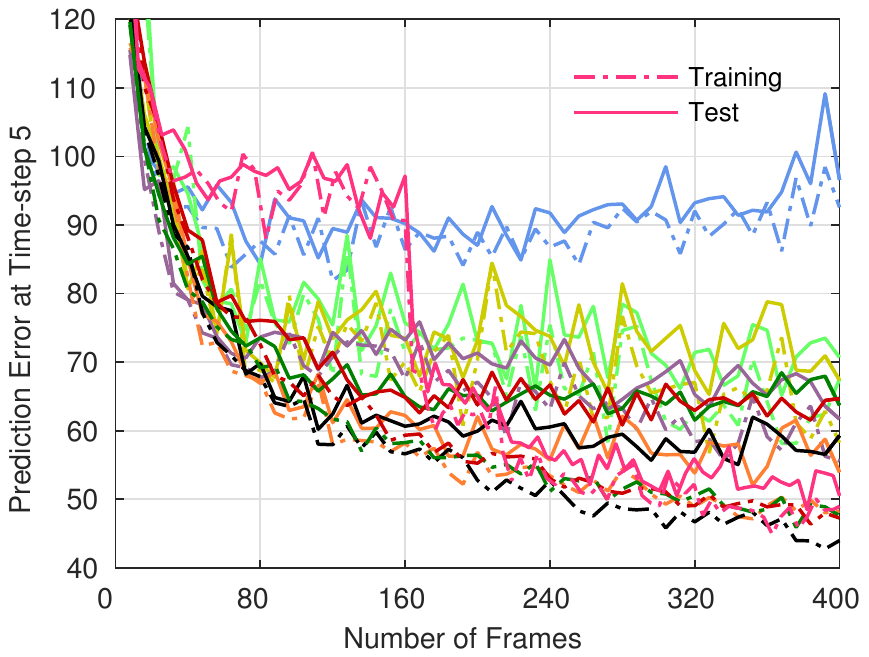}}
\subfigure[]{
\scalebox{0.79}{\includegraphics[]{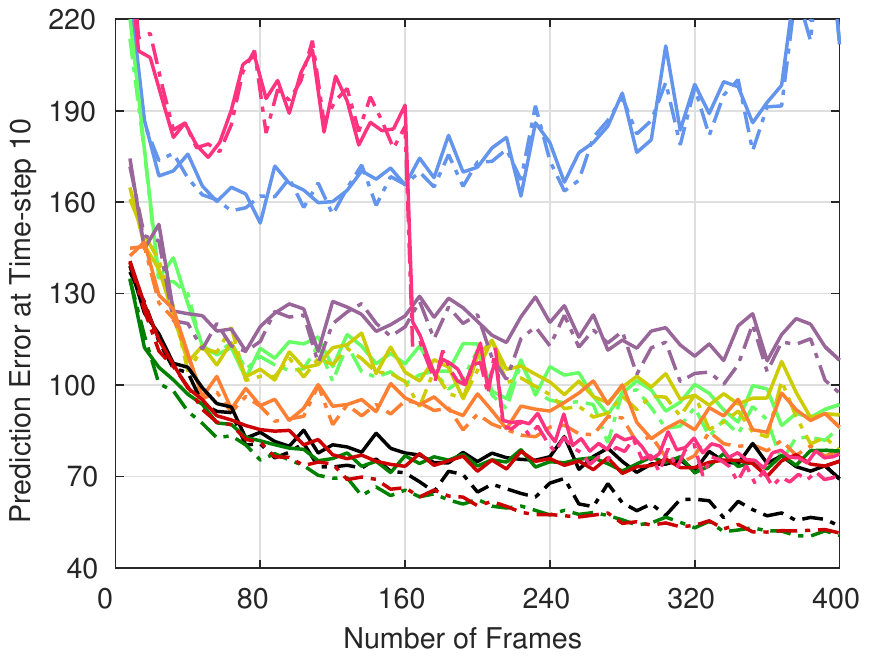}}
%\hskip0.1cm
\scalebox{0.79}{\includegraphics[]{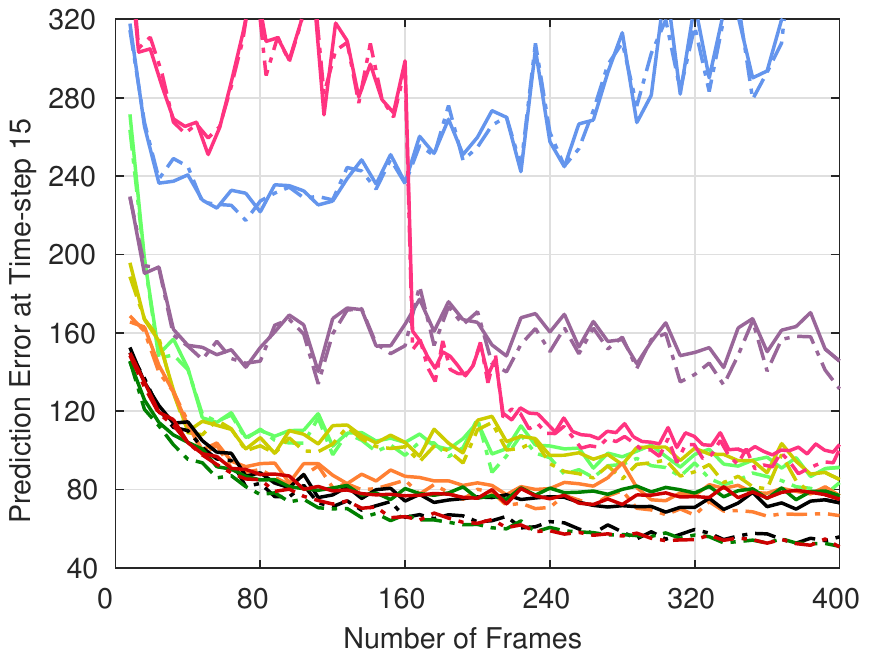}}}
\caption{Prediction error for different training schemes on (a) Qbert and (b) Riverraid.}
\label{fig:predErrFBTQbert-Riverraid}
\end{figure}

\begin{figure}[] % Figures obtained with predErrFBTypeAppendix
\vskip-0.5cm
\scalebox{0.79}{\includegraphics[]{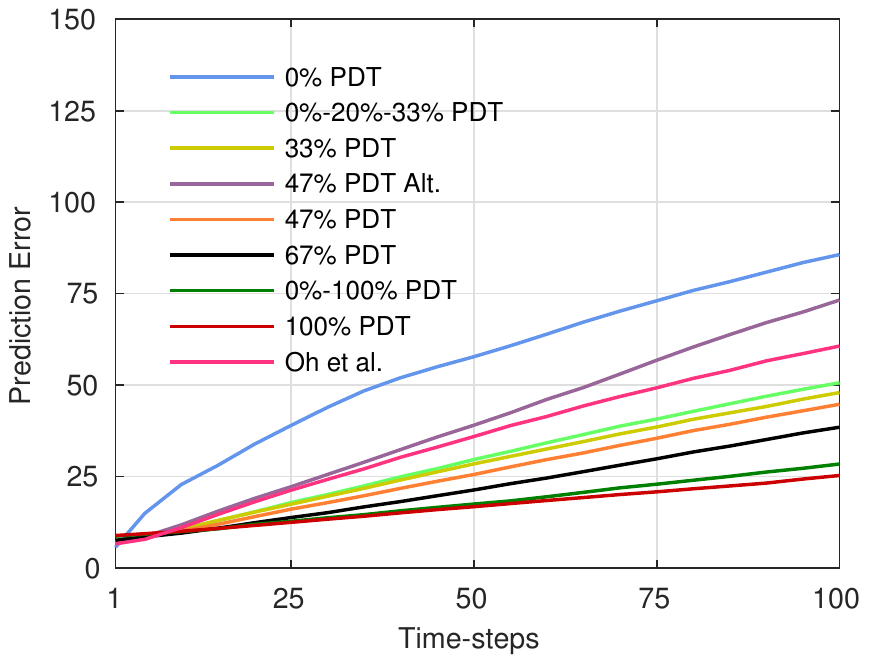}}
%\hskip0.1cm
\scalebox{0.79}{\includegraphics[]{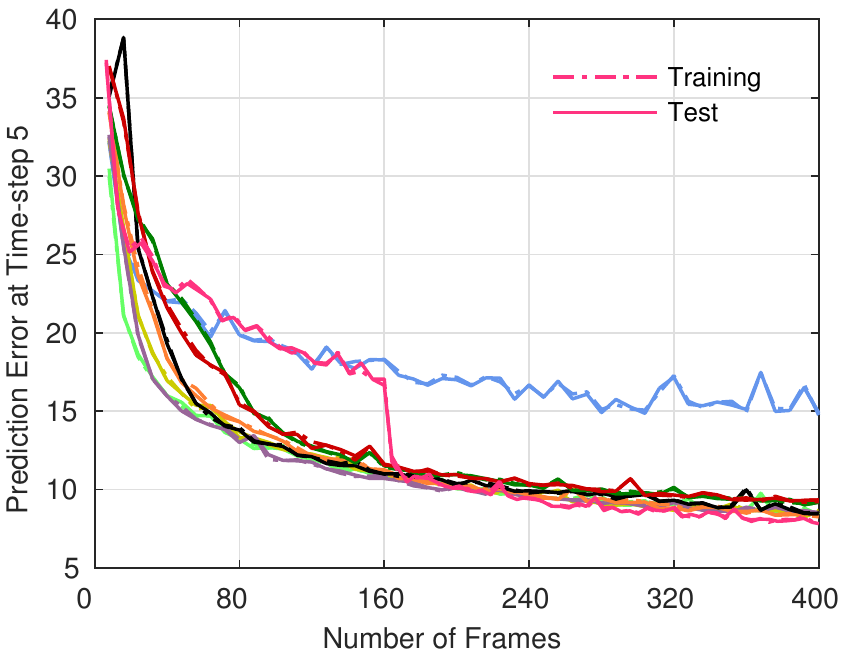}}
\subfigure[]{
\scalebox{0.79}{\includegraphics[]{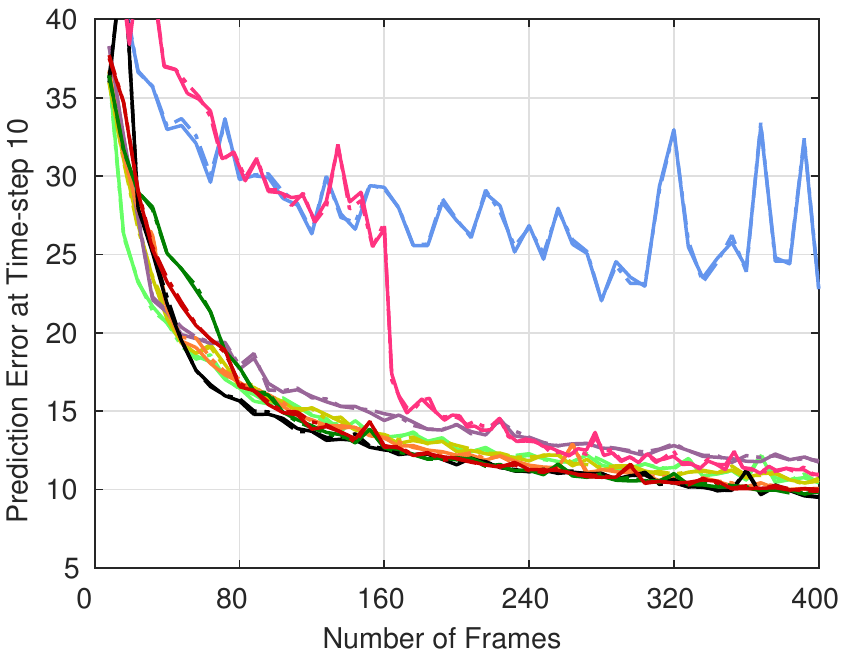}}
%\hskip0.1cm
\scalebox{0.79}{\includegraphics[]{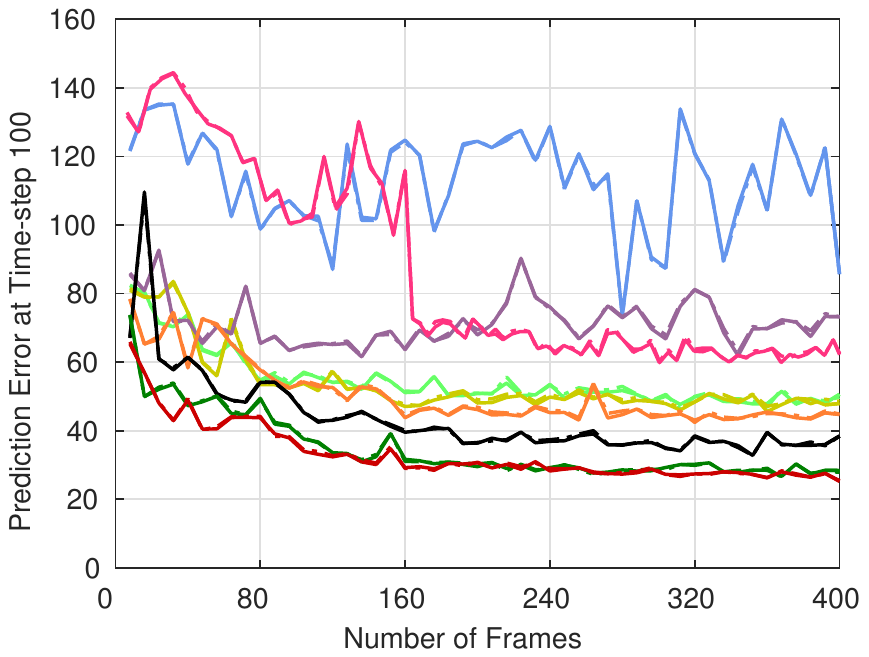}}}
\scalebox{0.79}{\includegraphics[]{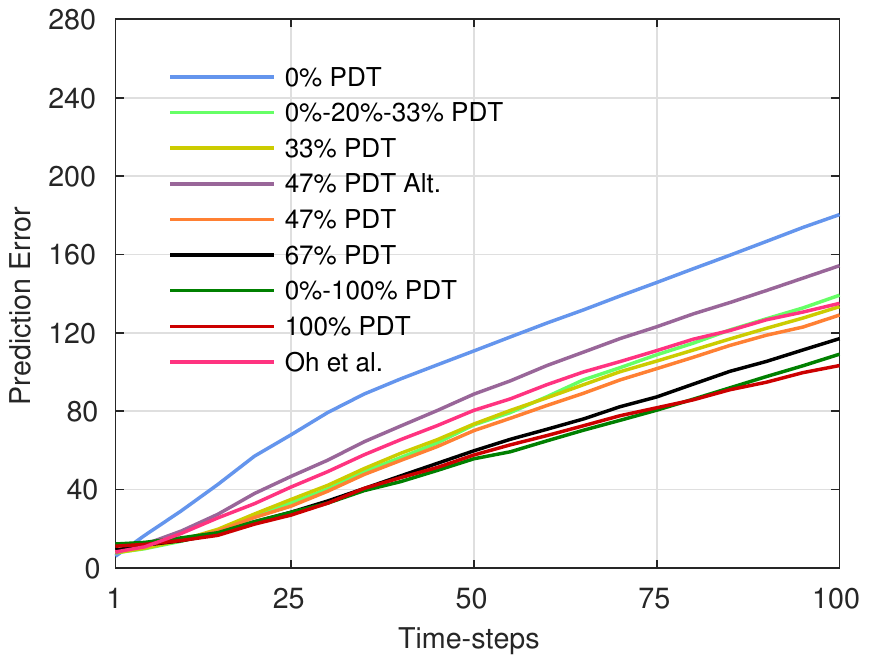}}
%\hskip0.1cm
\scalebox{0.79}{\includegraphics[]{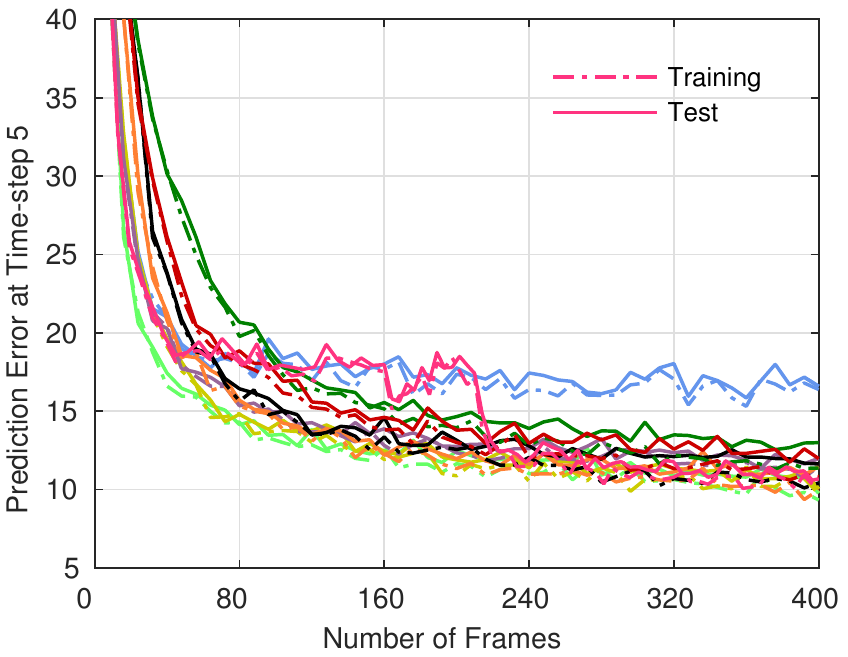}}
\subfigure[]{
\scalebox{0.79}{\includegraphics[]{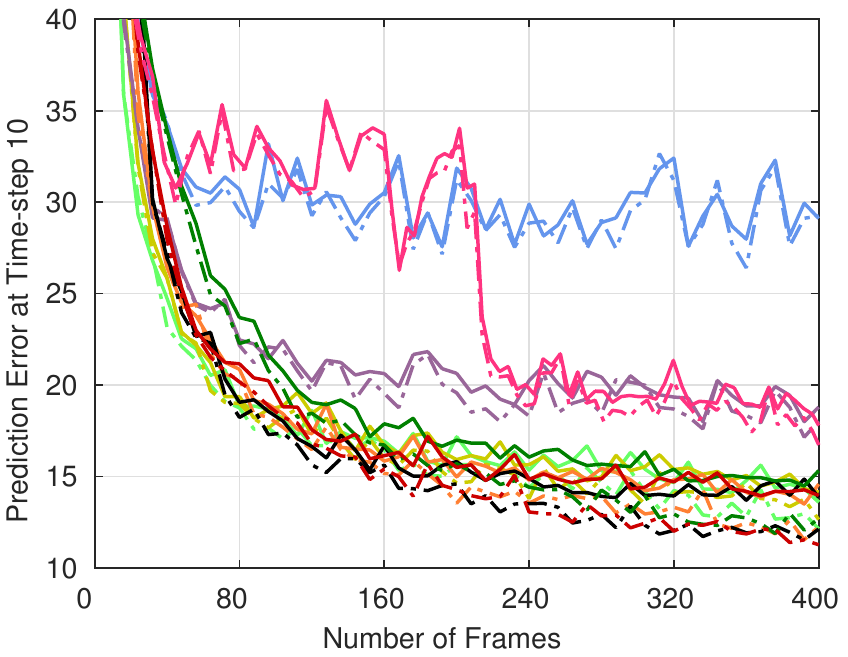}}
%\hskip0.1cm
\scalebox{0.79}{\includegraphics[]{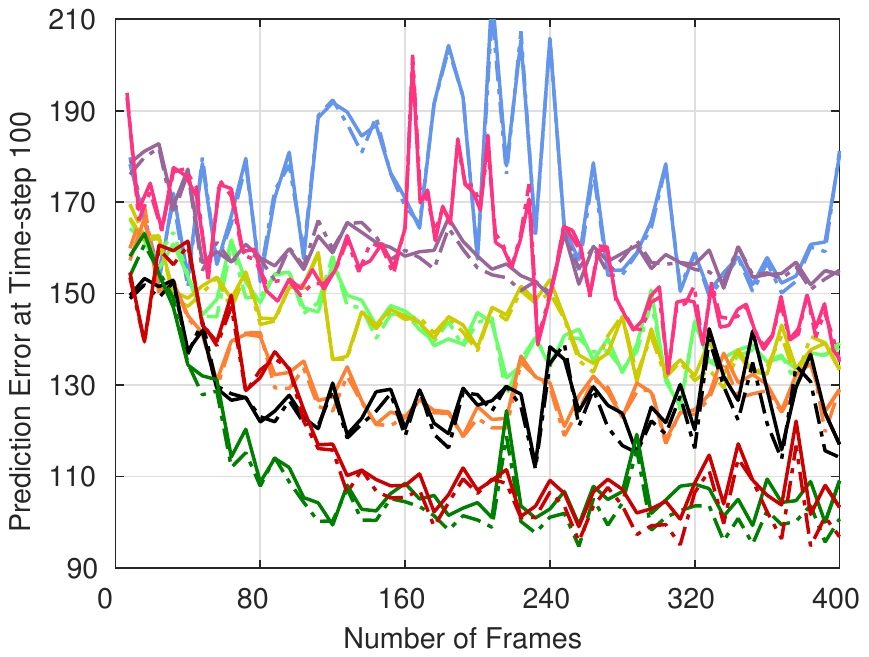}}}
\caption{Prediction error for different training schemes on (a) Seaquest and (b) Space Invaders.}
\label{fig:predErrFBTSeaquest-SpaceInvaders}
\end{figure}

%\vspace{-2mm}
In Figures \ref{fig:predErrFBTBowling-Breakout}-\ref{fig:predErrFBTSeaquest-SpaceInvaders}, 
we show the prediction error obtained with the training schemes described in \secref{sec:PDT} for all games. 
Below we discuss the main findings for each game.

\paragraph{Bowling.}
Bowling is one of the easiest games to model. A simulator trained using only \OD~transitions gives quite accurate predictions. 
However, using only \PD~transitions reduces the error in updating the score and predicting the ball direction.

\paragraph{Breakout.}
Breakout is a difficult game to model. A simulator trained with only \PD~transitions predicts the paddle movement very accurately but almost always fails to represent the ball. 
A simulator trained with only \OD~transitions struggles much less to represent the ball but 
does not predict the paddle and ball positions as accurately, and the ball also often disappears after hitting the paddle. 
Interestingly, the long-term prediction error (bottom-right of \figref{fig:predErrFBTBowling-Breakout}(b)) for the 100\%\PDT~training scheme is 
the lowest, as when not representing the ball the predicted frames look closer to the real frames
than when representing the ball incorrectly. 
A big improvement in the model ability to represent the ball could be obtained by pre-processing the frames with max-pooling as done for DQN, as this increases the ball size.  
We believe that a more sophisticated convolutional structure would be even more effective, but did not succeed in discovering such a structure.


\paragraph{Fishing Derby.}
In Fishing Derby, long-term accuracy is disastrous with the 0\%\PDT~training scheme and good with the 100\%\PDT~training scheme. 
Short-term accuracy is better with schemes using more \OD~transitions than in the 100\% or 0\%-100\%\PDT~training schemes, 
especially at low numbers of parameter updates. 

\paragraph{Freeway.}
With Bowling, Freeway is one of the easiest games to model, but more parameter updates are required for convergence than for Bowling. 
The 0\%\PDT~training scheme gives good accuracy, although sometimes the chicken disappears or its position is incorrectly predicted --
this happens extremely rarely with the 100\%\PDT~training scheme. In both schemes, the score is often wrongly updated in the warning phase.

\paragraph{Ms Pacman.}
Ms Pacman is a very difficult game to model and accurate prediction can only be obtained for a few time-steps into the future. 
The movement of the ghosts, especially when in frightened mode, is regulated by the position of Ms Pacman according to complex rules.
Furthermore, the DQN $\epsilon=0.2$-greedy policy does not enable the agent to explore certain regions of the state space.
As a result, the simulator can predict well the movement of Ms Pacman, but fails to predict long-term the movement of the ghosts when in frightened 
mode or when in chase mode later in the episodes.

\paragraph{Pong.}
With the 0\%\PDT~training scheme, the model often incorrectly predicts the direction of the ball when hit by the agent or by the opponent. Quite rarely, the ball disappears when hit by the agent.
With the 100\%\PDT~training scheme, the direction the ball is much more accurately predicted, but the ball more often disappears when hit by the agent, and the ball and paddles are generally less sharp.

\paragraph{Qbert.}
Qbert is a game for which the 0\%\PDT~training scheme is unable to predict accurately beyond very short-term, as after a few frames only the background is predicted. 
The more \PD~transitions are used, the less sharply the agent and the moving objects are represented. 

\paragraph{Riverraid.}
In Riverraid, prediction with the 0\%\PDT~training scheme is very poor, as this scheme causes no generation of new objects or background. 
With all schemes, the model fails to predict the frames that follow a jet loss -- that's why the prediction error increases sharply after around time-step 13 in \figref{fig:predErrFBTQbert-Riverraid}(b).
The long-term prediction error is lower with the 100\%\PDT~training scheme, as with this scheme the simulator is more accurate before, and sometimes after, a jet loss. 
The problem of incorrect prediction after a jet loss disappears when using BBTT(15,2) with \PD~transitions.

\paragraph{Seaquest.}
In Seaquest, with the 0\%\PDT~training scheme, the existing fish disappears after a few time-steps and no new fish ever appears from the sides of the frame. 
The higher the number of \PD~transitions the less sharply the fish is represented, but the more accurately its dynamics and appearance from the sides of the frame can be predicted.

\paragraph{Space Invaders}
Space Invaders is a very difficult game to model and accurate prediction can only be obtained for a few time-steps into the future. 
The 0\%\PDT~training scheme is unable to predict accurately beyond very short-term. The 100\%\PDT~training scheme struggles to represent the bullets.

\begin{figure}[htbp] % Figures obtained with predErrSeqLengthAppendix
\vskip-0.5cm
\scalebox{0.79}{\includegraphics[]{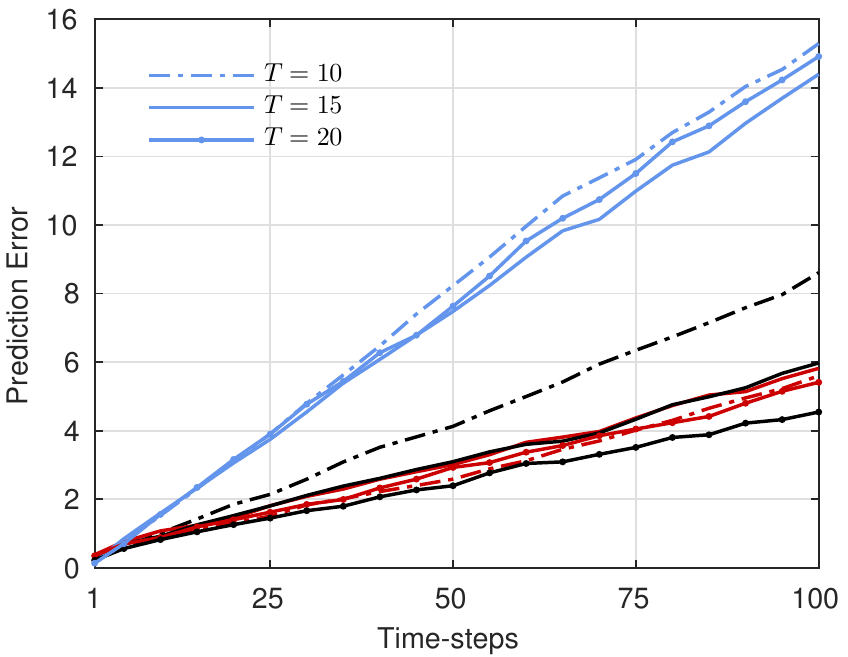}}
%\hskip0.1cm
\scalebox{0.79}{\includegraphics[]{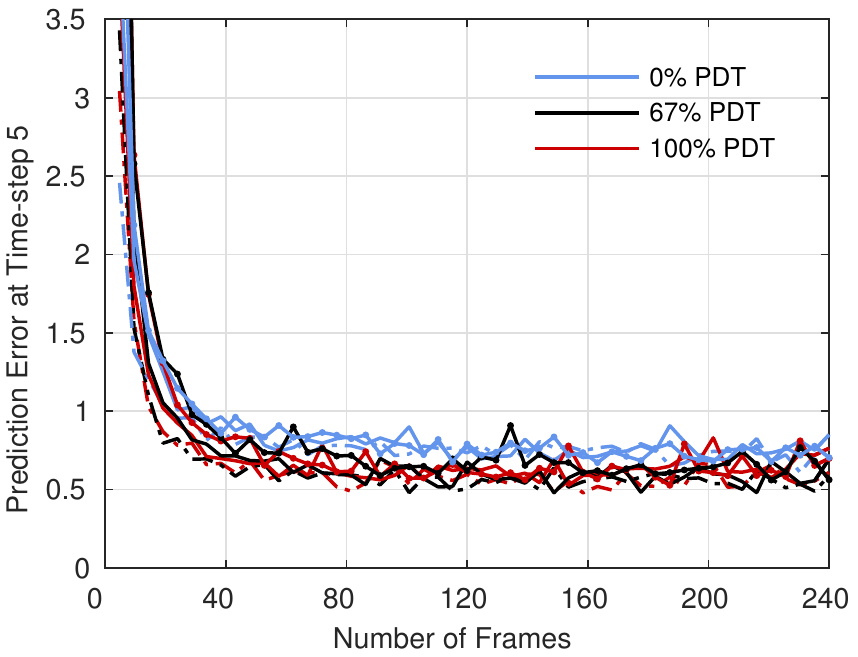}}\\
\subfigure[]{\scalebox{0.79}{\includegraphics[]{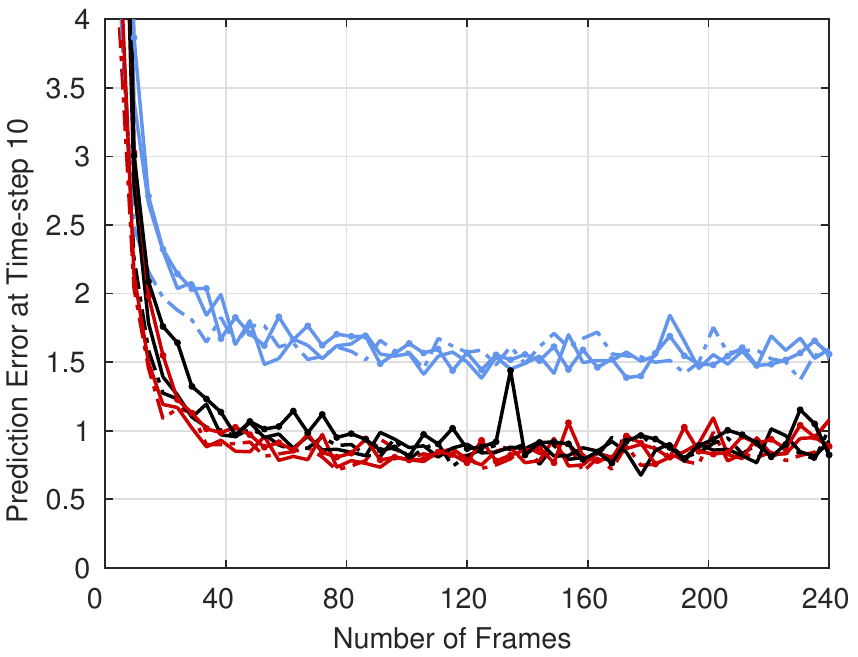}}
%\hskip0.1cm
\scalebox{0.79}{\includegraphics[]{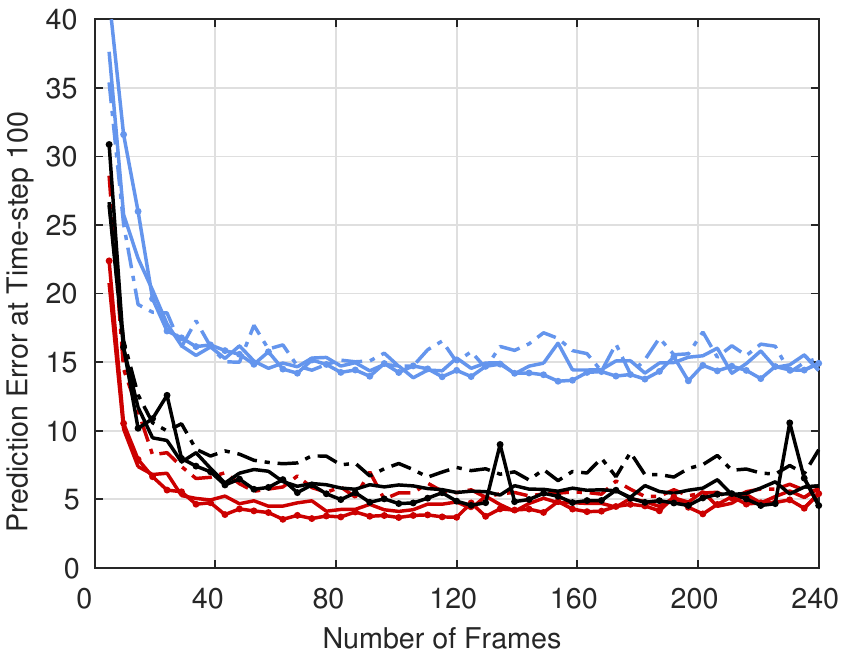}}}
\scalebox{0.79}{\includegraphics[]{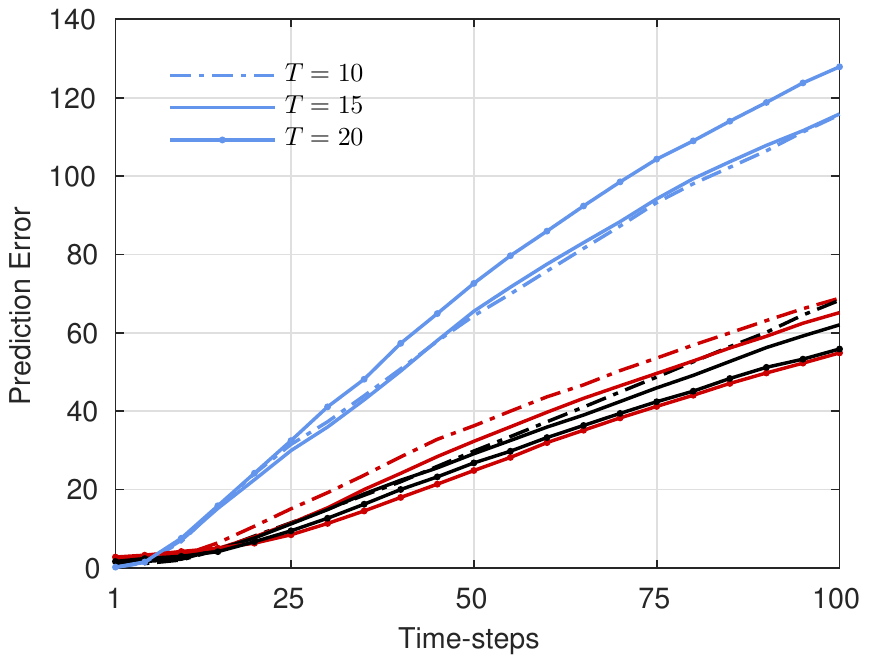}}
%\hskip0.1cm
\scalebox{0.79}{\includegraphics[]{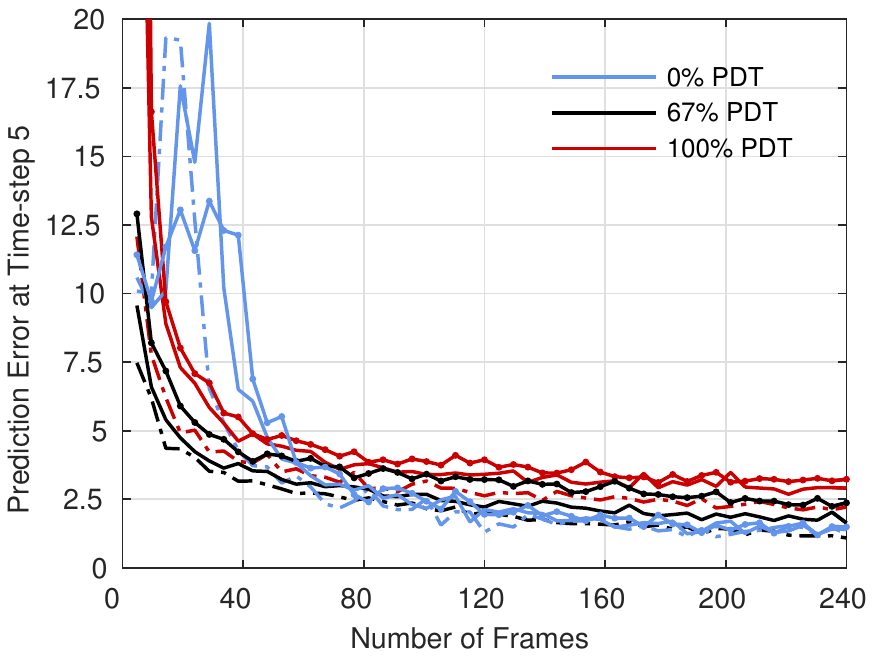}}
\subfigure[]{
\scalebox{0.79}{\includegraphics[]{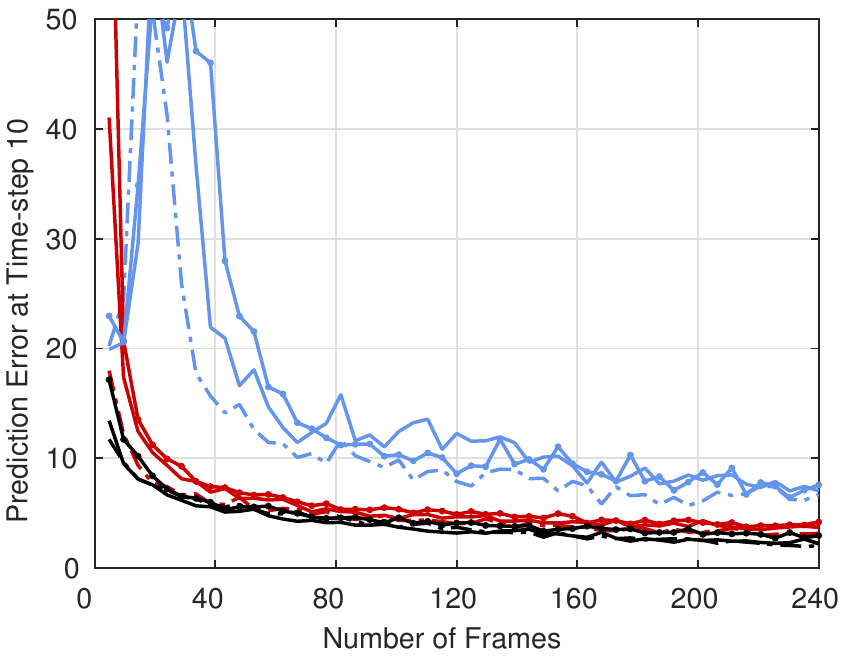}}
%\hskip0.1cm
\scalebox{0.79}{\includegraphics[]{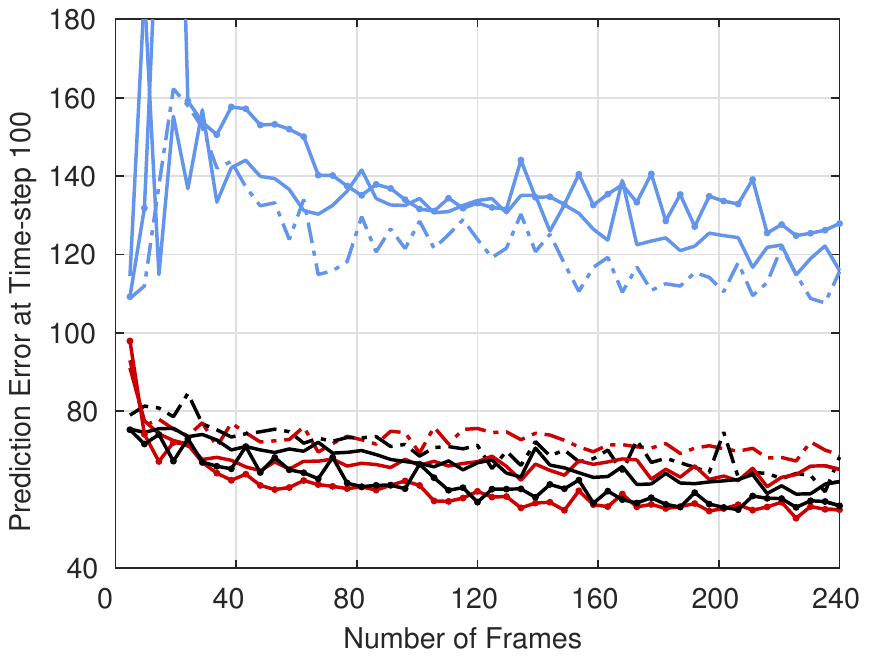}}}
\vskip-0.3cm
\caption{Prediction error (average over 10,000 sequences) for different prediction lengths $T\leq 20$ on (a) Bowling and (b) Breakout. Number of frames is in millions and excludes warm-up frames.}
\label{fig:predErrSeqLengthBowling-Breakout}
\end{figure}

\begin{figure}[htbp] % Figures obtained with predErrSeqLengthAppendix
\vskip-0.5cm
\scalebox{0.79}{\includegraphics[]{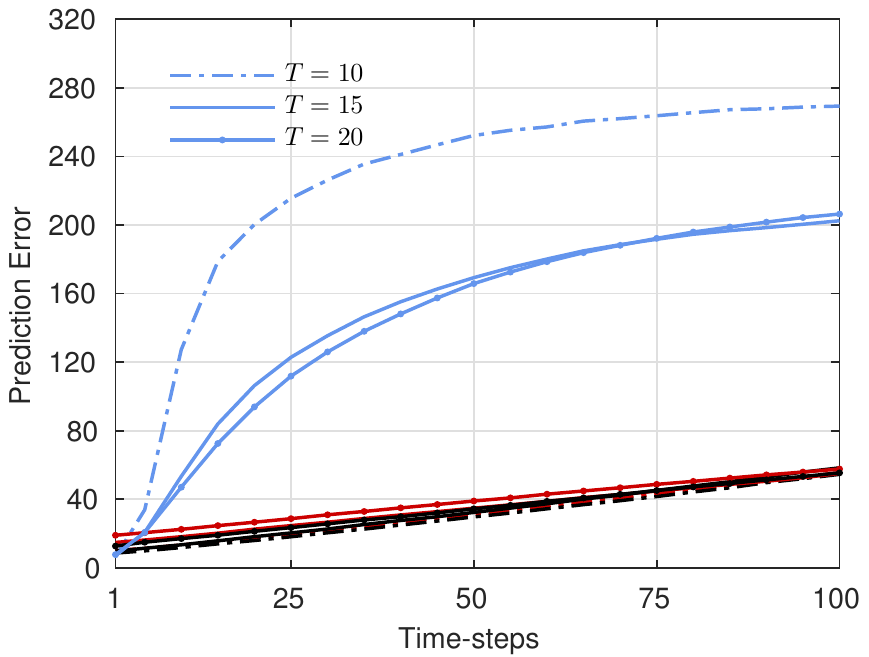}}
%\hskip0.1cm
\scalebox{0.79}{\includegraphics[]{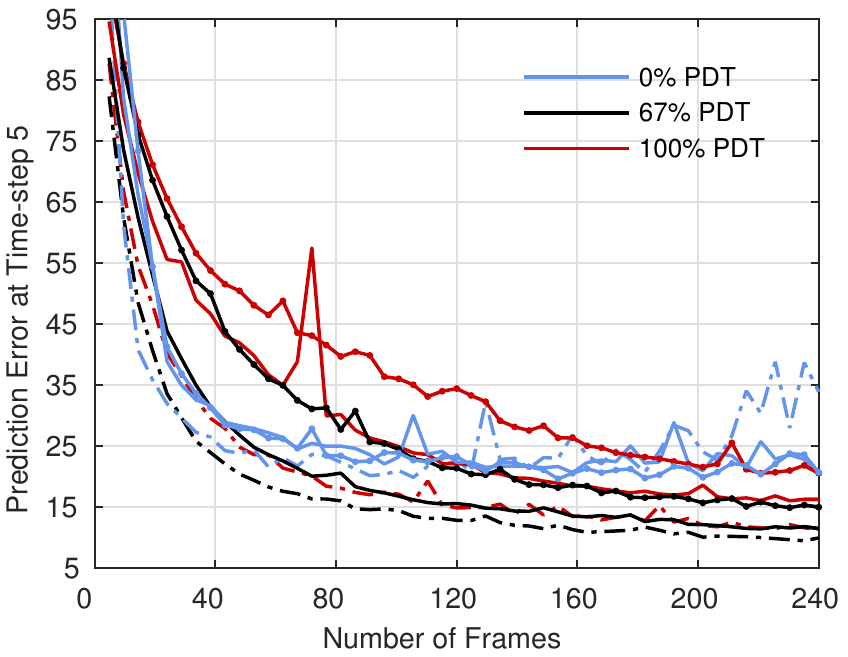}}
\subfigure[]{
\scalebox{0.79}{\includegraphics[]{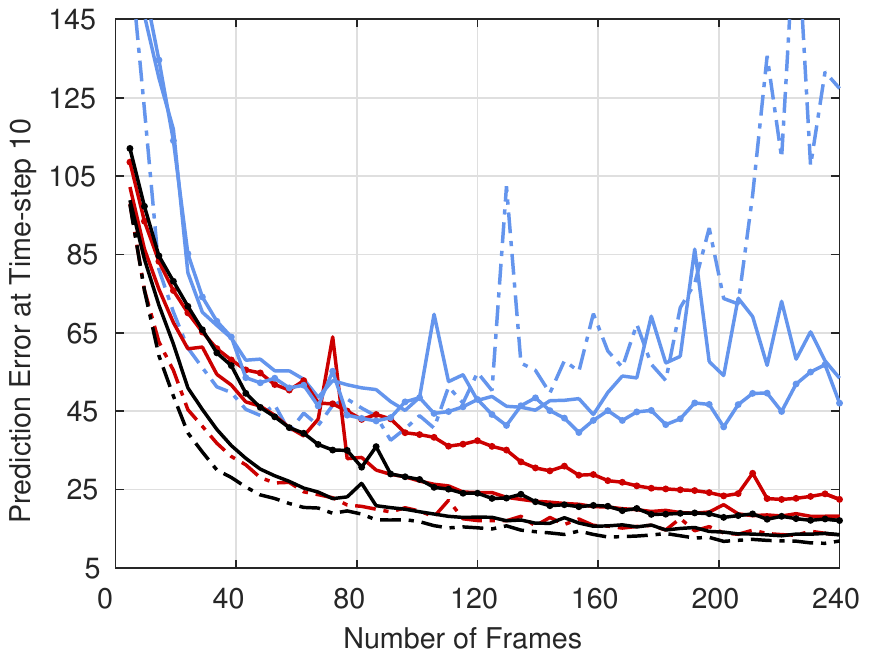}}
%\hskip0.1cm
\scalebox{0.79}{\includegraphics[]{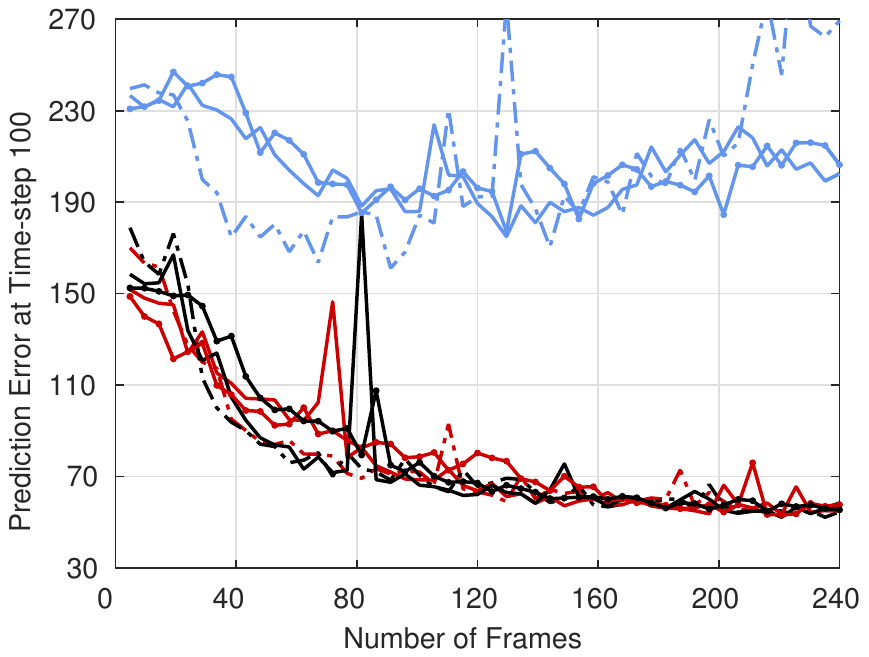}}}
\scalebox{0.79}{\includegraphics[]{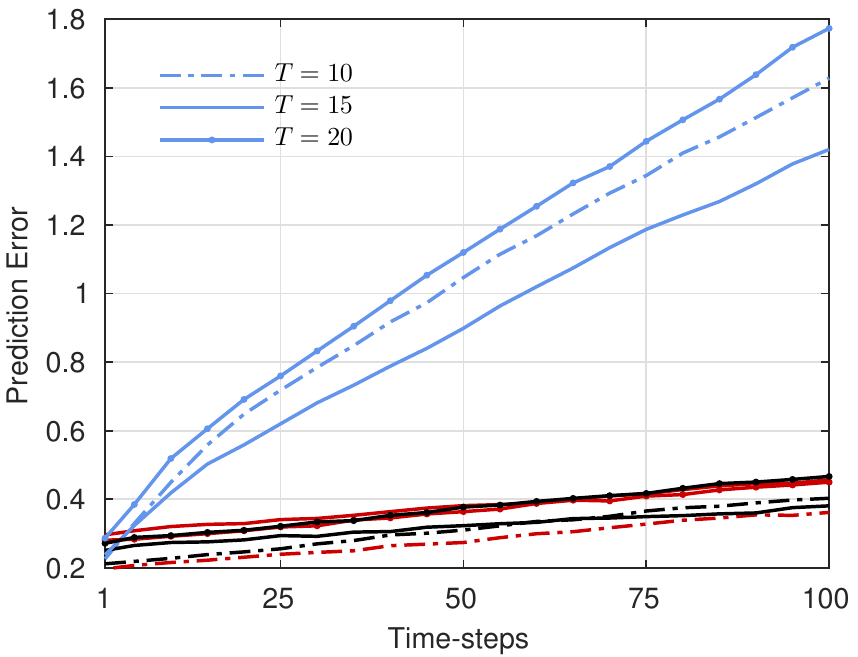}}
%\hskip0.1cm
\scalebox{0.79}{\includegraphics[]{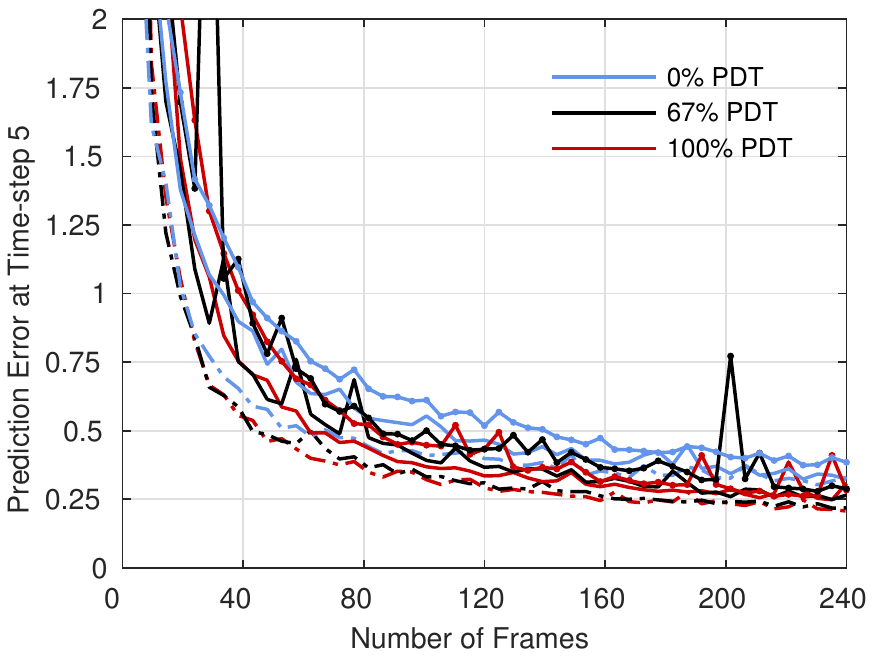}}
\subfigure[]{
\scalebox{0.79}{\includegraphics[]{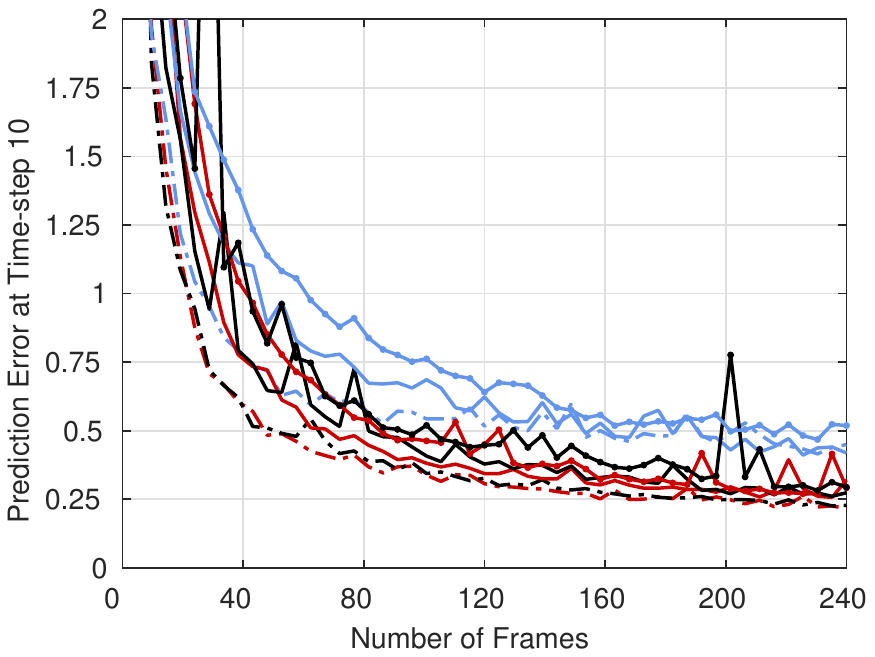}}
%\hskip0.1cm
\scalebox{0.79}{\includegraphics[]{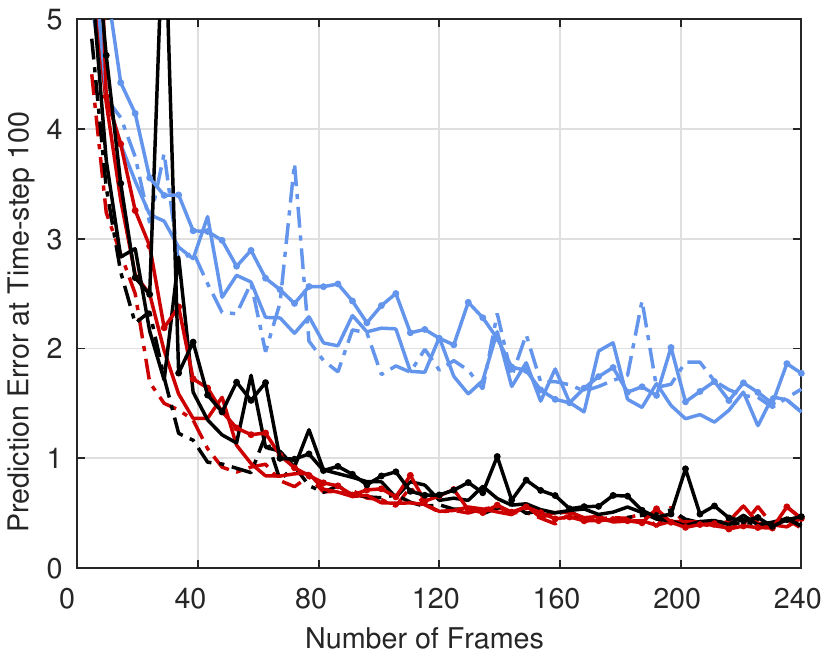}}}
\caption{Prediction error for different prediction lengths $T\leq 20$ on (a) Fishing Derby and (b) Freeway.}
\label{fig:predErrSeqLengthFishingDerby-Freeway}
\end{figure}

\begin{figure}[htbp] % Figures obtained with predErrSeqLengthAppendix
\vskip-0.5cm
\scalebox{0.79}{\includegraphics[]{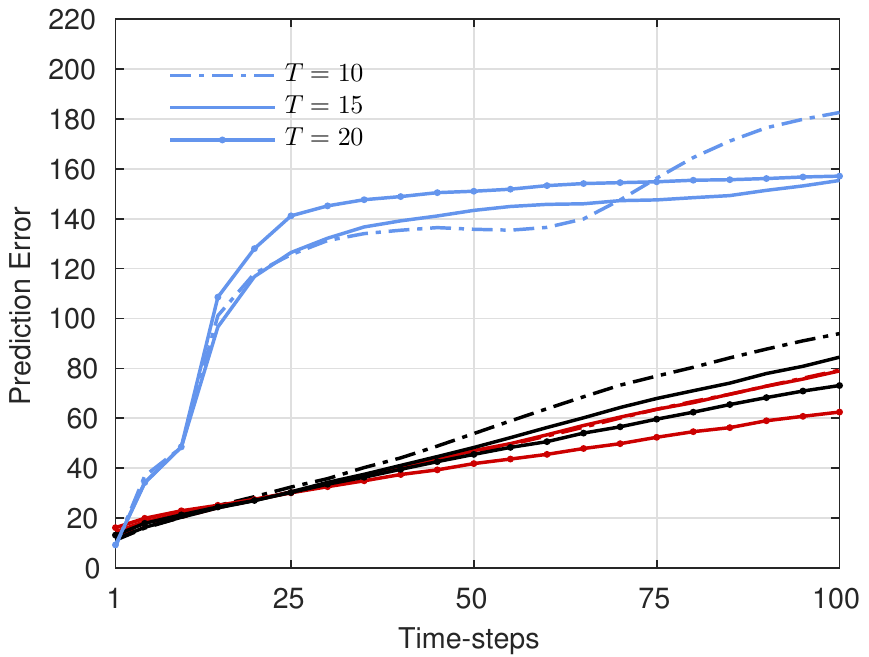}}
%\hskip0.1cm
\scalebox{0.79}{\includegraphics[]{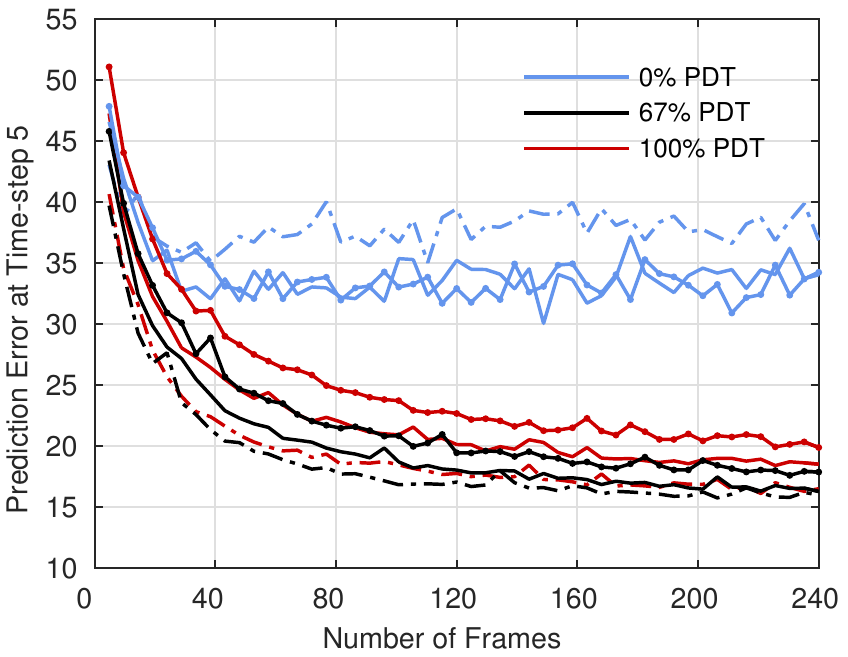}}
\subfigure[]{
\scalebox{0.79}{\includegraphics[]{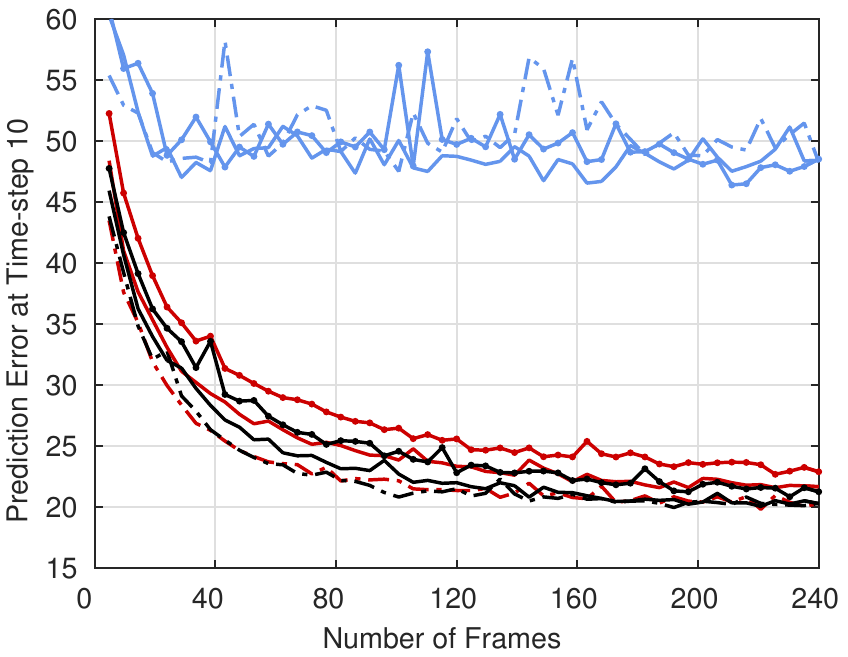}}
%\hskip0.1cm
\scalebox{0.79}{\includegraphics[]{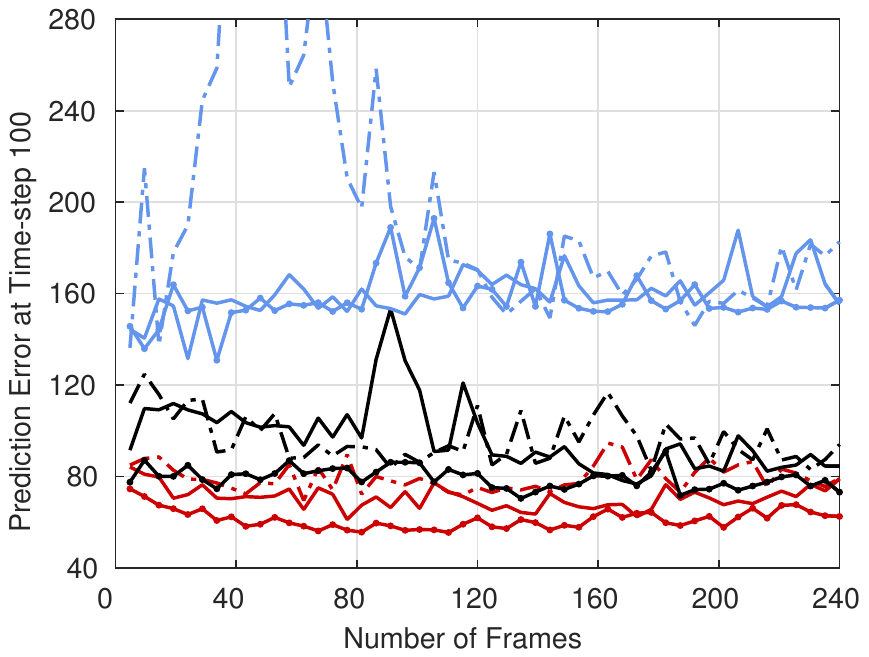}}}
\scalebox{0.79}{\includegraphics[]{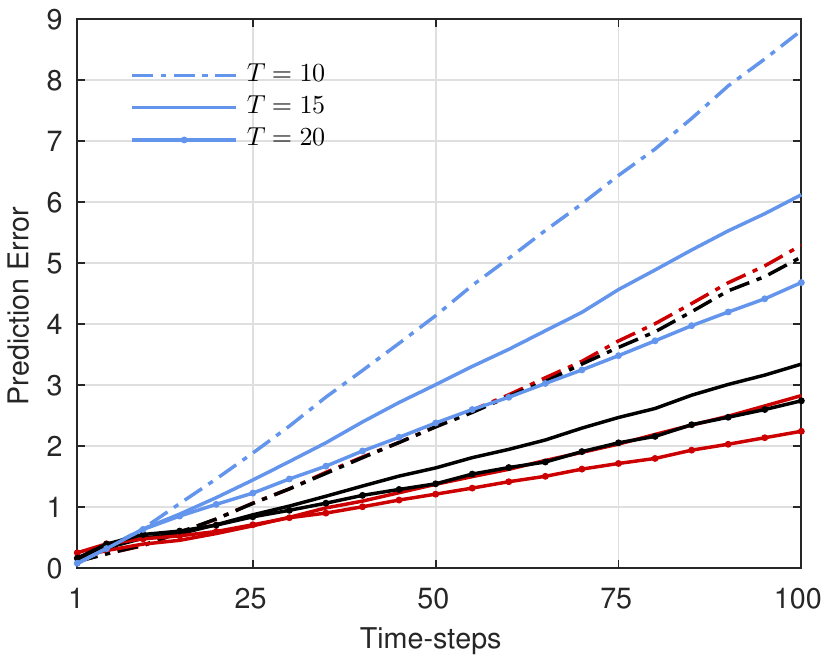}}
%\hskip0.1cm
\scalebox{0.79}{\includegraphics[]{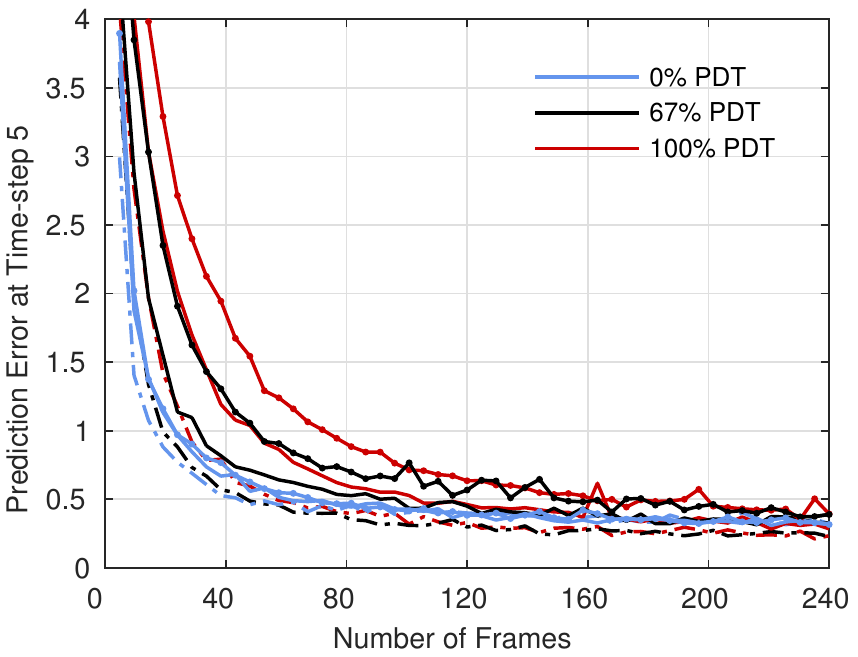}}
\subfigure[]{
\scalebox{0.79}{\includegraphics[]{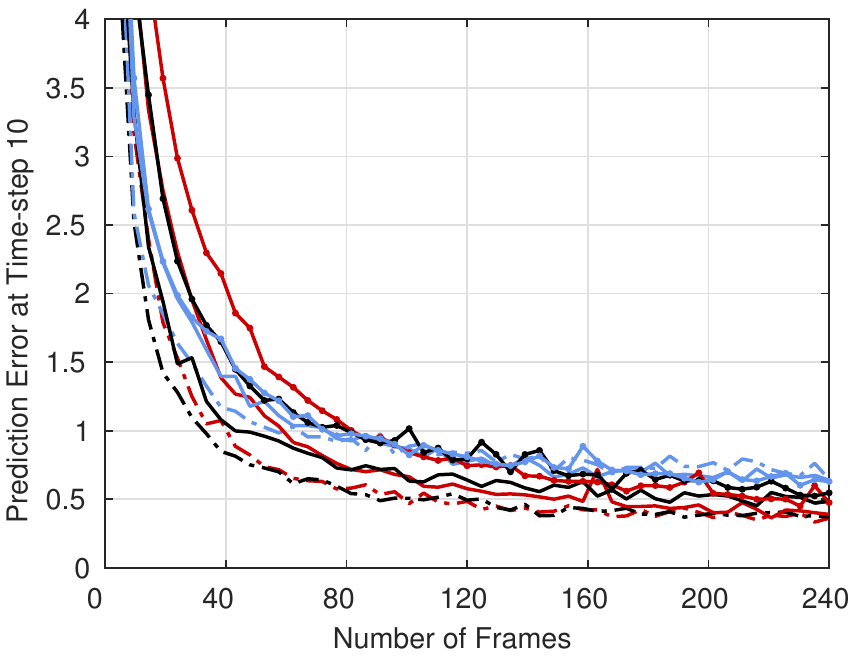}}
%\hskip0.1cm
\scalebox{0.79}{\includegraphics[]{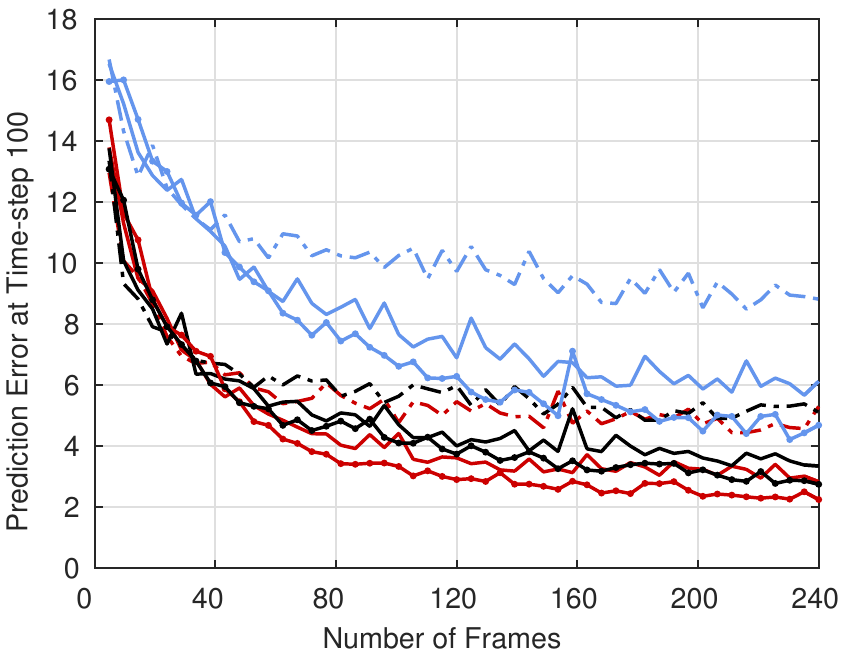}}}
\caption{Prediction error for different prediction lengths $T\leq 20$ on (a) Ms Pacman and (b) Pong.}
\label{fig:predErrSeqLengthMsPacman-Pong}
\end{figure}

\begin{figure}[htbp] % Figures obtained with predErrSeqLengthAppendix
\vskip-0.5cm
\scalebox{0.79}{\includegraphics[]{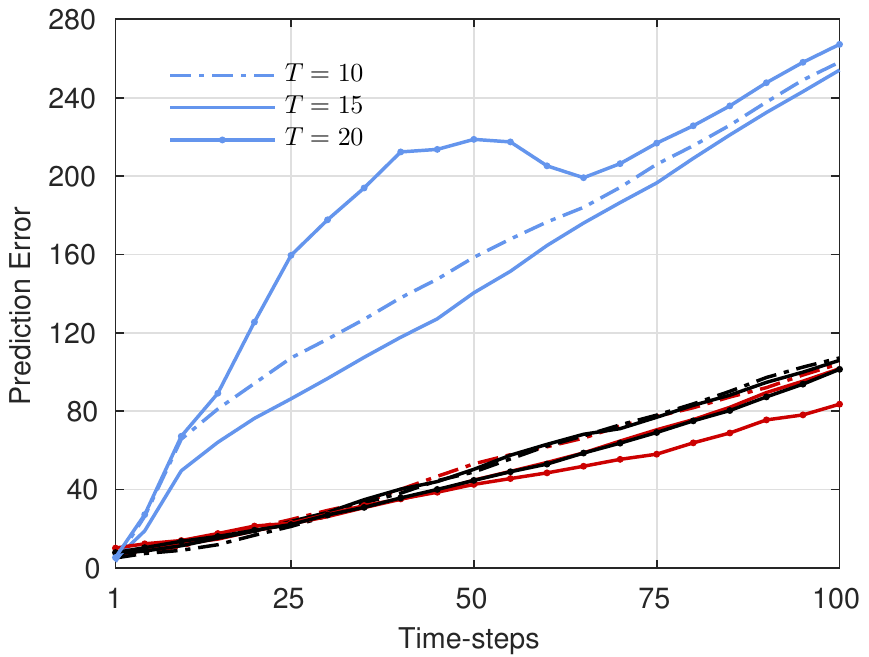}}
%\hskip0.1cm
\scalebox{0.79}{\includegraphics[]{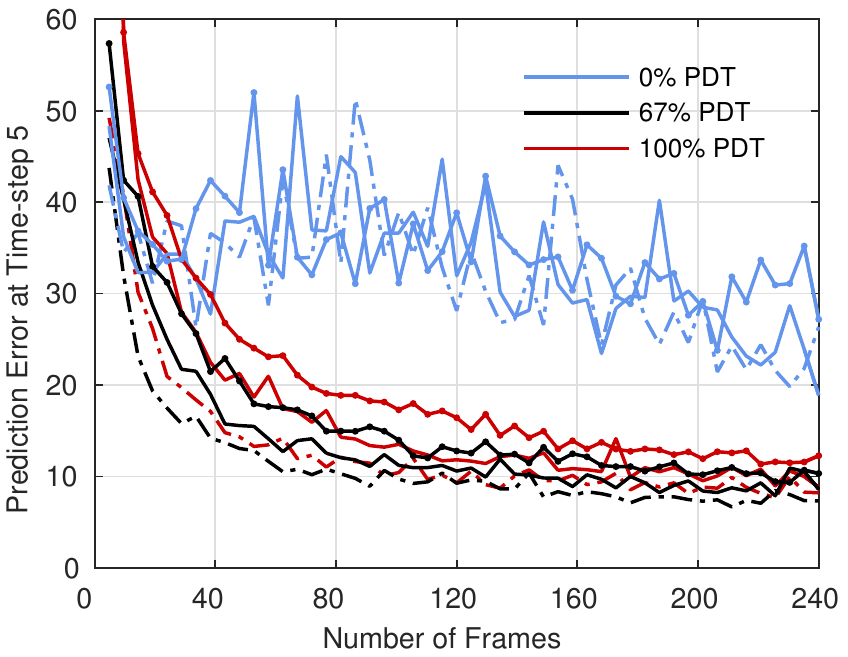}}
\subfigure[]{
\scalebox{0.79}{\includegraphics[]{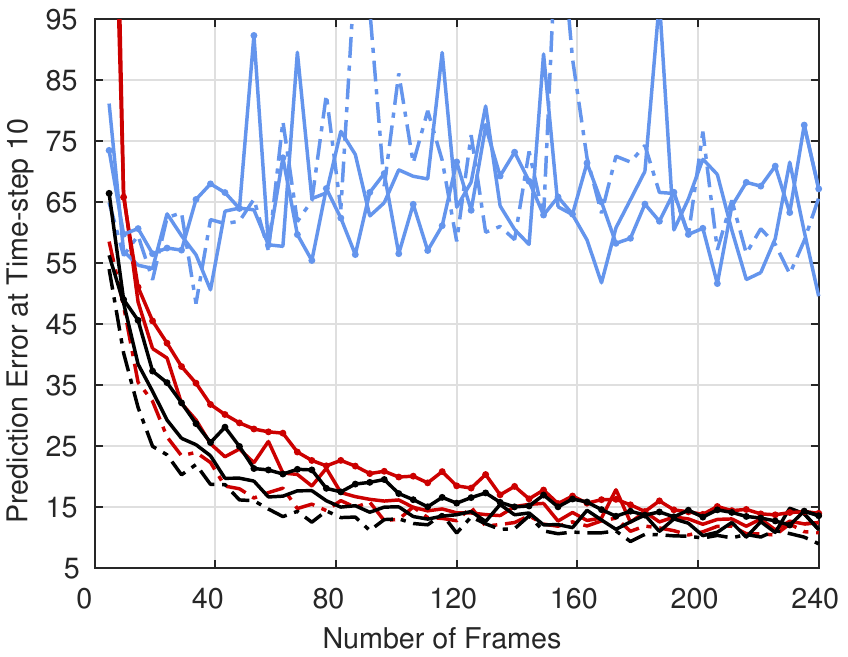}}
%\hskip0.1cm
\scalebox{0.79}{\includegraphics[]{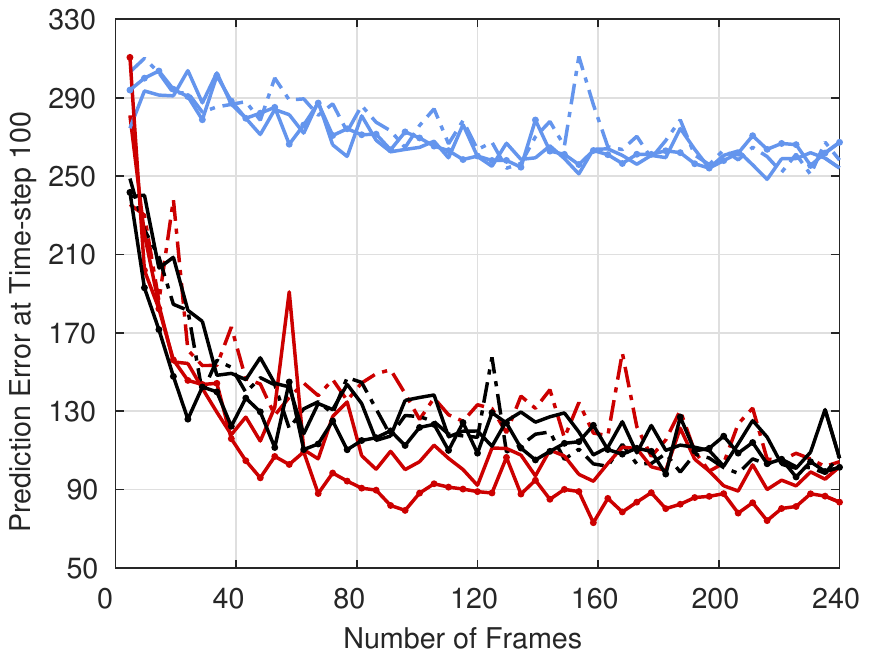}}}
\scalebox{0.79}{\includegraphics[]{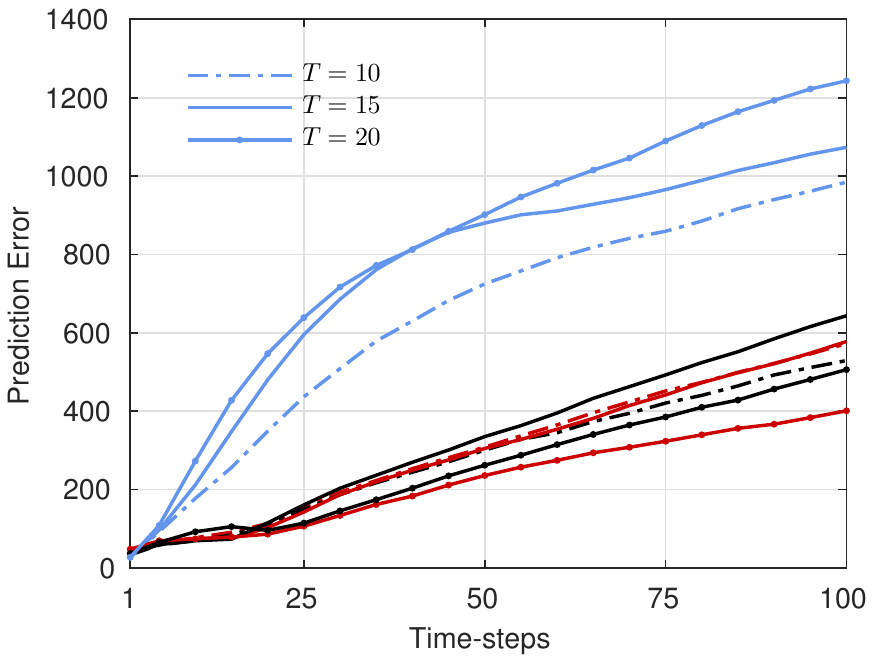}}
%\hskip0.1cm
\scalebox{0.79}{\includegraphics[]{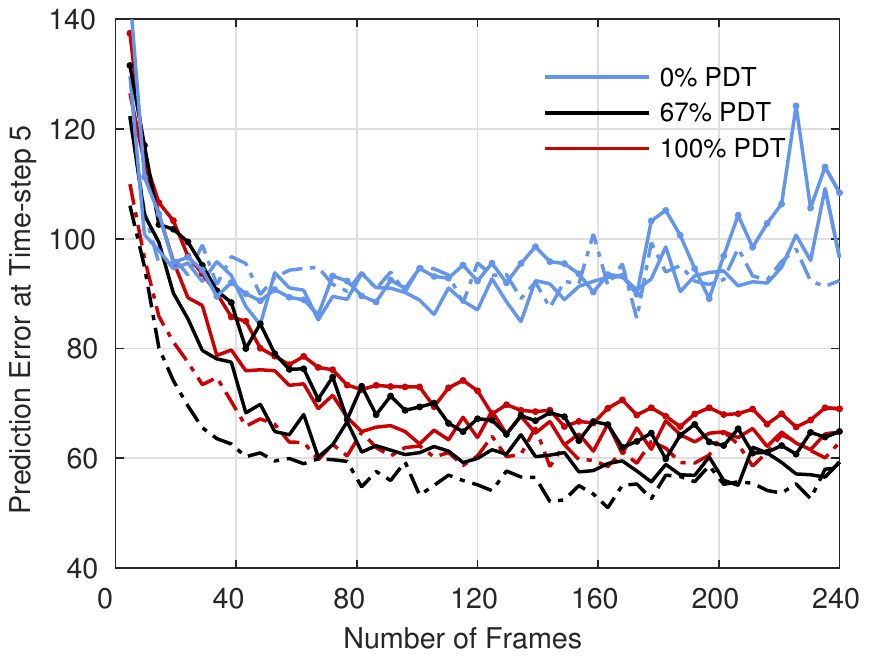}}
\subfigure[]{
\scalebox{0.79}{\includegraphics[]{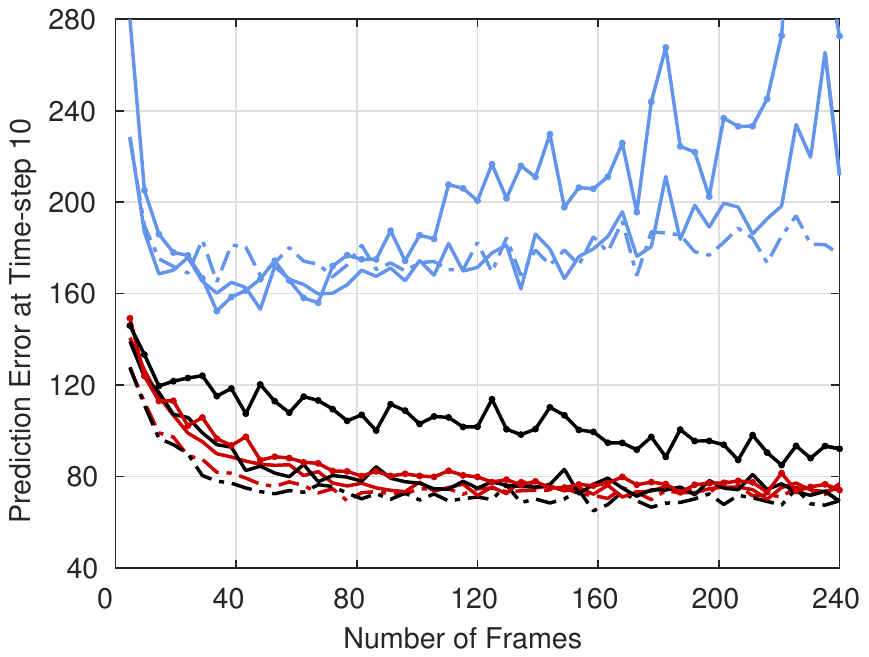}}
%\hskip0.1cm
\scalebox{0.79}{\includegraphics[]{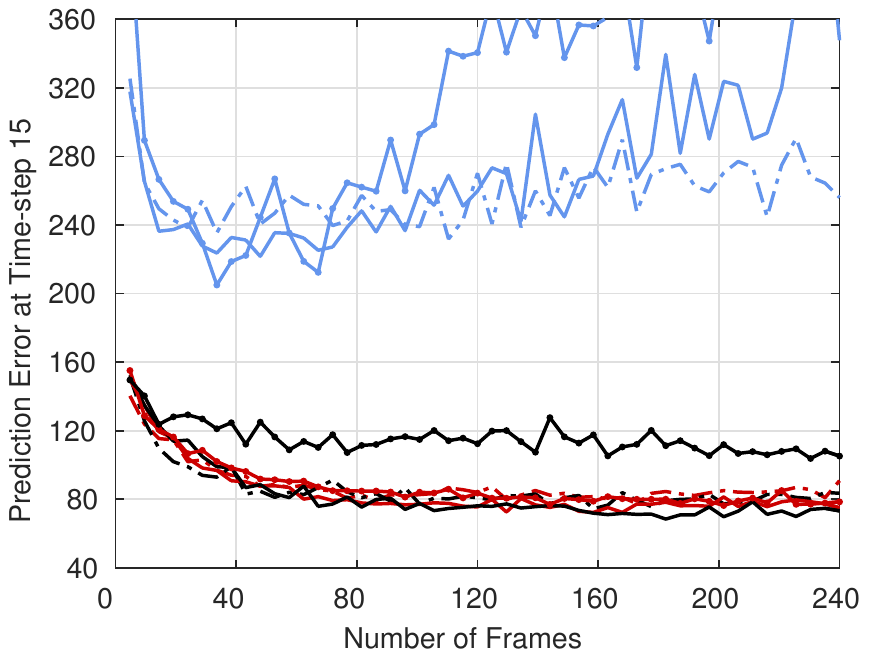}}}
\caption{Prediction error for different prediction lengths $T\leq 20$ on (a) Qbert and (b) Riverraid.}
\label{fig:predErrSeqLengthQbert-Riverraid}
\end{figure}

\begin{figure}[htbp] % Figures obtained with predErrSeqLengthAppendix
\vskip-0.5cm
\scalebox{0.79}{\includegraphics[]{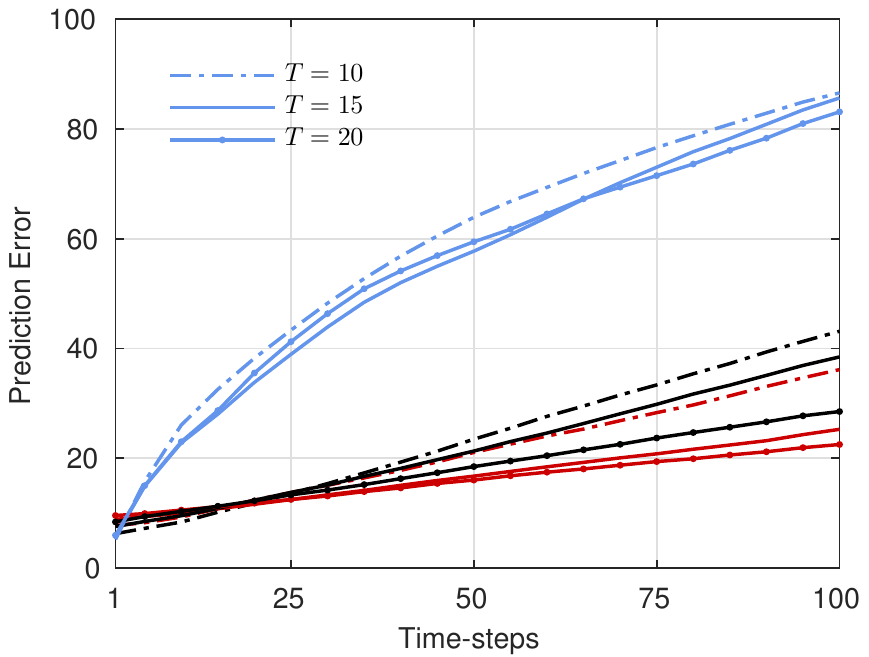}}
%\hskip0.1cm
\scalebox{0.79}{\includegraphics[]{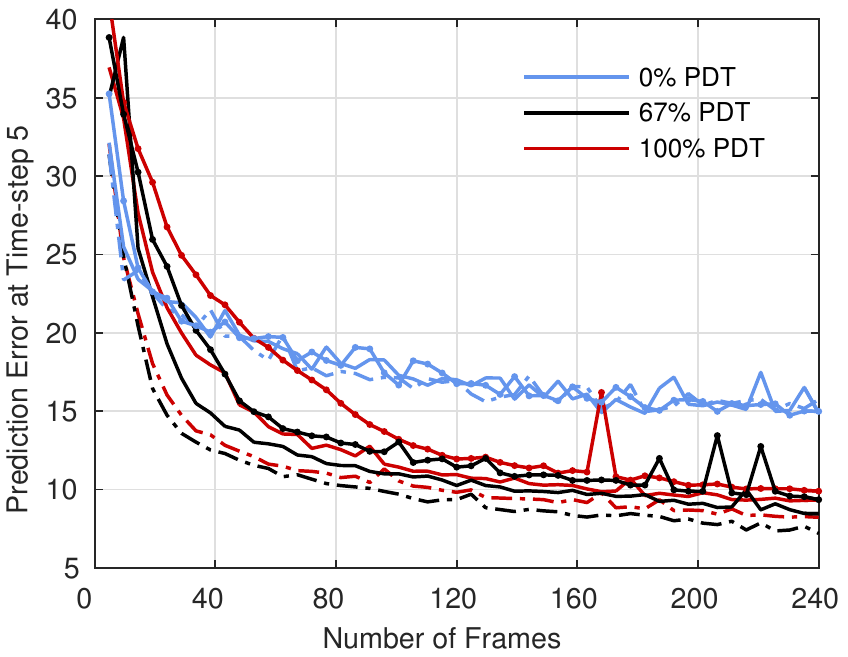}}
\subfigure[]{
\scalebox{0.79}{\includegraphics[]{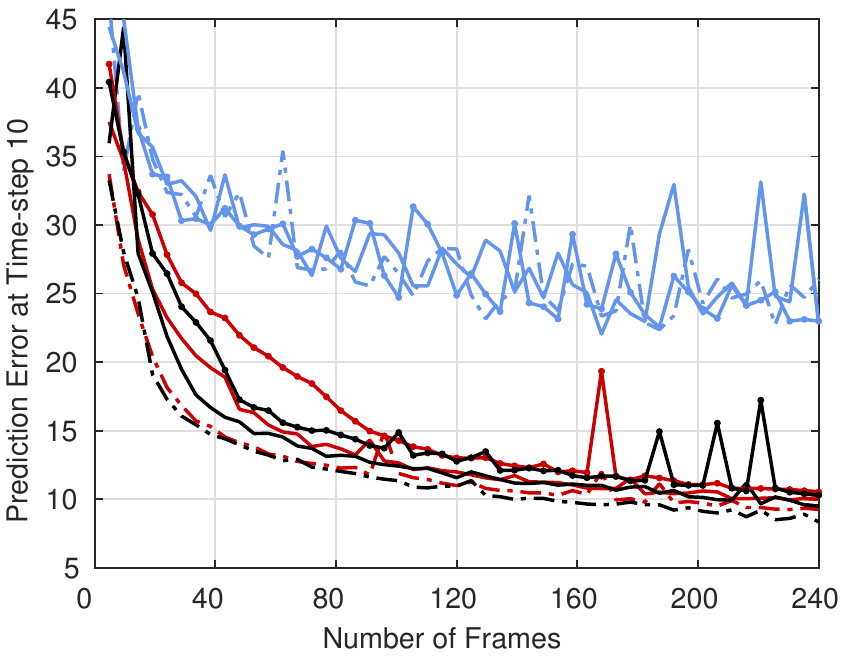}}
%\hskip0.1cm
\scalebox{0.79}{\includegraphics[]{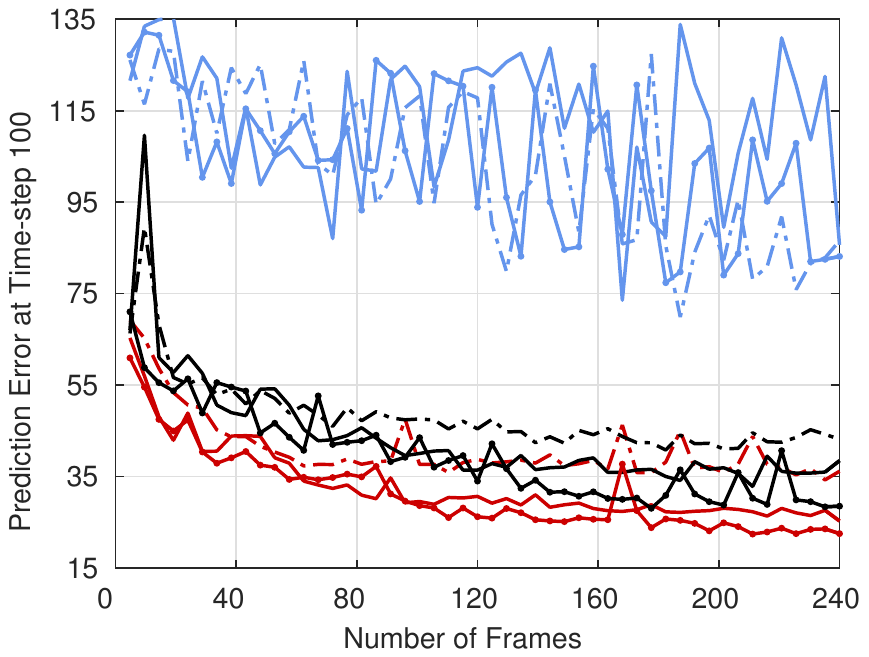}}}
\scalebox{0.79}{\includegraphics[]{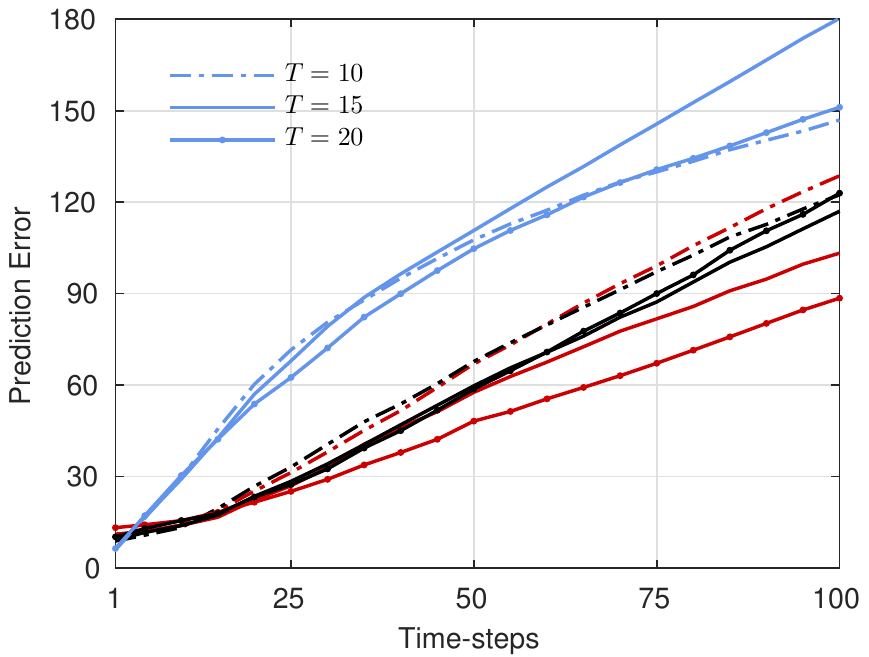}}
%\hskip0.1cm
\scalebox{0.79}{\includegraphics[]{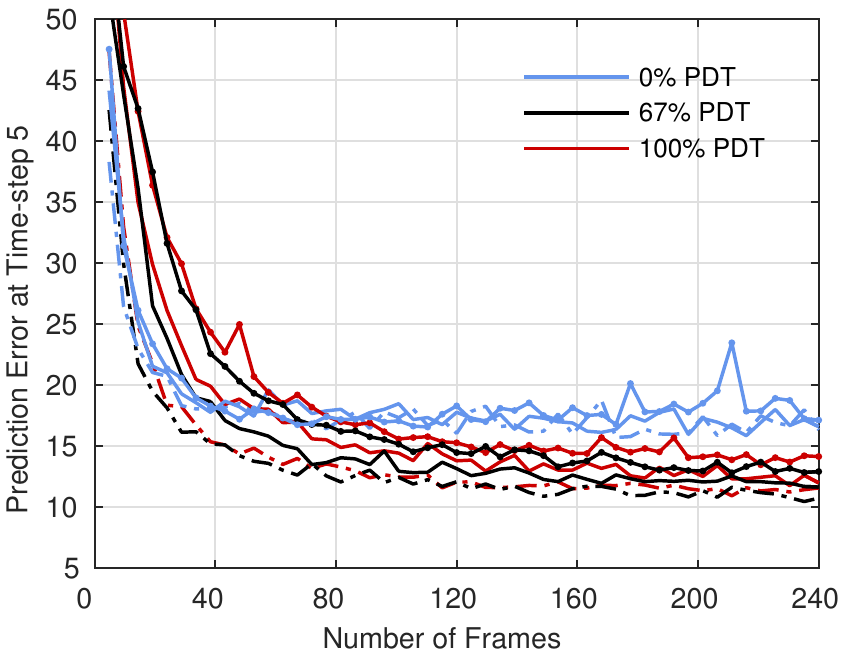}}
\subfigure[]{
\scalebox{0.79}{\includegraphics[]{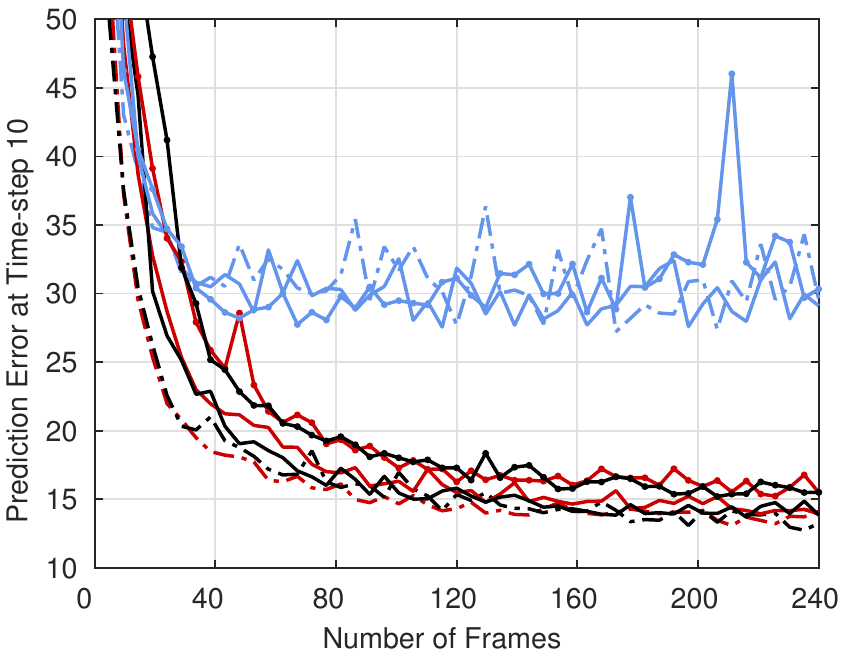}}
%\hskip0.1cm
\scalebox{0.79}{\includegraphics[]{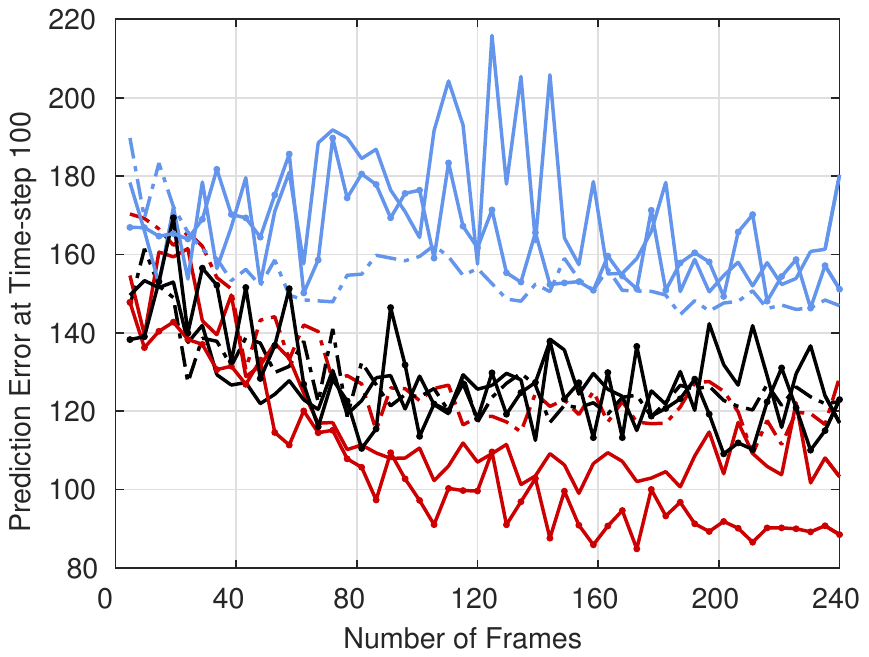}}}
\caption{Prediction error for different prediction lengths $T\leq 20$ on (a) Seaquest and (b) Space Invaders.}
\label{fig:predErrSeqLengthSeaquest-SpaceInvaders}
\end{figure}

In Figs. \ref{fig:predErrSeqLengthBowling-Breakout}-\ref{fig:predErrSeqLengthSeaquest-SpaceInvaders} 
we show the effect of using different prediction lengths $T\leq 20$ with the training schemes 0\%\PDT, 67\%\PDT, and 100\%\PDT~for all games.

\begin{figure}[htbp] % Figures obtained with predErrSeqNumAppendix
\vskip-0.5cm
\scalebox{0.79}{\includegraphics[]{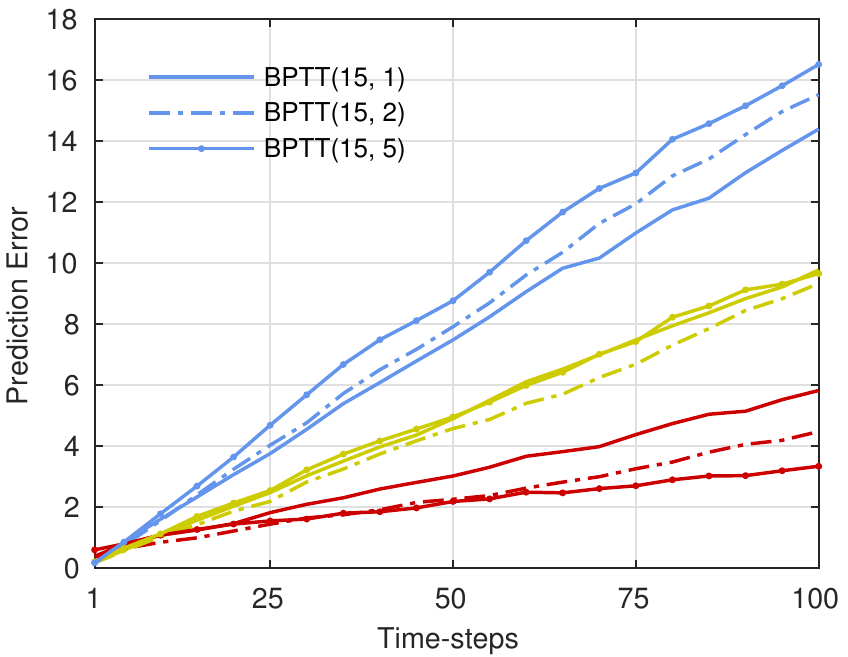}}
%\hskip0.1cm
\scalebox{0.79}{\includegraphics[]{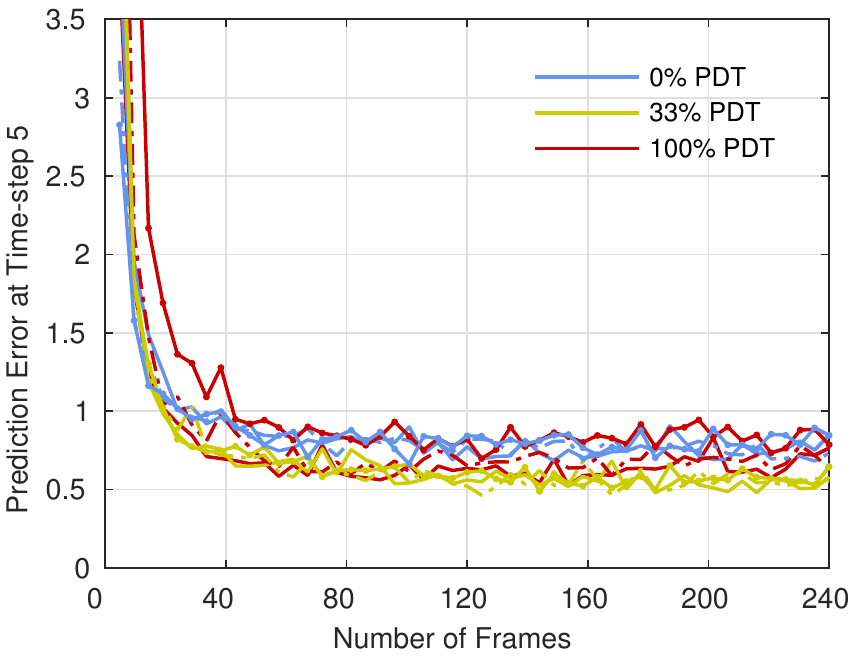}}\\
\subfigure[]{\scalebox{0.79}{\includegraphics[]{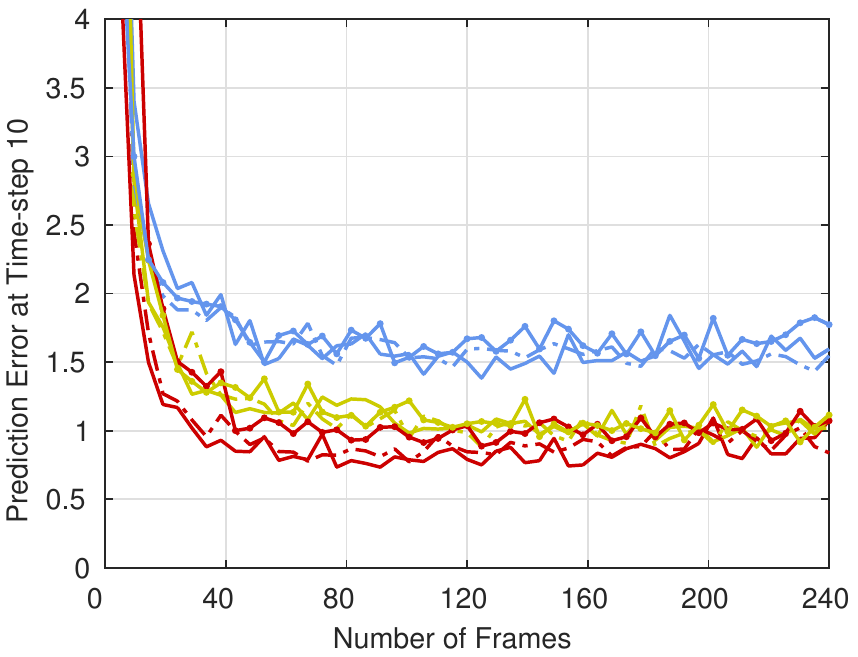}}
%\hskip0.1cm
\scalebox{0.79}{\includegraphics[]{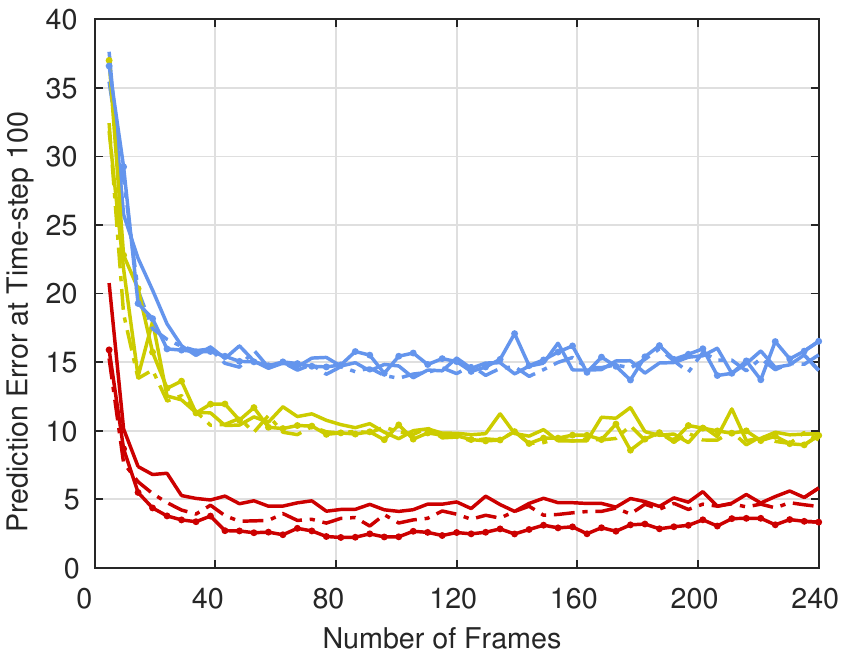}}}
\scalebox{0.79}{\includegraphics[]{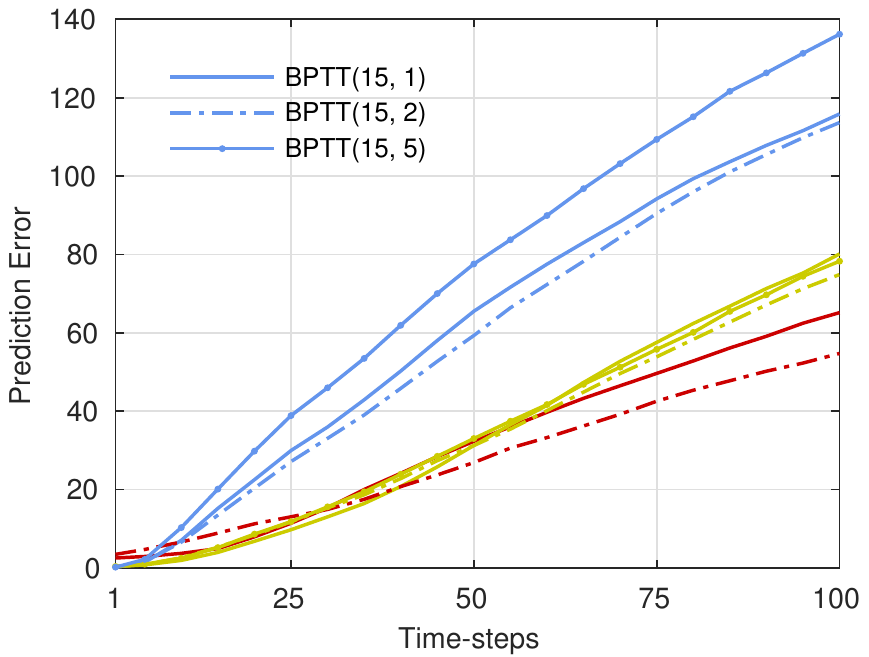}}
%\hskip0.1cm
\scalebox{0.79}{\includegraphics[]{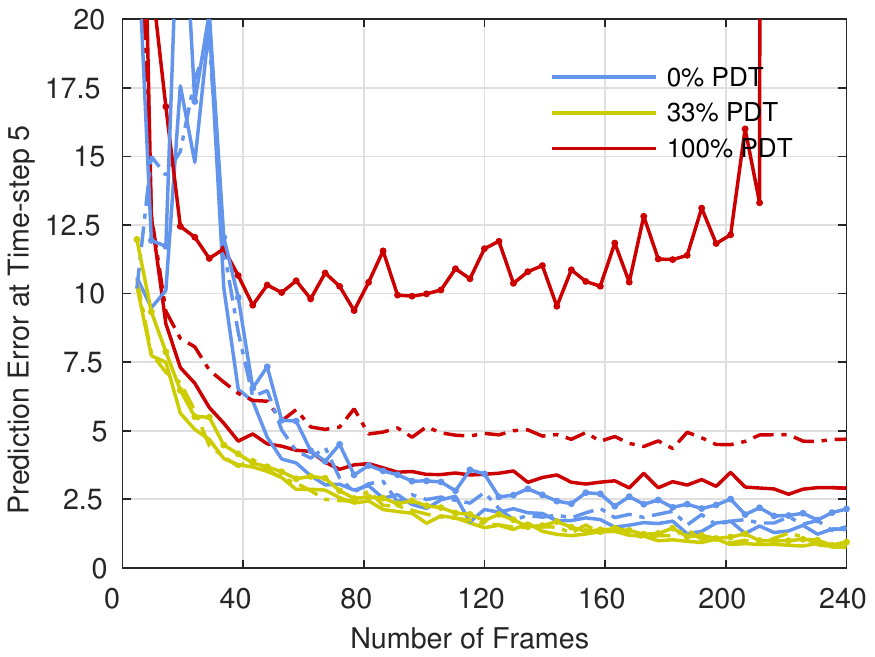}}
\subfigure[]{
\scalebox{0.79}{\includegraphics[]{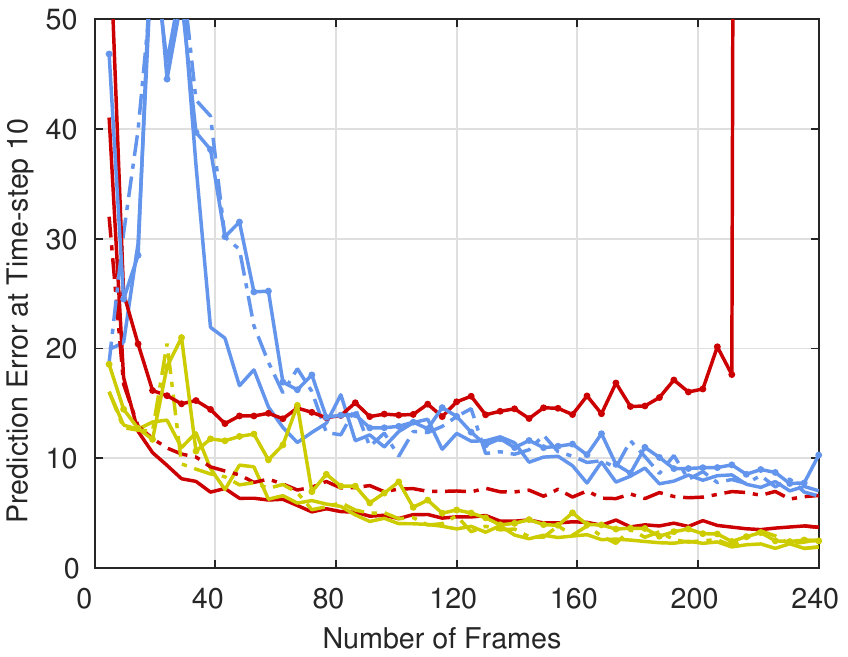}}
%\hskip0.1cm
\scalebox{0.79}{\includegraphics[]{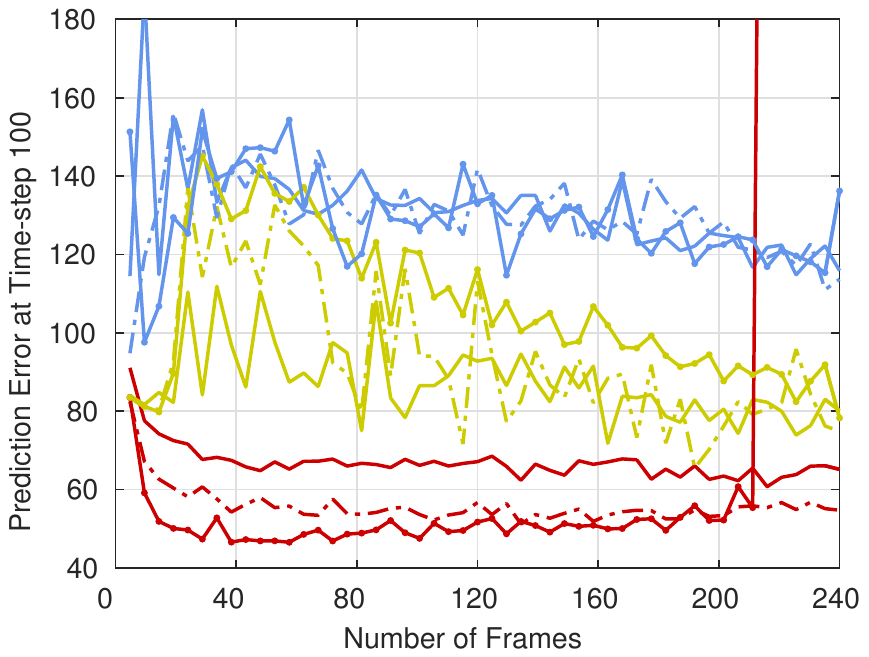}}}
\caption{Prediction error (average over 10,000 sequences) for different prediction lengths through truncated BPTT on (a) Bowling and (b) Breakout. Number of frames is in millions and excludes warm-up frames.}
\label{fig:predErrSeqNumBowling-Breakout}
\end{figure}

\begin{figure}[htbp] % Figures obtained with predErrSeqNumAppendix
\vskip-0.5cm
\scalebox{0.79}{\includegraphics[]{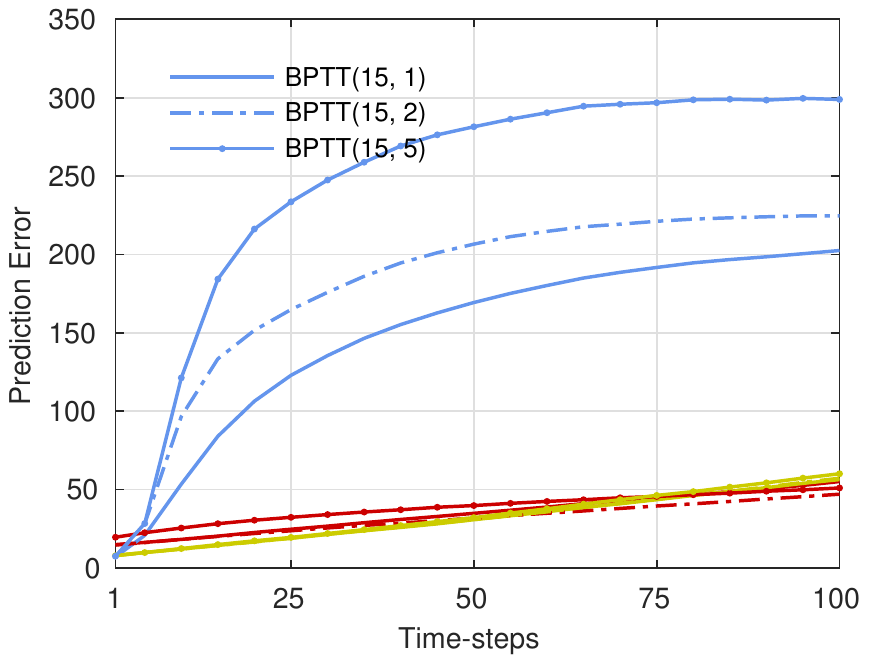}}
%\hskip0.1cm
\scalebox{0.79}{\includegraphics[]{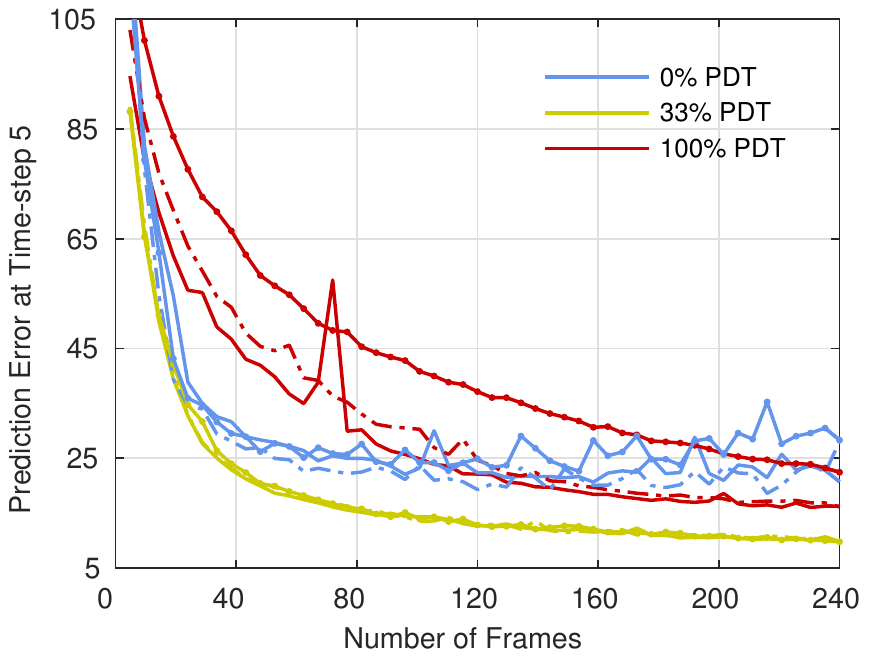}}
\subfigure[]{
\scalebox{0.79}{\includegraphics[]{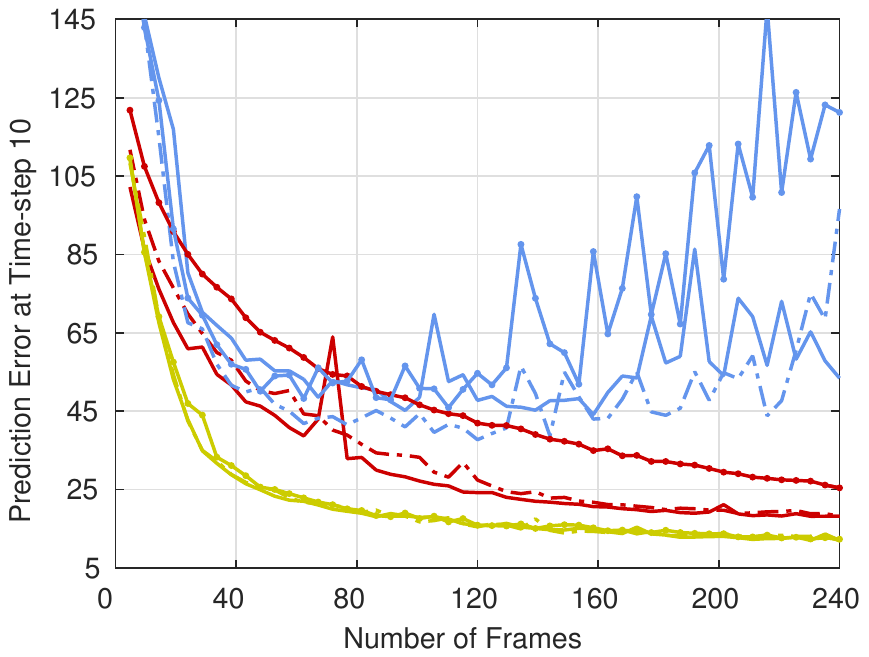}}
%\hskip0.1cm
\scalebox{0.79}{\includegraphics[]{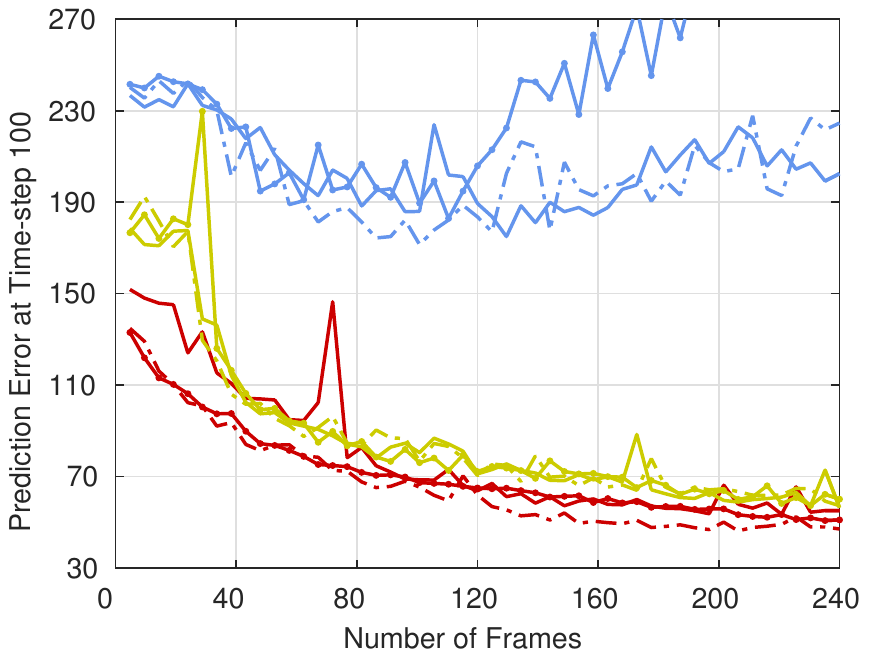}}}
\scalebox{0.79}{\includegraphics[]{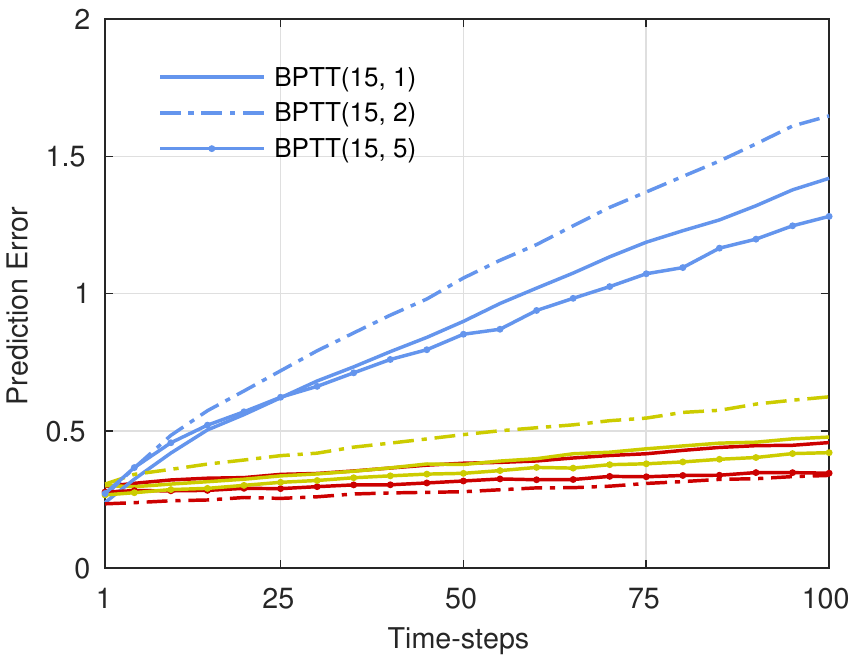}}
%\hskip0.1cm
\scalebox{0.79}{\includegraphics[]{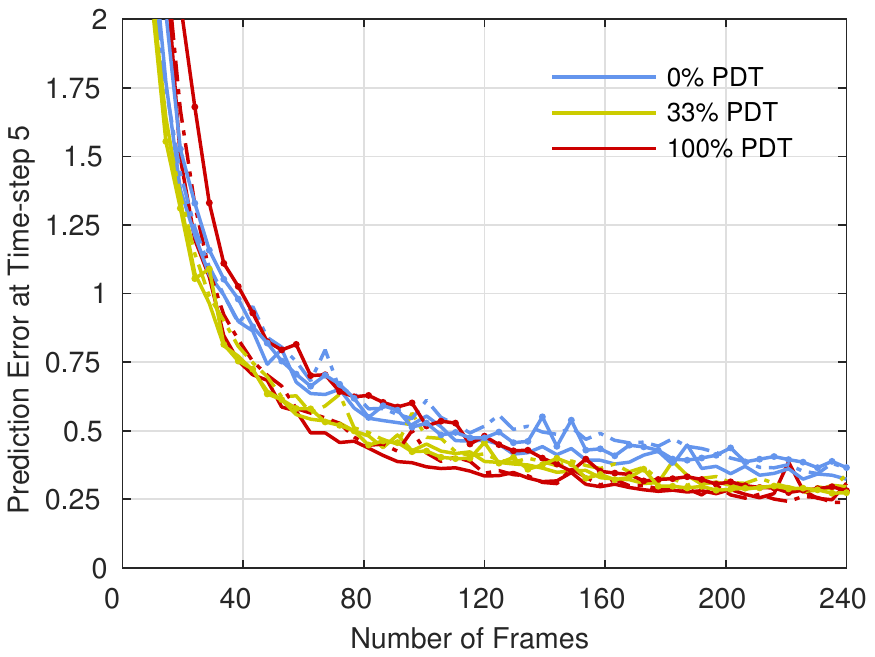}}
\subfigure[]{
\scalebox{0.79}{\includegraphics[]{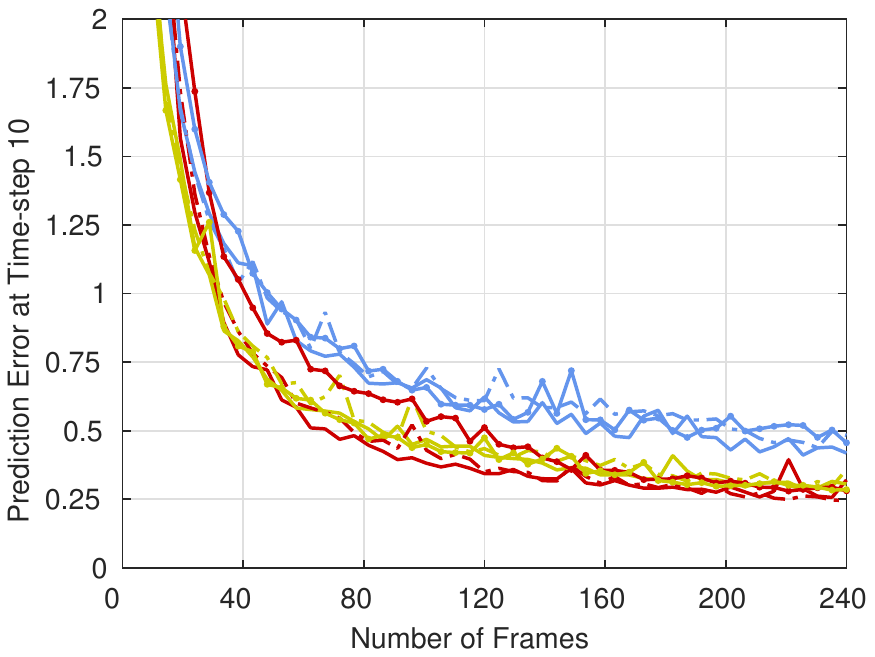}}
%\hskip0.1cm
\scalebox{0.79}{\includegraphics[]{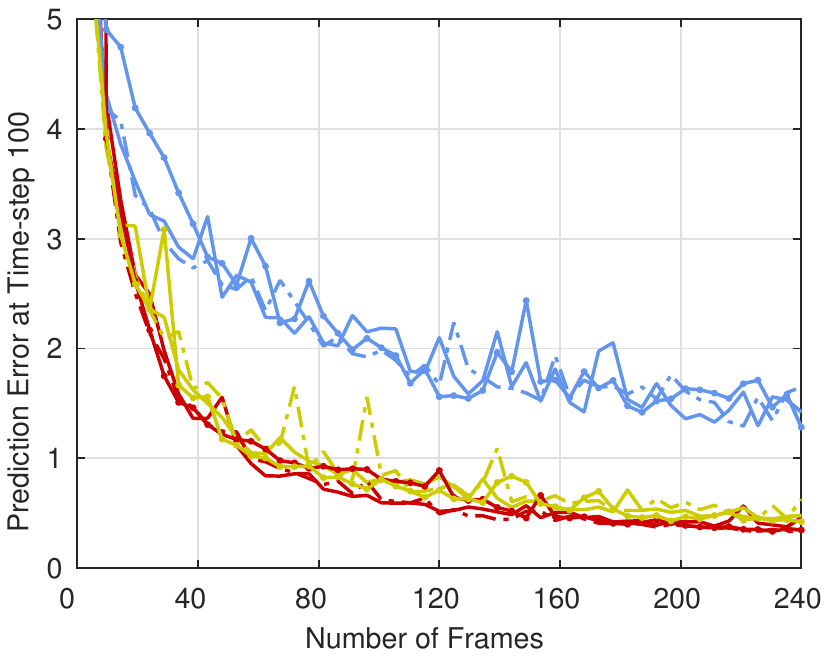}}}
\caption{Prediction error for different prediction lengths through truncated BPTT on (a) Fishing Derby and (b) Freeway.}
\label{fig:predErrSeqNumFishingDerby-Freeway}
\end{figure}

\begin{figure}[htbp] % Figures obtained with predErrSeqNumAppendix
\vskip-0.5cm
\scalebox{0.79}{\includegraphics[]{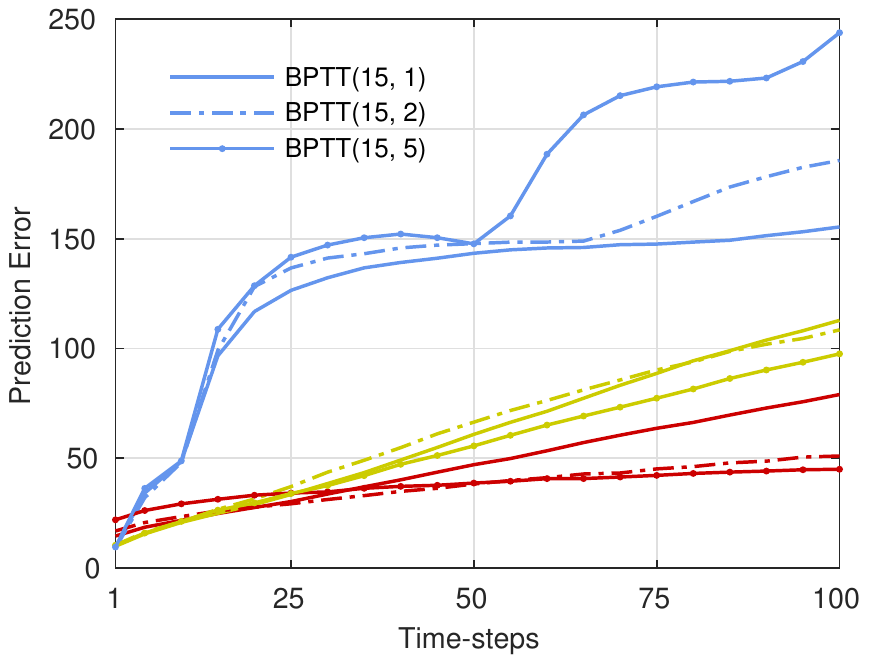}}
%\hskip0.1cm
\scalebox{0.79}{\includegraphics[]{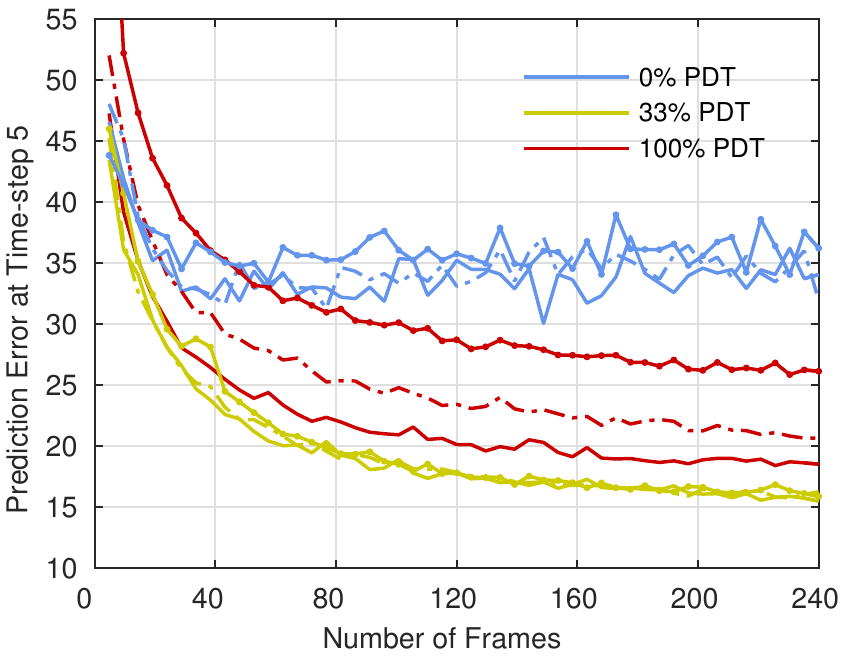}}
\subfigure[]{
\scalebox{0.79}{\includegraphics[]{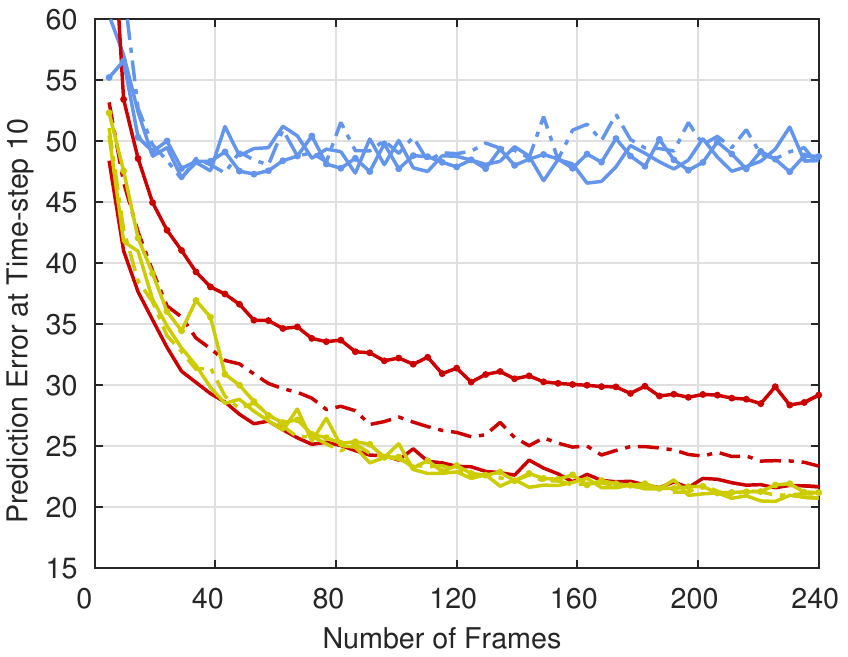}}
%\hskip0.1cm
\scalebox{0.79}{\includegraphics[]{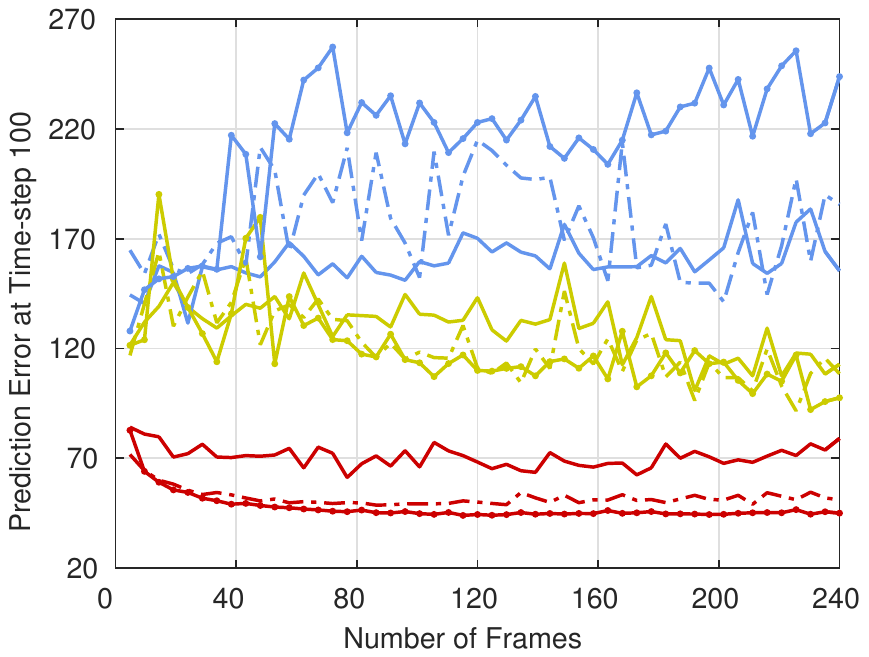}}}
\scalebox{0.79}{\includegraphics[]{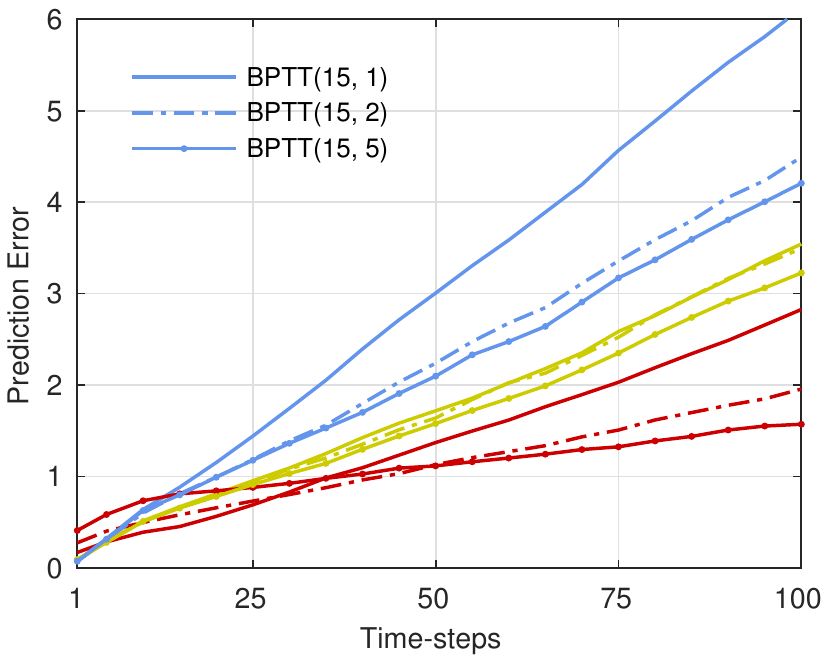}}
%\hskip0.1cm
\scalebox{0.79}{\includegraphics[]{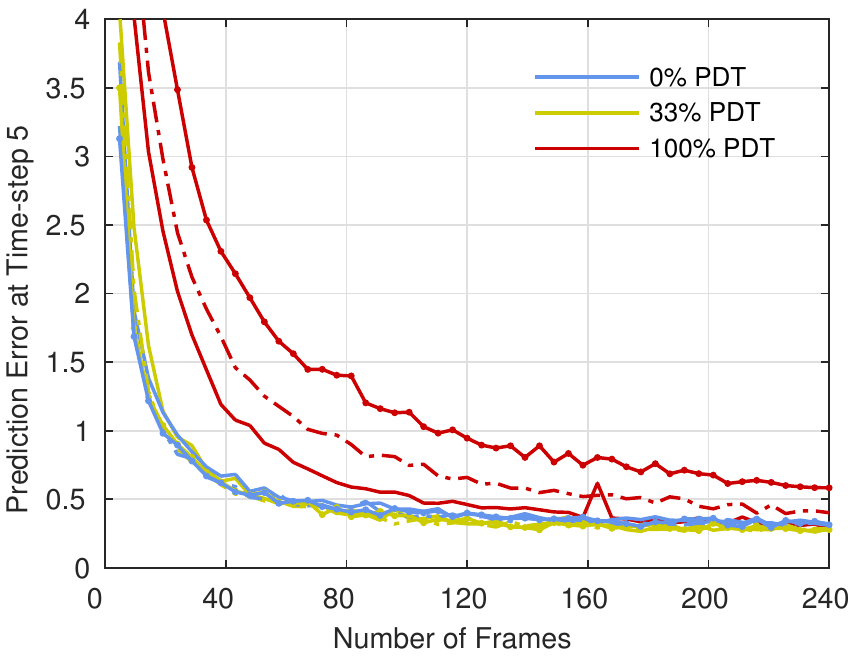}}
\subfigure[]{
\scalebox{0.79}{\includegraphics[]{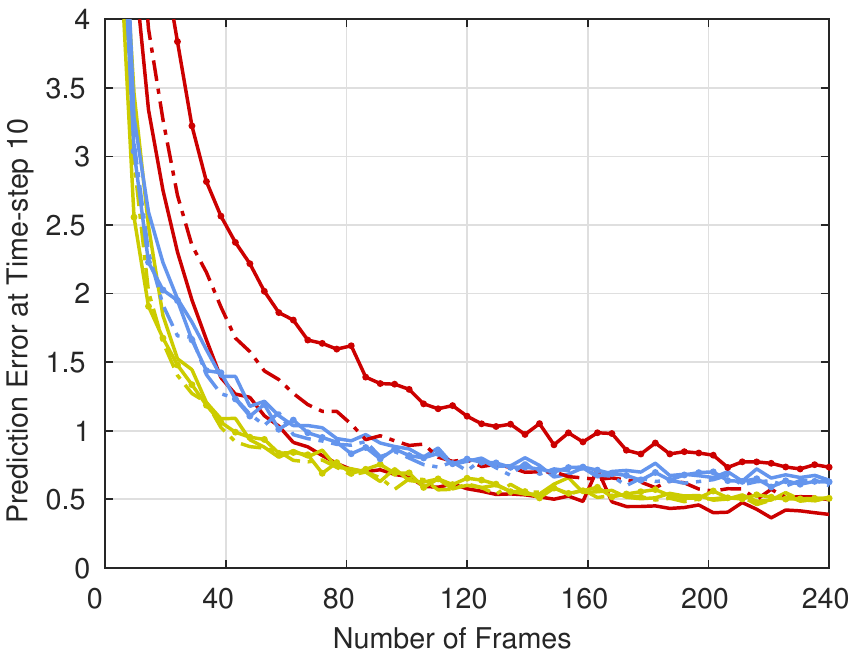}}
%\hskip0.1cm
\scalebox{0.79}{\includegraphics[]{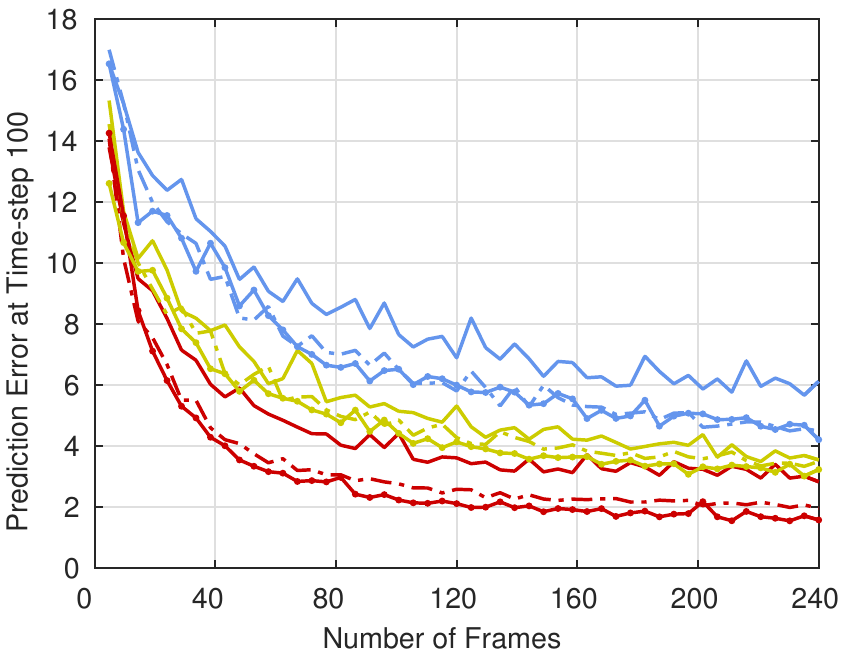}}}
\caption{Prediction error for different prediction lengths through truncated BPTT on (a) Ms Pacman and (b) Pong.}
\label{fig:predErrSeqNumMsPacman-Pong}
\end{figure}

\begin{figure}[htbp] % Figures obtained with predErrSeqNumAppendix
\vskip-0.5cm
\scalebox{0.79}{\includegraphics[]{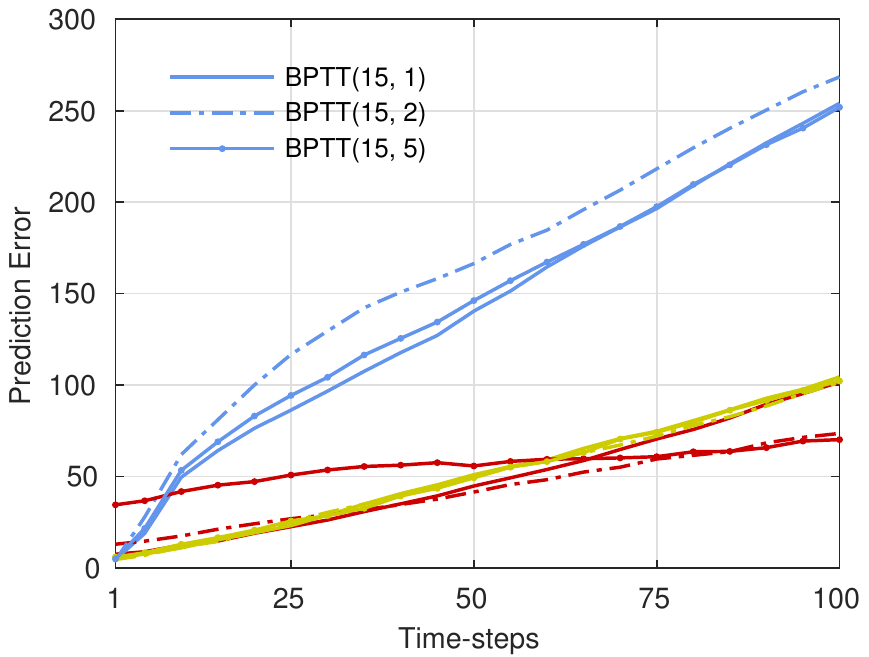}}
%\hskip0.1cm
\scalebox{0.79}{\includegraphics[]{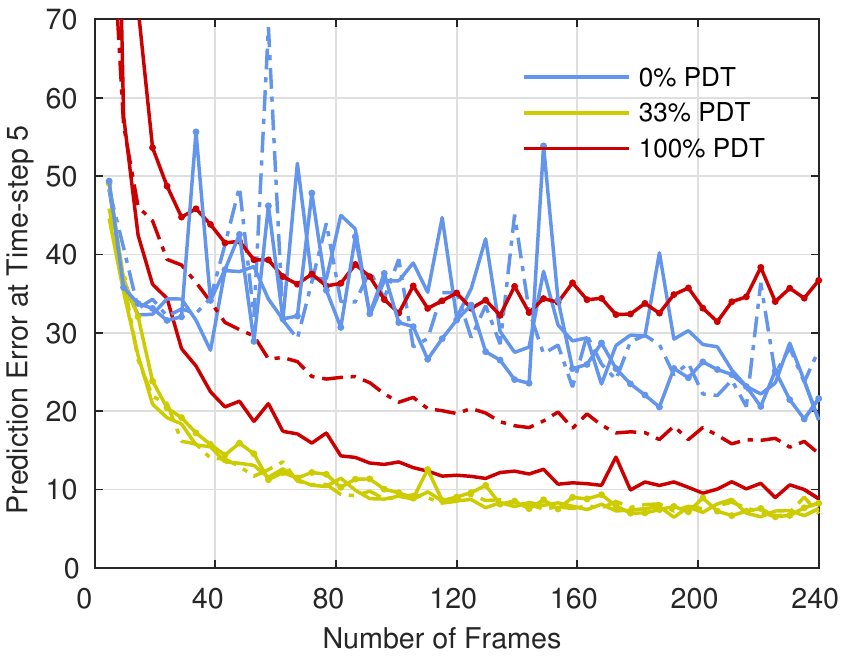}}
\subfigure[]{
\scalebox{0.79}{\includegraphics[]{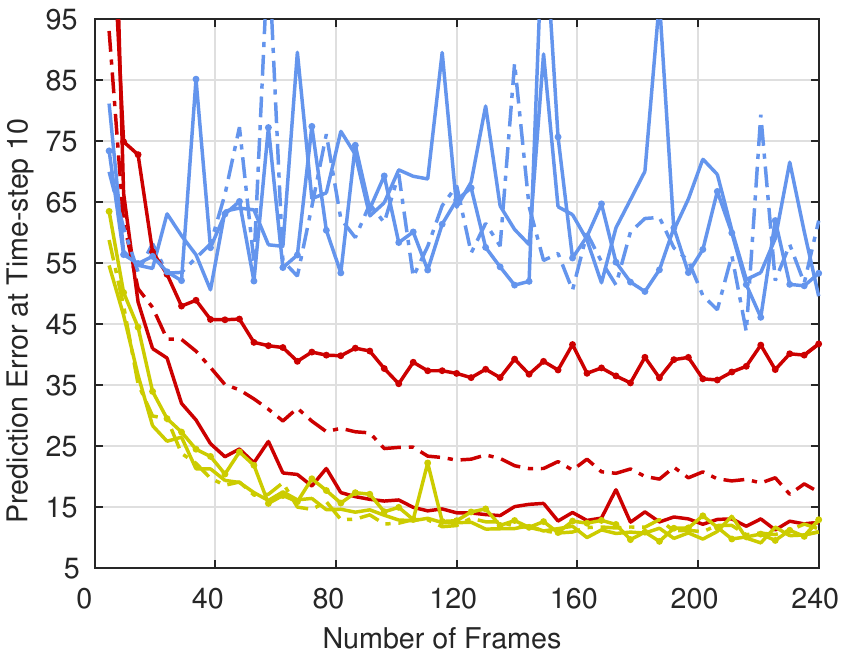}}
%\hskip0.1cm
\scalebox{0.79}{\includegraphics[]{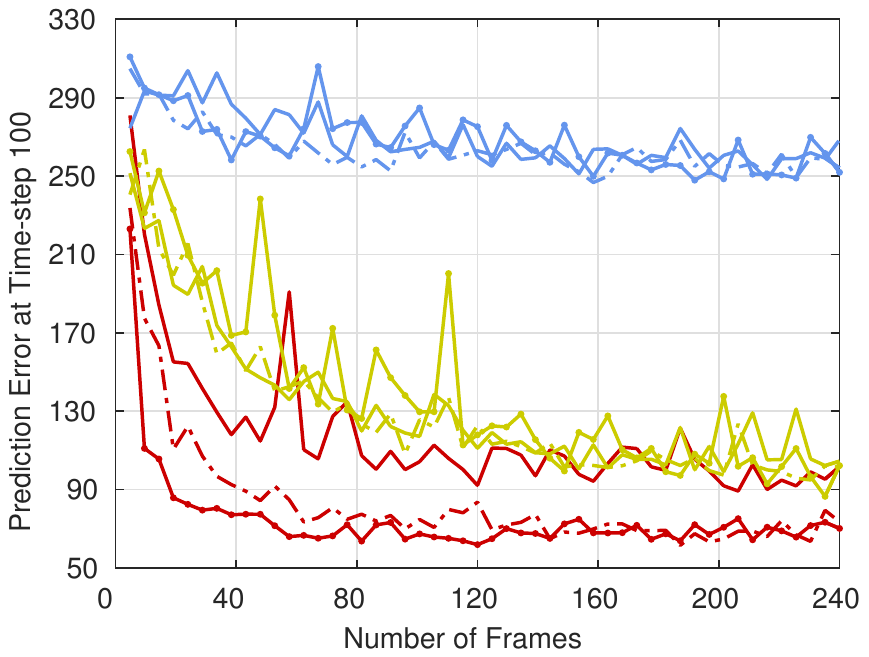}}}
\scalebox{0.79}{\includegraphics[]{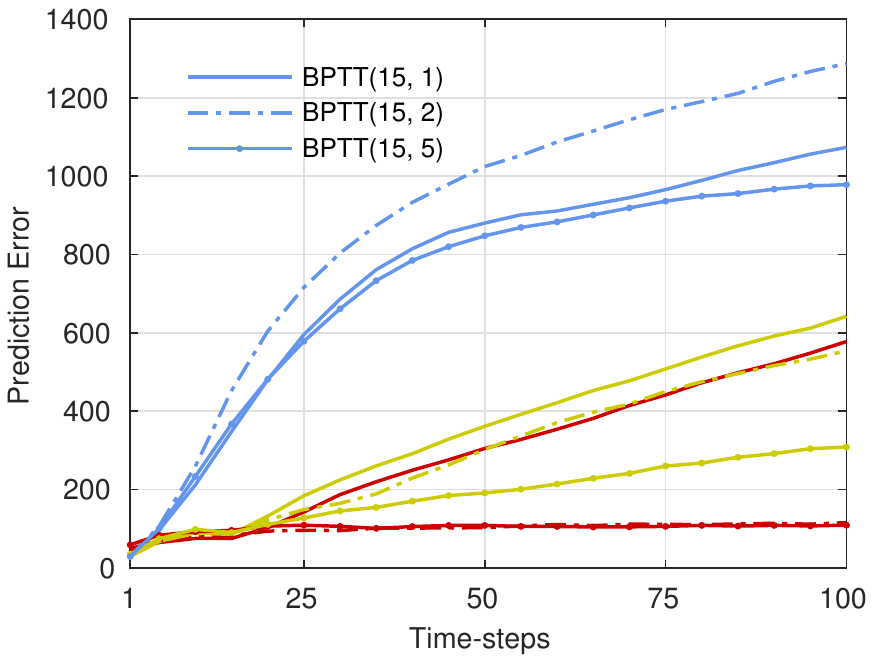}}
%\hskip0.1cm
\scalebox{0.79}{\includegraphics[]{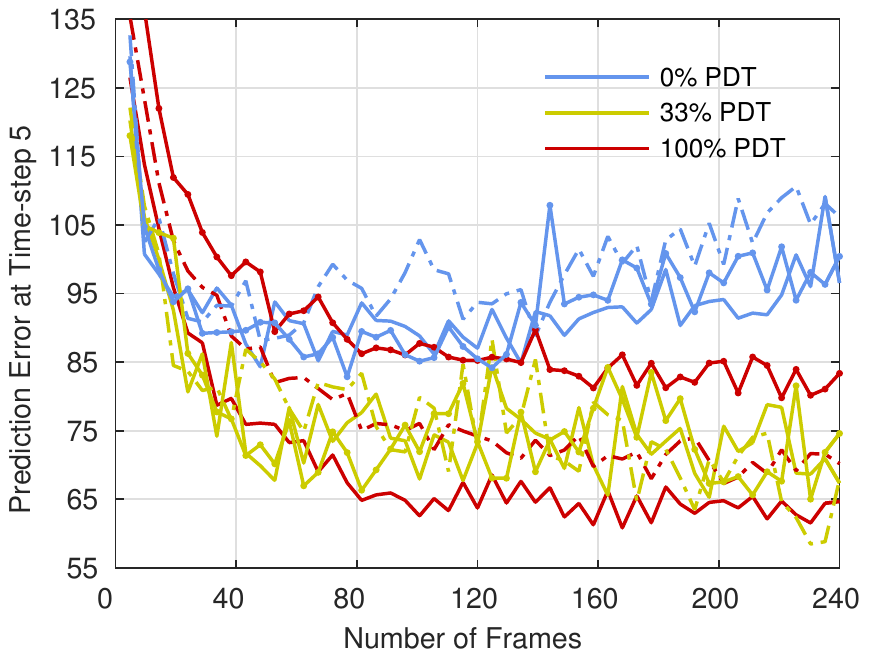}}
\subfigure[]{
\scalebox{0.79}{\includegraphics[]{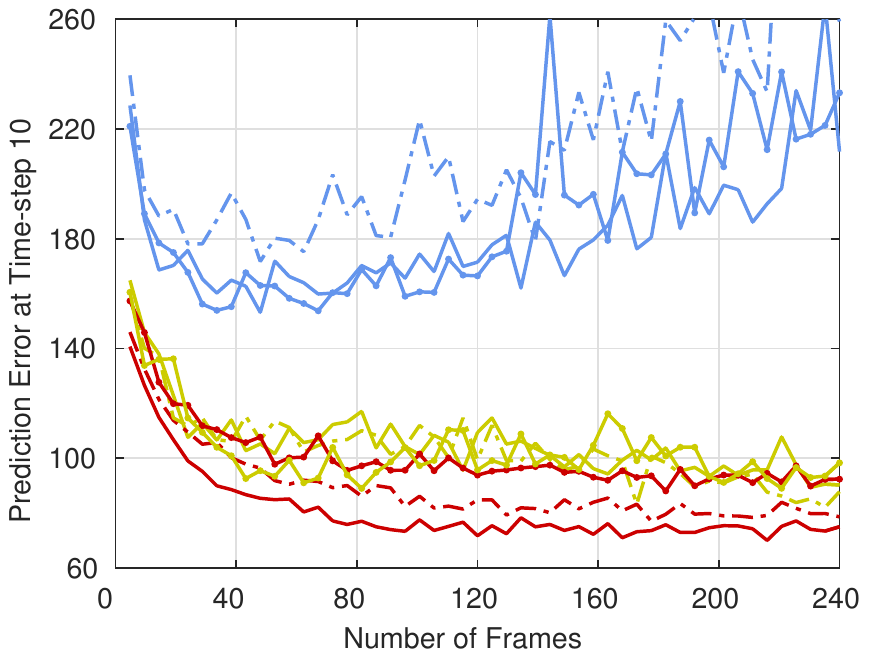}}
%\hskip0.1cm
\scalebox{0.79}{\includegraphics[]{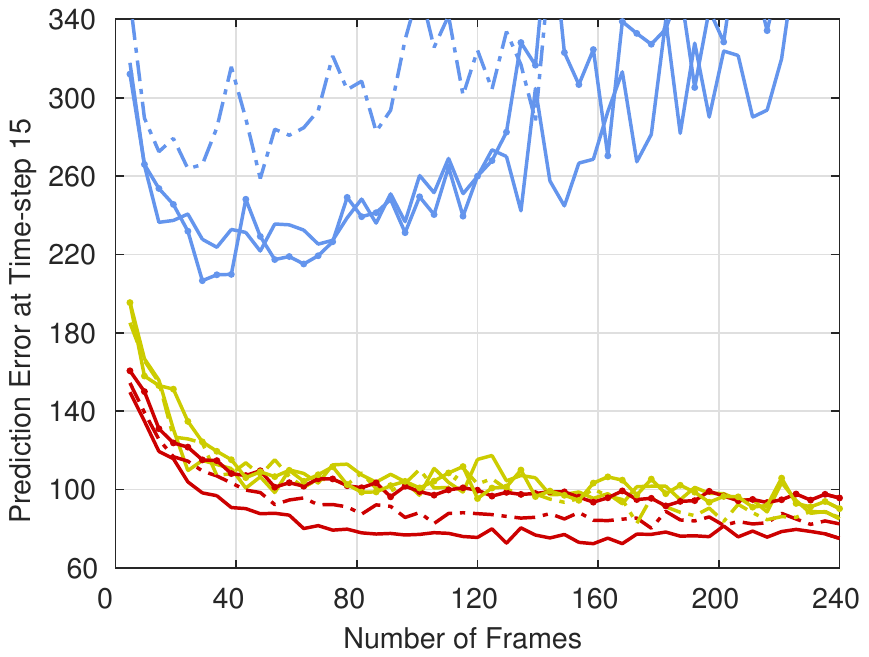}}}
\caption{Prediction error for different prediction lengths through truncated BPTT on (a) Qbert and (b) Riverraid.}
\label{fig:predErrSeqNumQbert-Riverraid}
\end{figure}

\begin{figure}[htbp] % Figures obtained with predErrSeqNumAppendix
\vskip-0.5cm
\scalebox{0.79}{\includegraphics[]{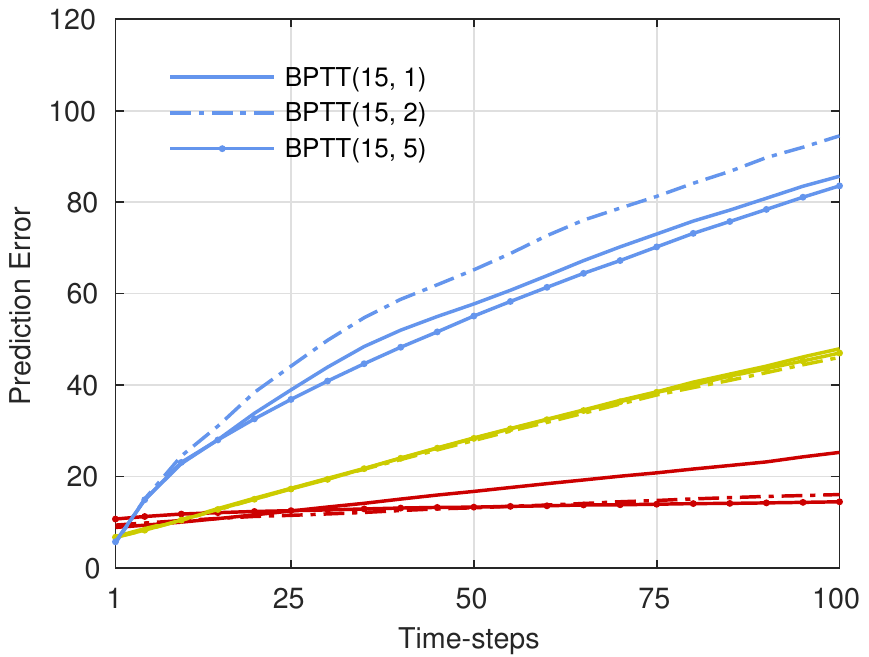}}
%\hskip0.1cm
\scalebox{0.79}{\includegraphics[]{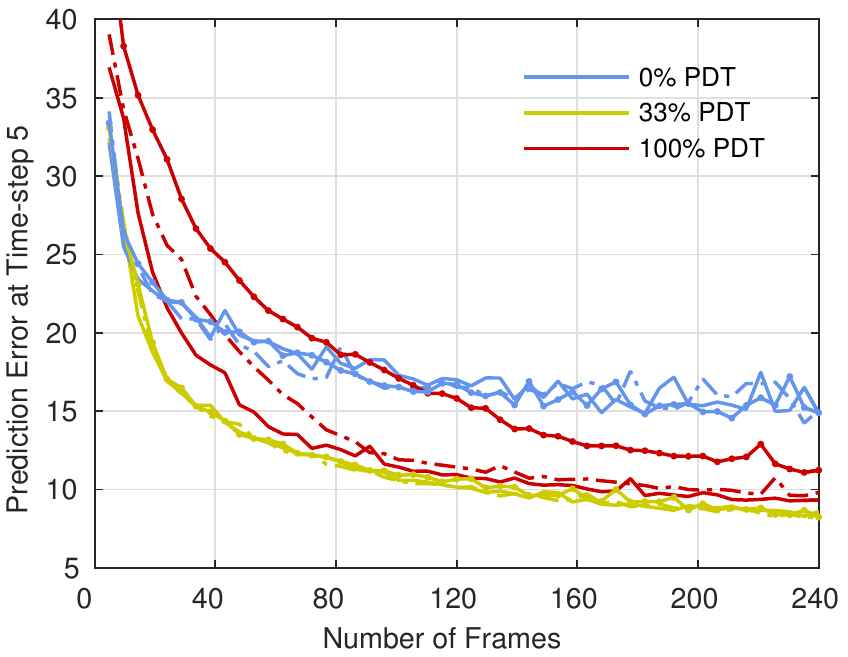}}
\subfigure[]{
\scalebox{0.79}{\includegraphics[]{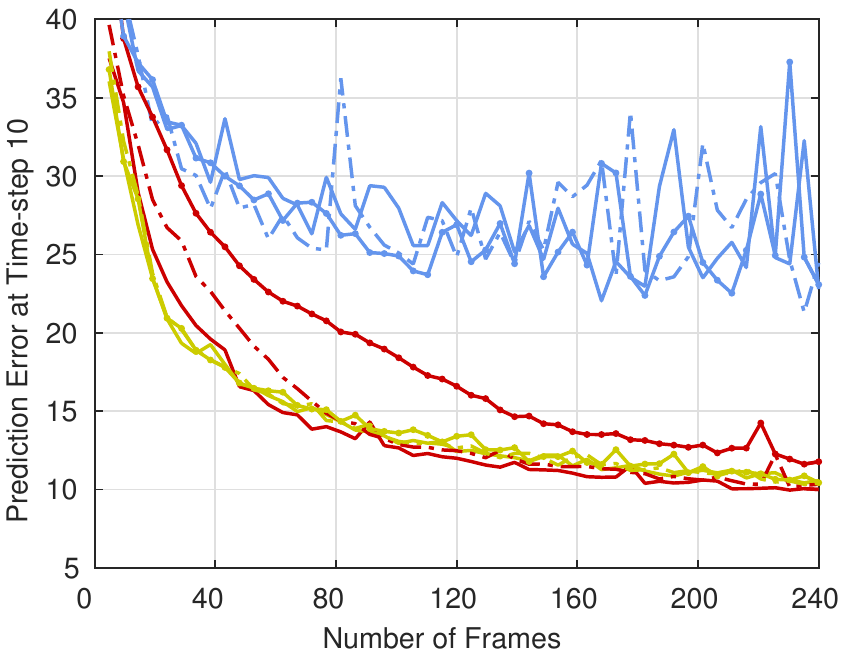}}
%\hskip0.1cm
\scalebox{0.79}{\includegraphics[]{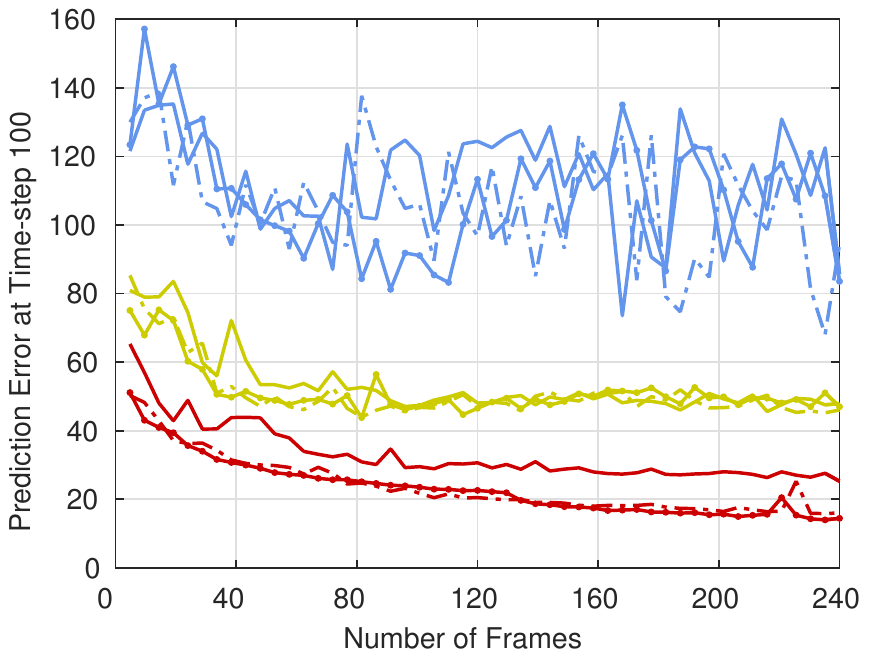}}}
\scalebox{0.79}{\includegraphics[]{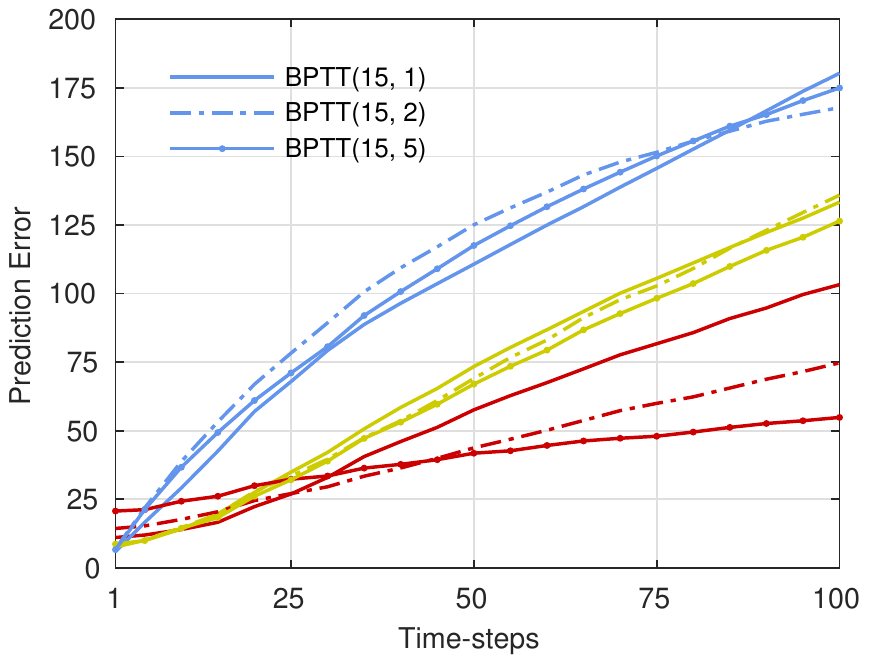}}
%\hskip0.1cm
\scalebox{0.79}{\includegraphics[]{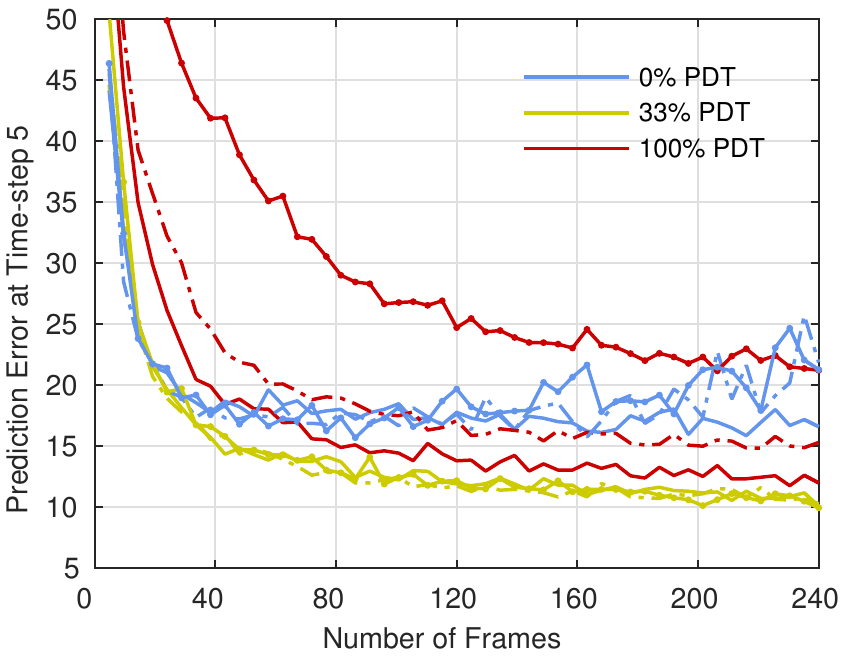}}
\subfigure[]{
\scalebox{0.79}{\includegraphics[]{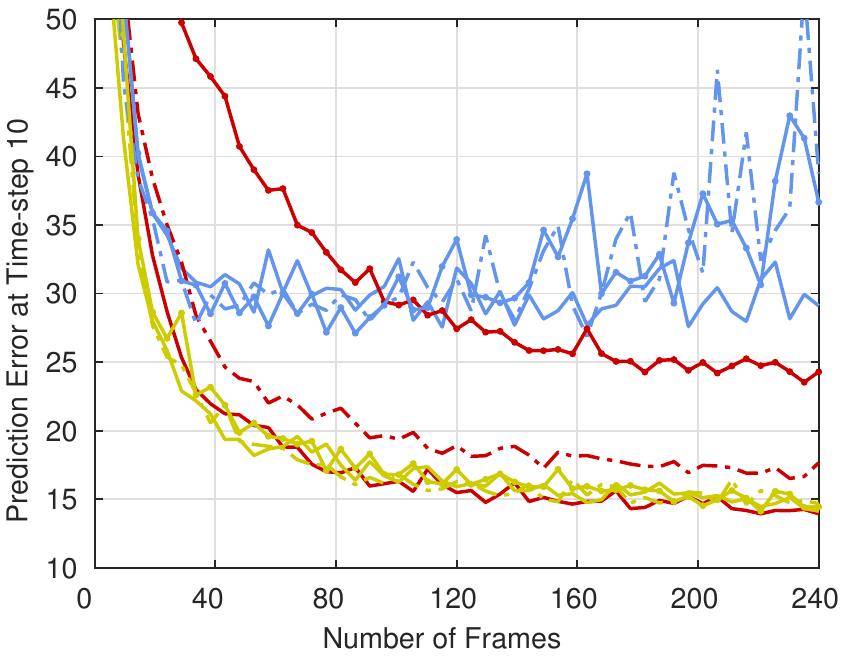}}
%\hskip0.1cm
\scalebox{0.79}{\includegraphics[]{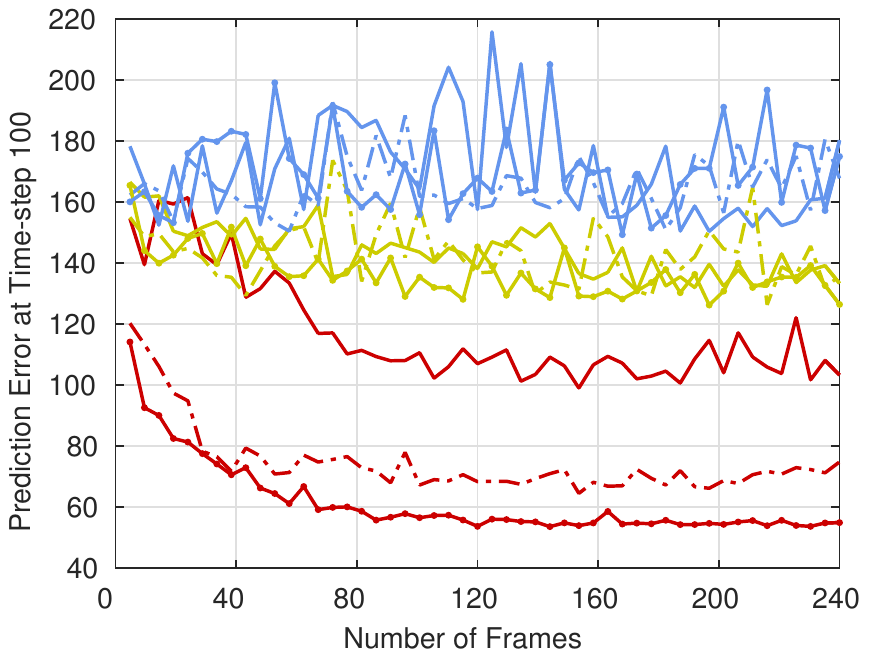}}}
\caption{Prediction error for different prediction lengths through truncated BPTT on (a) Seaquest and (b) Space Invaders.}
\label{fig:predErrSeqNumSeaquest-SpaceInvaders}
\end{figure}

In Figs. \ref{fig:predErrSeqNumBowling-Breakout}-\ref{fig:predErrSeqNumSeaquest-SpaceInvaders} 
we show the effect of using different prediction lengths $T>20$ through truncated backpropagation with the training schemes 0\%\PDT, 33\%\PDT, and 100\%\PDT~for all games.

\clearpage
\clearpage

\subsubsection{Different Action-Dependent State Transitions\label{sec:AppStructures}}
%\vspace{-2mm}
In this section we compare the baseline state transition  
\begin{align*}
\textrm{Encoding: } & \vz_{t-1} = {\cal C}(\mathbb{I}(\hat \vx_{t-1}, \vx_{t-1}))\,,\\
\textrm{Action fusion: } & \ha_{t} = \vW^h \vh_{t-1}\otimes \vW^a \va_{t-1}\,,\\
\textrm{Gate update: } & \vi_t = {\sigma}(\vW^{iv}\ha_{t} +\vW^{iz}\vz_{t-1})\,, \hskip0.1cm \vf_t = {\sigma}(\vW^{fv}\ha_t +\vW^{fz}\vz_{t-1})\,,\\ 
& \vo_t = {\sigma}(\vW^{ov}\ha_t + \vW^{oz}\vz_{t-1})\,,\\
\textrm{Cell update: } & \vc_t = \vf_t\otimes \vc_{t-1} + \vi_t\otimes \textrm{tanh}(\vW^{cv}\ha_{t}+\vW^{cz}\vz_{t-1})\,,\\
\textrm{State update: } & \vh_t = \vo_t\otimes \textrm{tanh}(\vc_t)\,, 
\end{align*}
where the vectors $\vh_{t-1}$ and $\ha_t$ have dimension 1024 and 2048 respectively (this model has around 25 millions (25M) parameters),
with alternatives using unconstrained or convolutional transformations, for prediction length $T=15$ and the 0\%-100\%\PDT~training scheme.
 
\begin{figure}[htbp] % Figures obtained with predErrStructAppendix
\vskip-0.5cm
\scalebox{0.79}{\includegraphics[]{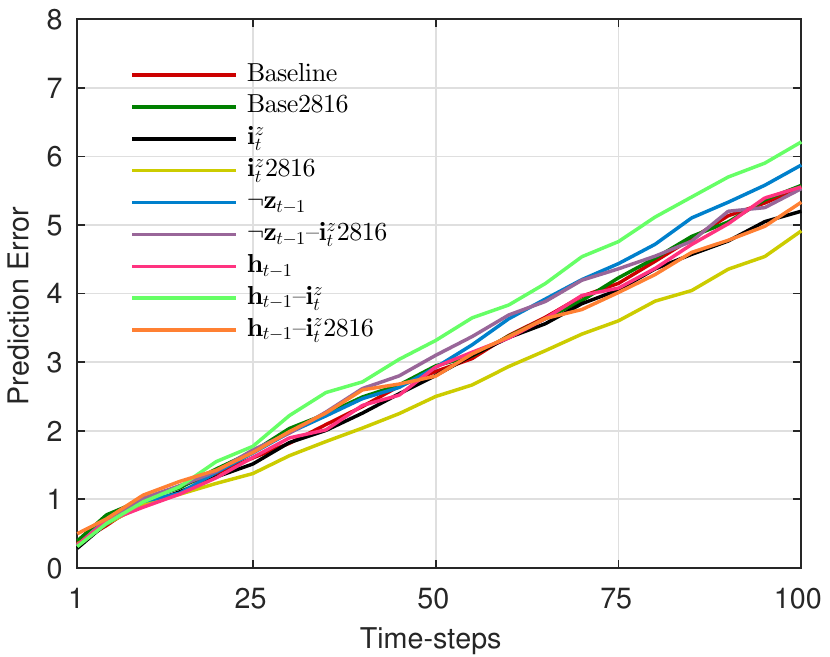}}
%\hskip0.1cm
\scalebox{0.79}{\includegraphics[]{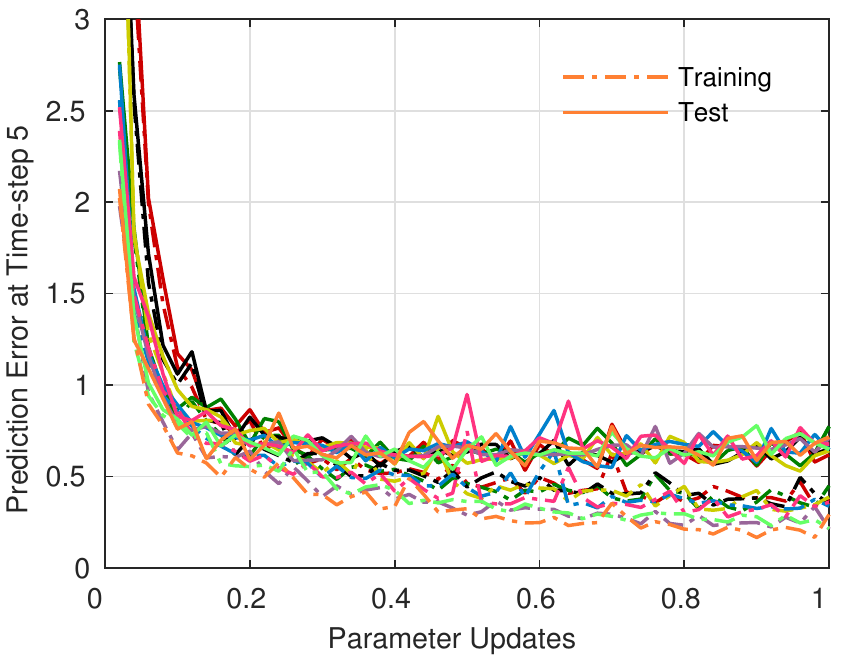}}\\
\subfigure[]{\scalebox{0.79}{\includegraphics[]{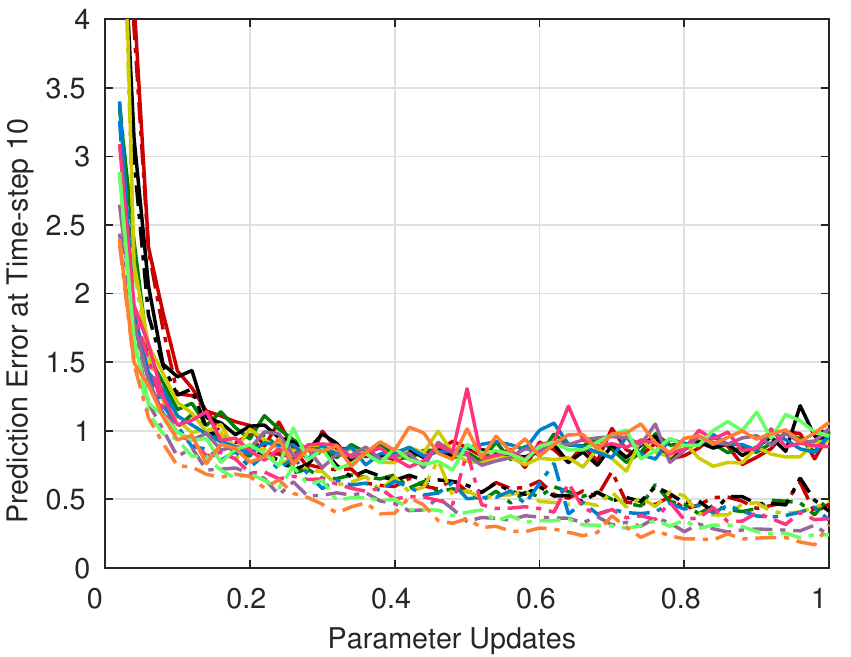}}
%\hskip0.1cm
\scalebox{0.79}{\includegraphics[]{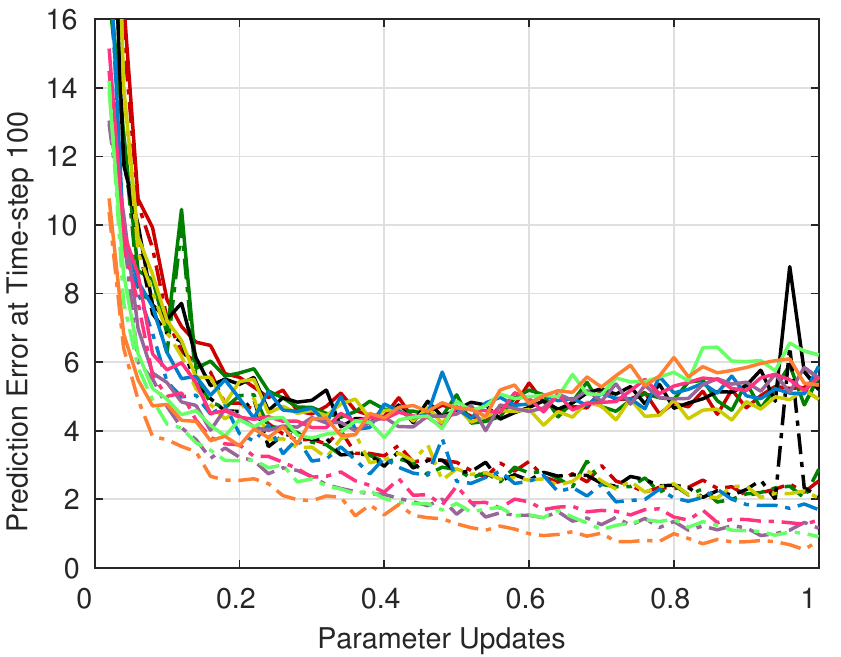}}}
\scalebox{0.79}{\includegraphics[]{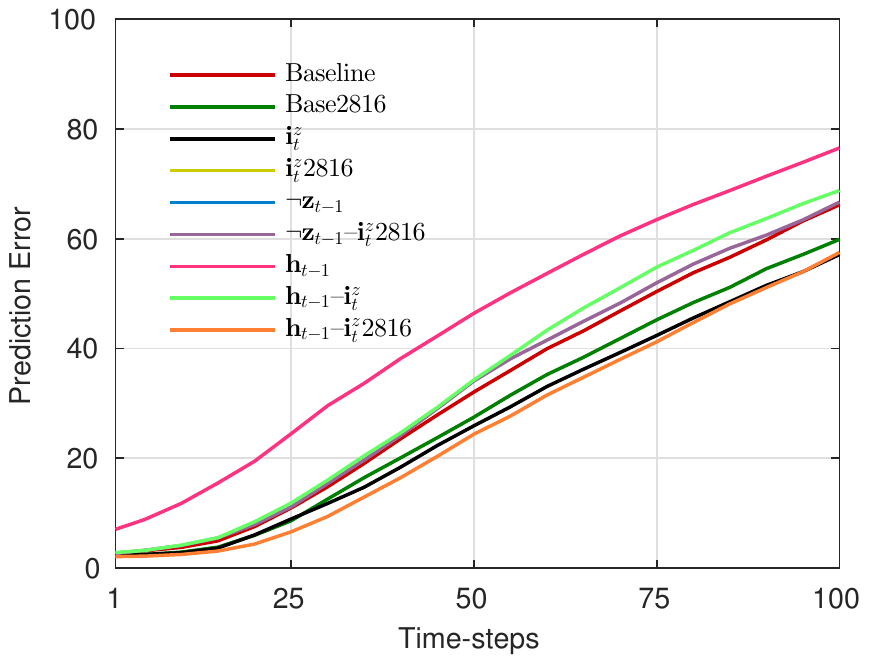}}
%\hskip0.1cm
\scalebox{0.79}{\includegraphics[]{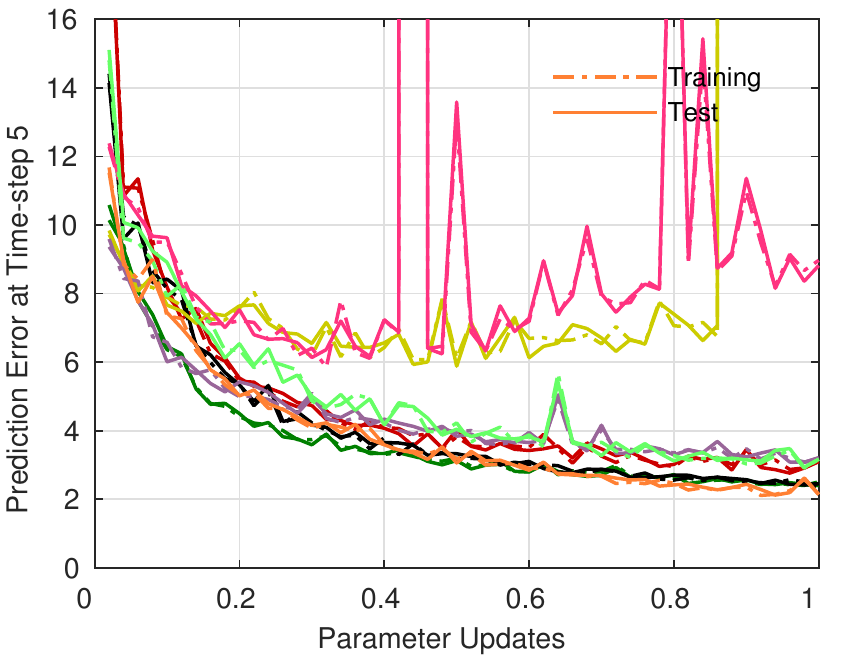}}
\subfigure[]{
\scalebox{0.79}{\includegraphics[]{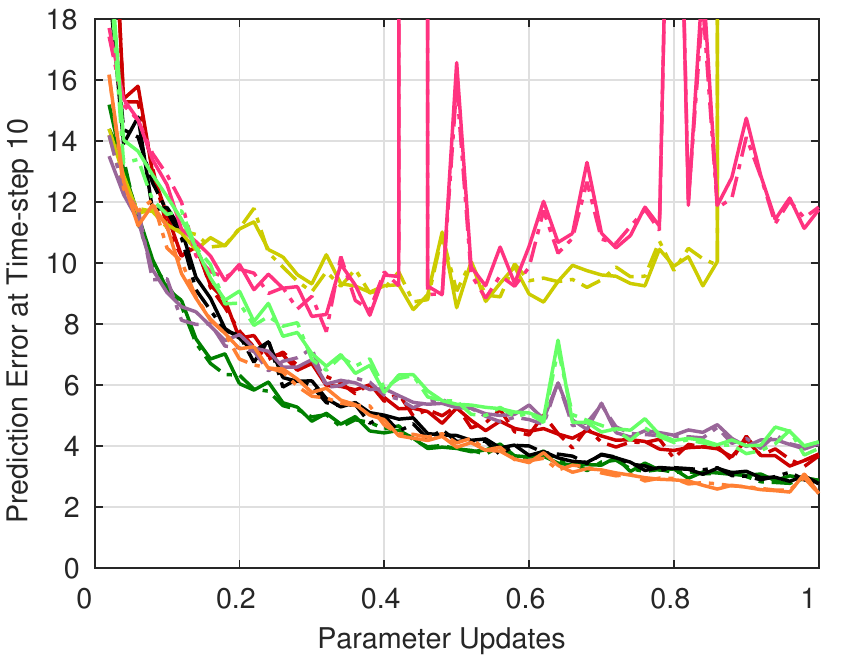}}
%\hskip0.1cm
\scalebox{0.79}{\includegraphics[]{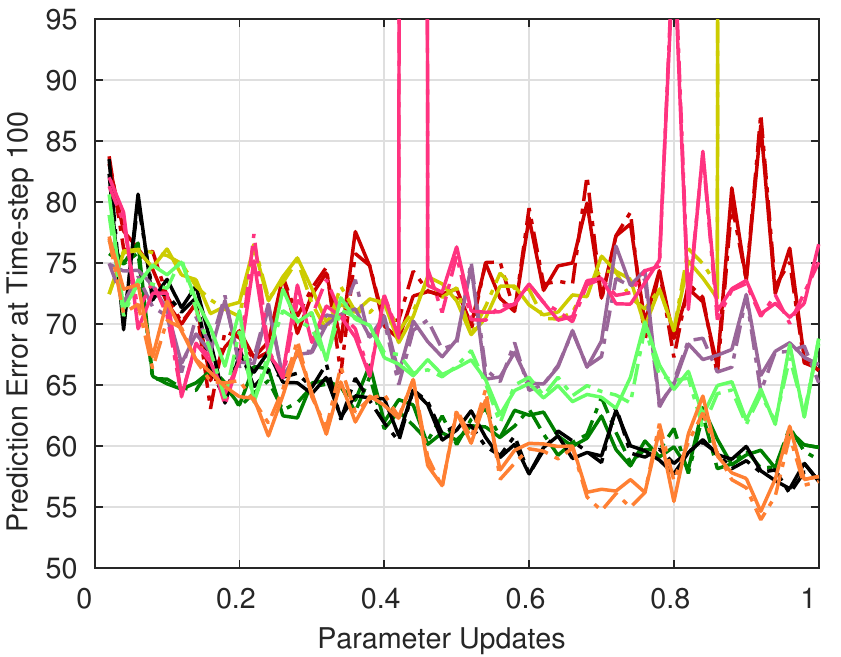}}}
%\vskip-0.3cm
\caption{Prediction error (average over 10,000 sequences) for different  action-dependent state transitions on (a) Bowling and (b) Breakout. Parameter updates are in millions.}
\label{fig:predErrStructBowling-Breakout}
\end{figure}

\begin{figure}[htbp] % Figures obtained with predErrStructAppendix
\vskip-0.5cm
\scalebox{0.79}{\includegraphics[]{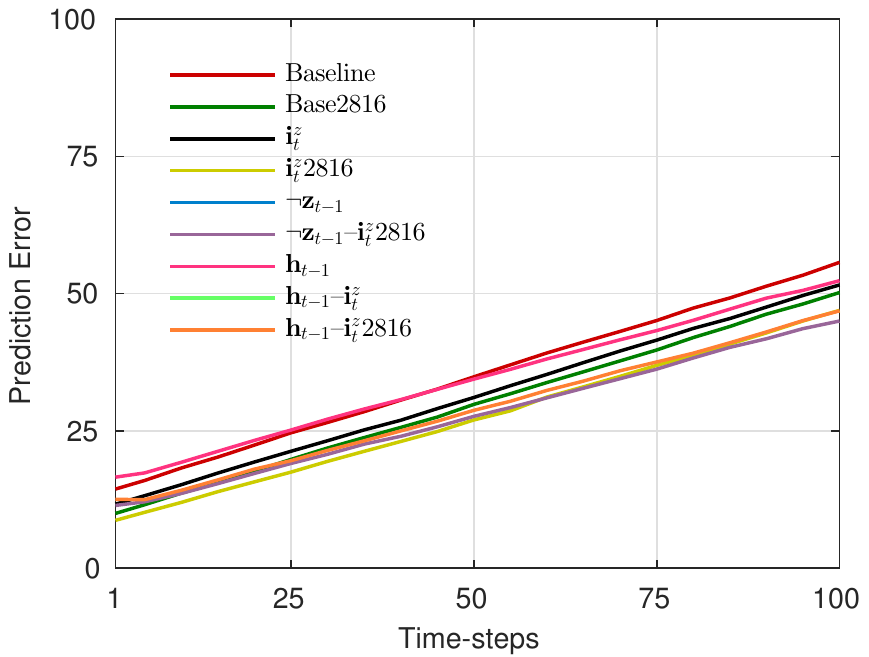}}
%\hskip0.1cm
\scalebox{0.79}{\includegraphics[]{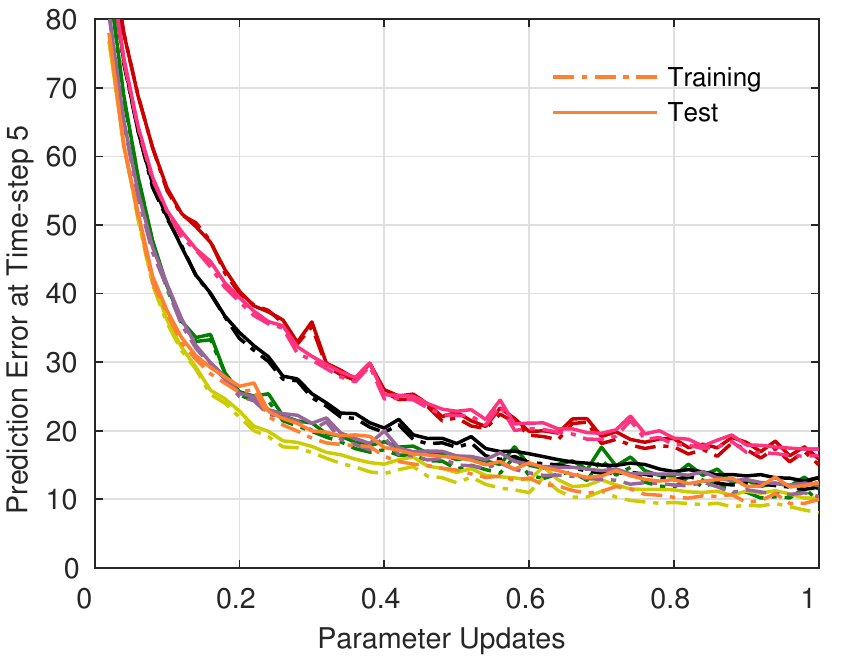}}
\subfigure[]{
\scalebox{0.79}{\includegraphics[]{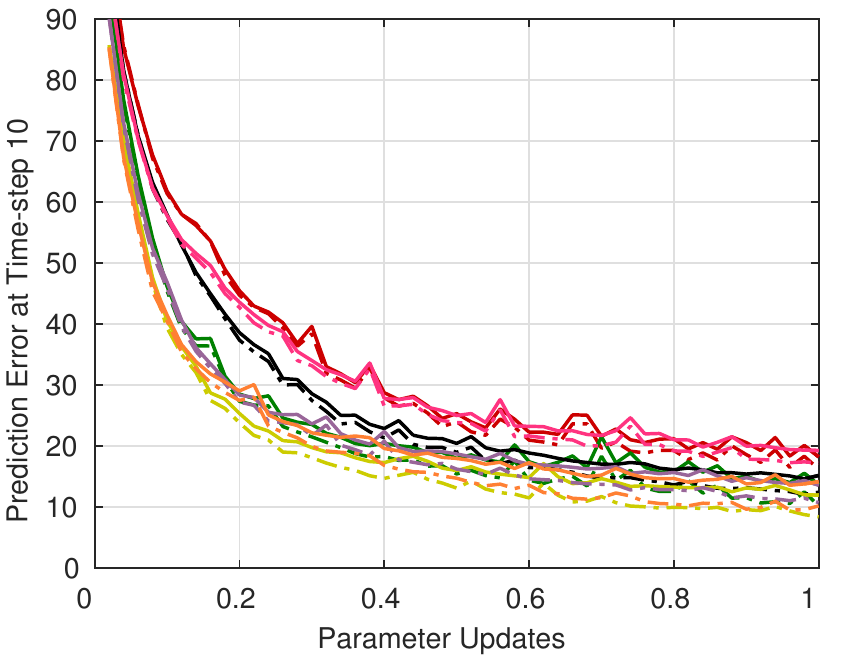}}
%\hskip0.1cm
\scalebox{0.79}{\includegraphics[]{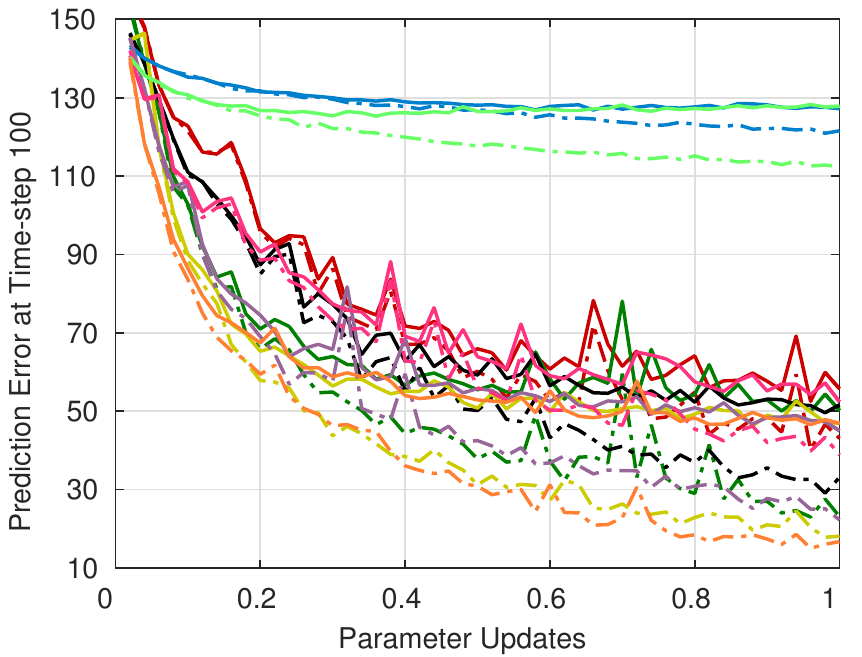}}}
\scalebox{0.79}{\includegraphics[]{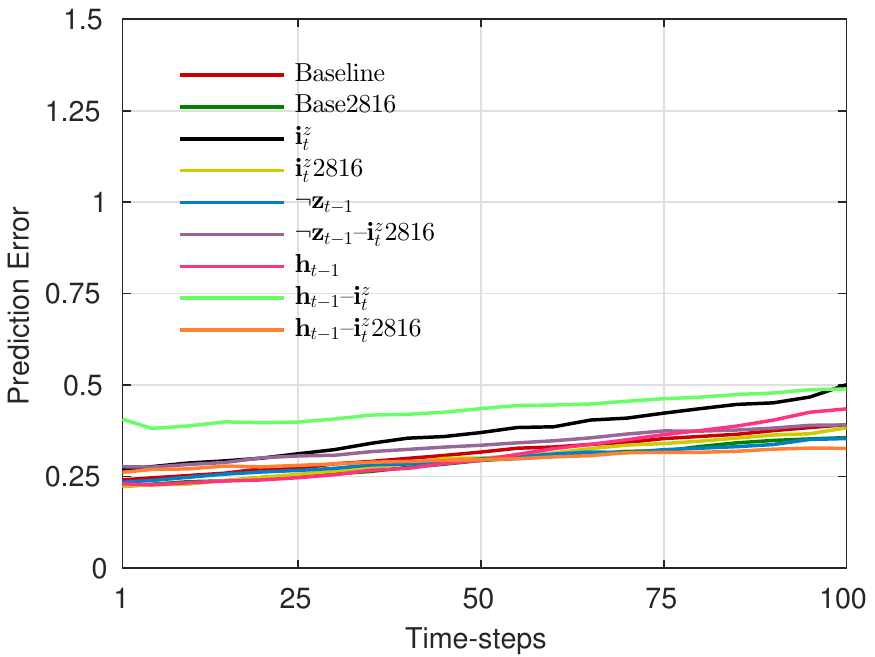}}
%\hskip0.1cm
\scalebox{0.79}{\includegraphics[]{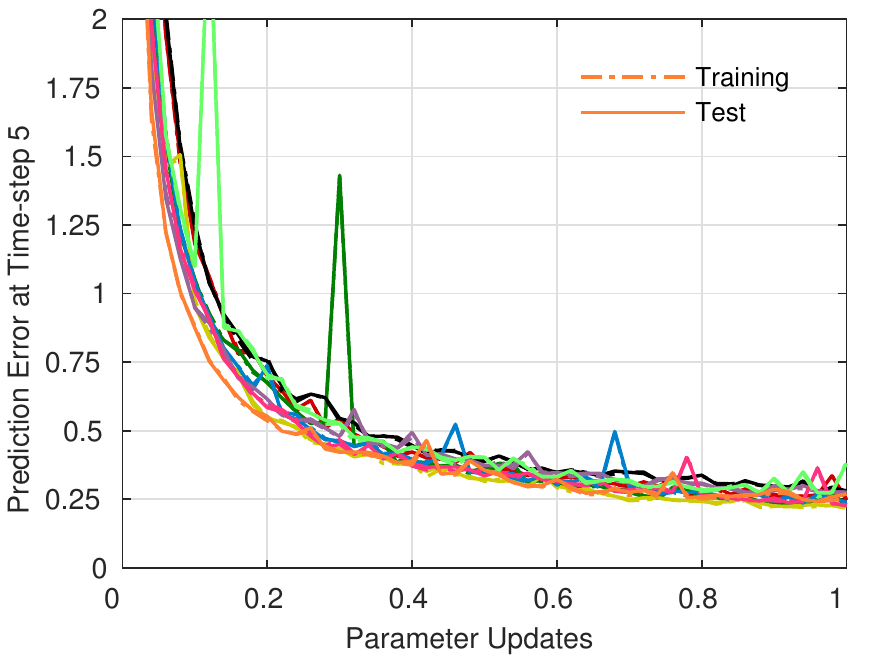}}
\subfigure[]{
\scalebox{0.79}{\includegraphics[]{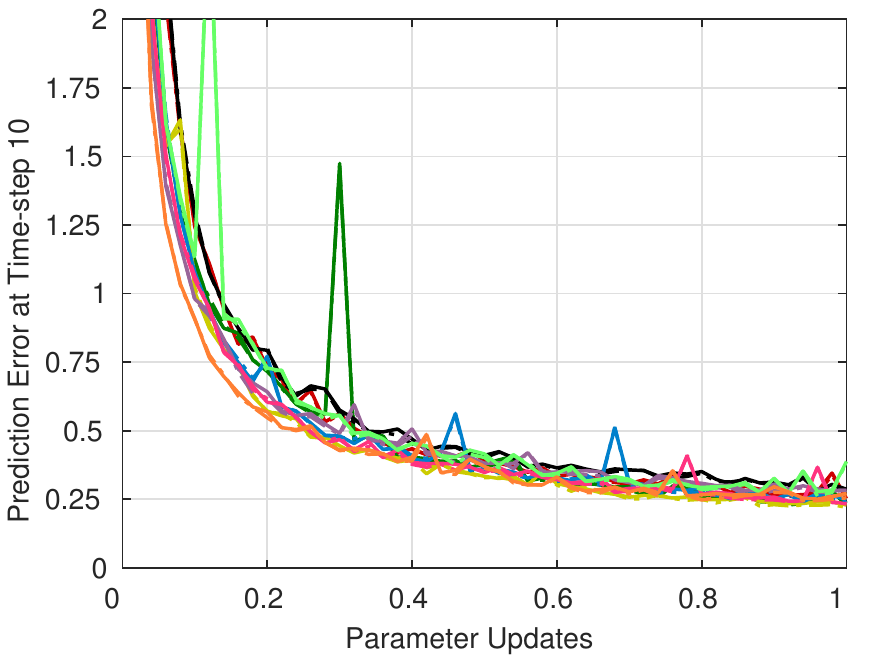}}
%\hskip0.1cm
\scalebox{0.79}{\includegraphics[]{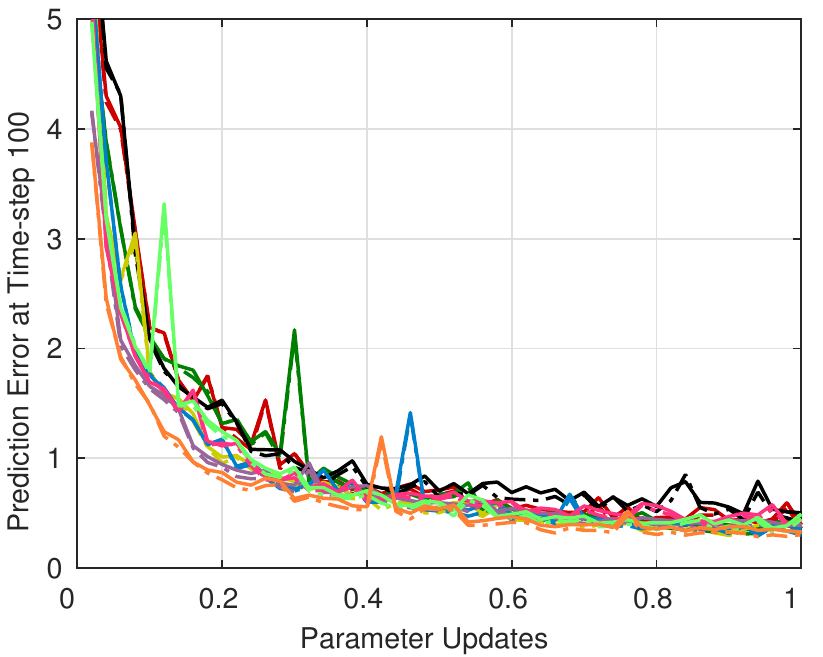}}}
\caption{Prediction error for different action-dependent state transitions on (a) Fishing Derby and (b) Freeway.}
\label{fig:predErrStructFishingDerby-Freeway}
\end{figure}

\begin{figure}[htbp] % Figures obtained with predErrStructAppendix
\vskip-0.5cm
\scalebox{0.79}{\includegraphics[]{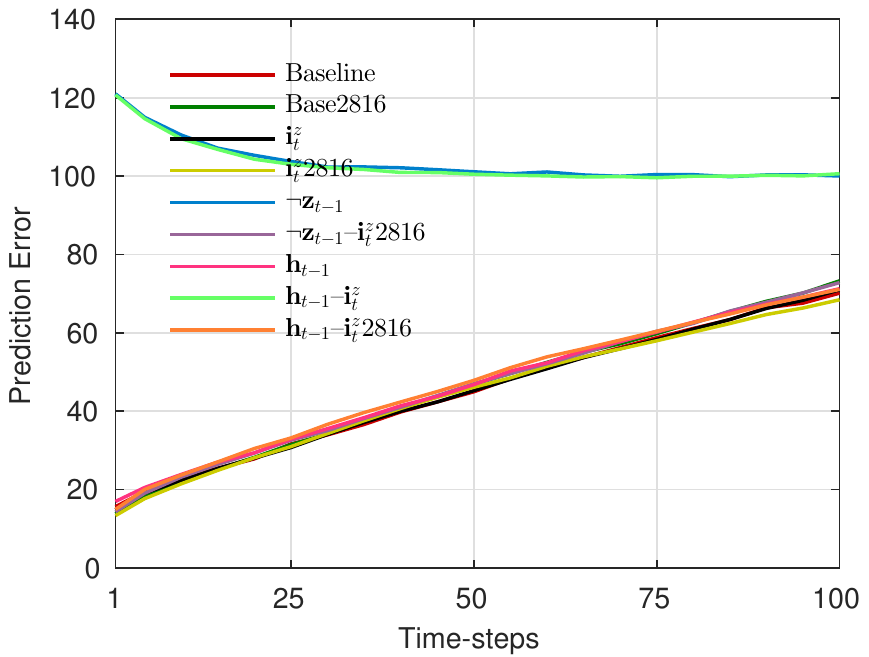}}
%\hskip0.1cm
\scalebox{0.79}{\includegraphics[]{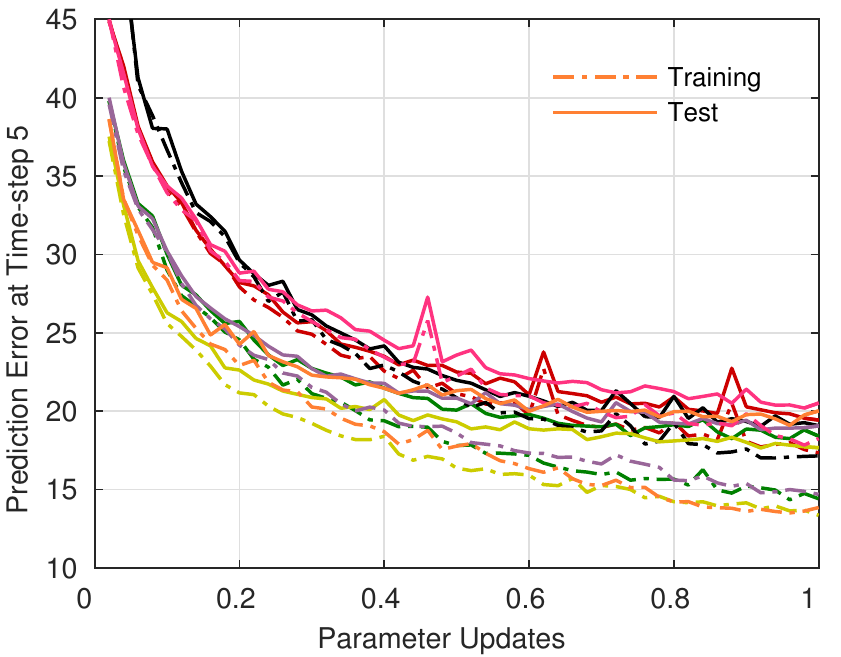}}
\subfigure[]{
\scalebox{0.79}{\includegraphics[]{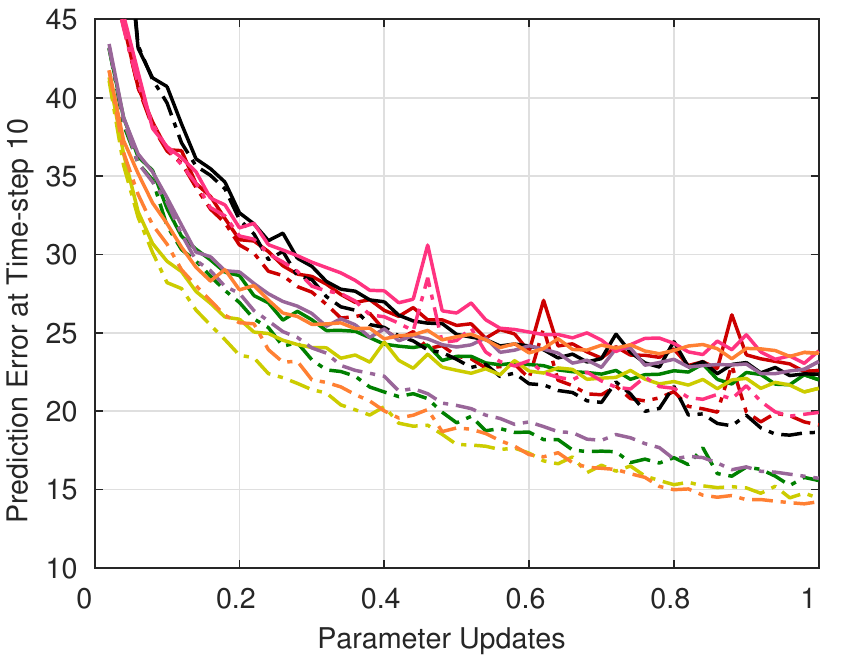}}
%\hskip0.1cm
\scalebox{0.79}{\includegraphics[]{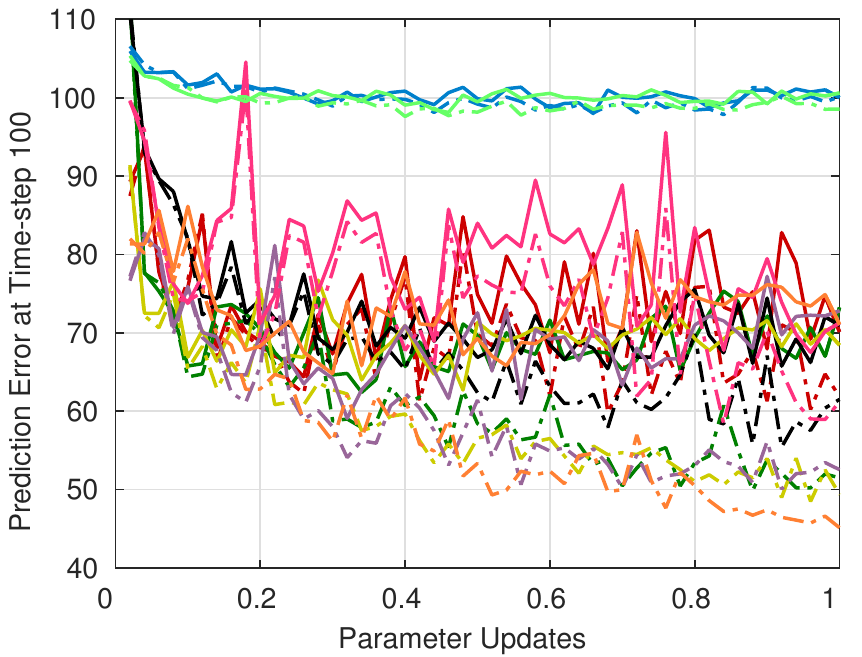}}}
\scalebox{0.79}{\includegraphics[]{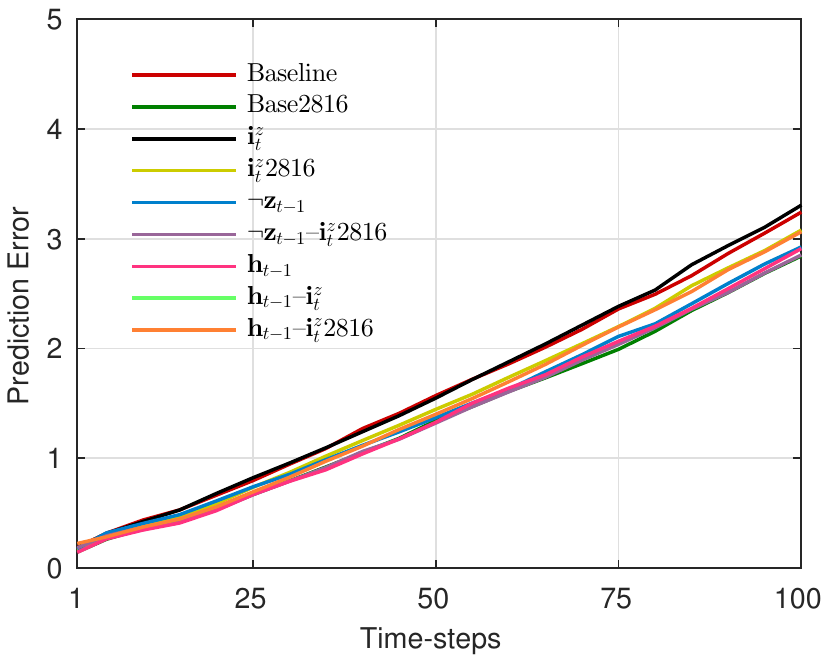}}
%\hskip0.1cm
\scalebox{0.79}{\includegraphics[]{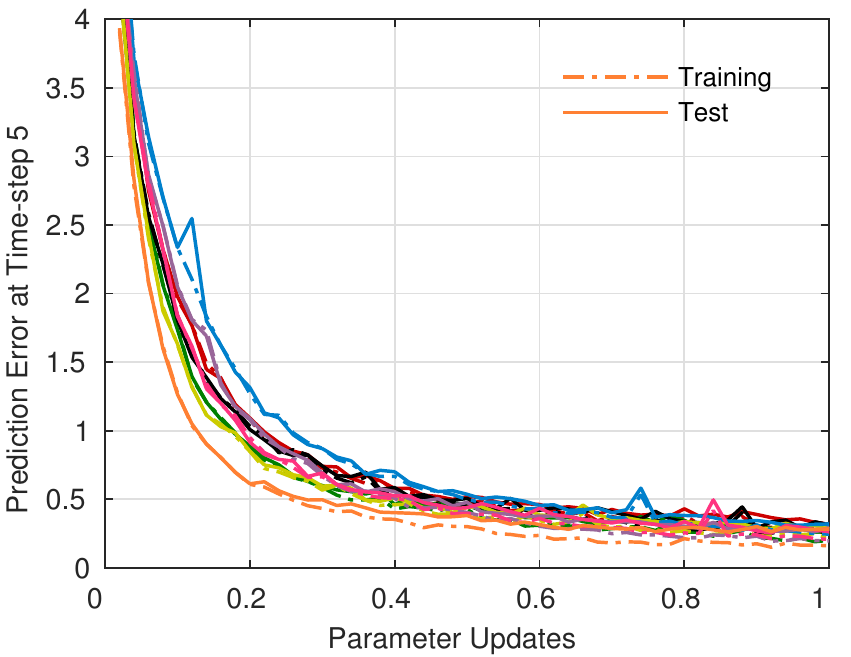}}
\subfigure[]{
\scalebox{0.79}{\includegraphics[]{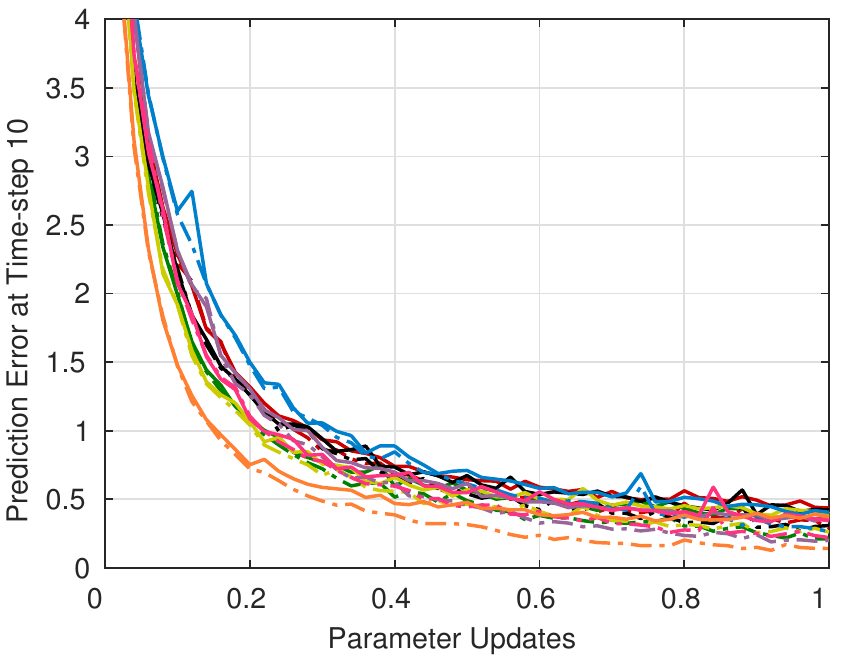}}
%\hskip0.1cm
\scalebox{0.79}{\includegraphics[]{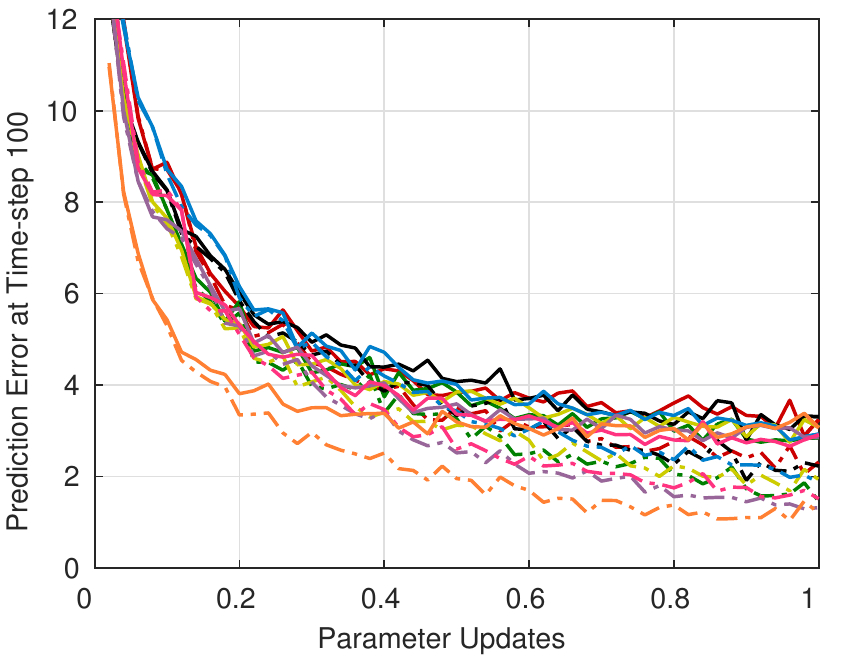}}}
\caption{Prediction error for different action-dependent state transitions on (a) Ms Pacman and (b) Pong.}
\label{fig:predErrStructMsPacman-Pong}
\end{figure}

\begin{figure}[htbp] % Figures obtained with predErrStructAppendix
\vskip-0.5cm
\scalebox{0.79}{\includegraphics[]{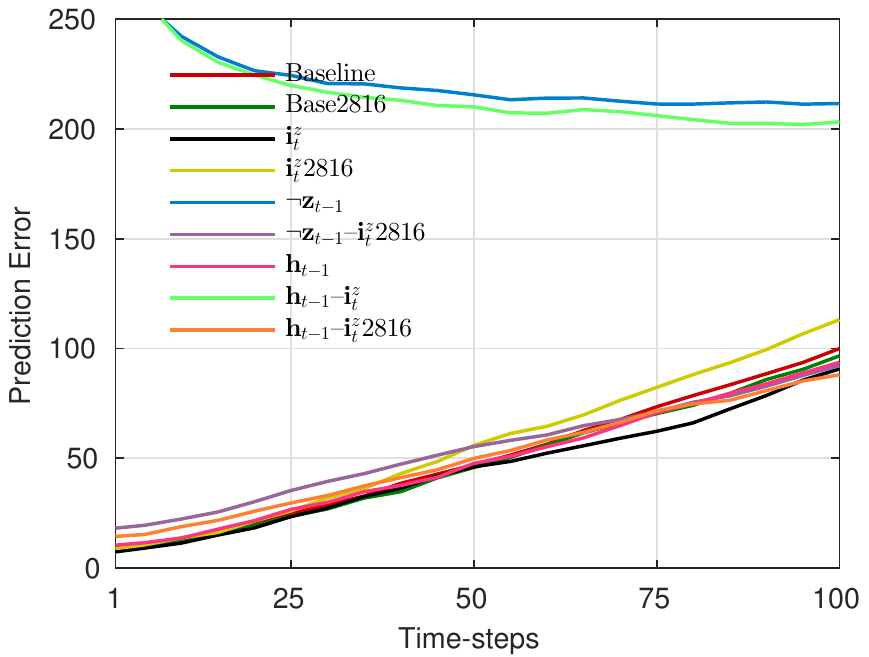}}
%\hskip0.1cm
\scalebox{0.79}{\includegraphics[]{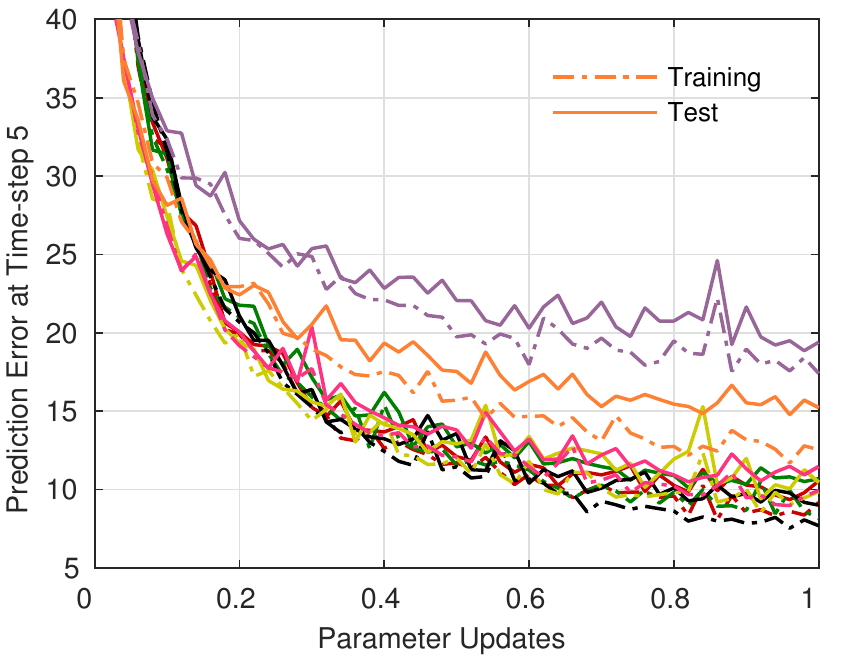}}
\subfigure[]{
\scalebox{0.79}{\includegraphics[]{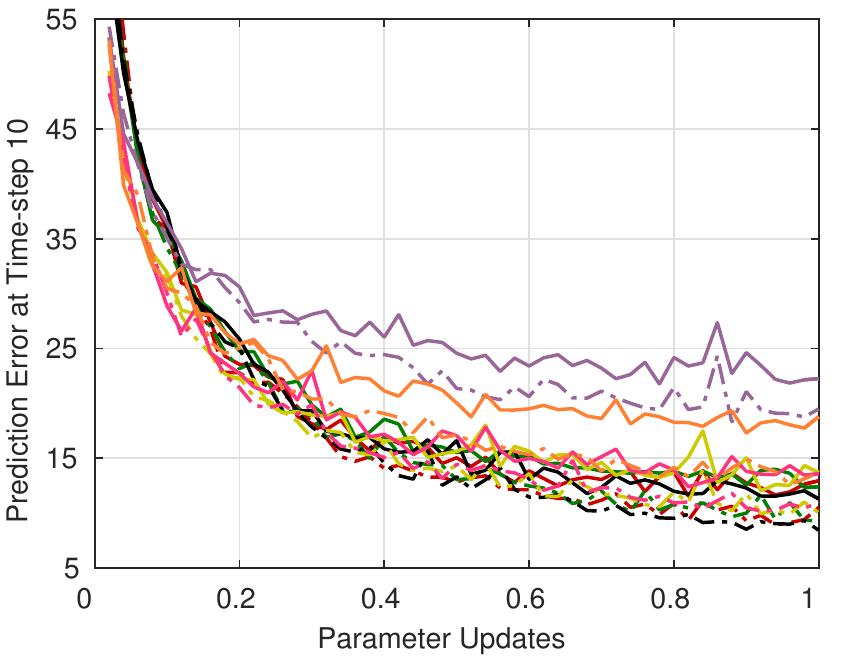}}
%\hskip0.1cm
\scalebox{0.79}{\includegraphics[]{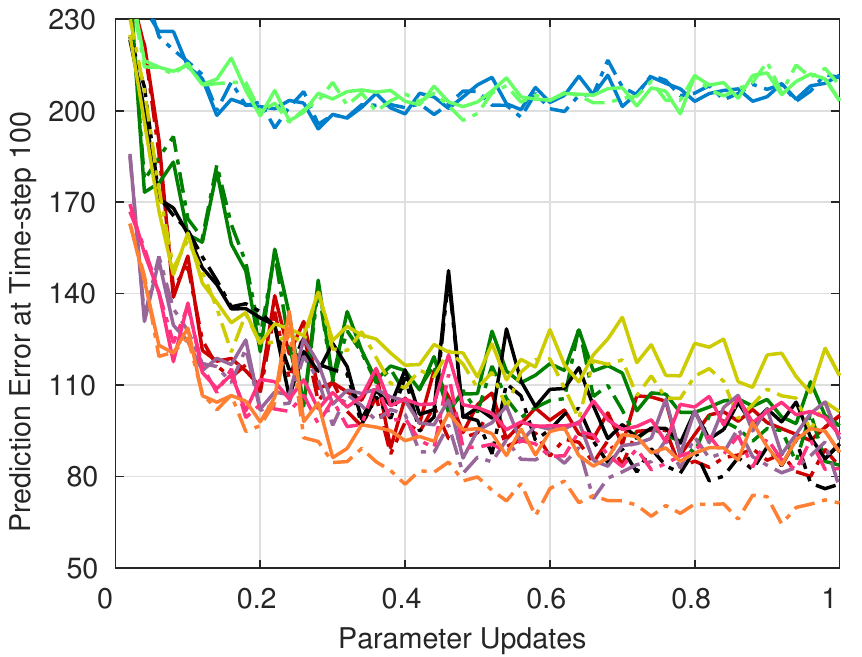}}}
\scalebox{0.79}{\includegraphics[]{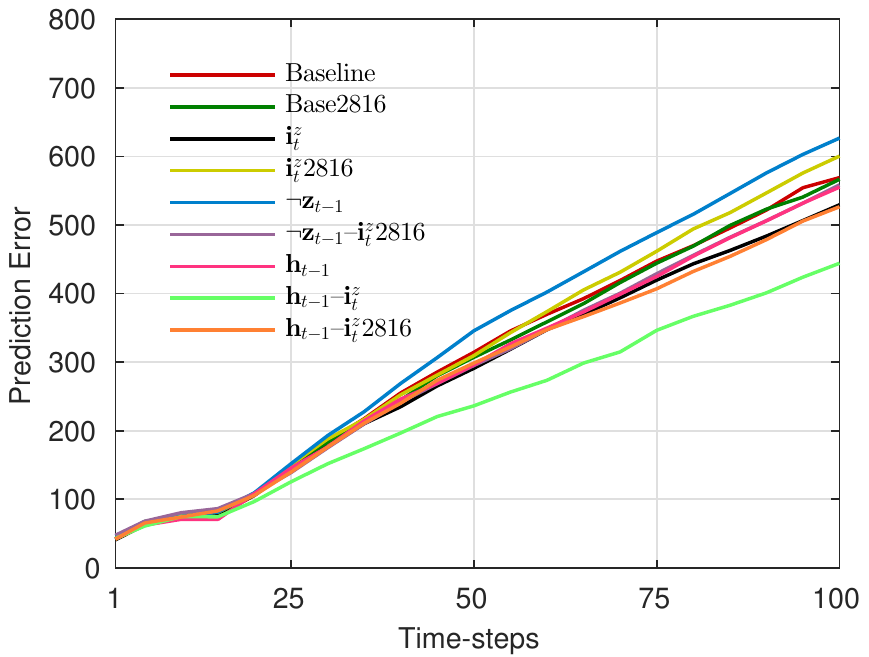}}
%\hskip0.1cm
\scalebox{0.79}{\includegraphics[]{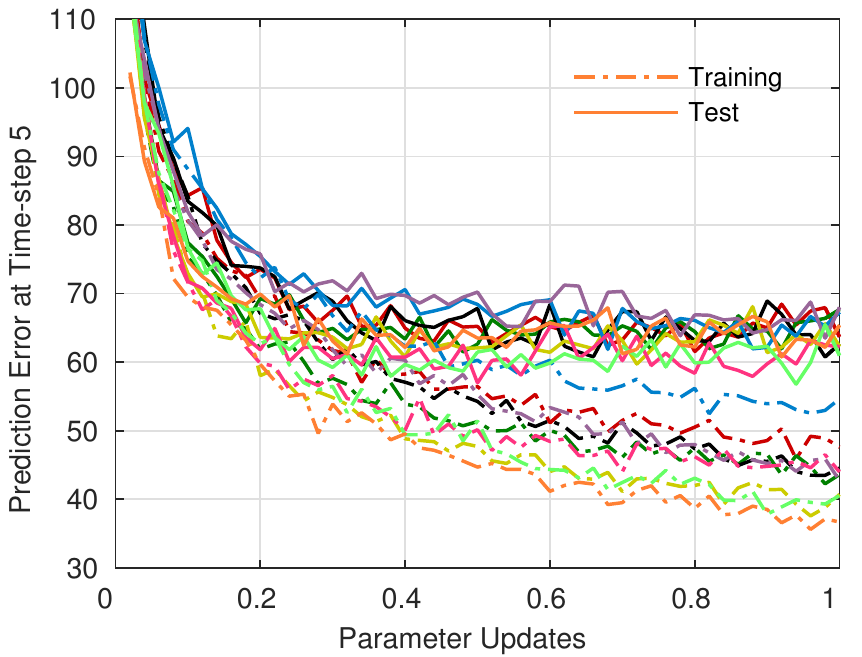}}
\subfigure[]{
\scalebox{0.79}{\includegraphics[]{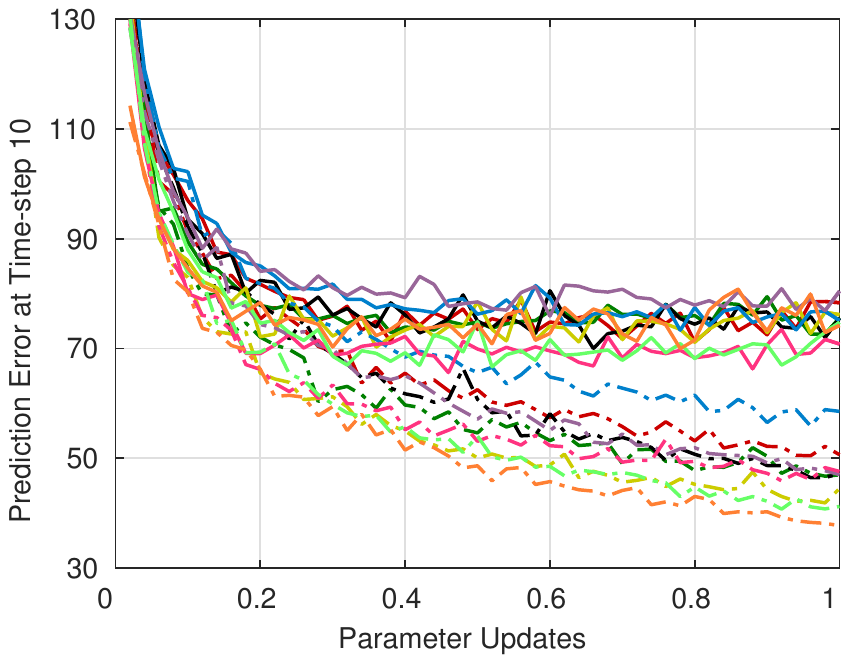}}
%\hskip0.1cm
\scalebox{0.79}{\includegraphics[]{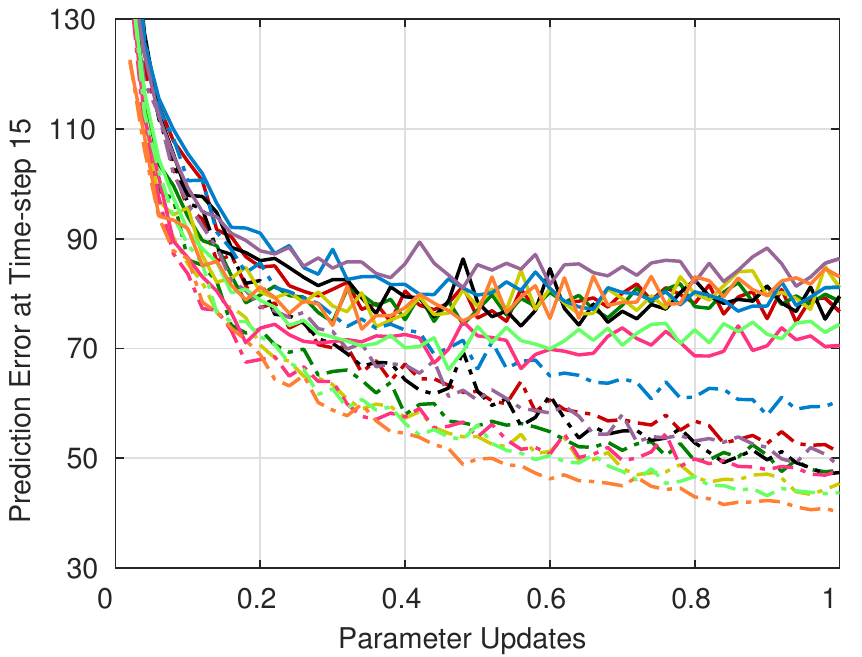}}}
\caption{Prediction error for different action-dependent state transitions on (a) Qbert and (b) Riverraid.}
\label{fig:predErrStructQbert-Riverraid}
\end{figure}

\begin{figure}[htbp] % Figures obtained with predErrStructAppendix
\vskip-0.5cm
\scalebox{0.79}{\includegraphics[]{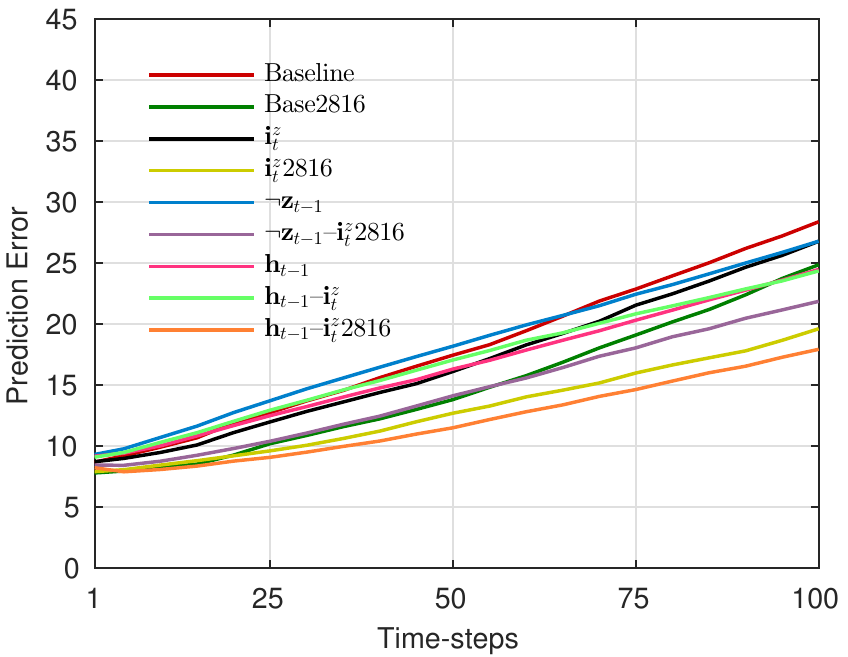}}
%\hskip0.1cm
\scalebox{0.79}{\includegraphics[]{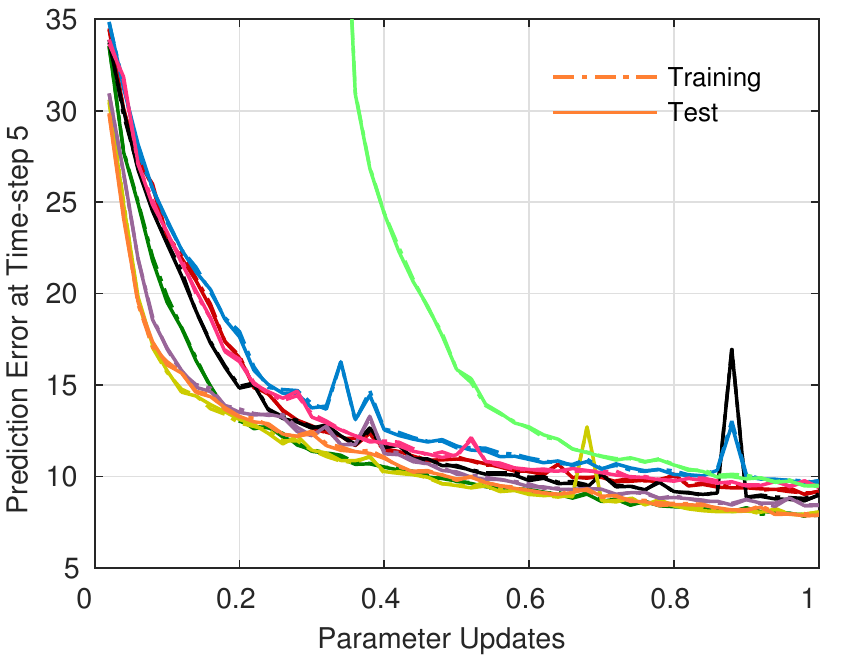}}
\subfigure[]{
\scalebox{0.79}{\includegraphics[]{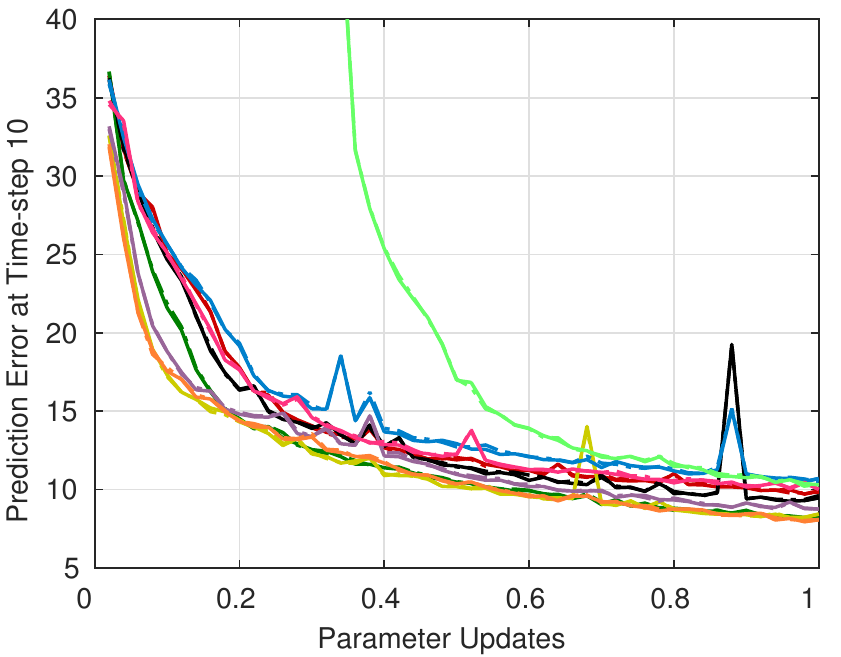}}
%\hskip0.1cm
\scalebox{0.79}{\includegraphics[]{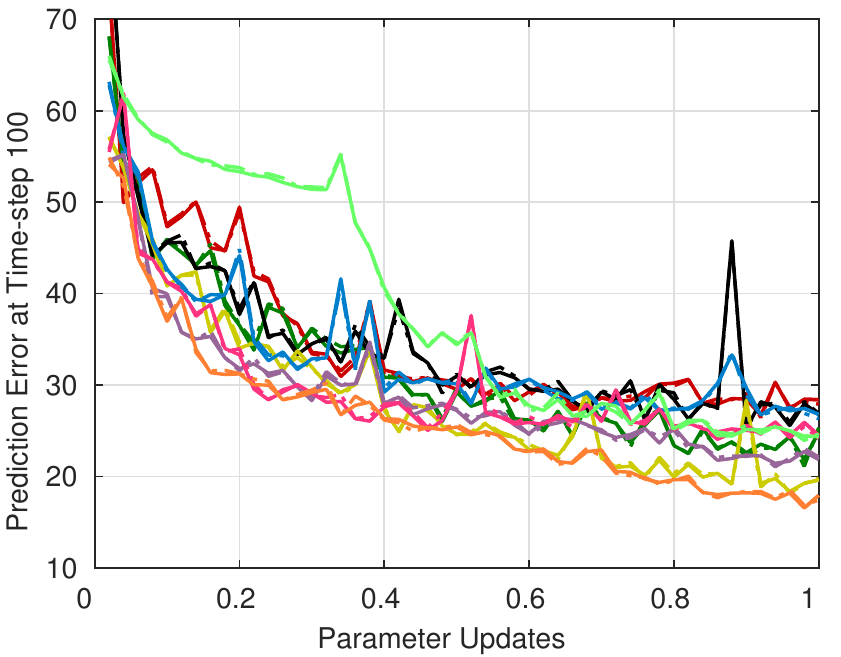}}}
\scalebox{0.79}{\includegraphics[]{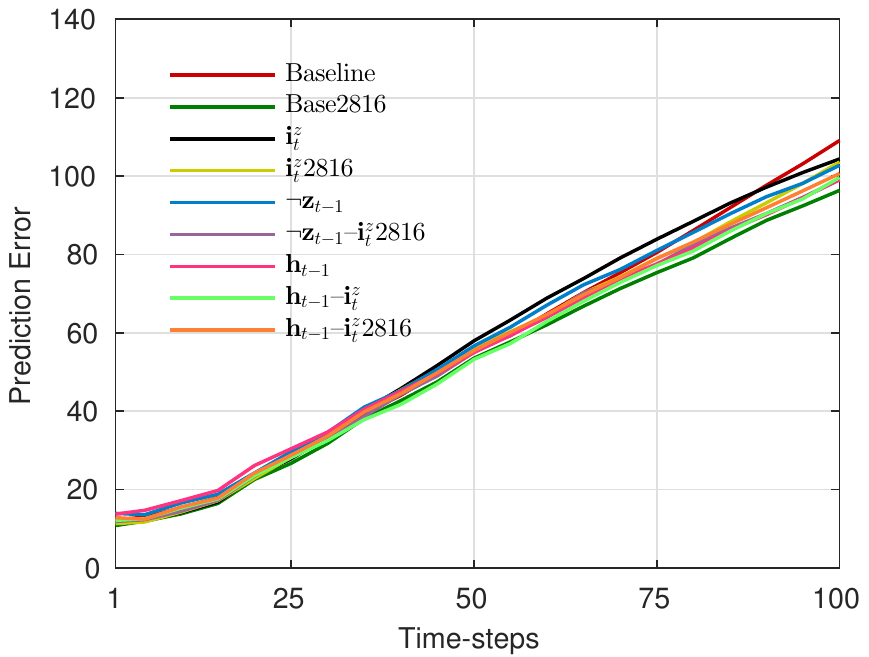}}
%\hskip0.1cm
\scalebox{0.79}{\includegraphics[]{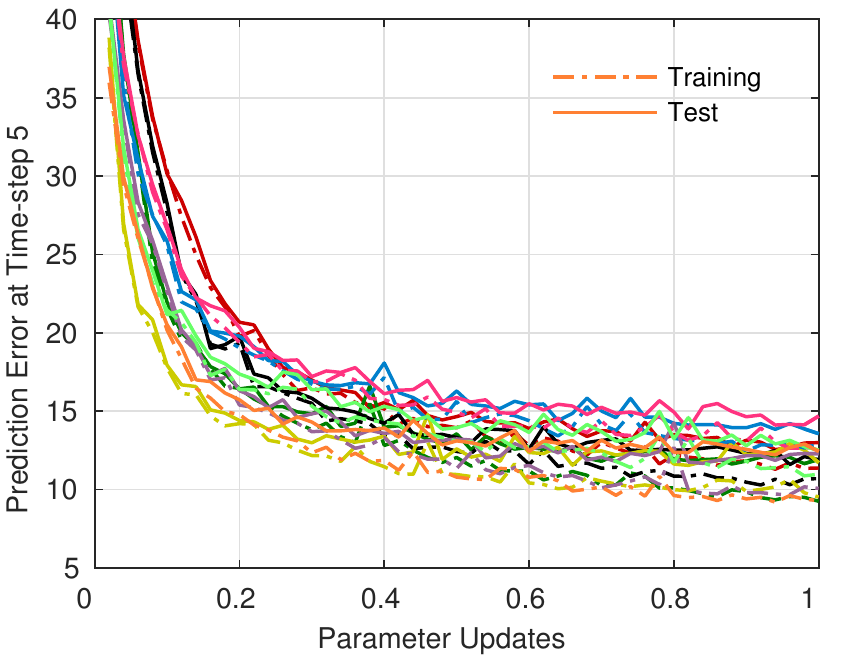}}
\subfigure[]{
\scalebox{0.79}{\includegraphics[]{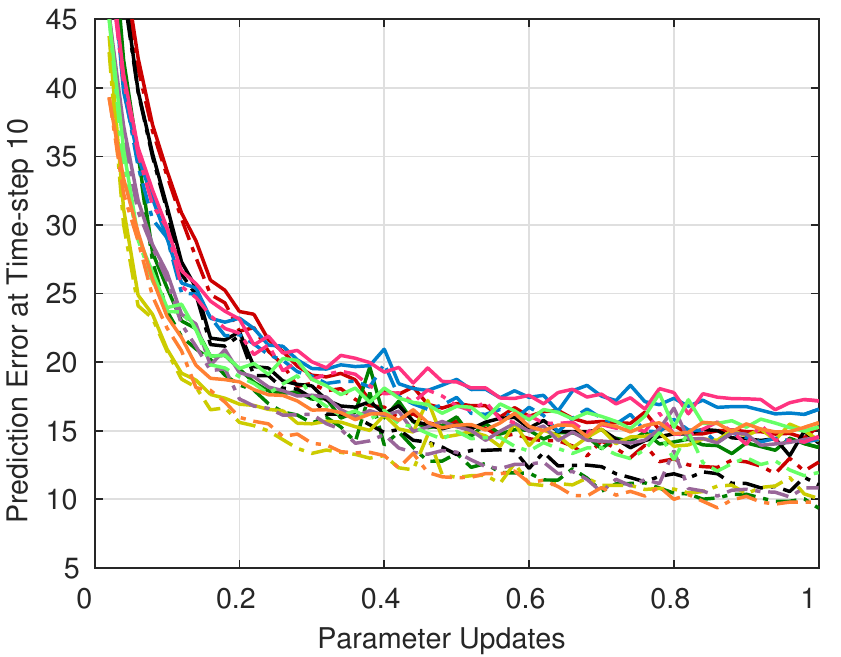}}
%\hskip0.1cm
\scalebox{0.79}{\includegraphics[]{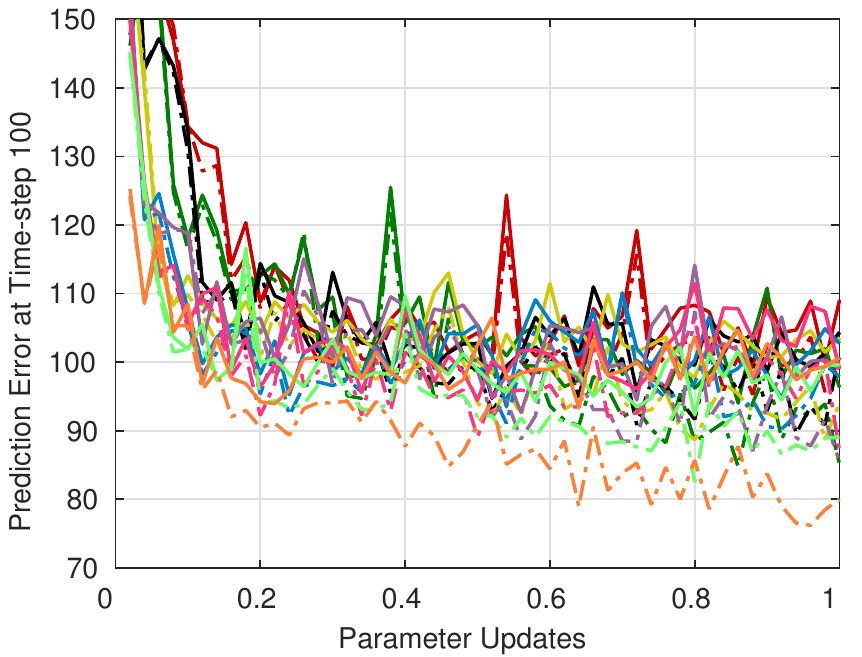}}}
\caption{Prediction error for different action-dependent state transitions on (a) Seaquest and (b) Space Invaders.}
\label{fig:predErrStructSeaquest-SpaceInvaders}
\end{figure}

More specifically, in Figs. \ref{fig:predErrStructBowling-Breakout}-\ref{fig:predErrStructSeaquest-SpaceInvaders} we compare the baseline transition with the following alternatives:
\begin{description}%[leftmargin=*]
% \item Remove the RReLU after the last convolution. %ATA5e6Tr1e6Tse0.2MPfalsebS16nH10nHid1024nLHid1024nF2048mask2-15-1.1-15-1SEnF64lR1e-5MNoRReLULC
\item[$\bullet$ Base2816:] The vectors $\vh_{t-1}$ and $\ha_t$ have the same dimension as $\vz_{t-1}$, namely 2816. %  ATA5e6Tr1e6Tse0.2MPfalsebS16nH10nHid2816nLHid2816nF2816mask2-15-1.1-15-1SEnF64lR1e-5M
This model has around 80M parameters. \\
% \item We also considered removing the linear projection at the beginning of the decoding. % ATA5e6Tr1e6Tse0.2MPfalsebS16nH10nHid2816nLHid2816nF2816mask2-15-1.1-15-1SEnF64lR1e-5MNoLD
% \item We also considered removing the $L^{*s}$ having tanh as nonlinearity in the last convolution. %ATA5e6Tr1e6Tse0.2MPfalsebS16nH10nHid2816nLHid2816nF2816mask2-15-1.1-15-1SEnF64lR1e-5MTanhX
\item[$\bullet$ $\vi^z_t$ and $\vi^z_t$2816:]
Have a separate gating for $\vz_{t-1}$ in the cell update, \ie %ATA5e6Tr1e6Tse0.2MPfalsebS16nH10nHid2816nLHid2816nF2816mask2-15-1.1-15-1SEnF64lR1e-5MSXL or 2816 %ATA5e6Tr1e6Tse0.2MPfalsebS16nH10nHid1024nLHid1024nF2048mask2-15-1.1-15-1SEnF64lR1e-5MSXL 
\begin{align*}
\vc_t = \vf_t\otimes \vc_{t-1}+\vi_t\otimes \textrm{tanh}(\vW^{cv}\ha_{t})+\vi^z_t\otimes \textrm{tanh}(\vW^{cz}\vz_{t-1})\,.
\end{align*}
This model has around 30 million parameters. We also considered removing the linear projection of $\vz_{t-1}$, \ie %ATA5e6Tr1e6Tse0.2MPfalsebS16nH10nHid2816nLHid2816nF2816mask2-15-1.1-15-1SEnF64lR1e-5MSX 
\begin{align*}
\vc_t = \vf_t\otimes \vc_{t-1}+\vi_t\otimes \textrm{tanh}(\vW^{cv}\ha_{t})+\vi^z_t\otimes \textrm{tanh}(\vz_{t-1})\,,
\end{align*}
without RReLU after the last convolution and with vectors $\vh_{t-1}$ and $\ha_t$ of dimensionality 2816.
This model has around 88M parameters.\\
%\item We also considered removing the linear projection before the decoding in the first case.  %ATA5e6Tr1e6Tse0.2MPfalsebS16nH10nHid2816nLHid2816nF2816mask2-15-1.1-15-1SEnF64lR1e-5MSXNoLD
%\item We also considered using the raw encoded frame in the gates for the fist case. %ATA5e6Tr1e6Tse0.2MPfalsebS16nH10nHid2816nLHid2816nF2816mask2-15-1.1-15-1SEnF64lR1e-5MRXGSX
%\item To discuss ATA5e6Tr1e6Tse0.2MPfalsebS16nH10nHid1024nLHid1024nF2048mask2-15-1.1-15-1SEnF64lR1e-5MNoXHGIXSI
\item[$\bullet$ $\lnot \vz_{t-1}$ and $\lnot \vz_{t-1}$--$\vi^z_t$2816:]
Remove $\vz_{t-1}$ in the gate updates, \ie
\begin{align*}
\vi_t = {\sigma}(\vW^{iv}\ha_{t})\,, \hskip0.3cm \vf_t = {\sigma}(\vW^{fv}\ha_t)\,,  \hskip0.3cm \vo_t = {\sigma}(\vW^{ov}\ha_t)\,,
\end{align*}
with one of the following cell updates  
\begin{align*}
%ATA5e6Tr1e6Tse0.2MPfalsebS16nH10nHid1024nLHid1024nF2048mask2-15-1.1-15-1SEnF64lR1e-5MNoXG 
&\vc_t = \vf_t\otimes \vc_{t-1}+\vi_t\otimes \textrm{tanh}(\vW^{cv}\ha_{t}+\vW^{cz}\vz_{t-1})\,,  \hskip0.1cm \textrm{17M parameters}\,,\\ 
%ATA5e6Tr1e6Tse0.2MPfalsebS16nH10nHid2816nLHid2816nF2816mask2-15-1.1-15-1SEnF64lR1e-5MNoXGSX 
&\vc_t = \vf_t\otimes \vc_{t-1}+\vi_t\otimes \textrm{tanh}(\vW^{cv}\ha_{t})+\vi^z_t\otimes \textrm{tanh}(\vz_{t-1})\,, \hskip0.1cm \textrm{56M parameters}\,.
\end{align*}
\\
\item[$\bullet$ $\vh_{t-1}$, $\vh_{t-1}$--$\vi^z_t$, and $\vh_{t-1}$--$\vi^z_t$2816:]
Substitute $\vz_{t-1}$ with $\vh_{t-1}$ in the gate updates, \ie
\begin{align*}
&\vi_t = {\sigma}(\vW^{iv}\ha_{t}+\vW^{ih}\vh_{t-1})\,, \hskip0.2cm \vf_t = {\sigma}(\vW^{fv}\ha_t+\vW^{fh}\vh_{t-1})\,, \\
& \vo_t = {\sigma}(\vW^{ov}\ha_t+\vW^{oh}\vh_{t-1})\,,
\end{align*}
with one of the following cell updates % ATA5e6Tr1e6Tse0.2MPfalsebS16nH10nHid1024nLHid1024nF2048mask2-15-1.1-15-1SEnF64lR1e-5MNoXHGH 
\begin{align*}
&\vc_t = \vf_t\otimes \vc_{t-1}+\vi_t\otimes \textrm{tanh}(\vW^{cv}\ha_{t}+\vW^{ch}\vh_{t-1}+\vW^{cz}\vz_{t-1})\,, \hskip0.1cm \textrm{21M parameters}\,,\\ %ATA5e6Tr1e6Tse0.2MPfalsebS16nH10nHid1024nLHid1024nF2048mask2-15-1.1-15-1SEnF64lR1e-5MNoXHGSXL
&\vc_t = \vf_t\otimes \vc_{t-1}+\vi_t\otimes \textrm{tanh}(\vW^{cv}\ha_{t}+\vW^{ch}\vh_{t-1})+\vi^z_t\otimes \textrm{tanh}(\vW^{cz}\vz_{t-1})\,, \hskip0.1cm \textrm{24M parameters}\,,\\ %ATA5e6Tr1e6Tse0.2MPfalsebS16nH10nHid2816nLHid2816nF2816mask2-15-1.1-15-1SEnF64lR1e-5MNoXHGHTanhX
%\begin{align*}
%\textrm{Cell update: } \vc_t = f_t\otimes \vc_{t-1}+i_t\otimes \textrm{tanh}(L^{cv}(\ha_{t})+\textrm{tanh}(\vs_{t-1}))\,,
%\end{align*}
%(this model has around 80 million parameters) 
&\vc_t = \vf_t\otimes \vc_{t-1}+\vi_t\otimes \textrm{tanh}(\vW^{cv}\ha_{t}+\vW^{ch}\vh_{t-1})+\vi^z_t\otimes \textrm{tanh}(\vz_{t-1})\,, \hskip0.1cm \textrm{95M parameters}\,. %ATA5e6Tr1e6Tse0.2MPfalsebS16nH10nHid2816nLHid2816nF2816mask2-15-1.1-15-1SEnF64lR1e-5MNoXHGSX
\end{align*}
%We also considered using 16 filters in the final convolution - first deconvolution to reduce the dimensionality of $\vh_{t-1}$ to 1408, %ATA5e6Tr1e6Tse0.2MPfalsebS16nH10nHid1408nLHid1408nF1408mask2-15-1.1-15-1SEnF64lR1e-5MNoXHGSX
%We also considered removing the linear projection at the beginning of the decoding.  %ATA5e6Tr1e6Tse0.2MPfalsebS16nH10nHid2816nLHid2816nF2816mask2-15-1.1-15-1SEnF64lR1e-5MNoXHGSXNoLD 
\end{description}
As we can see from the figures, there is no other transition that is clearly preferable to the baseline,
with the exception of Fishing Derby, for which transitions with 2816 hidden dimensionality perform better and converge earlier in terms of number of parameter updates.

\begin{figure}[htbp] % Figures obtained with predErrConvStructAppendix
\vskip-0.5cm
\scalebox{0.79}{\includegraphics[]{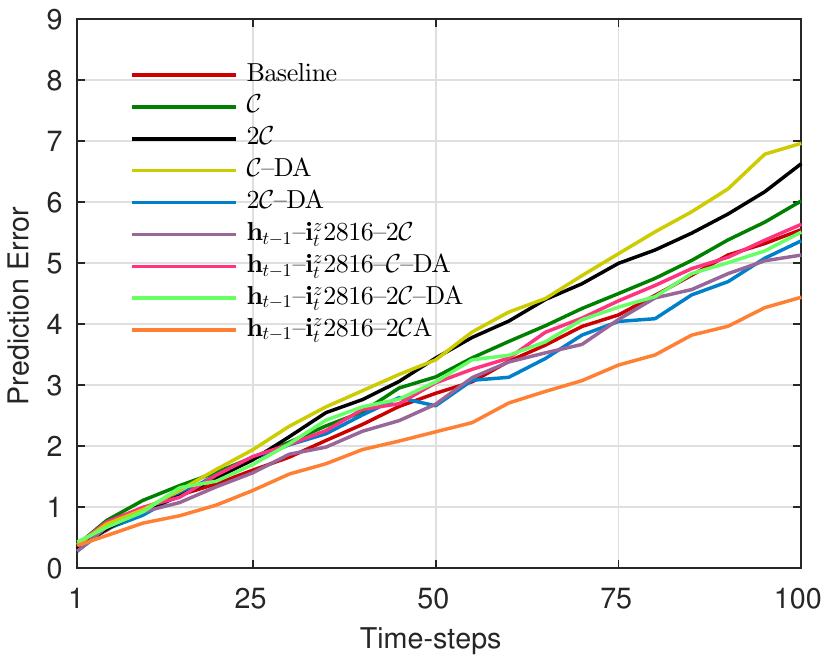}}
%\hskip0.1cm
\scalebox{0.79}{\includegraphics[]{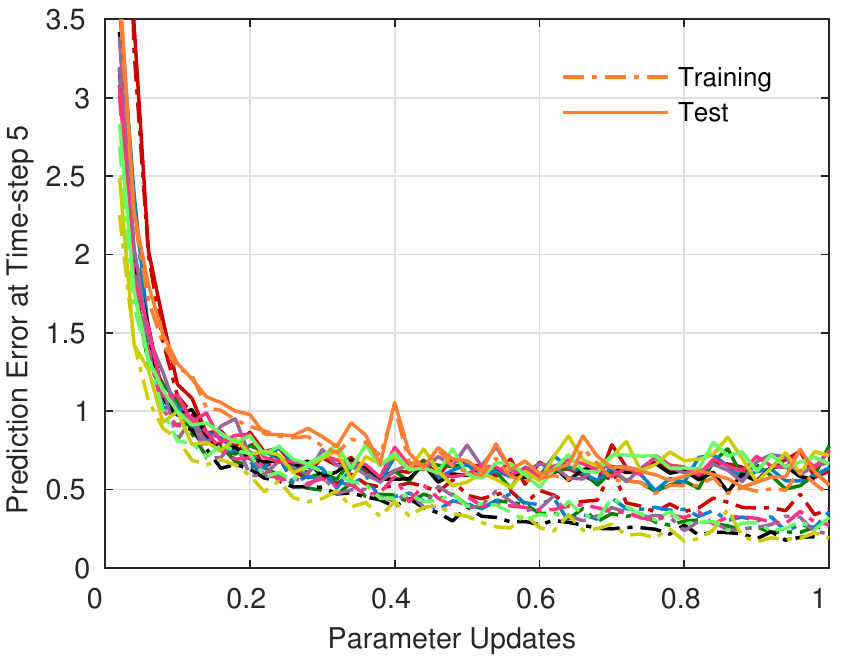}}\\
\subfigure[]{\scalebox{0.79}{\includegraphics[]{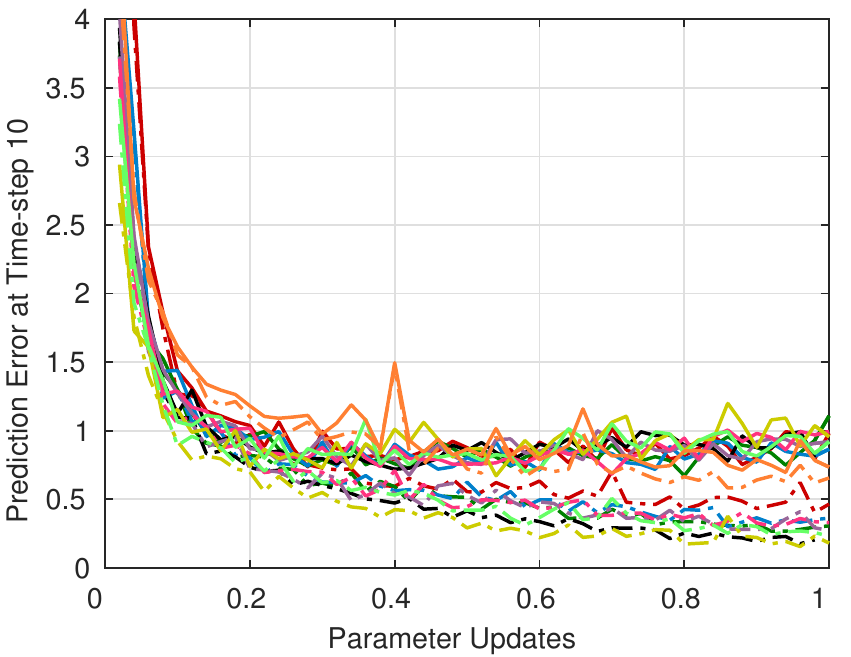}}
%\hskip0.1cm
\scalebox{0.79}{\includegraphics[]{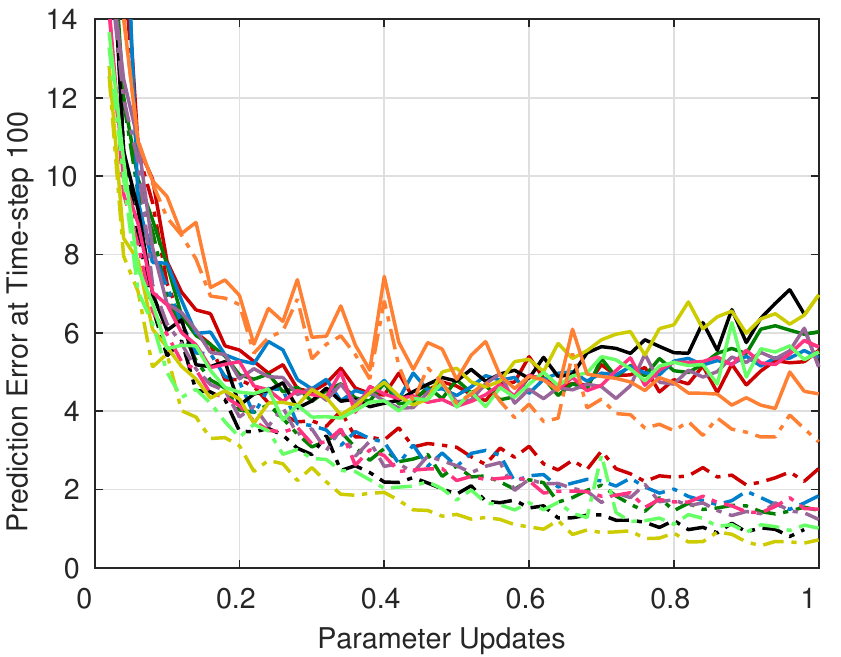}}}
\scalebox{0.79}{\includegraphics[]{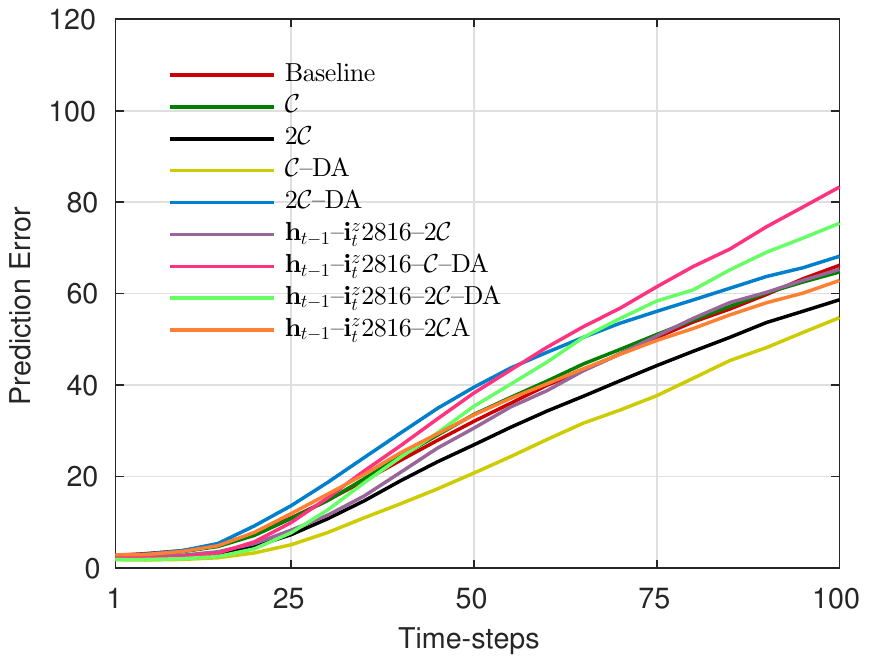}}
%\hskip0.1cm
\scalebox{0.79}{\includegraphics[]{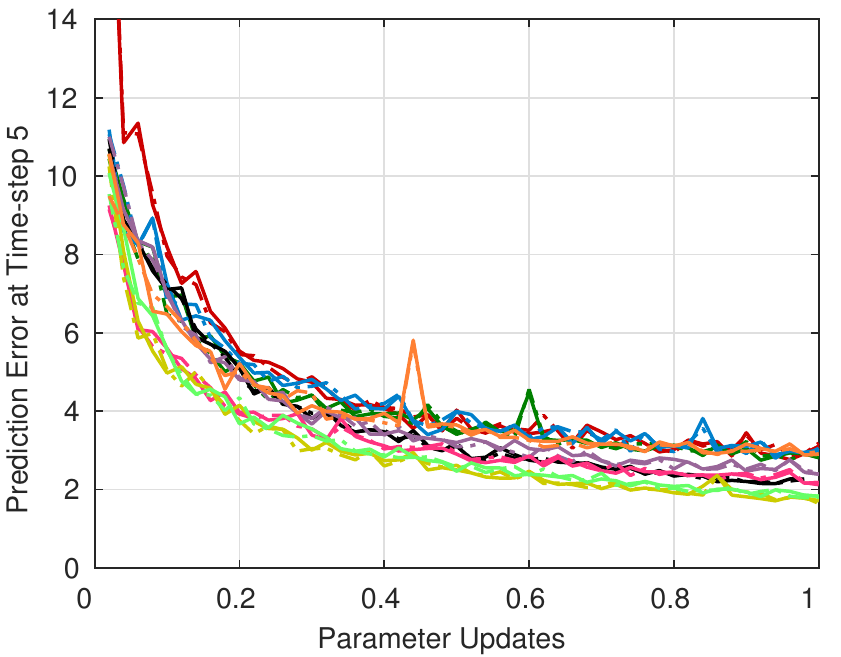}}
\subfigure[]{
\scalebox{0.79}{\includegraphics[]{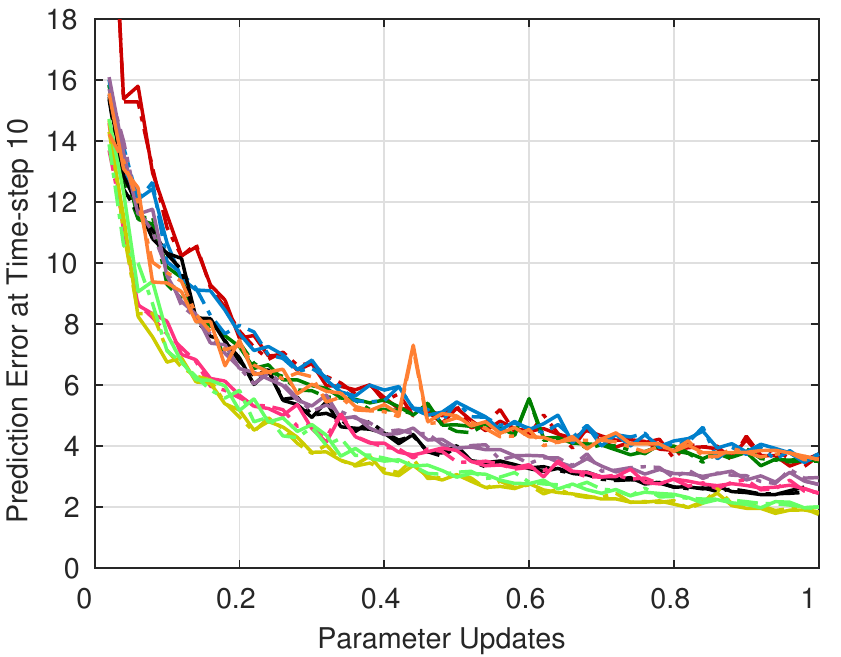}}
%\hskip0.1cm
\scalebox{0.79}{\includegraphics[]{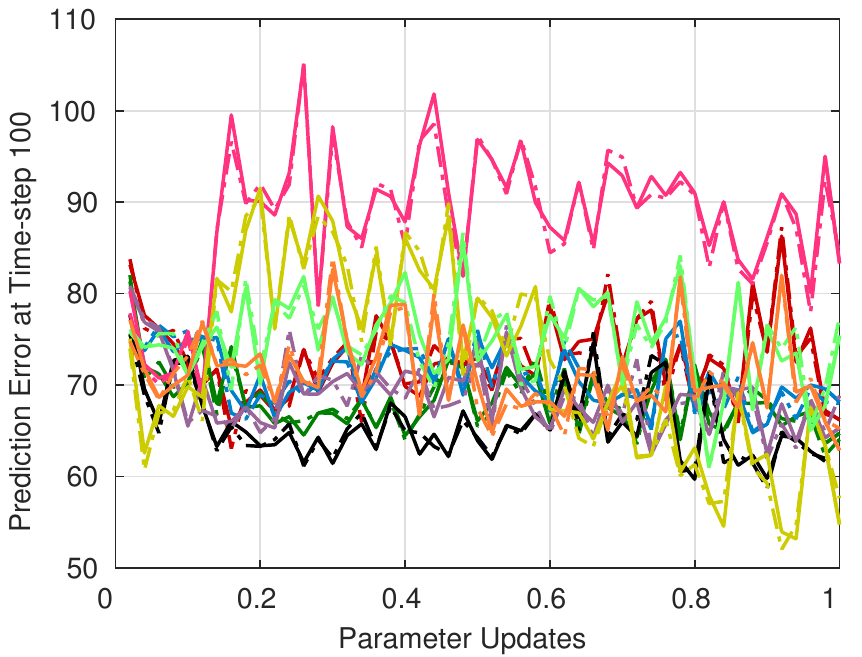}}}
%\vskip-0.3cm
\caption{Prediction error (average over 10,000 sequences) for different convolutional action-dependent state transitions on (a) Bowling and (b) Breakout. Parameter updates are in millions.}
\label{fig:predErrConvStructBowling-Breakout}
\end{figure}

\begin{figure}[htbp] % Figures obtained with predErrConvStructAppendix
\vskip-0.5cm
\scalebox{0.79}{\includegraphics[]{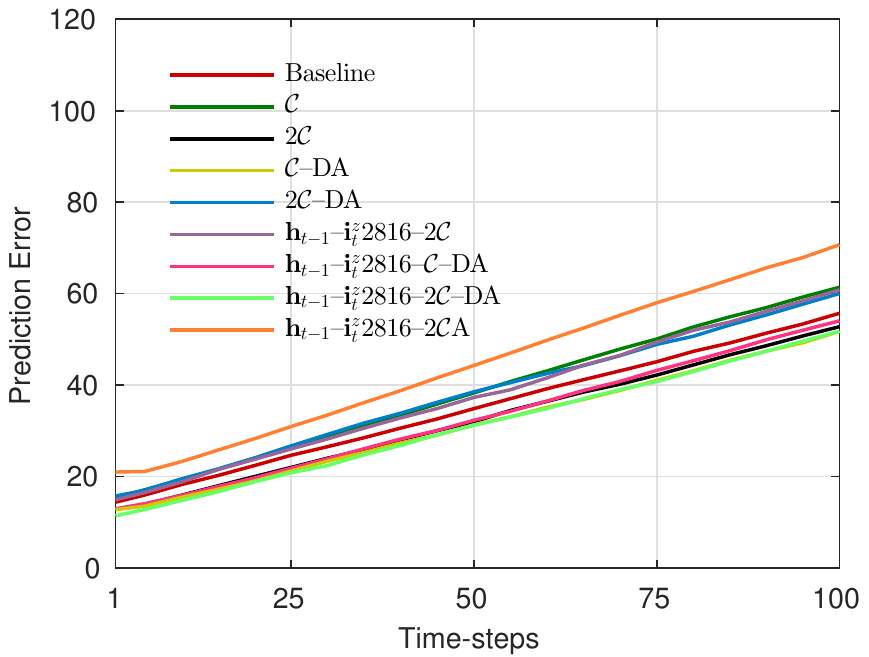}}
%\hskip0.1cm
\scalebox{0.79}{\includegraphics[]{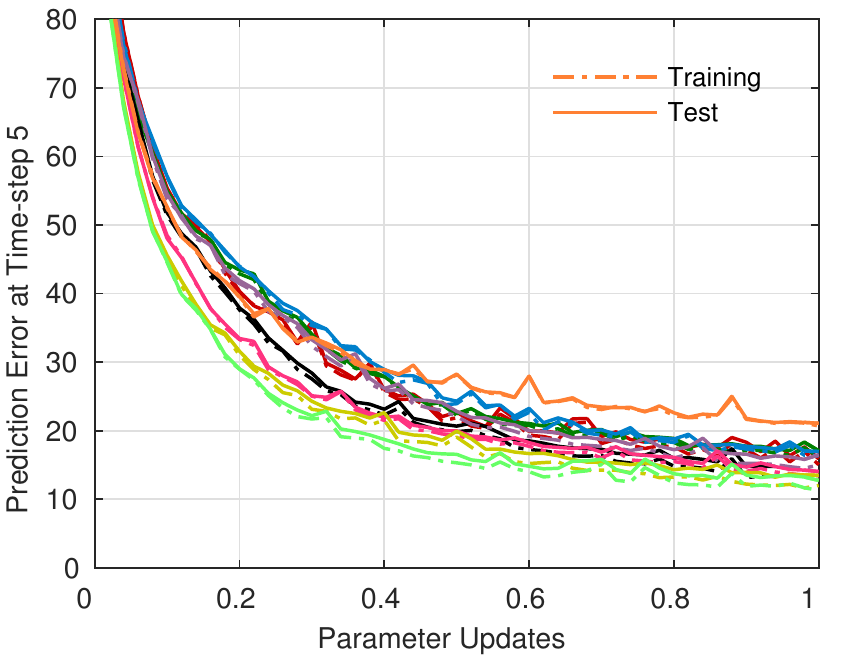}}
\subfigure[]{
\scalebox{0.79}{\includegraphics[]{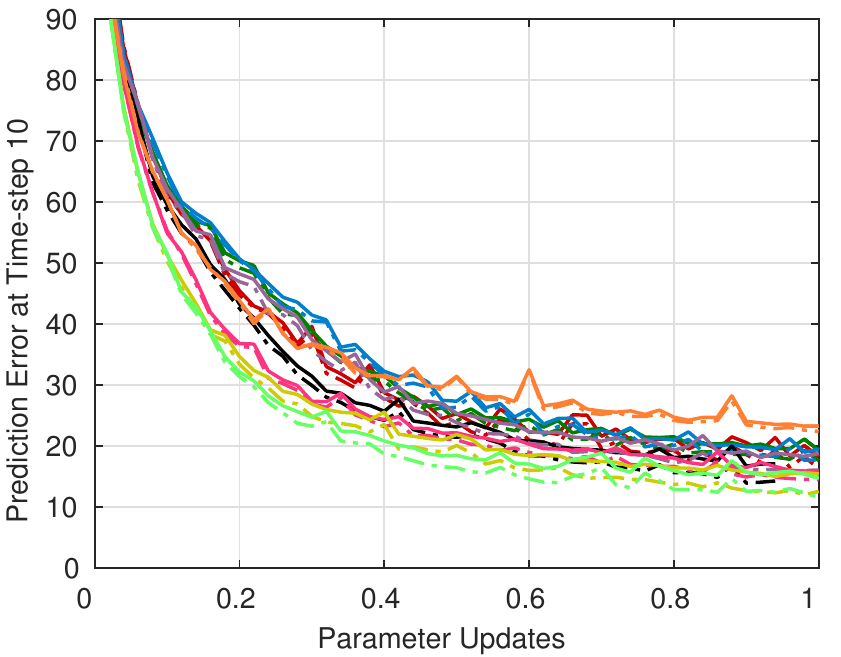}}
%\hskip0.1cm
\scalebox{0.79}{\includegraphics[]{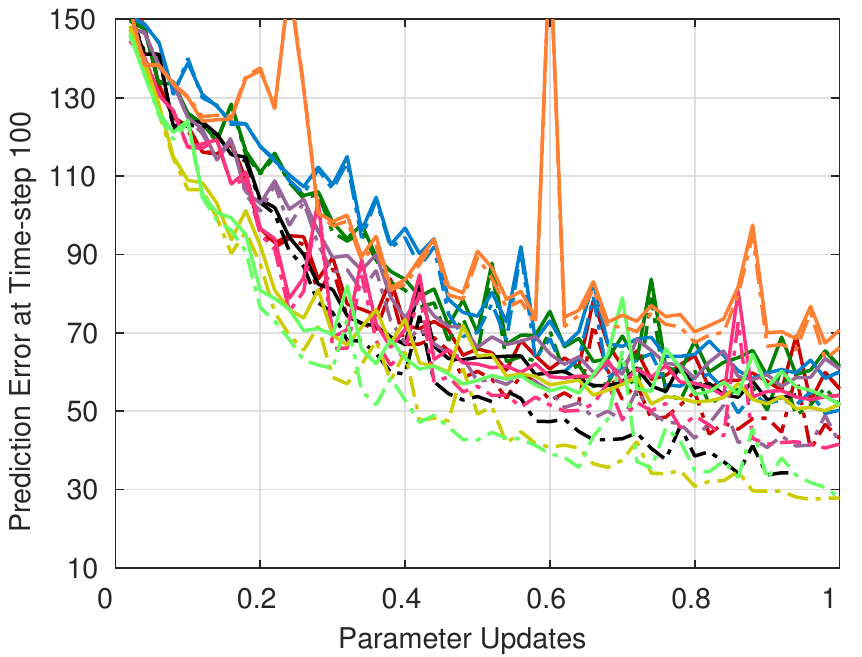}}}
\scalebox{0.79}{\includegraphics[]{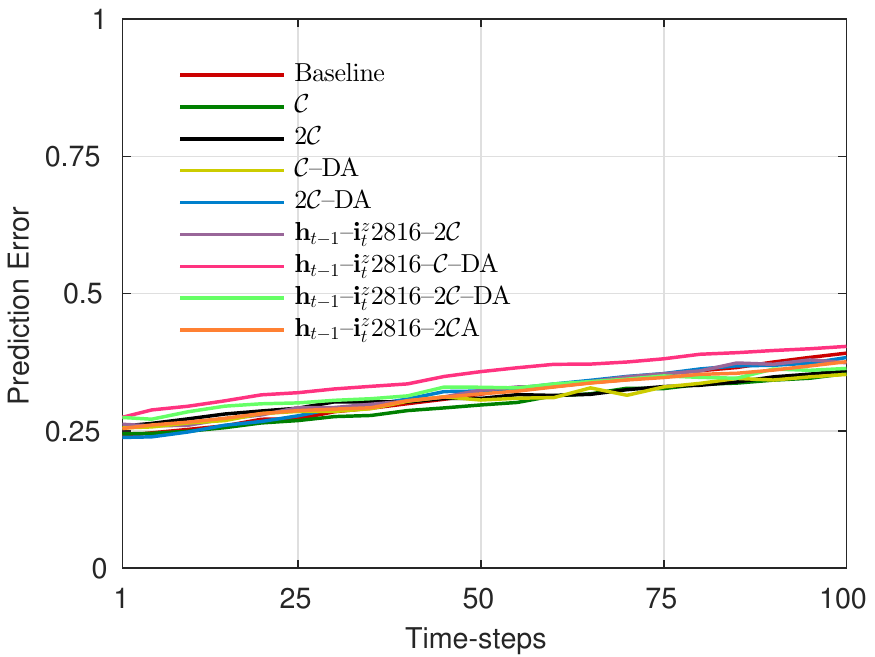}}
%\hskip0.1cm
\scalebox{0.79}{\includegraphics[]{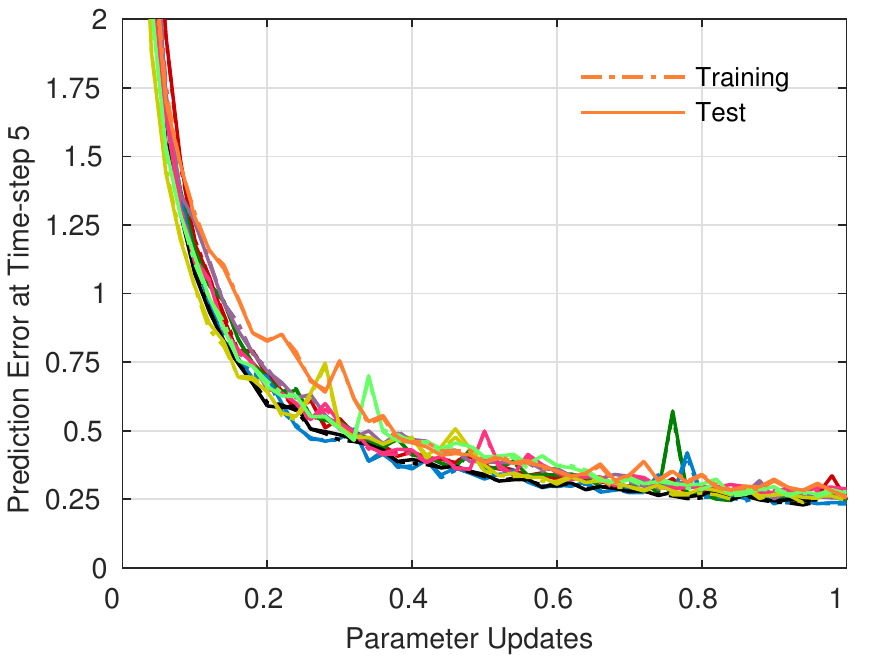}}
\subfigure[]{
\scalebox{0.79}{\includegraphics[]{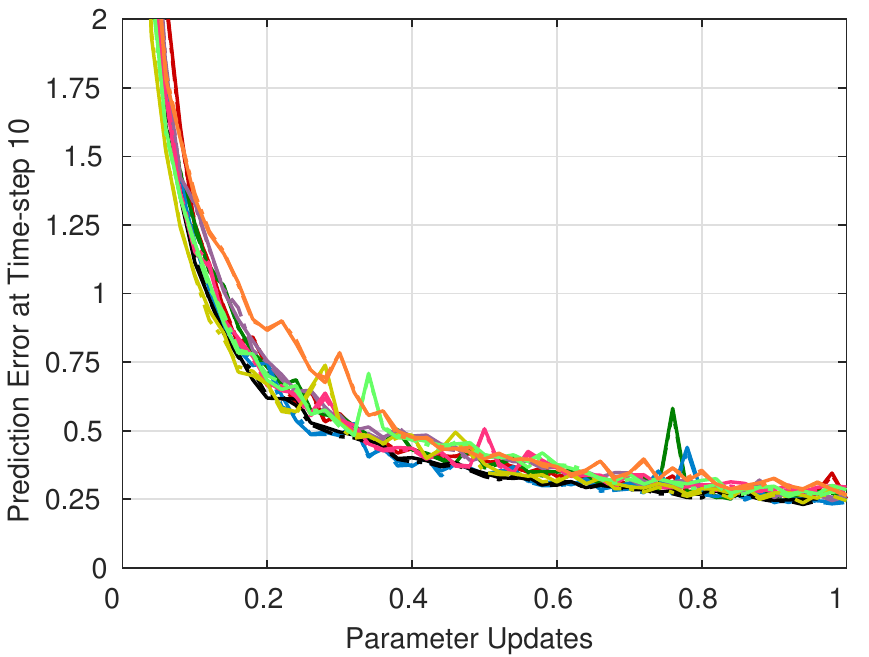}}
%\hskip0.1cm
\scalebox{0.79}{\includegraphics[]{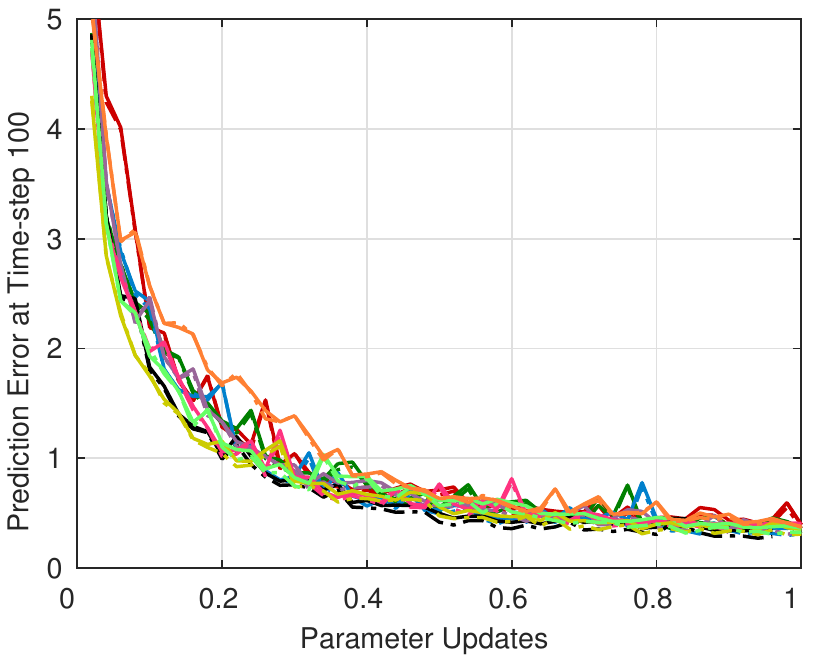}}}
\caption{Prediction error for different convolutional action-dependent state transitions on (a) Fishing Derby and (b) Freeway.}
\label{fig:predErrConvStructFishingDerby-Freeway}
\end{figure}

\begin{figure}[htbp] % Figures obtained with predErrConvStructAppendix
\vskip-0.5cm
\scalebox{0.79}{\includegraphics[]{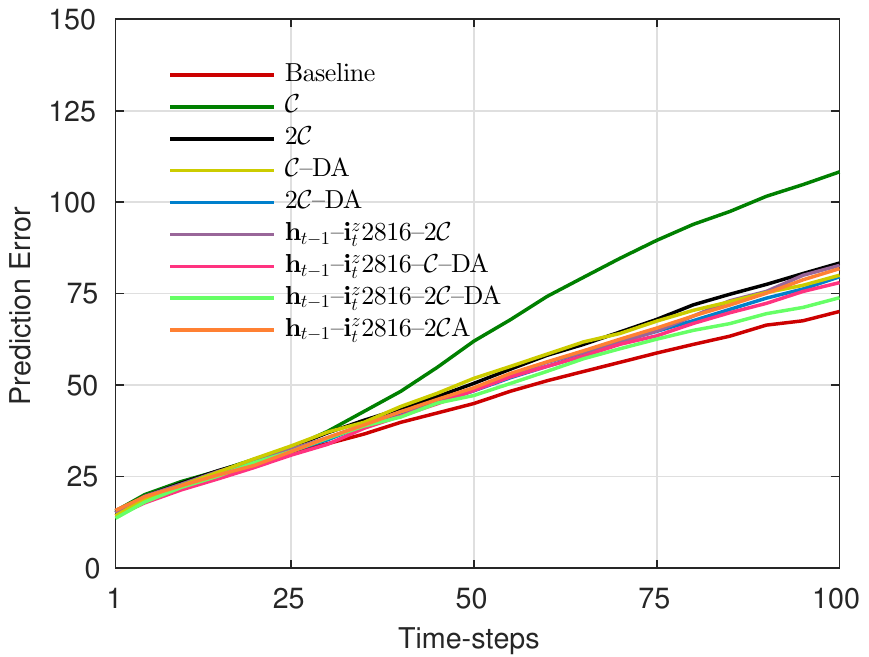}}
%\hskip0.1cm
\scalebox{0.79}{\includegraphics[]{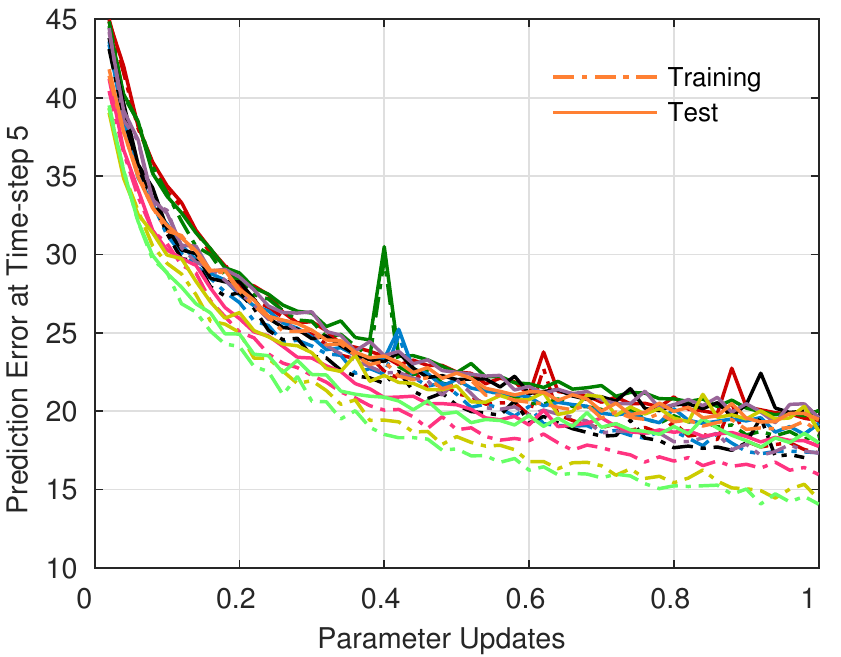}}
\subfigure[]{
\scalebox{0.79}{\includegraphics[]{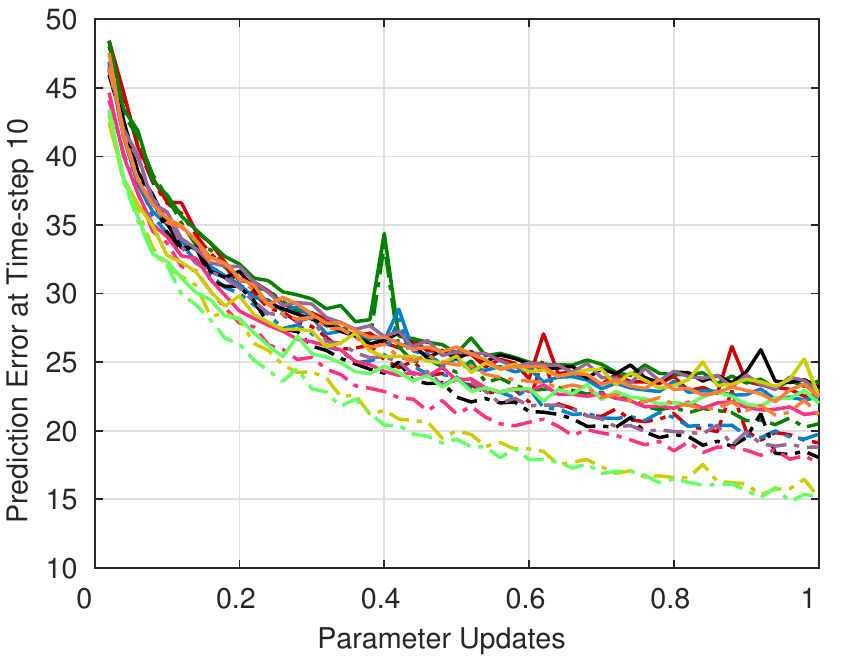}}
%\hskip0.1cm
\scalebox{0.79}{\includegraphics[]{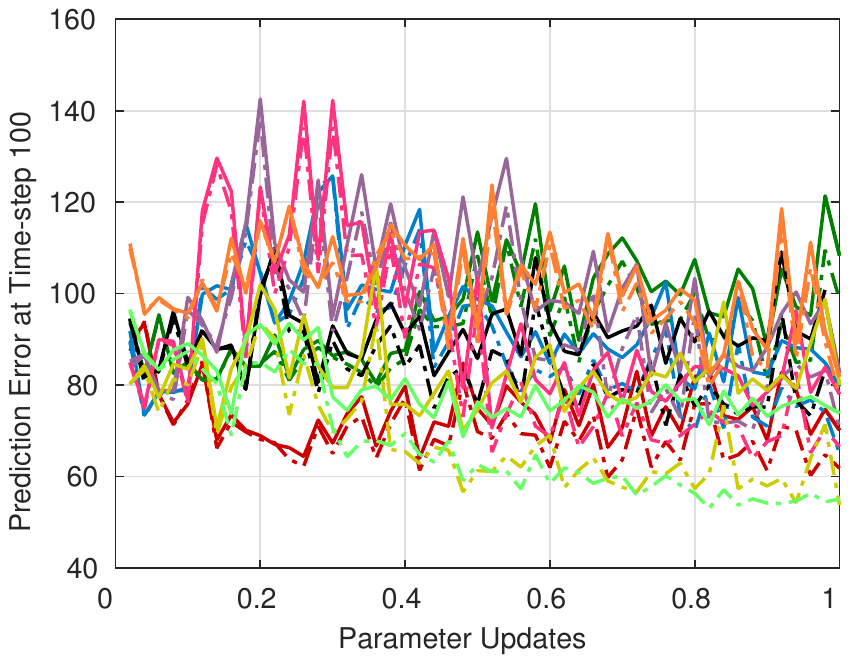}}}
\scalebox{0.79}{\includegraphics[]{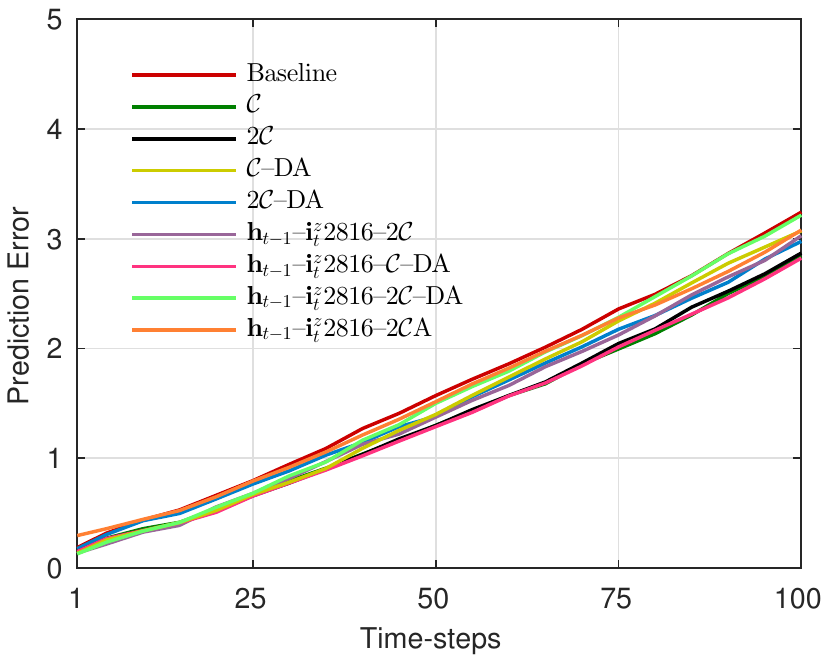}}
%\hskip0.1cm
\scalebox{0.79}{\includegraphics[]{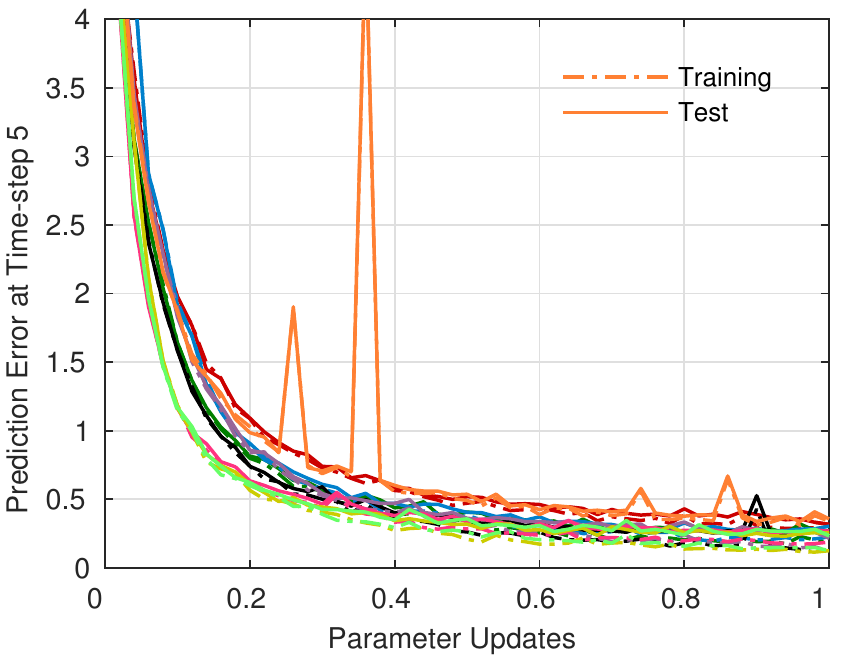}}
\subfigure[]{
\scalebox{0.79}{\includegraphics[]{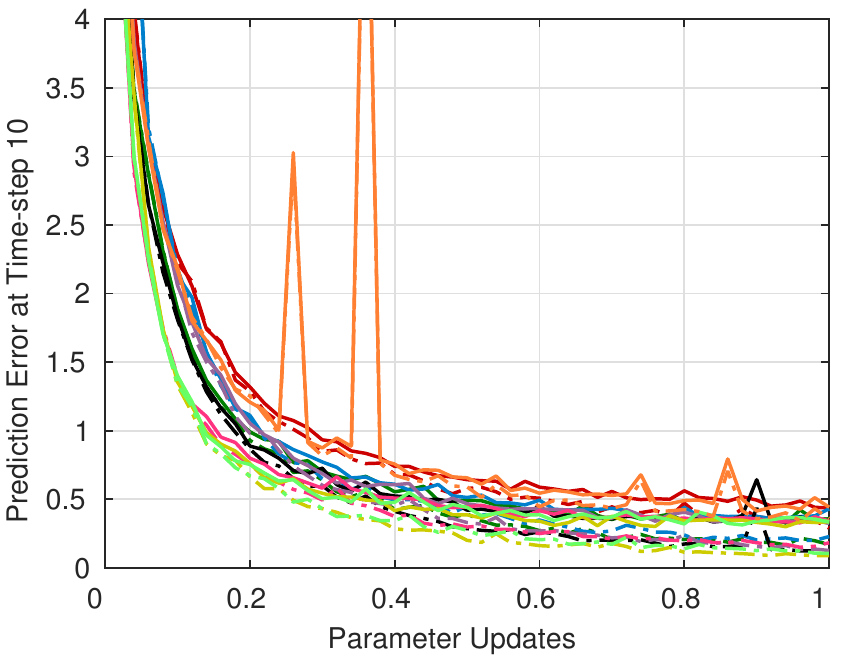}}
%\hskip0.1cm
\scalebox{0.79}{\includegraphics[]{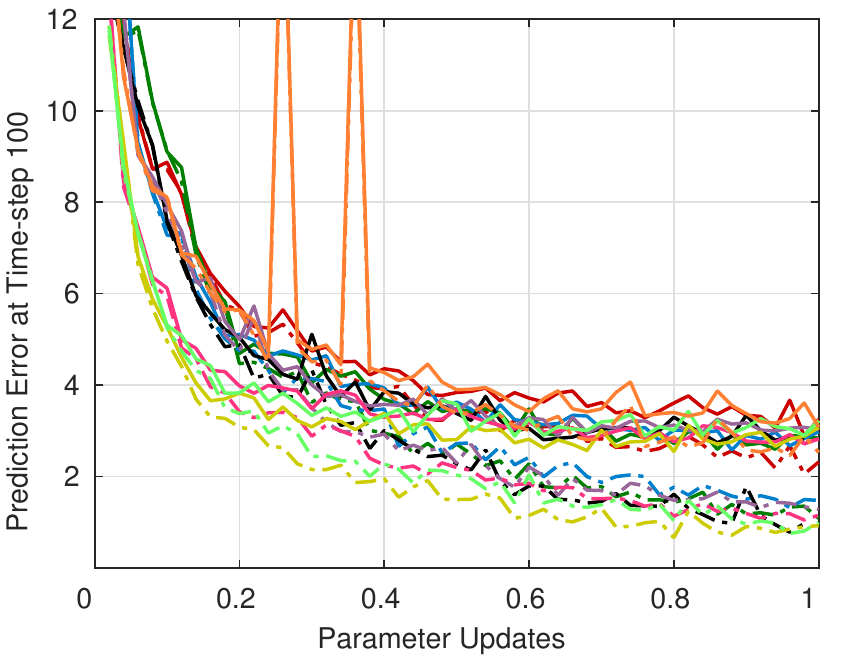}}}
\caption{Prediction error for different convolutional action-dependent state transitions on (a) Ms Pacman and (b) Pong.}
\label{fig:predErrConvStructMsPacman-Pong}
\end{figure}

\begin{figure}[htbp] % Figures obtained with predErrConvStructAppendix
\vskip-0.5cm
\scalebox{0.79}{\includegraphics[]{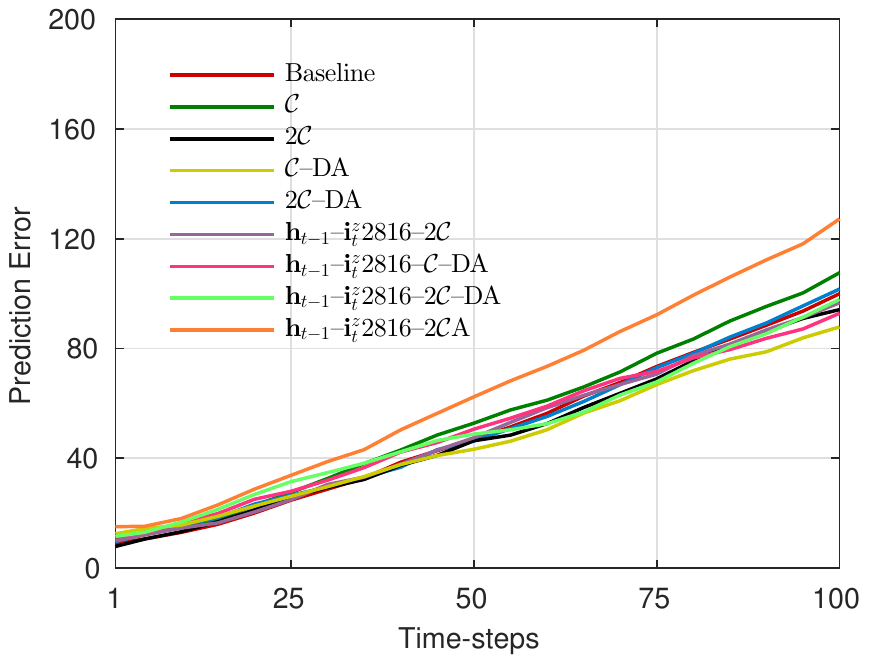}}
%\hskip0.1cm
\scalebox{0.79}{\includegraphics[]{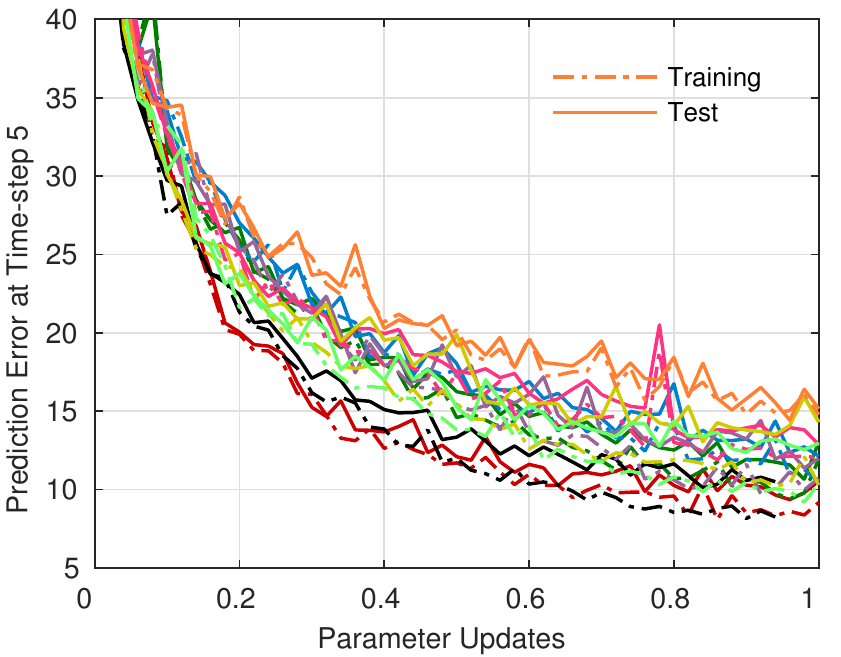}}
\subfigure[]{
\scalebox{0.79}{\includegraphics[]{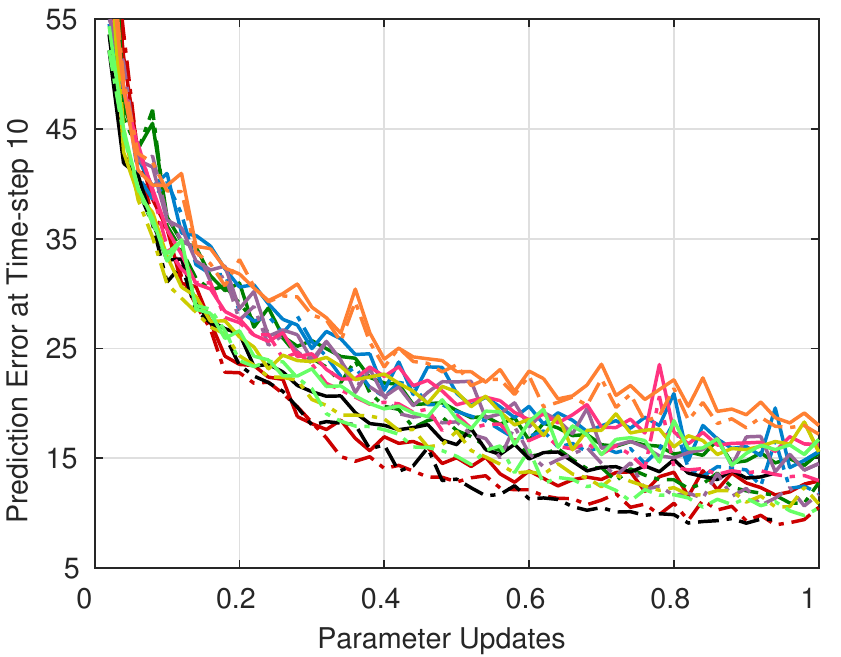}}
%\hskip0.1cm
\scalebox{0.79}{\includegraphics[]{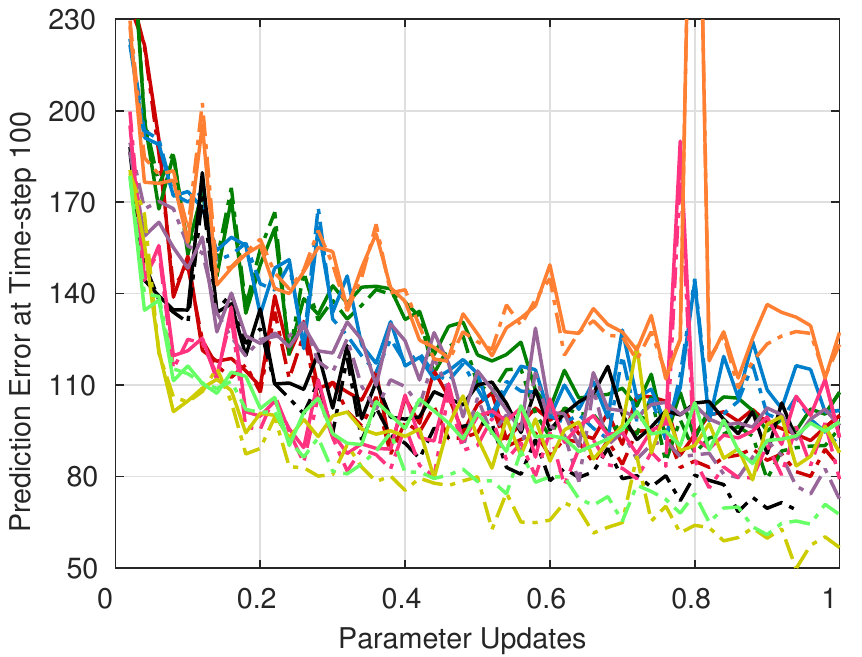}}}
\scalebox{0.79}{\includegraphics[]{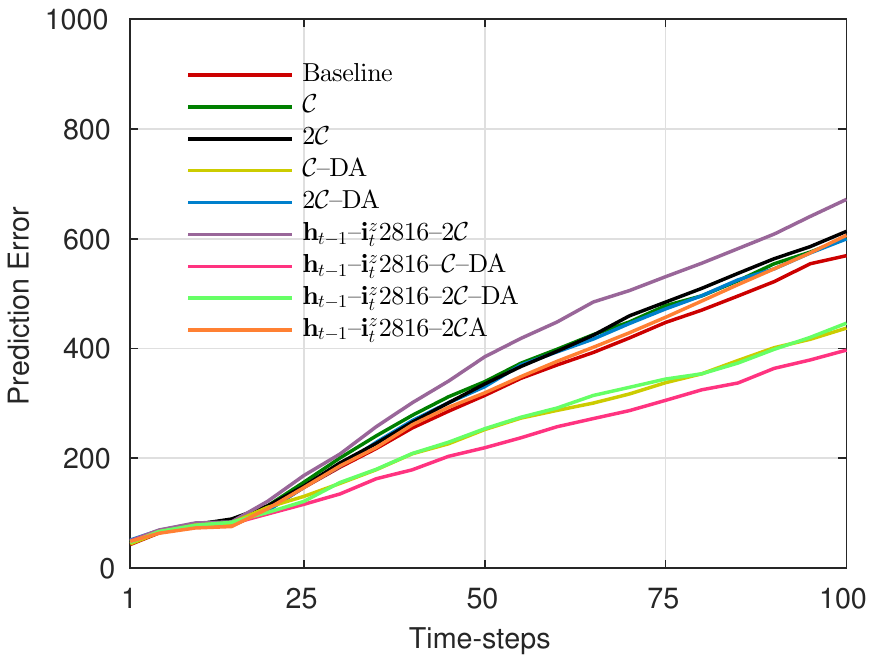}}
%\hskip0.1cm
\scalebox{0.79}{\includegraphics[]{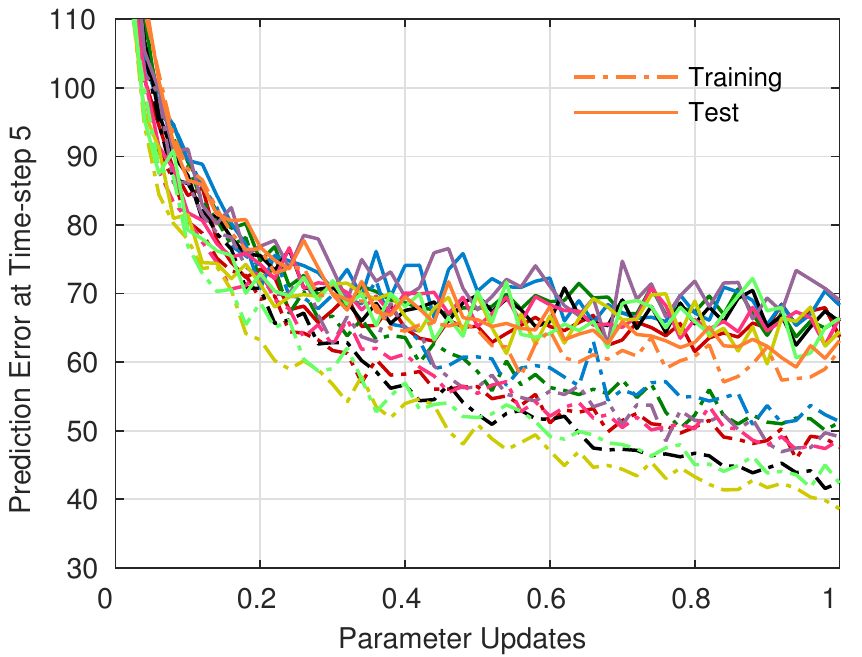}}
\subfigure[]{
\scalebox{0.79}{\includegraphics[]{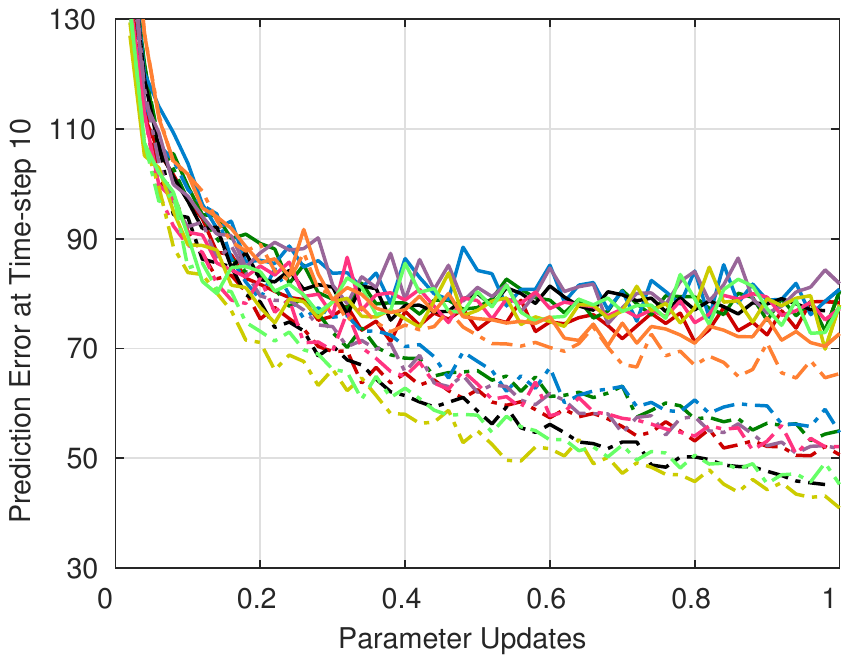}}
%\hskip0.1cm
\scalebox{0.79}{\includegraphics[]{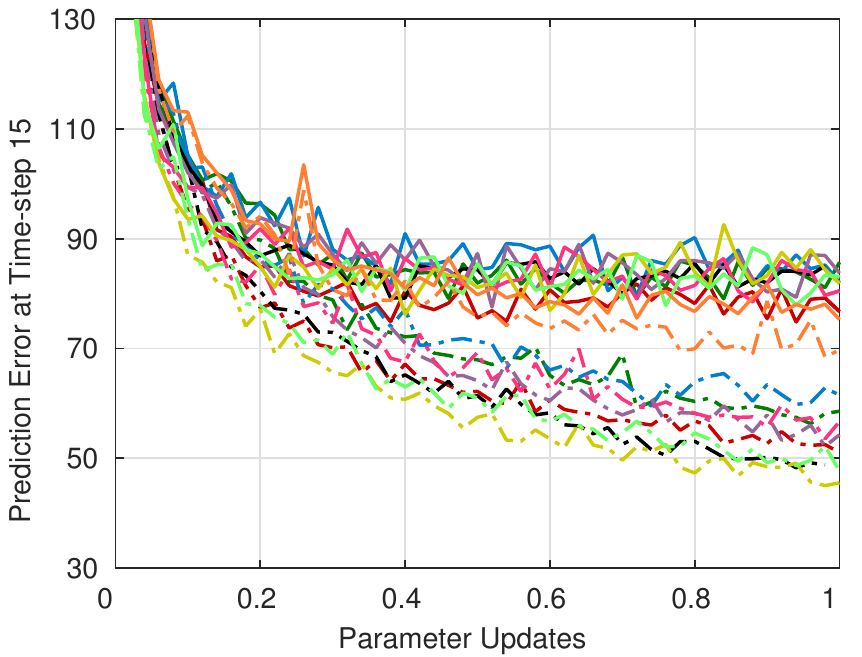}}}
\caption{Prediction error for different convolutional action-dependent state transitions on (a) Qbert and (b) Riverraid.}
\label{fig:predErrConvStructQbert-Riverraid}
\end{figure}

\begin{figure}[htbp] % Figures obtained with predErrConvStructAppendix
\vskip-0.5cm
\scalebox{0.79}{\includegraphics[]{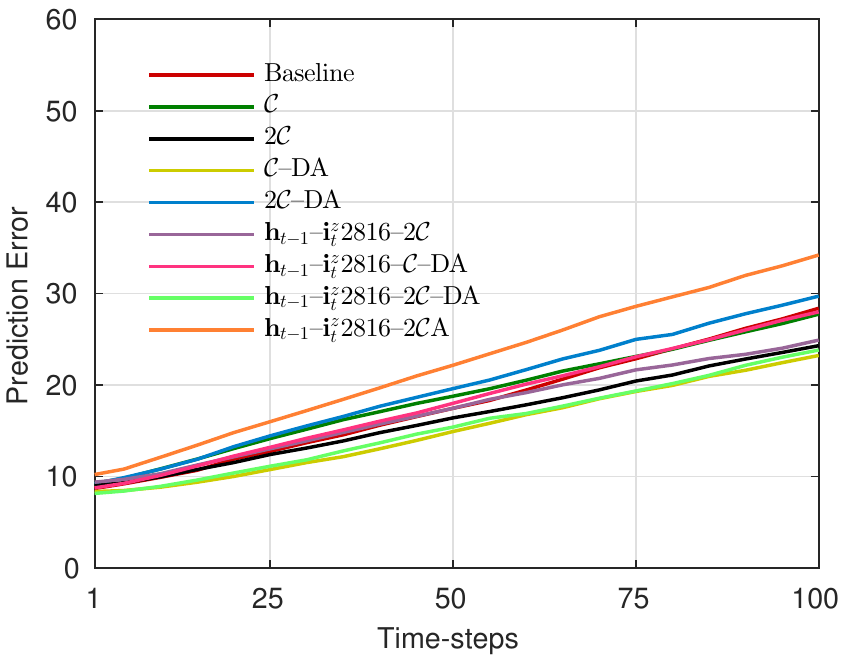}}
%\hskip0.1cm
\scalebox{0.79}{\includegraphics[]{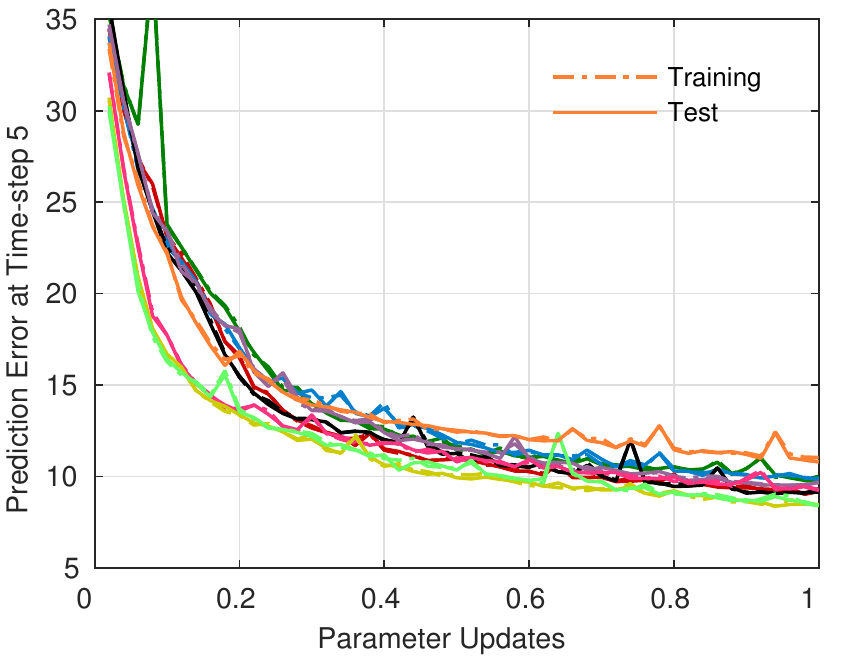}}
\subfigure[]{
\scalebox{0.79}{\includegraphics[]{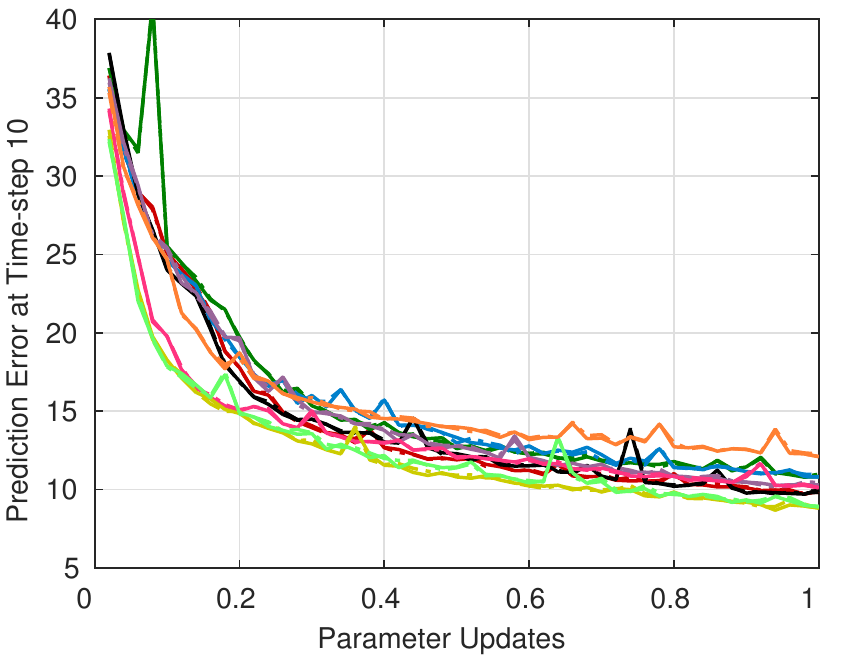}}
%\hskip0.1cm
\scalebox{0.79}{\includegraphics[]{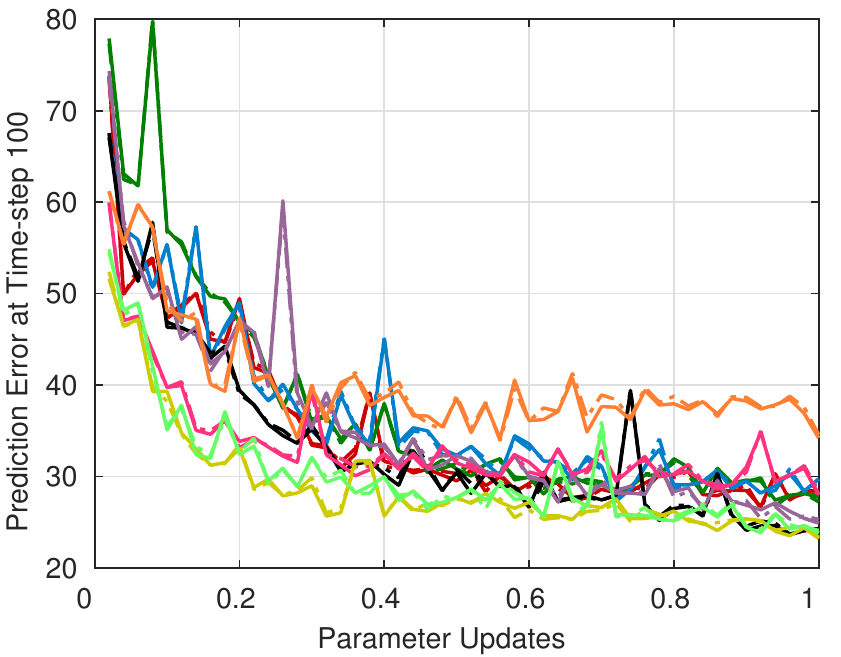}}}
\scalebox{0.79}{\includegraphics[]{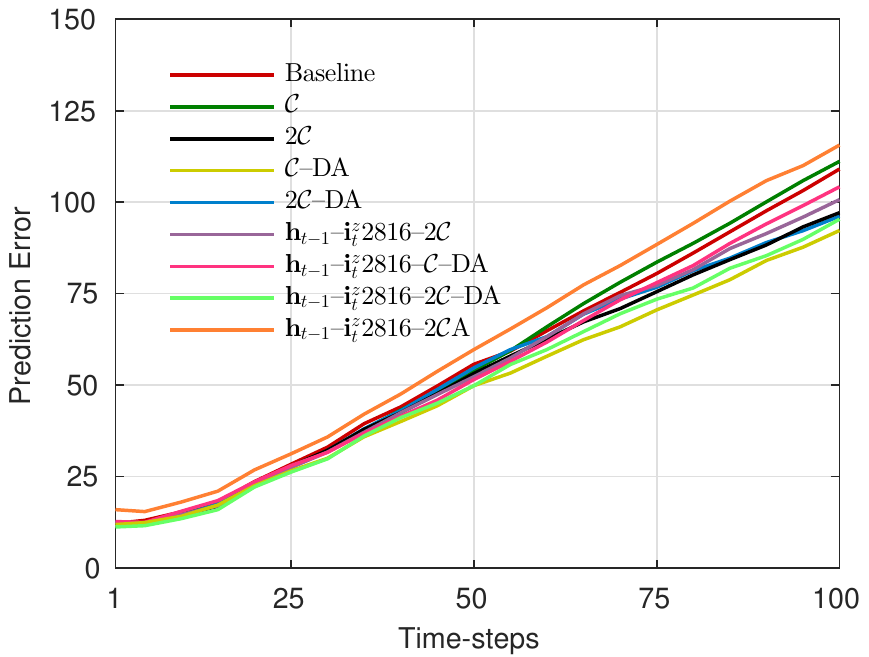}}
%\hskip0.1cm
\scalebox{0.79}{\includegraphics[]{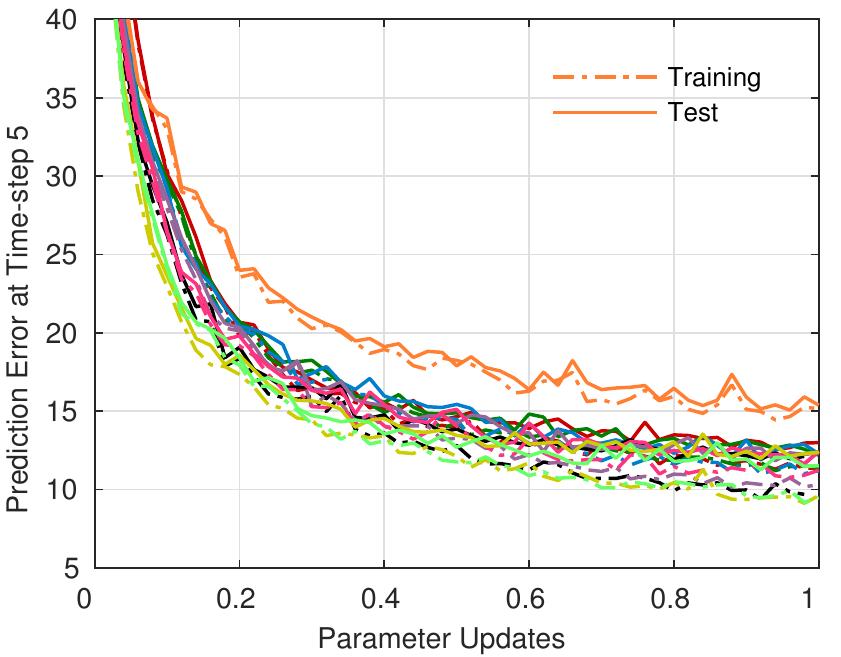}}
\subfigure[]{
\scalebox{0.79}{\includegraphics[]{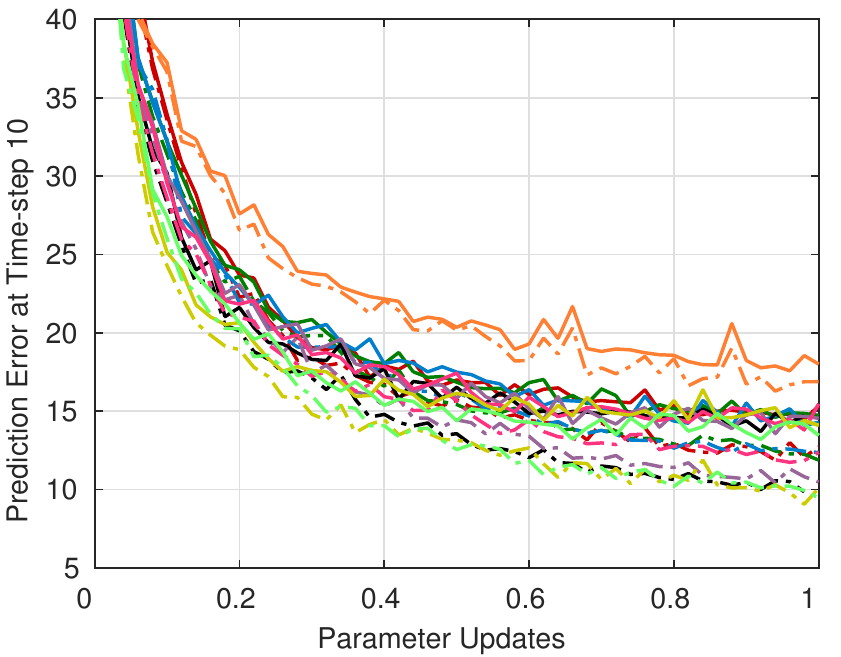}}
%\hskip0.1cm
\scalebox{0.79}{\includegraphics[]{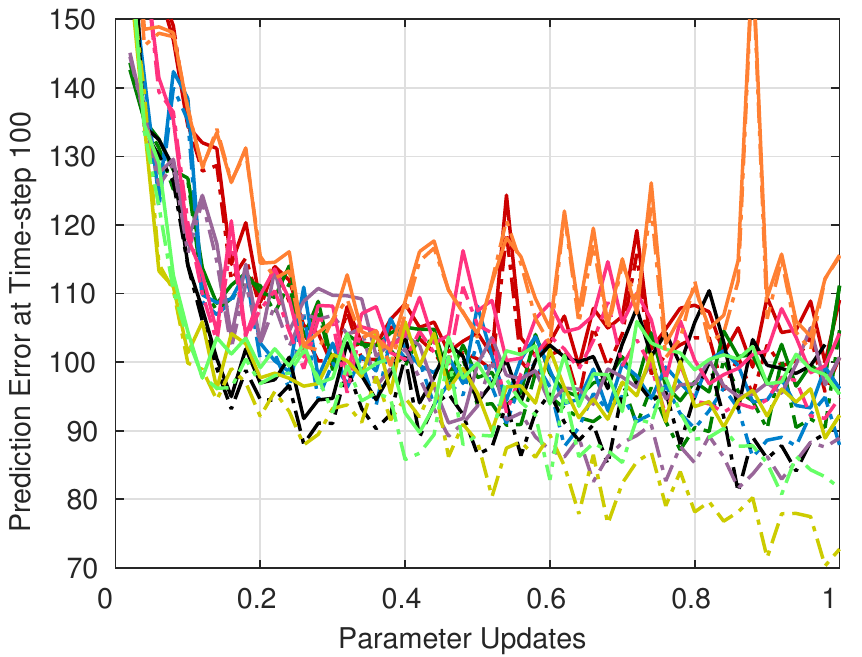}}}
\caption{Prediction error for different convolutional action-dependent state transitions on (a) Seaquest and (b) Space Invaders.}
\label{fig:predErrConvStructSeaquest-SpaceInvaders}
\end{figure}

In Figs. \ref{fig:predErrConvStructBowling-Breakout}-\ref{fig:predErrConvStructSeaquest-SpaceInvaders} we compare the baseline 
transition with the following convolutional alternatives (where to apply the convolutional transformations the vectors $\vz_{t-1}$ and $\ha_{t}$ 
of dimensionality 2816 are reshaped into tensors of dimension $32\times11\times8$)
\begin{description} 
\item[$\bullet$ ${\cal C}$ and 2${\cal C}$:] Convolutional gate and cell updates, \ie 
%ATA5e6Tr1e6Tse0.2MPfalsebS16nH10nHid2816nLHid2816nF2816mask2-15-1.1-15-1SEnF64lR1e-5MConv3 
%ATA5e6Tr1e6Tse0.2MPfalsebS16nH10nHid2816nLHid2816nF2816mask2-15-1.1-15-1SEnF64lR1e-5M2Conv3
\begin{align*}
&\vi_t = {\sigma}({\cal C}^{iv}(\ha_{t}) +{\cal C}^{iz}(\vz_{t-1}))\,, \hskip0.3cm \vf_t = {\sigma}({\cal C}^{fv}(\ha_t) +{\cal C}^{fz}(\vz_{t-1}))\,,\\
&\vo_t = {\sigma}({\cal C}^{ov}(\ha_t) + {\cal C}^{oz}(\vz_{t-1}))\,,\\
&\vc_t = \vf_t\otimes \vc_{t-1}+\vi_t\otimes \textrm{tanh}({\cal C}^{cv}(\ha_{t})+{\cal C}^{cz}(\vz_{t-1}))\,,
\end{align*}
where ${\cal C}$ denotes either one convolution with 32 filters of size 3$\times$3, with stride 1 and padding 1 (as to preserve the input size),
or two such convolutions with RReLU nonlinearity in between. These two models have around 16M parameters.\\
\item[$\bullet$ ${\cal C}$DA and 2${\cal C}$DA:] As above but with different action fusion parameters for the gate and cell updates, \ie  
%ATA5e6Tr1e6Tse0.2MPfalsebS16nH10nHid2816nLHid2816nF2816mask2-15-1.1-15-1SEnF64lR1e-5MConv3SI 
%ATA5e6Tr1e6Tse0.2MPfalsebS16nH10nHid2816nLHid2816nF2816mask2-15-1.1-15-1SEnF64lR1e-5M2Conv3SI
\begin{align*}
&\ha^i_{t} = \vW^{ih}\vh_{t-1}\otimes \vW^{ia}\va_{t-1}\,, \hskip0.3cm \ha^f_{t} = \vW^{fh}\vh_{t-1}\otimes \vW^{fa}\va_{t-1}\,,\\
& \ha^o_{t} = \vW^{oh}\vh_{t-1}\otimes \vW^{oa}\va_{t-1}\,, \hskip0.3cm \ha^c_{t} = \vW^{ch}\vh_{t-1}\otimes \vW^{ca}\va_{t-1}\,,\\
& \vi_t = {\sigma}({\cal C}^{iv}(\ha^i_{t}) +{\cal C}^{iz}(\vz_{t-1}))\,, \hskip0.3cm \vf_t = {\sigma}({\cal C}^{fv}(\ha^f_t) +{\cal C}^{fz}(\vz_{t-1}))\,,\\
&\vo_t = {\sigma}({\cal C}^{ov}(\ha^o_t) + {\cal C}^{oz}(\vz_{t-1}))\,,\\
& \vc_t = \vf_t\otimes \vc_{t-1}+\vi_t\otimes \textrm{tanh}({\cal C}^{cv}(\ha^c_{t})+{\cal C}^{cz}(\vz_{t-1}))\,.
\end{align*}
These two models have around 40M parameters.\\
\item[$\bullet$ $\vh_{t-1}$--$\vi^z_t$2816--2${\cal C}$:] As '$\vh_{t-1}$--$\vi^z_t$2816' with convolutional gate and cell updates, \ie 
% ATA5e6Tr1e6Tse0.2MPfalsebS16nH10nHid2816nLHid2816nF2816mask2-15-1.1-15-1SEnF64lR1e-5M2Conv3NoXHGSX
\begin{align*}
&\vi_t = {\sigma}({\cal C}^{iv}(\ha_{t})+{\cal C}^{ih}(\vh_{t-1}))\,, \hskip0.3cm \vf_t = {\sigma}({\cal C}^{fv}(\ha_t)+{\cal C}^{fh}(\vh_{t-1}))\,,\\
&\vo_t = {\sigma}({\cal C}^{ov}(\ha_t)+{\cal C}^{oh}(\vh_{t-1}))\,,\\
&\vc_t = \vf_t\otimes \vc_{t-1}+\vi_t\otimes \textrm{tanh}({\cal C}^{cv}(\ha_{t})+{\cal C}^{ch}(\vh_{t-1}))+\vi^z_t\otimes \textrm{tanh}(\vz_{t-1})\,,
\end{align*}
where ${\cal C}$ denotes two convolutions as above. This model has around 16M parameters.\\
\item[$\bullet$ $\vh_{t-1}$--$\vi^z_t$2816--${\cal C}$DA and $\vh_{t-1}$--$\vi^z_t$2816--2${\cal C}$DA:]
As above but with different parameters for the gate and cell updates, and one or two convolutions. 
% ATA5e6Tr1e6Tse0.2MPfalsebS16nH10nHid2816nLHid2816nF2816mask2-15-1.1-15-1SEnF64lR1e-5MConv3NoXHGSXSI 
% ATA5e6Tr1e6Tse0.2MPfalsebS16nH10nHid2816nLHid2816nF2816mask2-15-1.1-15-1SEnF64lR1e-5M2Conv3NoXHGSXSI
These two models have around 48M parameters.
%This model has around 95 million parameters.\\
\item[$\bullet$ $\vh_{t-1}$--$\vi^z_t$2816--2${\cal C}$A:] As '$\vh_{t-1}$--$\vi^z_t$2816' with convolutional action fusion, gate and cell updates, \ie
\begin{align*}
& \ha_{t} = {\cal C}^h(\vh_{t-1})\otimes \vW^{a}\va_{t-1}\,,\\
&\vi_t = {\sigma}({\cal C}^{iv}(\ha_{t})+{\cal C}^{ih}(\vh_{t-1}))\,, \hskip0.3cm \vf_t = {\sigma}({\cal C}^{fv}(\ha_t)+{\cal C}^{fh}(\vh_{t-1}))\,,\\
&\vo_t = {\sigma}({\cal C}^{ov}(\ha_t)+{\cal C}^{oh}(\vh_{t-1}))\,,\\
& \vc_t = \vf_t\otimes \vc_{t-1}+\vi_t\otimes \textrm{tanh}({\cal C}^{cv}(\ha_{t})+{\cal C}^{ch}(\vh_{t-1}))+\vi^z_t\otimes \textrm{tanh}(\vz_{t-1})\,,
\end{align*}
where ${\cal C}$ indicates two convolutions as above.
This model has around 8M parameters. 
\end{description}
\subsubsection{Action Incorporation \label{sec:AppAction}}
\begin{figure}[htbp] % Figures obtained with predErrActionAppendix
\vskip-0.5cm
\scalebox{0.79}{\includegraphics[]{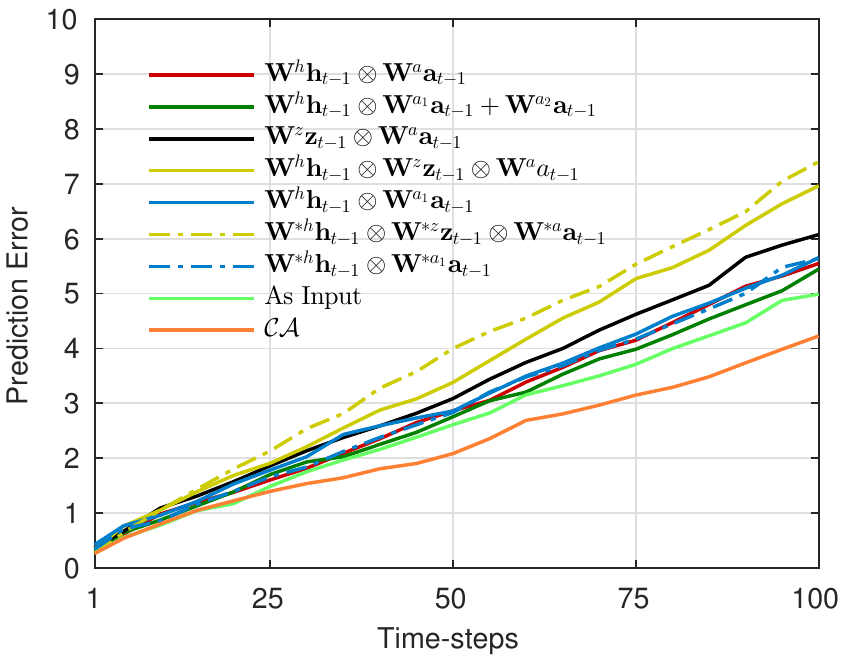}}
%\hskip0.1cm
\scalebox{0.79}{\includegraphics[]{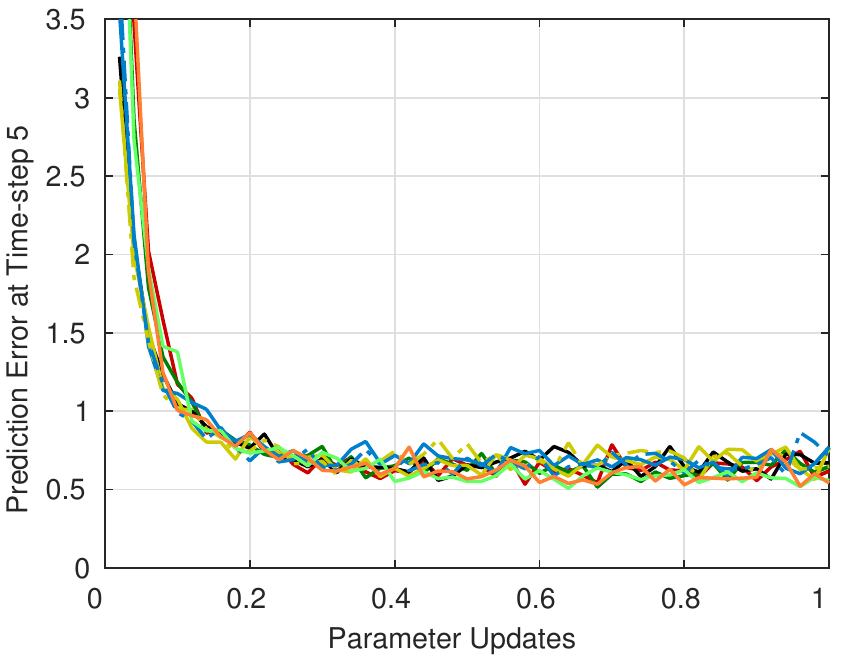}}\\
\subfigure[]{\scalebox{0.79}{\includegraphics[]{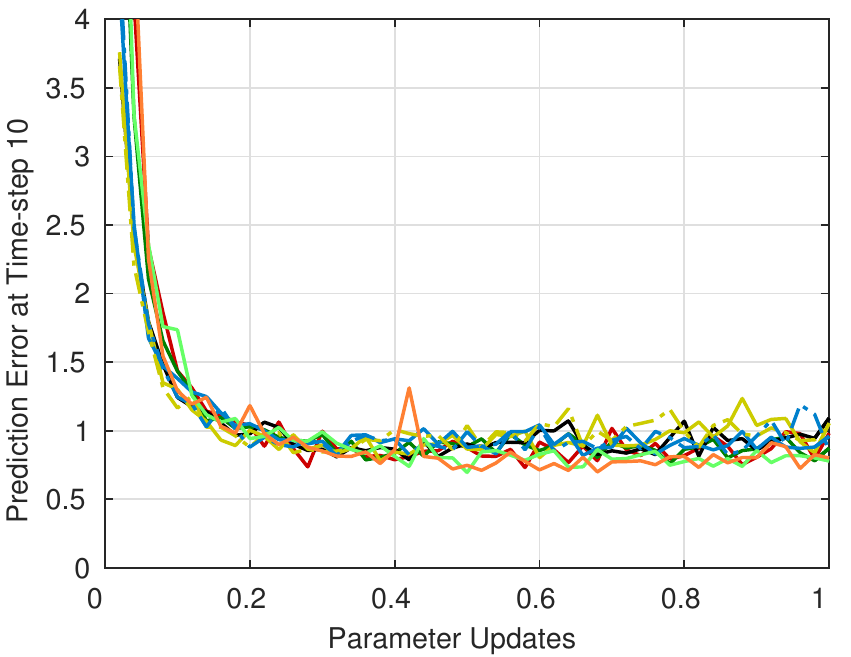}}
%\hskip0.1cm
\scalebox{0.79}{\includegraphics[]{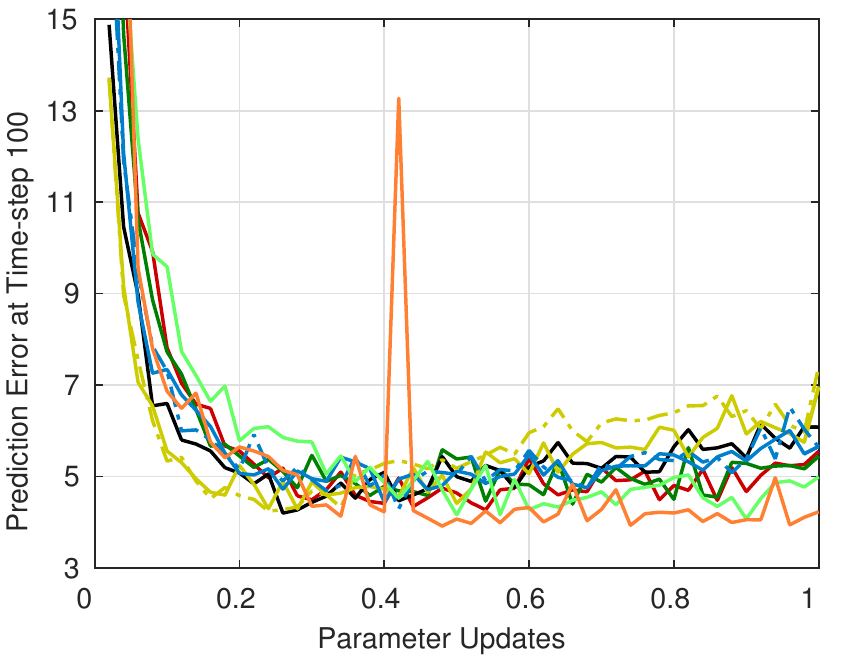}}}
\scalebox{0.79}{\includegraphics[]{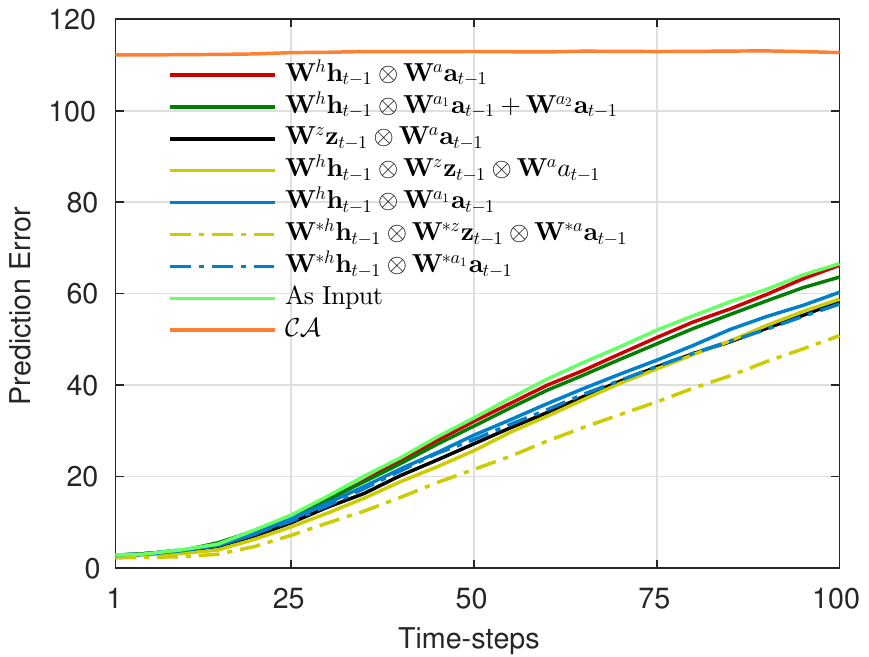}}
%\hskip0.1cm
\scalebox{0.79}{\includegraphics[]{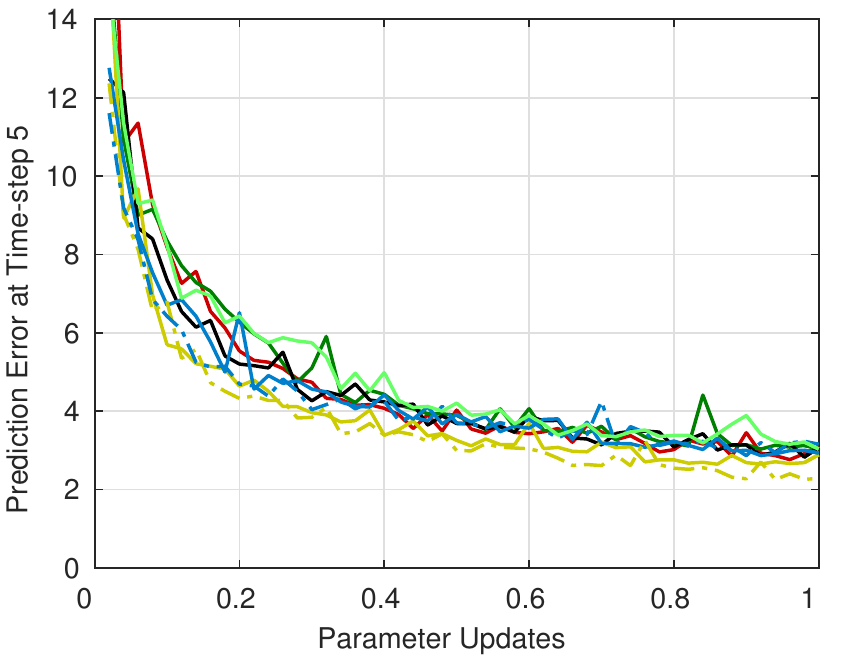}}
\subfigure[]{
\scalebox{0.79}{\includegraphics[]{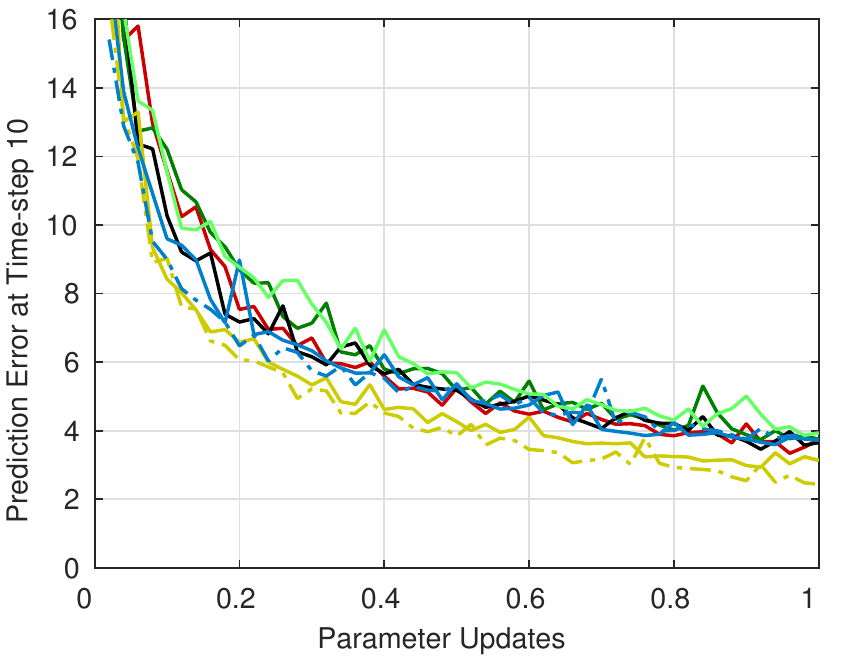}}
%\hskip0.1cm
\scalebox{0.79}{\includegraphics[]{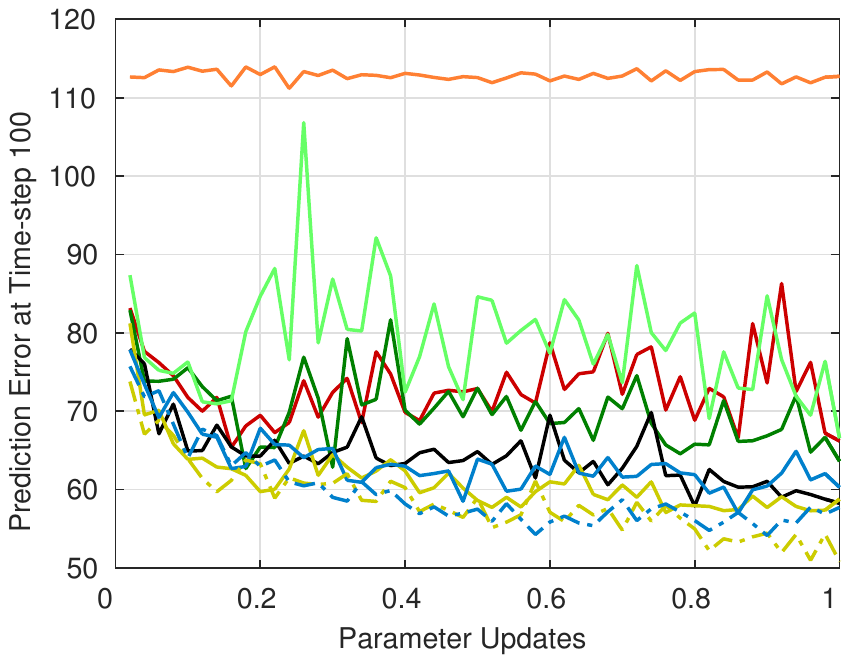}}}
%\vskip-0.3cm
\caption{Prediction error (average over 10,000 sequences) for different ways of incorporating the action on (a) Bowling and (b) Breakout. Parameter updates are in millions.}
\label{fig:predErrActionBowling-Breakout}
\end{figure}

\begin{figure}[htbp] % Figures obtained with predErrActionAppendix
\vskip-0.5cm
\scalebox{0.79}{\includegraphics[]{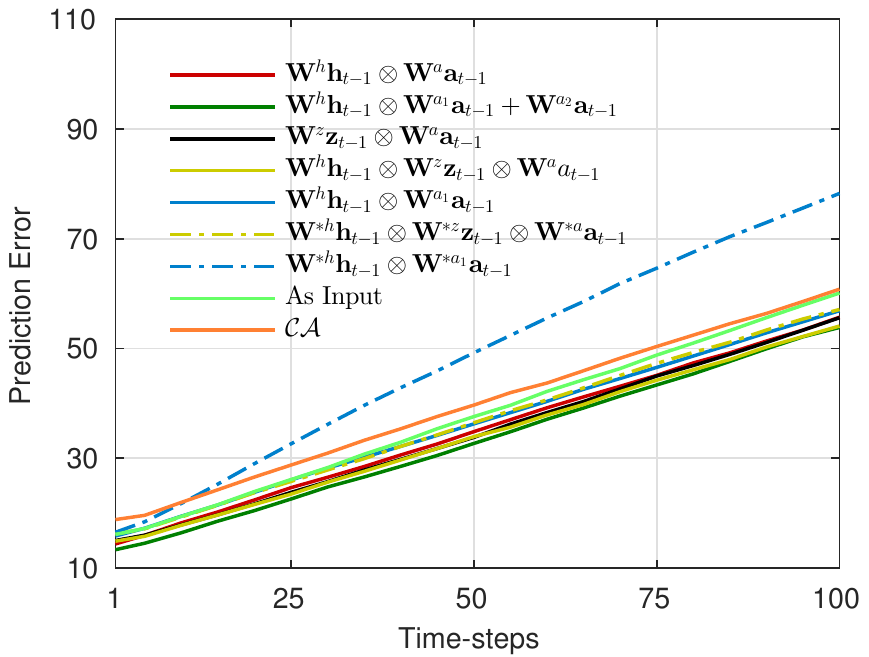}}
%\hskip0.1cm
\scalebox{0.79}{\includegraphics[]{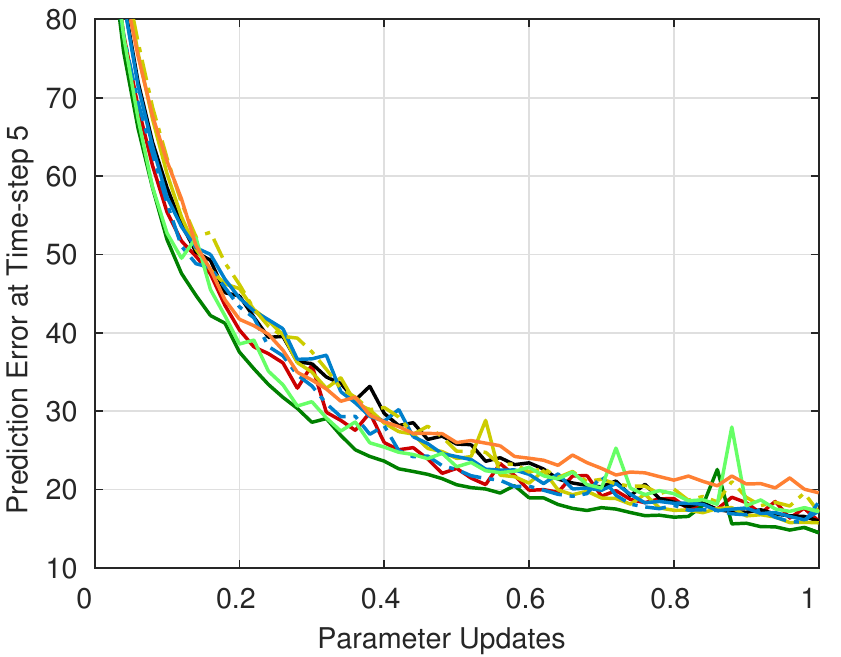}}
\subfigure[]{
\scalebox{0.79}{\includegraphics[]{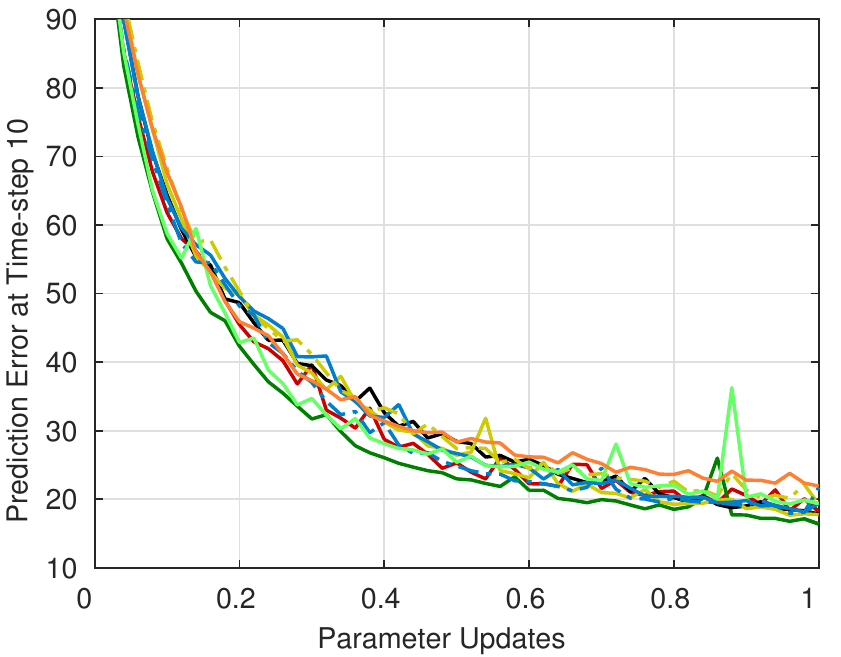}}
%\hskip0.1cm
\scalebox{0.79}{\includegraphics[]{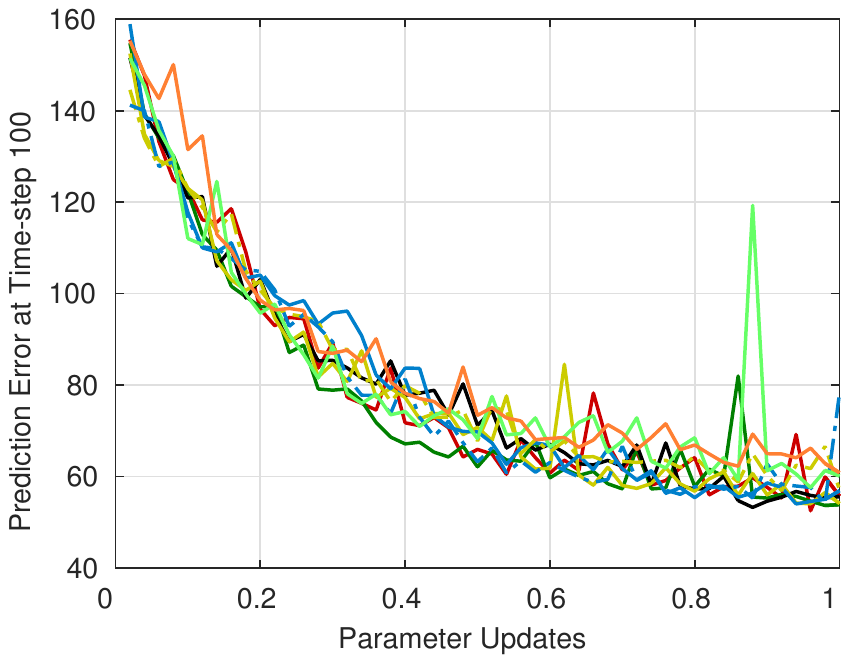}}}
\scalebox{0.79}{\includegraphics[]{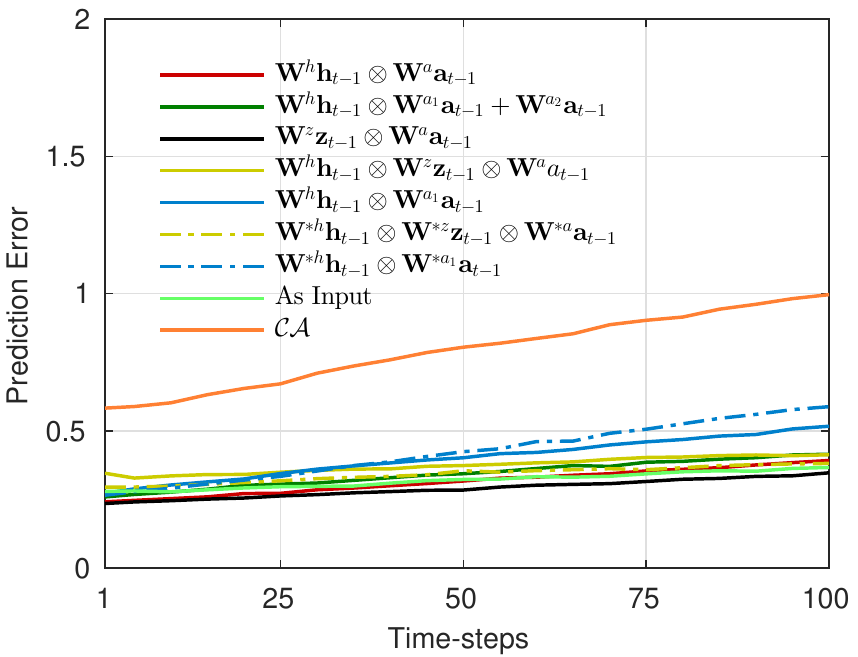}}
%\hskip0.1cm
\scalebox{0.79}{\includegraphics[]{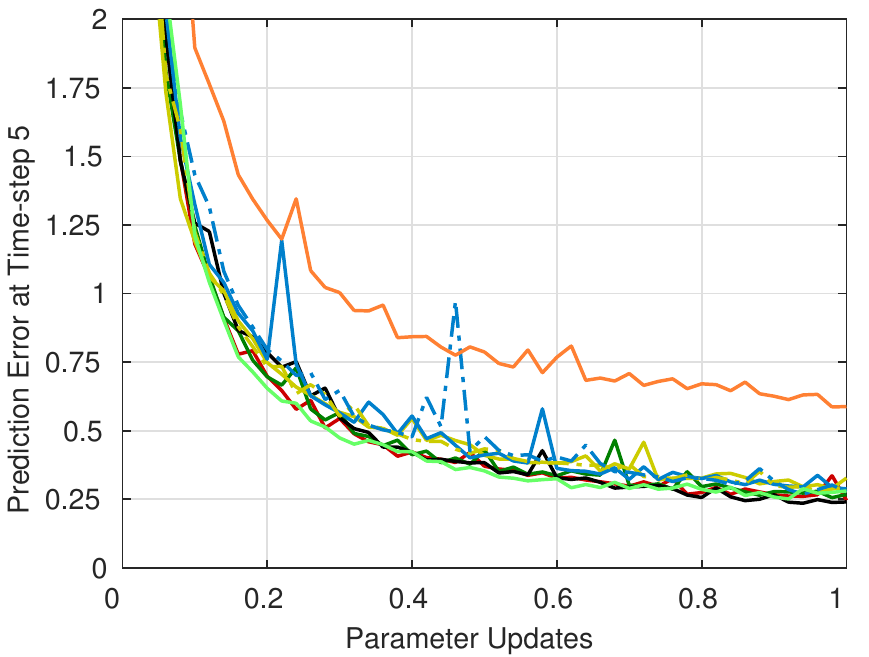}}
\subfigure[]{
\scalebox{0.79}{\includegraphics[]{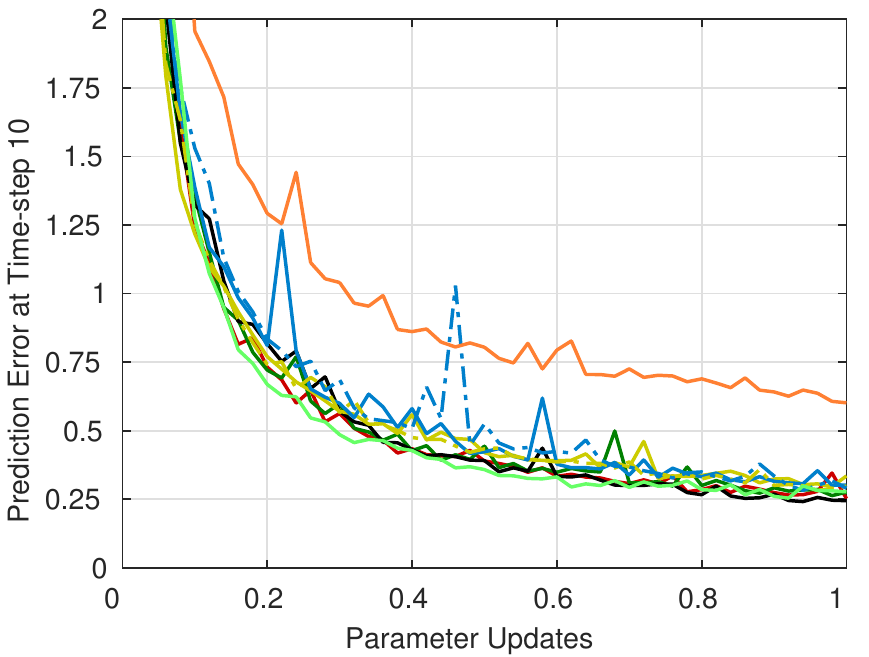}}
%\hskip0.1cm
\scalebox{0.79}{\includegraphics[]{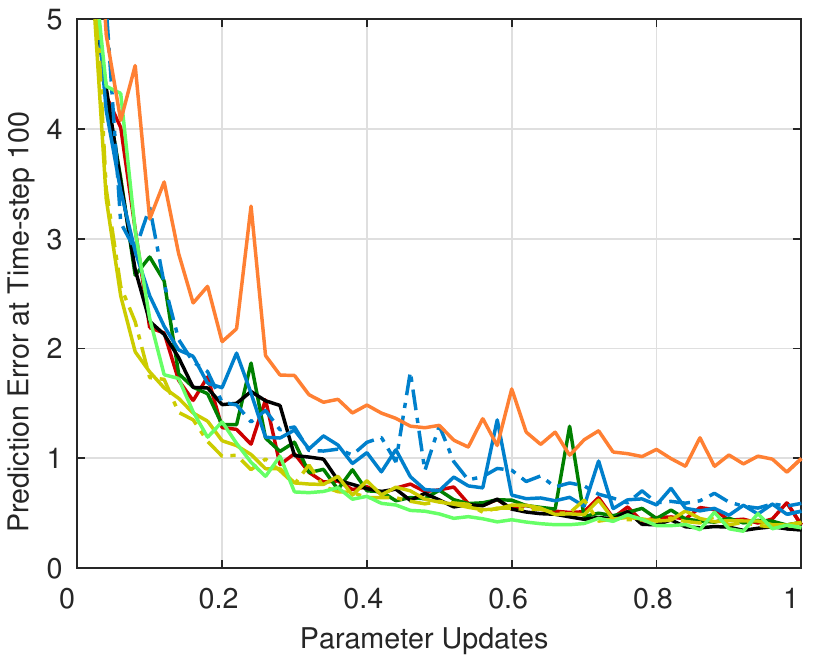}}}
\caption{Prediction error for different ways of incorporating the action on (a) Fishing Derby and (b) Freeway.}
\label{fig:predErrActionFishingDerby-Freeway}
\end{figure}

\begin{figure}[htbp] % Figures obtained with predErrActionAppendix
\vskip-0.5cm
\scalebox{0.79}{\includegraphics[]{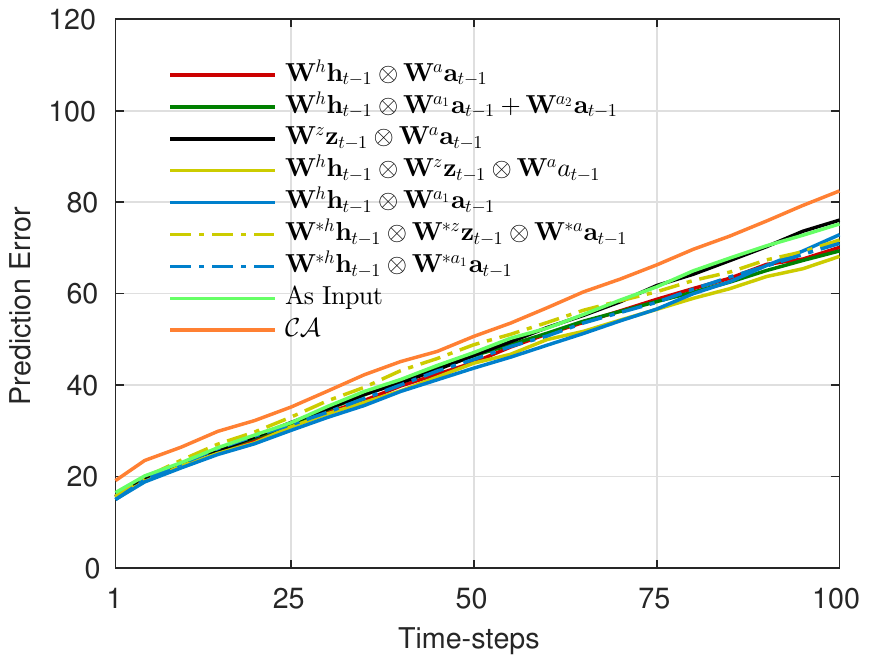}}
%\hskip0.1cm
\scalebox{0.79}{\includegraphics[]{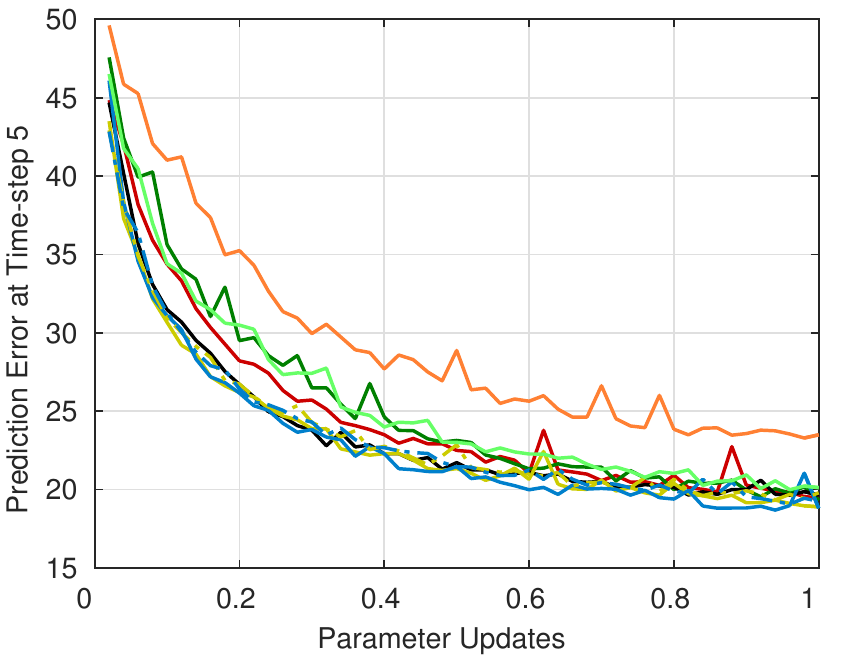}}
\subfigure[]{
\scalebox{0.79}{\includegraphics[]{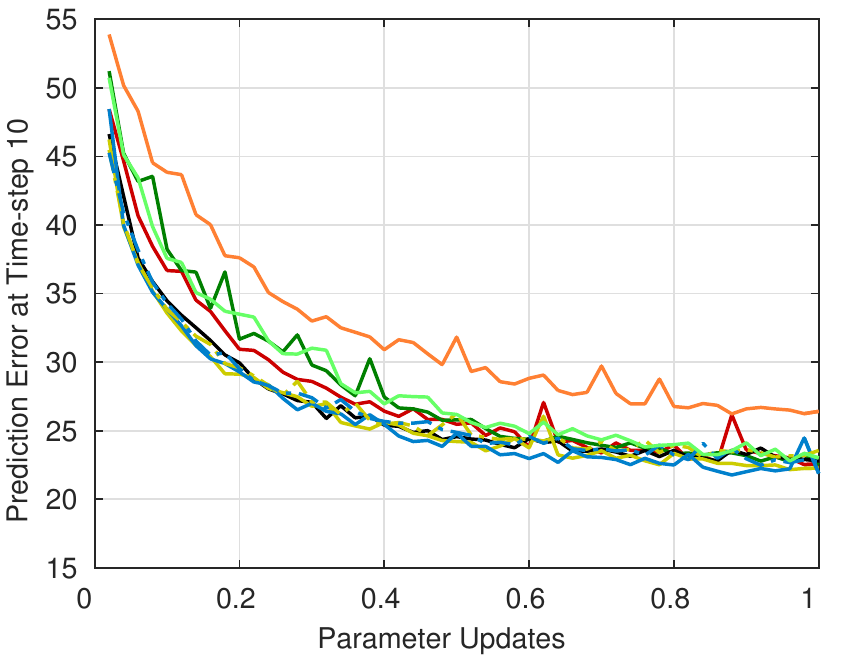}}
%\hskip0.1cm
\scalebox{0.79}{\includegraphics[]{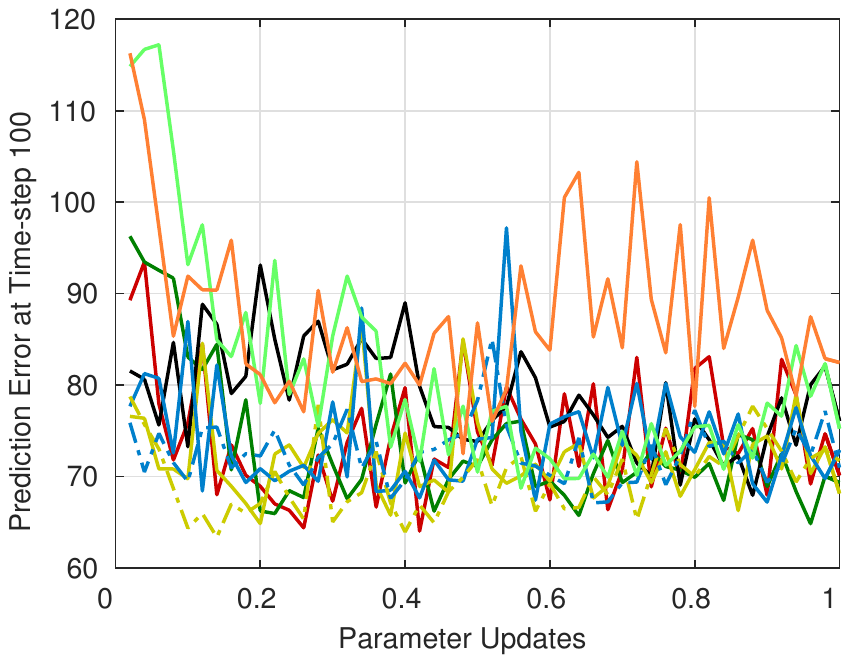}}}
\scalebox{0.79}{\includegraphics[]{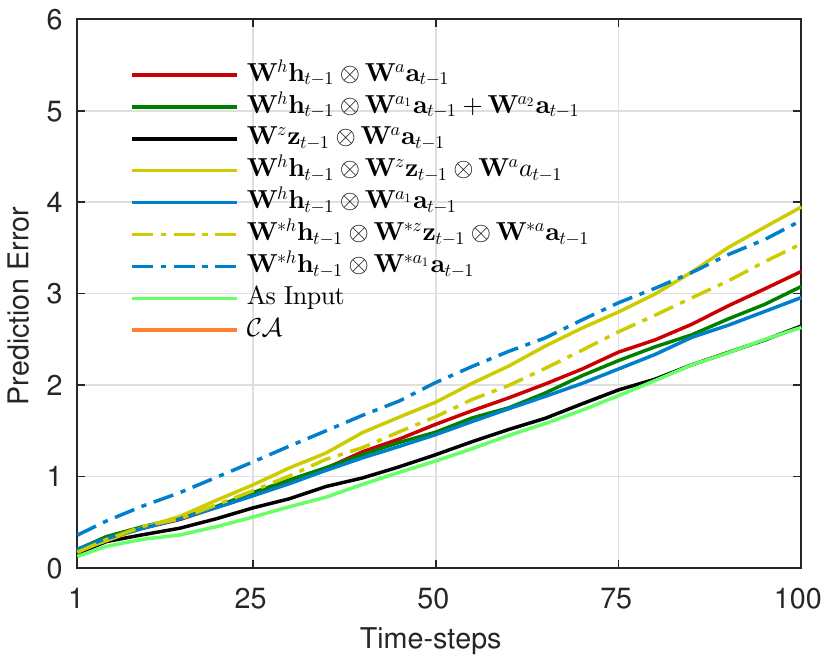}}
%\hskip0.1cm
\scalebox{0.79}{\includegraphics[]{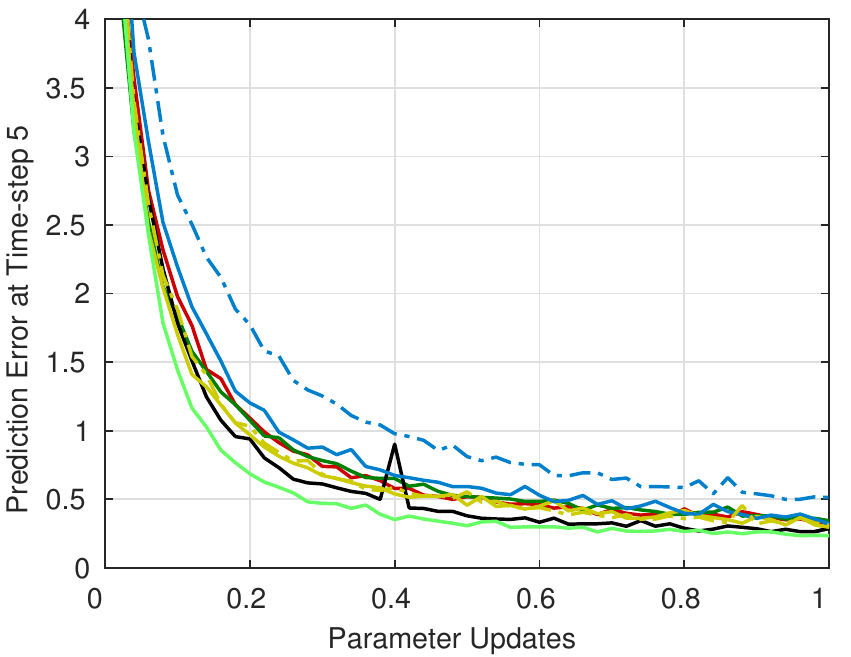}}
\subfigure[]{
\scalebox{0.79}{\includegraphics[]{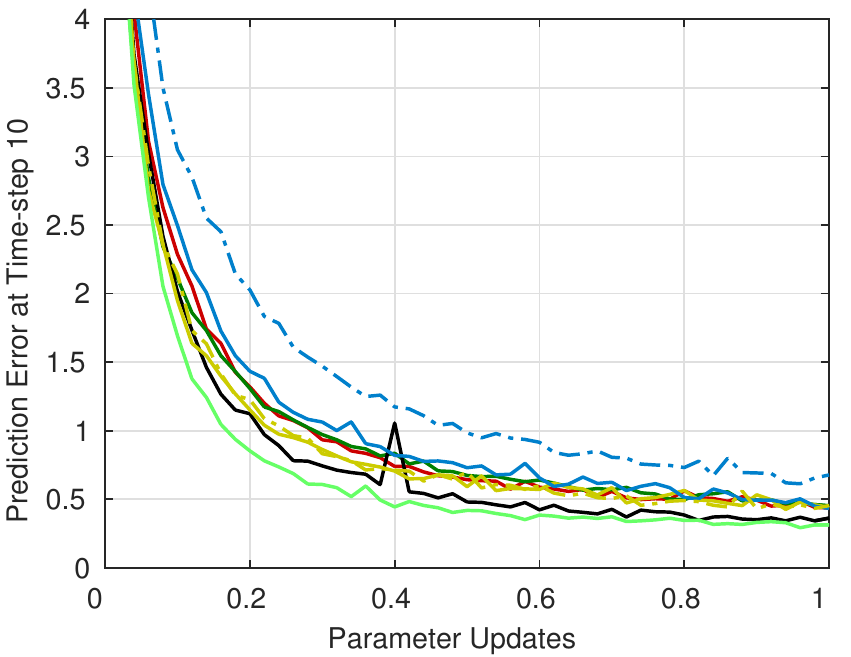}}
%\hskip0.1cm
\scalebox{0.79}{\includegraphics[]{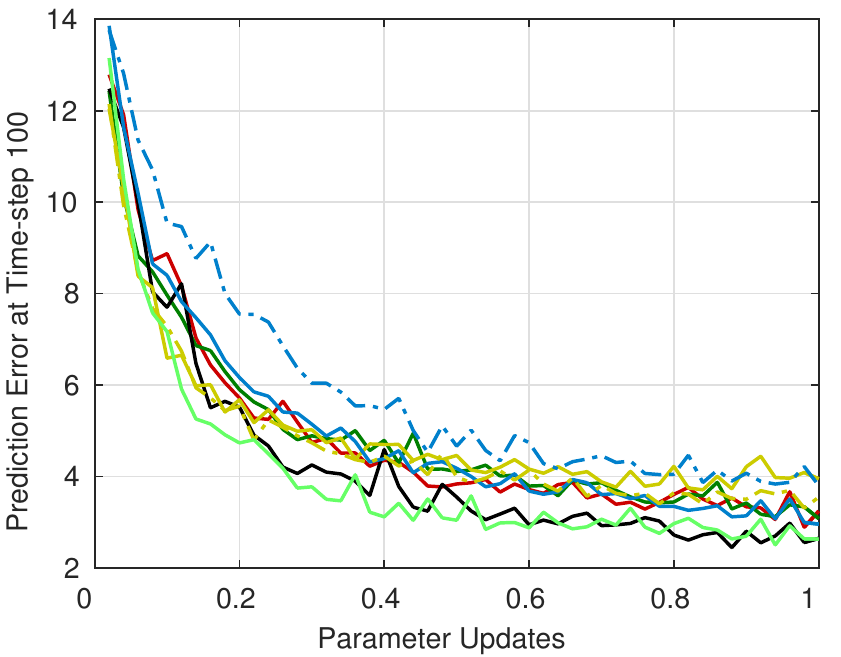}}}
\caption{Prediction error for different ways of incorporating the action on (a) Ms Pacman and (b) Pong.}
\label{fig:predErrActionMsPacman-Pong}
\end{figure}

\begin{figure}[htbp] % Figures obtained with predErrActionAppendix
\vskip-0.5cm
\scalebox{0.79}{\includegraphics[]{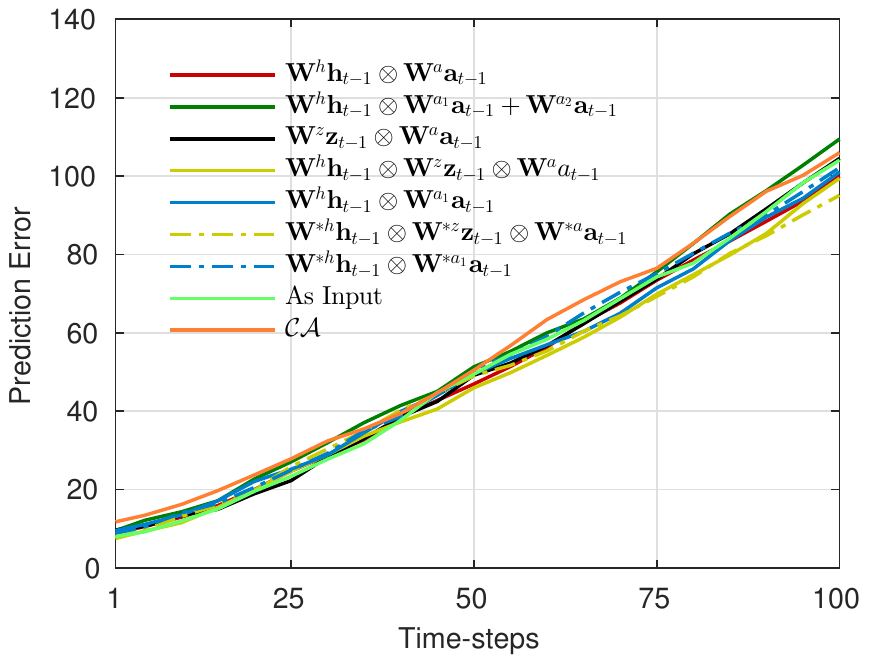}}
%\hskip0.1cm
\scalebox{0.79}{\includegraphics[]{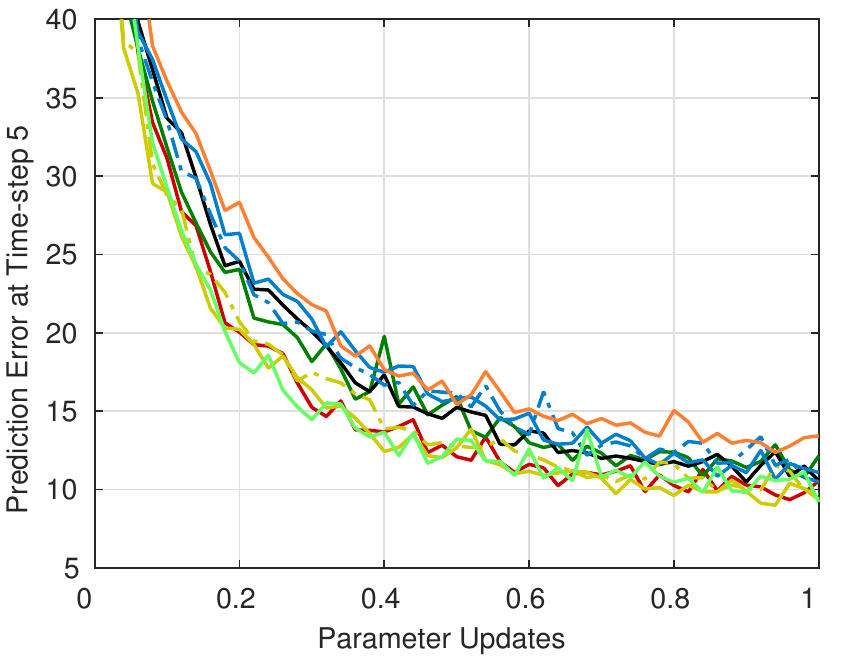}}
\subfigure[]{
\scalebox{0.79}{\includegraphics[]{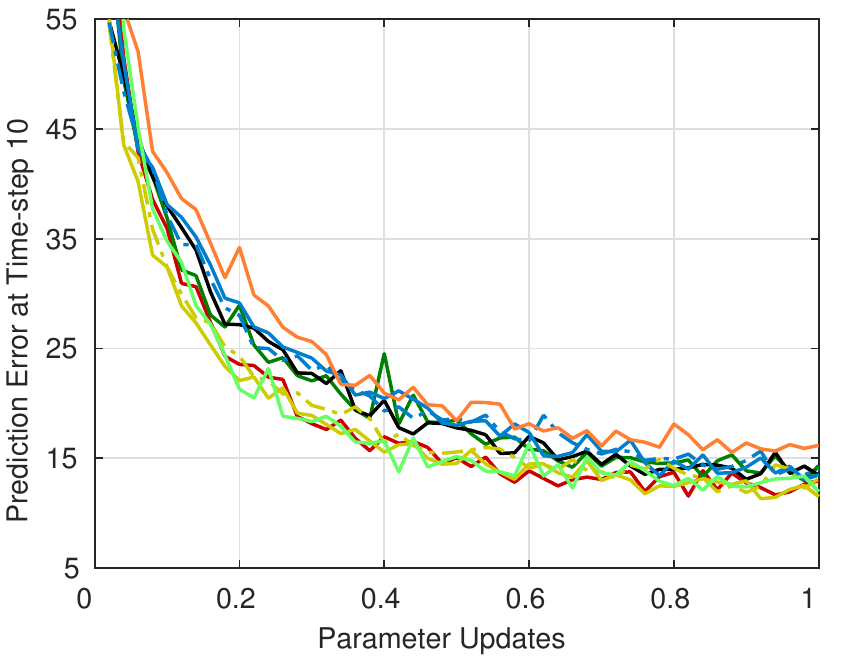}}
%\hskip0.1cm
\scalebox{0.79}{\includegraphics[]{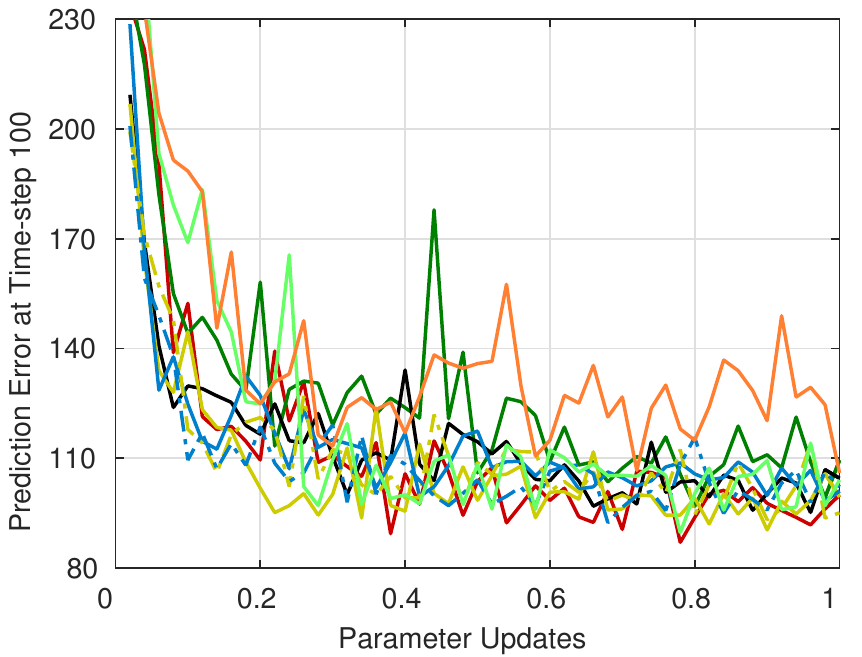}}}
\scalebox{0.79}{\includegraphics[]{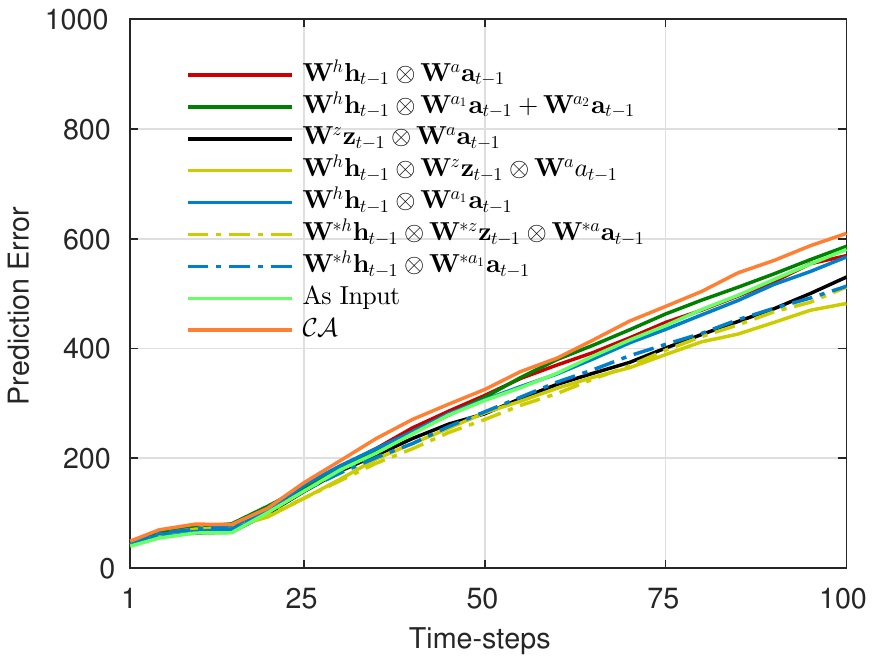}}
%\hskip0.1cm
\scalebox{0.79}{\includegraphics[]{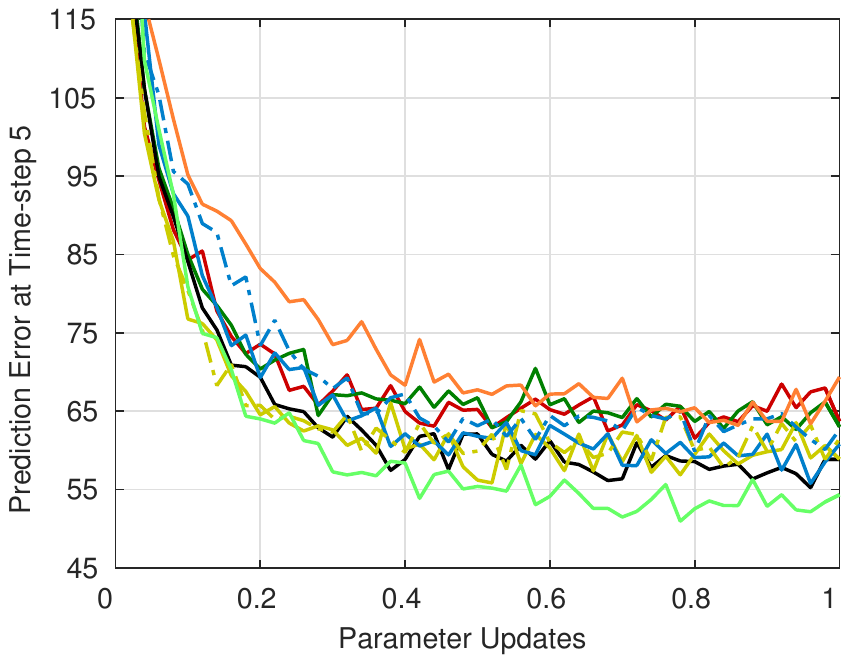}}
\subfigure[]{
\scalebox{0.79}{\includegraphics[]{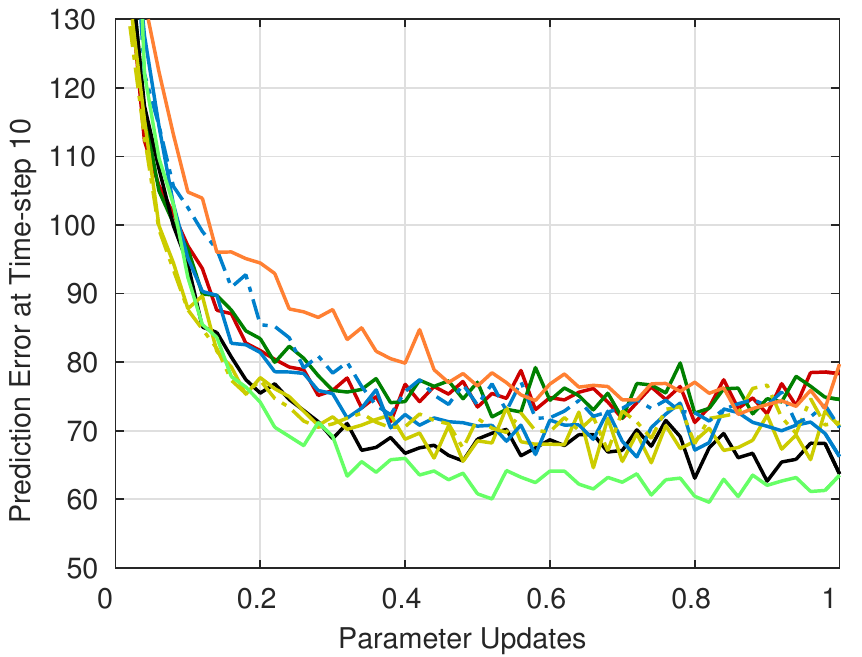}}
%\hskip0.1cm
\scalebox{0.79}{\includegraphics[]{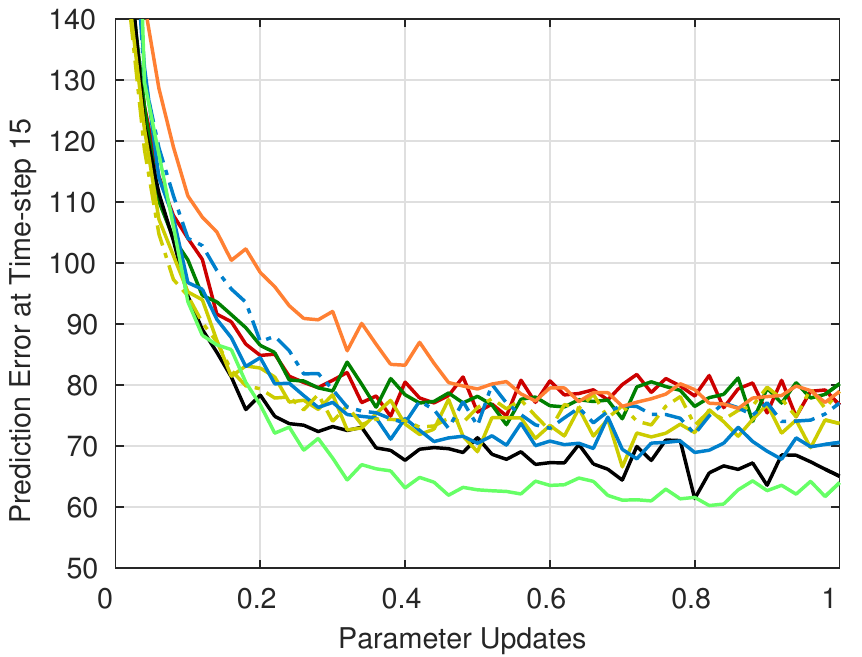}}}
\caption{Prediction error for different ways of incorporating the action on (a) Qbert and (b) Riverraid.}
\label{fig:predErrActionQbert-Riverraid}
\end{figure}

\begin{figure}[htbp] % Figures obtained with predErrActionAppendix
\vskip-0.5cm
\scalebox{0.79}{\includegraphics[]{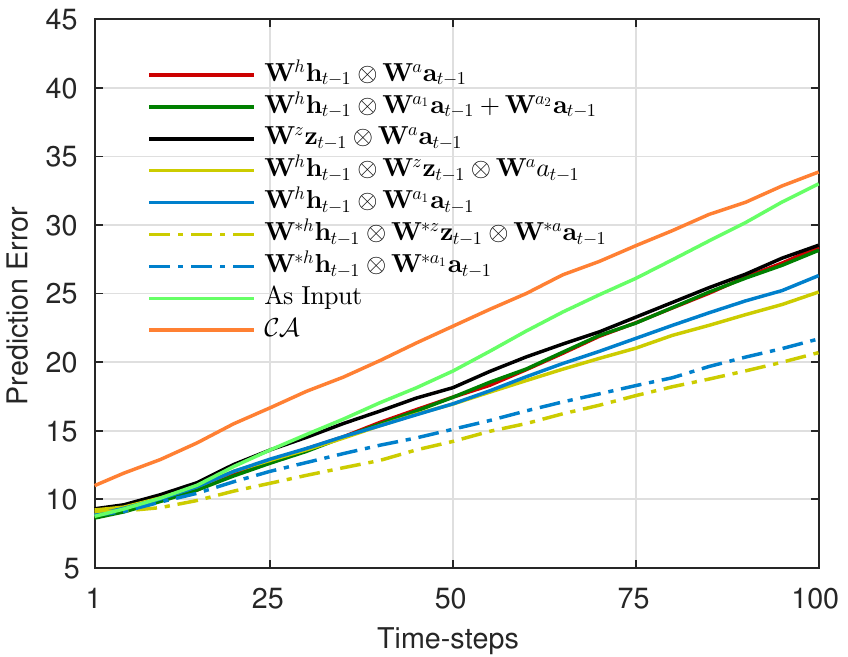}}
%\hskip0.1cm
\scalebox{0.79}{\includegraphics[]{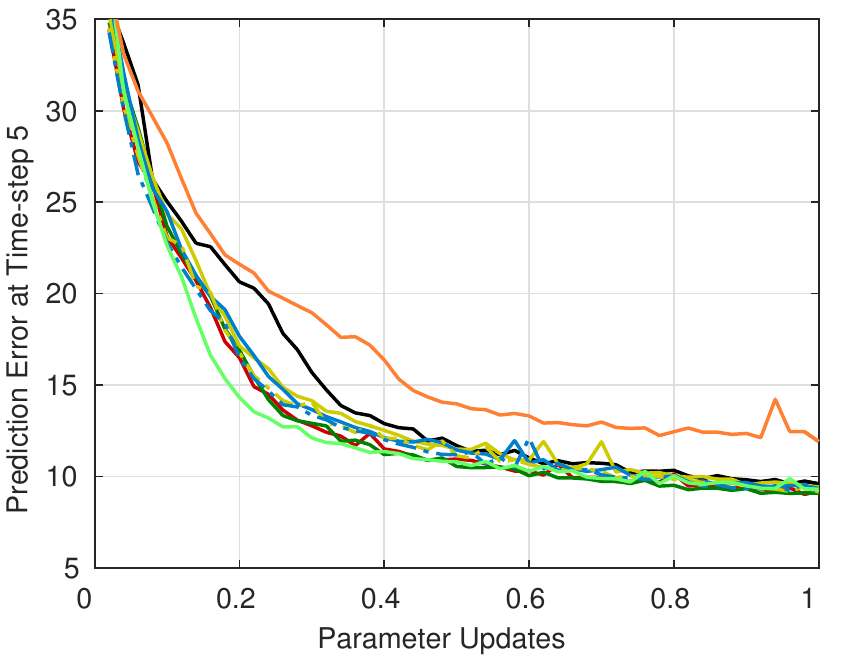}}
\subfigure[]{
\scalebox{0.79}{\includegraphics[]{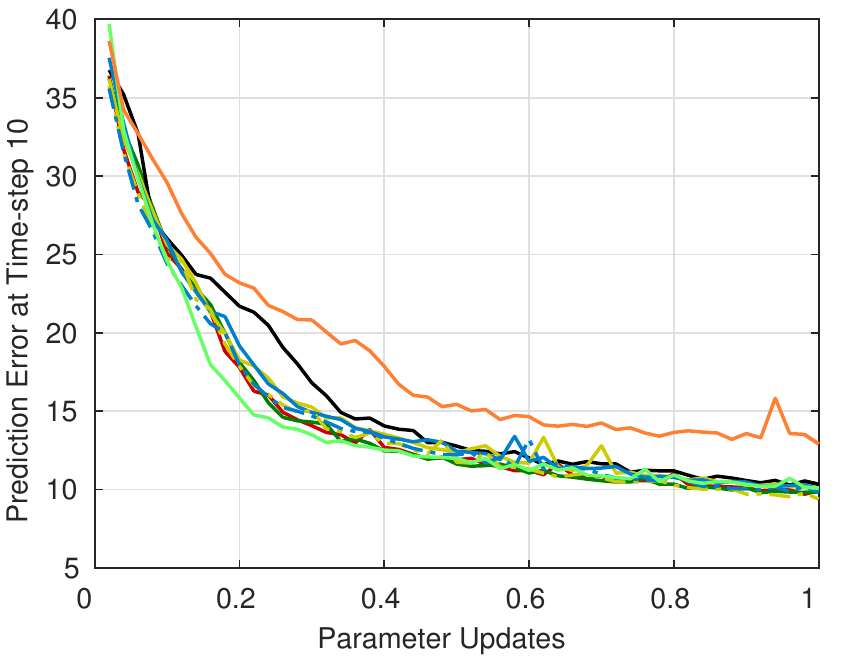}}
%\hskip0.1cm
\scalebox{0.79}{\includegraphics[]{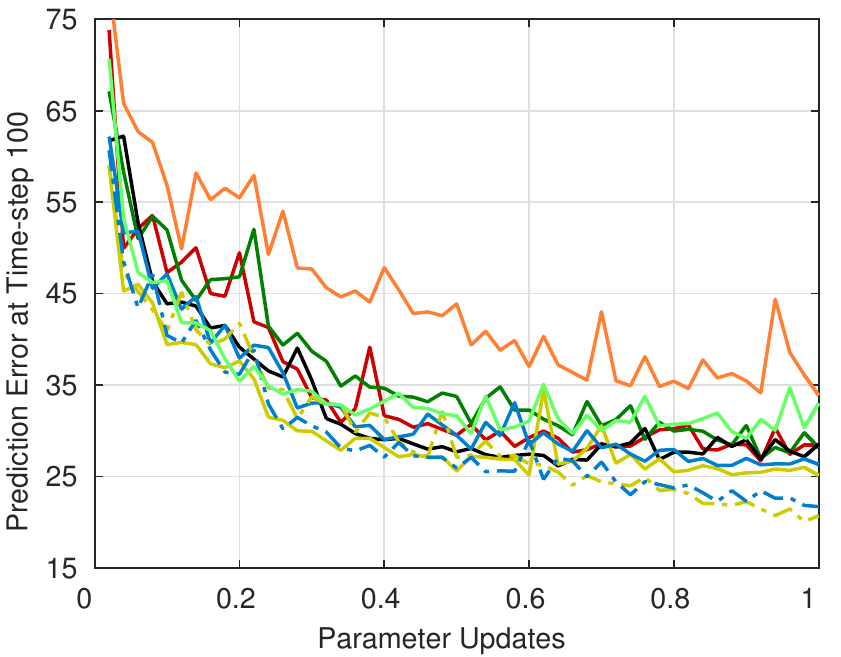}}}
\scalebox{0.79}{\includegraphics[]{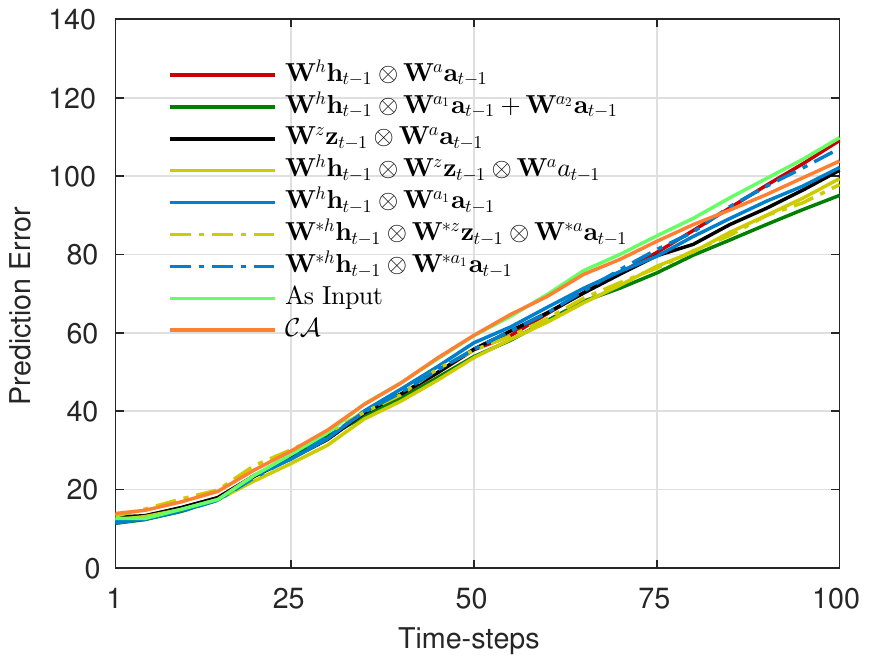}}
%\hskip0.1cm
\scalebox{0.79}{\includegraphics[]{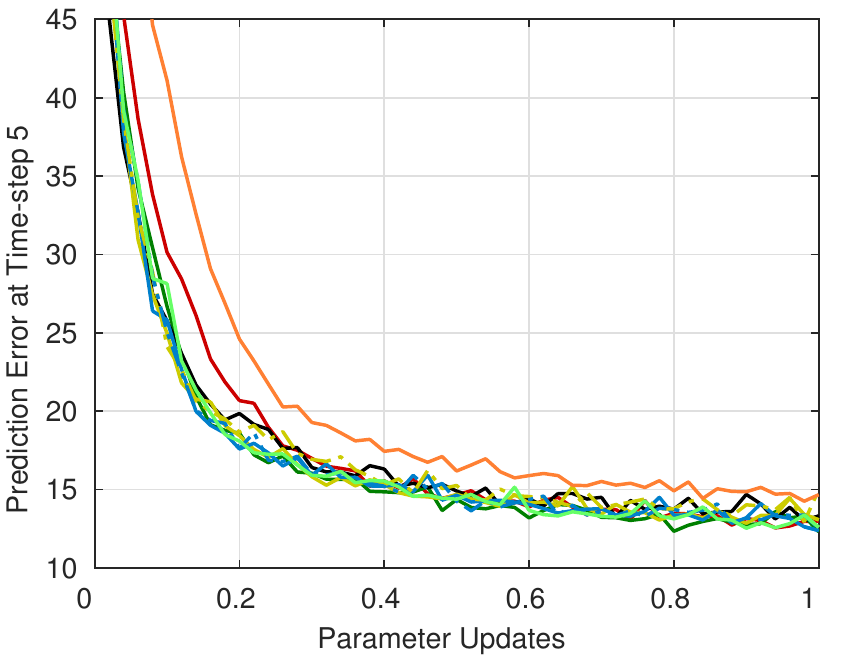}}
\subfigure[]{
\scalebox{0.79}{\includegraphics[]{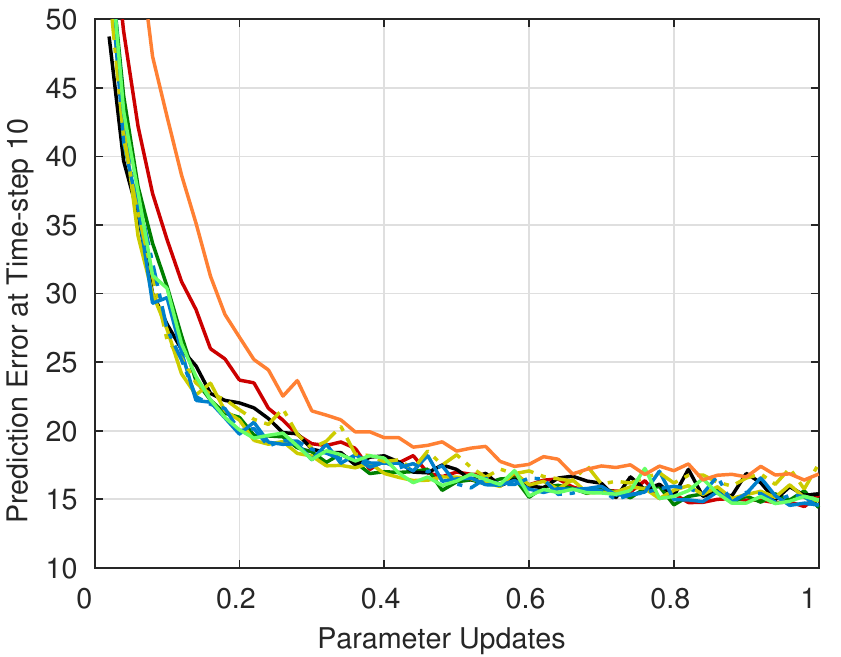}}
%\hskip0.1cm
\scalebox{0.79}{\includegraphics[]{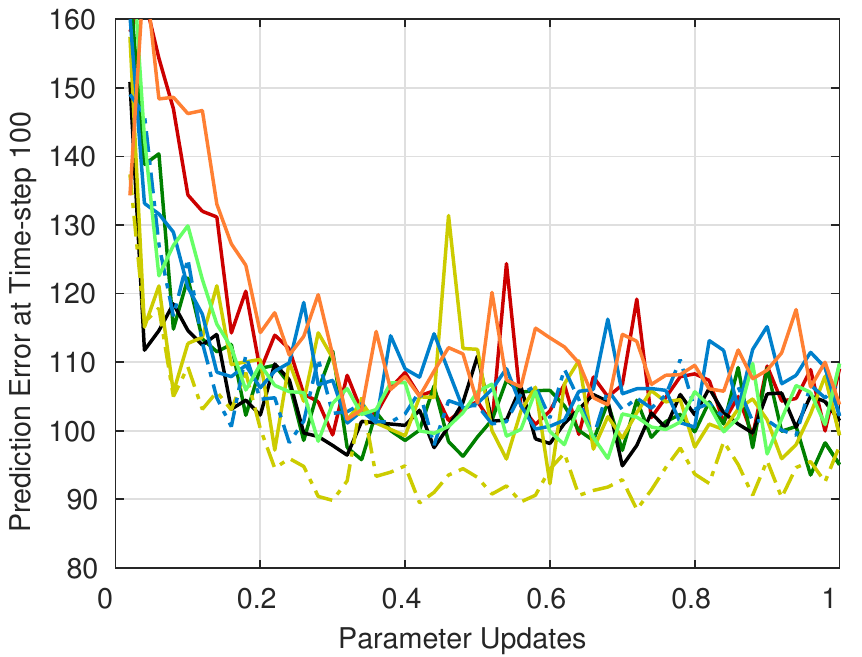}}}
\caption{Prediction error for different ways of incorporating the action on (a) Seaquest and (b) Space Invaders.}
\label{fig:predErrActionSeaquest-SpaceInvaders}
\end{figure}

In Figs. \ref{fig:predErrActionBowling-Breakout}-\ref{fig:predErrActionSeaquest-SpaceInvaders} we compare different ways of incorporating the action for action-dependent state transitions, 
using prediction length $T=15$ and the 0\%-100\%\PDT~training scheme.
More specifically, we compare the baseline structure (denoted as '$\vW^h \vh_{t-1}\otimes \vW^{a}\va_{t-1}$' in the figures) with the following alternatives: 
\begin{description} 
\item[$\bullet$ $\vW^h \vh_{t-1}\otimes \vW^{a_1}\va_{t-1}+\vW^{a_2}\va_{t-1}$:] %ATA5e6Tr1e6Tse0.2MPfalsebS16nH10nHid1024nLHid1024nF2048mask2-15-1.1-15-1SEnF64lR1e-5MMAI
Multiplicative/additive interaction of the action with $\vh_{t-1}$, \ie~
$\ha_t=\vW^h \vh_{t-1}\otimes \vW^{a_1}\va_{t-1}+\vW^{a_2}\va_{t-1}$.
This model has around 25M parameters.
\item[$\bullet$ $\vW^s \vs_{t-1}\otimes \vW^{a}\va_{t-1}$:] %ATA5e6Tr1e6Tse0.2MPfalsebS16nH10nHid1024nLHid1024nF2048mask2-15-1.1-15-1SEnF64lR1e-5MIXO
Multiplicative interaction of the action with the encoded frame $\vz_{t-1}$, \ie
\begin{align*}
& \ha_{t}=\vW^z \vz_{t-1}\otimes \vW^{a}\va_{t-1}\,, \\
& \vi_t = {\sigma}(\vW^{ih}\vh_{t-1} +\vW^{iv}\ha_{t})\,, \hskip0.3cm \vf_t = {\sigma}(\vW^{fh}\vh_{t-1} +\vW^{fv}\ha_{t}) \,,\\ 
& \vo_t = {\sigma}(\vW^{oh}\vh_{t-1} + \vW^{ov}\ha_{t})\,,\\
& \vc_t = \vf_t\otimes \vc_{t-1}+\vi_t\otimes \textrm{tanh}(\vW^{ch}\vh_{t-1}+\vW^{cv}\ha_{t}) \,.
\end{align*}
This model has around 22M parameters. 
\item[$\bullet$ $\vW^h \vh_{t-1}\otimes \vW^z \vz_{t-1}\otimes \vW^{a}\va_{t-1}$:] %ATA5e6Tr1e6Tse0.2MPfalsebS16nH10nHid1024nLHid1024nF2048mask2-15-1.1-15-1SEnF64lR1e-5MNoXGIX
Multiplicative interaction of the action with both $\vh_{t-1}$ and $\vz_{t-1}$ in the following way
\begin{align*}
&\ha_{t}=\vW^h \vh_{t-1}\otimes \vW^z \vz_{t-1}\otimes \vW^{a}\va_{t-1}\,, \\
&\vi_t = {\sigma}(\vW^{iv}\ha_{t})\,, \hskip0.3cm \vf_t = {\sigma}(\vW^{fv}\ha_{t})\,, \hskip0.3cm \vo_t = {\sigma}(\vW^{ov}\ha_{t})\,,\\
&\vc_t = \vf_t\otimes \vc_{t-1}+\vi_t\otimes \textrm{tanh}(\vW^{cv}\ha_{t}) \,.
\end{align*}
This model has around 19M parameters. 
%ATA5e6Tr1e6Tse0.2MPfalsebS16nH10nHid1024nLHid1024nF2048mask2-15-1.1-15-1SEnF64lR1e-5MNoXGIXSI
We also considered having different matrices for the gate and cell updates (denoted in the figures as '$\vW^{*h} \vh_{t-1}\otimes \vW^{*z} \vz_{t-1}\otimes \vW^{*a}\va_{t-1}$'). 
This model has around 43M parameters.
\item[$\bullet$ $\vW^h\vh_{t-1}\otimes \vW^{a_1}\va_{t-1}$:] %ATA5e6Tr1e6Tse0.2MPfalsebS16nH10nHid1024nLHid1024nF2048mask2-15-1.1-15-1SEnF64lR1e-5MIX
Alternative multiplicative interaction of the action with $\vh_{t-1}$ and $\vz_{t-1}$ 
\begin{align*}
&\ha^1_{t} = \vW^h \vh_{t-1}\otimes \vW^{a_1}\va_{t-1}\,, \hskip0.3cm \ha^2_{t} = \vW^z \vz_{t-1}\otimes \vW^{a_2}\va_{t-1}\,, \\
&\vi_t = {\sigma}(\vW^{iv_1}\ha^1_{t} +\vW^{iv_2}\ha^2_{t})\,, \hskip0.3cm \vf_t = {\sigma}(\vW^{fv_1}\ha^1_{t}+\vW^{fv_2}\ha^2_{t})\,,\\
&\vo_t = {\sigma}(\vW^{ov^1}\ha^1_{t}+\vW^{ov^2}\ha^2_{t})\,,\\
&\vc_t = \vf_t\otimes \vc_{t-1}+\vi_t\otimes \textrm{tanh}(\vW^{cv^1}\ha^1_{t}+\vW^{cv^2}\ha^2_{t}) \,.
\end{align*}
This model has around 28M parameters. % ATA5e6Tr1e6Tse0.2MPfalsebS16nH10nHid1024nLHid1024nF2048mask2-15-1.1-15-1SEnF64lR1e-5MIXSI
We also considered having different matrices for the gate and cell updates (denoted in the figures as '$\vW^{*h} \vh_{t-1}\otimes \vW^{*a_1}\va_{t-1}$'). 
This model has around 51M parameters.
\item[$\bullet$ As Input:] %ATA5e6Tr1e6Tse0.2MPfalsebS16nH10nHid1024nLHid1024nF2048mask2-15-1.1-15-1SEnF64lR1e-5MAI 
Consider the action as an additional input,  \ie~
\begin{align*}
&\vi_t = {\sigma}(\vW^{ih}\vh_{t-1} +\vW^{iz}\vz_{t-1} +\vW^{ia}\va_{t-1}) \,,\\  
&\vf_t = {\sigma}(\vW^{fh}\vh_{t-1} +\vW^{fz}\vz_{t-1}+\vW^{fa}\va_{t-1}) \,,\\ 
&\vo_t = {\sigma}(\vW^{oh}\vh_{t-1} + \vW^{oz}\vz_{t-1}+\vW^{oa}\va_{t-1})\,,\\
&\vc_t = \vf_t\otimes \vc_{t-1}+\vi_t\otimes \textrm{tanh}(\vW^{ch}\vh_{t-1}+\vW^{cz}\vz_{t-1}+\vW^{ca}\va_{t-1})\,.
\end{align*}
This model has around 19M parameters. 
\item[$\bullet$ ${\cal CA}$:] % ATA5e6Tr1e6Tse0.2MPfalsebS16nH10nHid1024nLHid1024nF2048mask2-15-1.1-15-1SEnF64lR1e-5MDanA 
Combine the action with the frame, by replacing the encoding with
\begin{align*}
\vz_{t-1}={\cal C}({\cal A}(\mathbb{I}(\hat \vx_{t-1}, \vx_{t-1}), \va_{t-1}))\,,
\end{align*}
where ${\cal A}$ indicates an augmenting operation: the frame of dimension $nC=3\times nH=210\times nW=160$ is augmented with $nA$ (number of actions) full-zero or full-one matrices of dimension $nH \times nW$,
producing a tensor of dimension $(nC+nA)\times nH\times nW$. 
As the output of the first convolution can be written as 
\begin{align*} 
\vy_{j,k,l} =\sum_{h=1}^{nH} \sum_{w=1}^{nW}\Big\{\sum_{i=1}^{nC}\vW^{i,j}_{h,w} \vx_{i,h+dH(k-1),w+dW(l-1)}+\vx_{nC+a,h+dH(k-1),w+dW(l-1)}\Big\}\,,
\end{align*} 
where $dH$ and $dW$ indicate the filter strides, with this augmentation the action has a local linear interaction. 
This model has around 19M parameters. 
\end{description}
As we can see from the figures, '${\cal C}{\cal A}$' is generally considerably worse than the other structures. %The structure with 
%multiplicative interaction of the action with both $\vh_{t-1}$ and the encoded frame ('$\vs_{t-1}$$\vW^h \vh_{t-1}\otimes \vW^s \vs_{t-1}\otimes \vW^{a}\va_{t-1}$') seems overall preferable to the '$\vW^h \vh_{t-1}\otimes \vW^{a}\va_{t-1}$' structure, followed by the structure '$\vW^s \vs_{t-1}\otimes \vW^{a}\va_{t-1}$'.
%Notice that these structures in which the action is combined with the encoded frame cannot be used in the jumpy model.
\subsubsection*{Action-Independent versus Action-Dependent State Transition\label{sec:AppNIPSA}}
\begin{figure}[htbp] % Obtained with predErrNIPSA1 and predErrNIPSA2
\vskip-0.5cm
%\centering
\scalebox{0.79}{\includegraphics[]{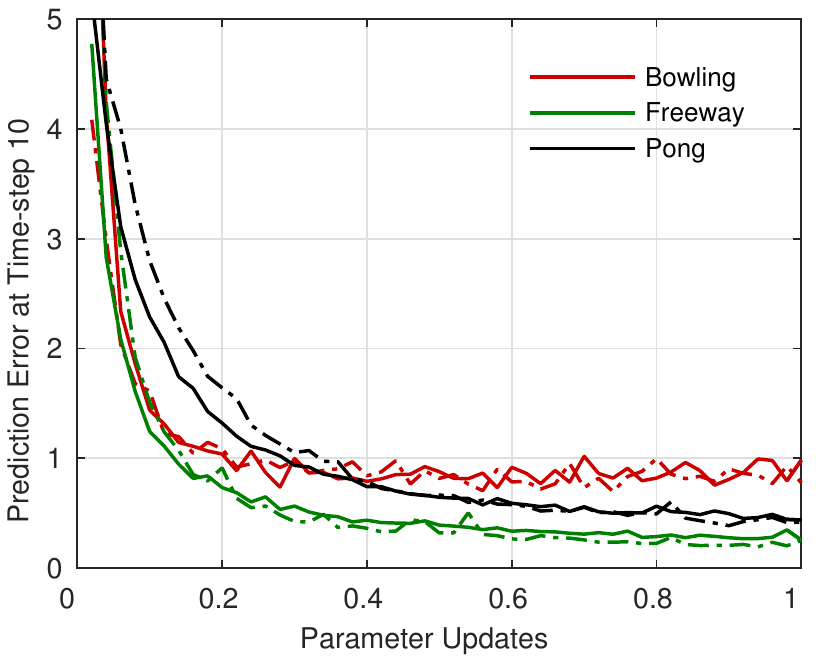}}
%\hskip0.1cm
\scalebox{0.79}{\includegraphics[]{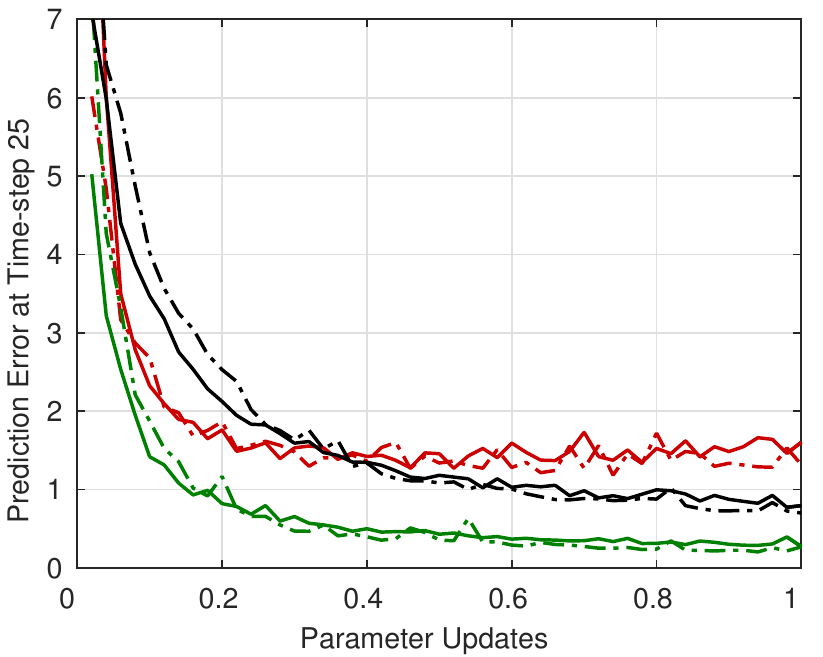}}
%\hskip0.1cm
\scalebox{0.79}{\includegraphics[]{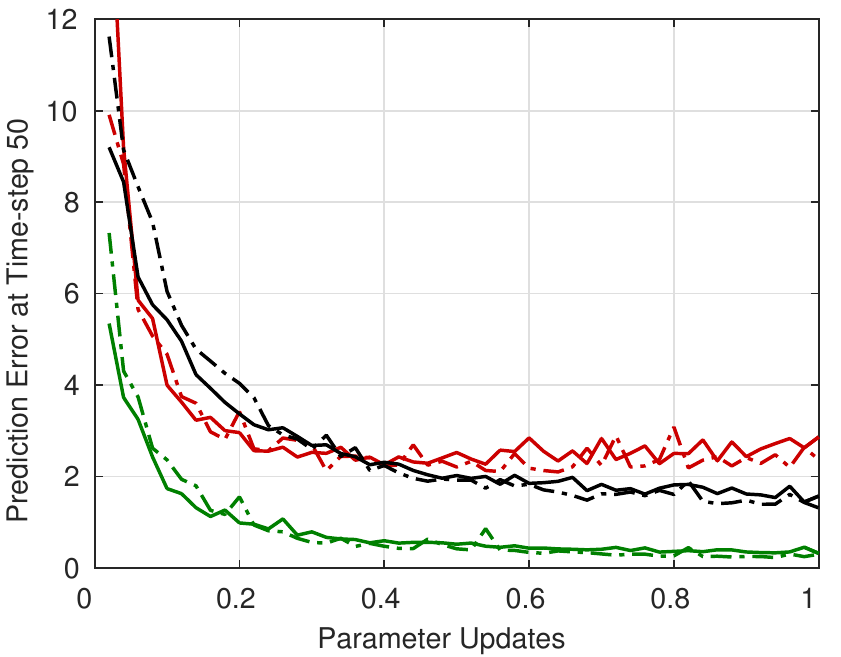}}
%\hskip0.1cm
\scalebox{0.79}{\includegraphics[]{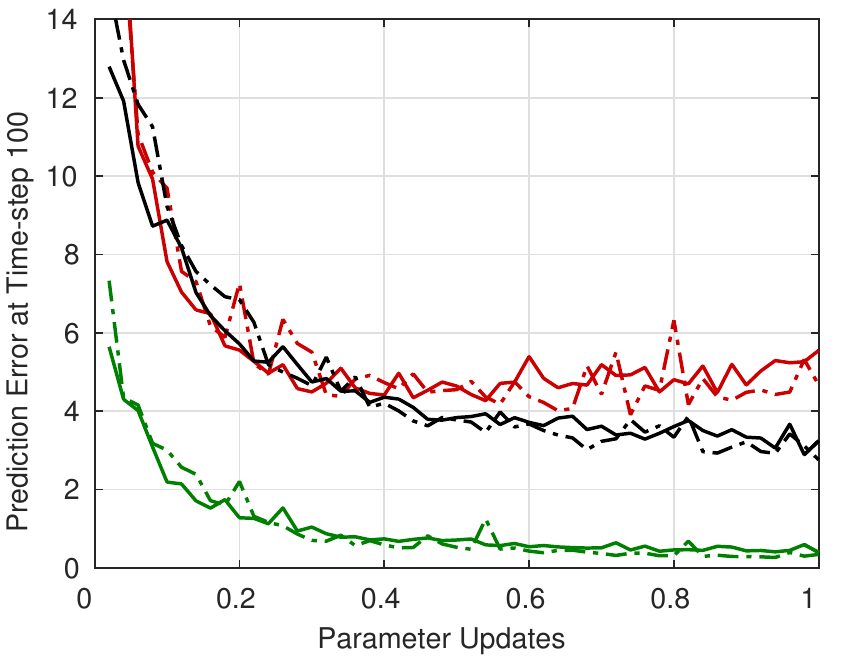}}
%\hskip0.1cm
\scalebox{0.79}{\includegraphics[]{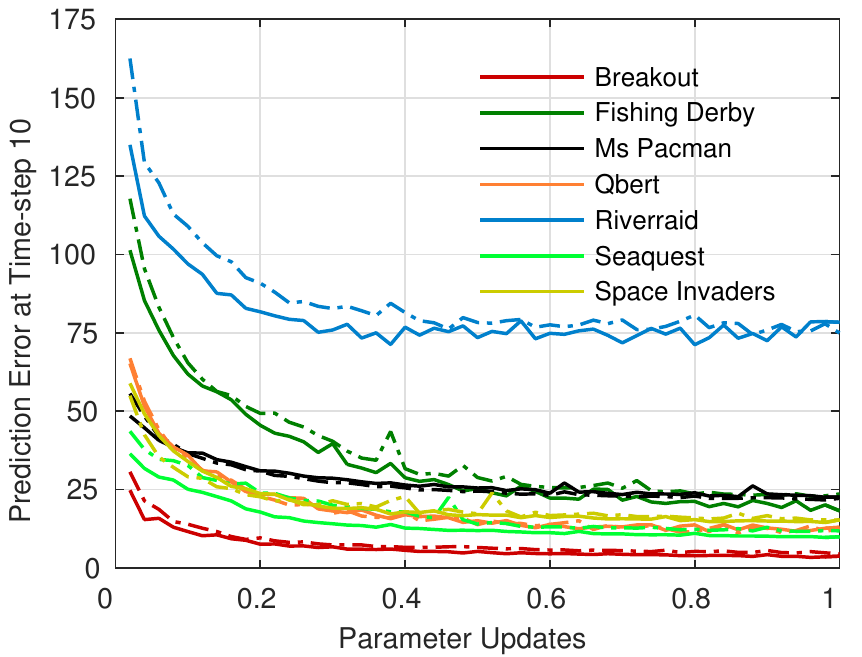}}
%\hskip0.1cm
\scalebox{0.79}{\includegraphics[]{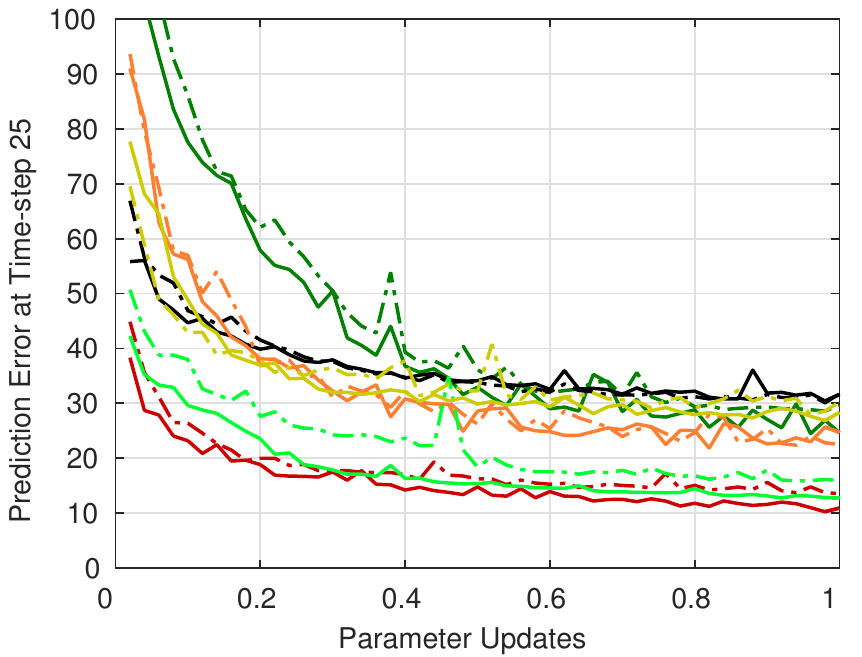}}
%\hskip0.1cm
\scalebox{0.79}{\includegraphics[]{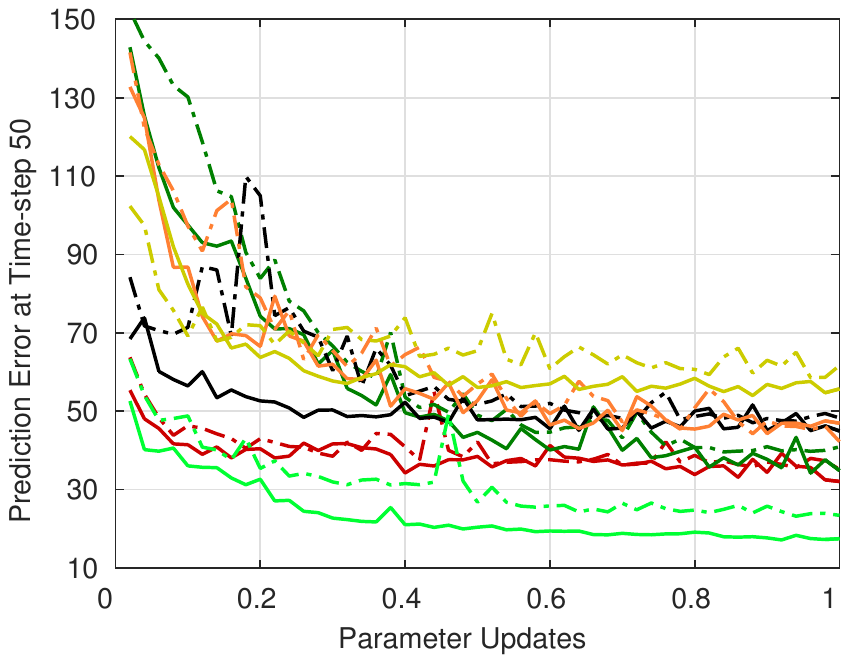}}
%\hskip0.1cm
\scalebox{0.79}{\includegraphics[]{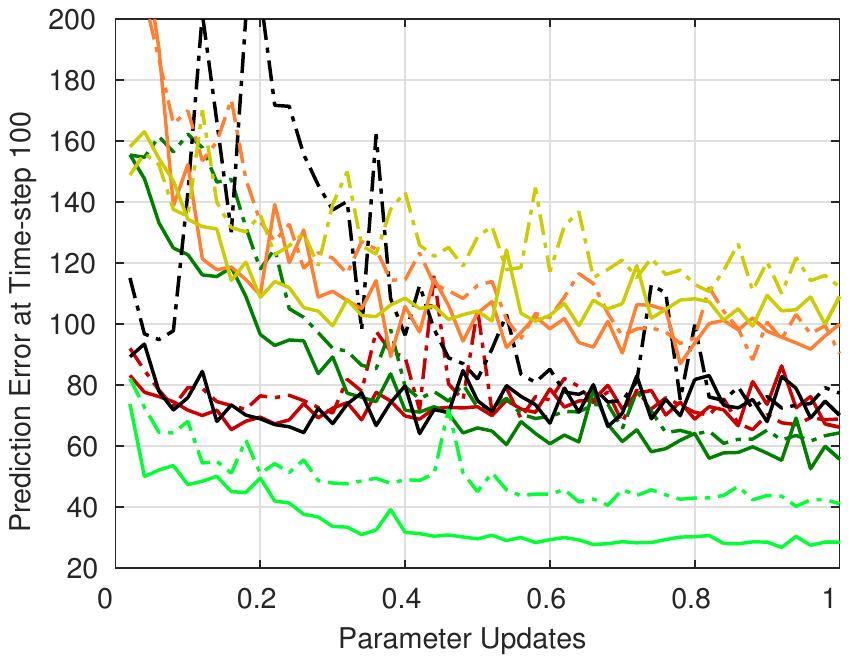}}
\caption{Prediction error (average over 10,000 sequences) with  (continuous lines) action-dependent and (dashed lines) action-independent state transition. Parameter updates are in millions.}
\label{fig:predErrAction}
\end{figure}

In \figref{fig:predErrAction}, we compare the baseline structure with one that is action-independent as in \cite{oh15action}, 
using prediction length $T=15$ and the 0\%-100\%\PDT~training scheme.

As we can see, having an an action-independent state transition generally gives worse performance in the games with higher error.
An interesting disadvantage of such a structure is its inability to predict the moving objects around the agent in Seaquest. 
This can be noticed in the videos in {\myblue \href{https://drive.google.com/file/d/0B_L2b7VHvBW2MG5LMy1Ud0V4Qk0/view?usp=sharing}{Seaquest}}, 
which show poor modelling of the fish. This structure also makes it more difficult to correctly update 
the score in some games such as Seaquest and Fishing Derby. 

% STRUCTURE
% CONVOLUTIONAL STRUCTURE
%\clearpage
\subsubsection{Human Play}
\begin{figure}[t]
\centering
\resizebox{\textwidth}{!}{\includegraphics[width=6.8cm,height=5cm]{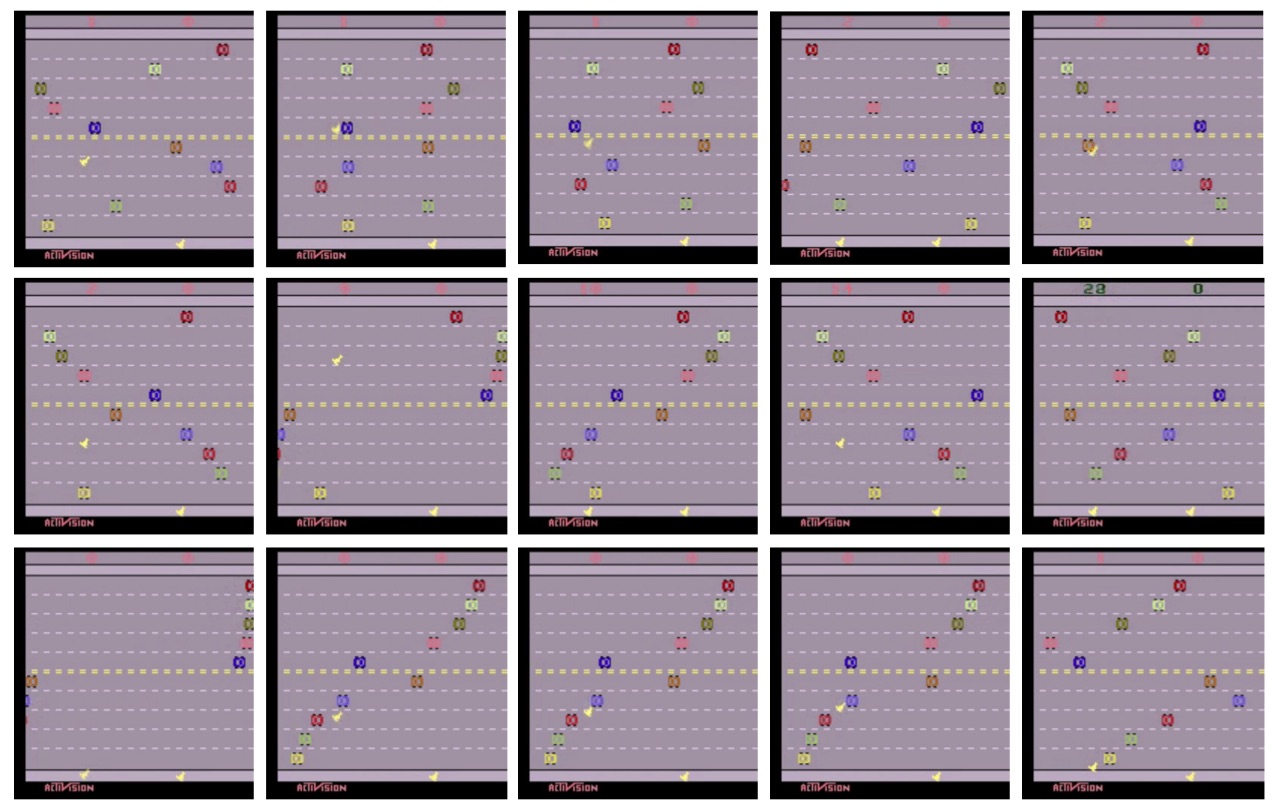}}
\caption{Salient frames extracted from 2000 frames of Freeway generated using our simulator with actions chosen by a human player.} 
\label{fig:freeway-hplay}
\end{figure}

In \figref{fig:freeway-hplay}, we show the results of a human playing Freeway for 2000 time-steps (the corresponding video is available at {\myblue \href{https://drive.google.com/open?id=0B_L2b7VHvBW2clBzNXJ6R0NOVEU}{Freeway-HPlay}}). 
The model is able to update the score correctly up to (14,0). At that point
the score starts flashing and to change color as a warn to the resetting of
the game. The model is not able to predict the 
score correctly in this warning phase, due to the bias in the data (DQN always achieves score above 20 at this point in the game), but flashing starts at the right time
as does the resetting of the game.

\begin{figure}[t]
\centering
\resizebox{\textwidth}{!}{\includegraphics[width=6.8cm,height=5.8cm]{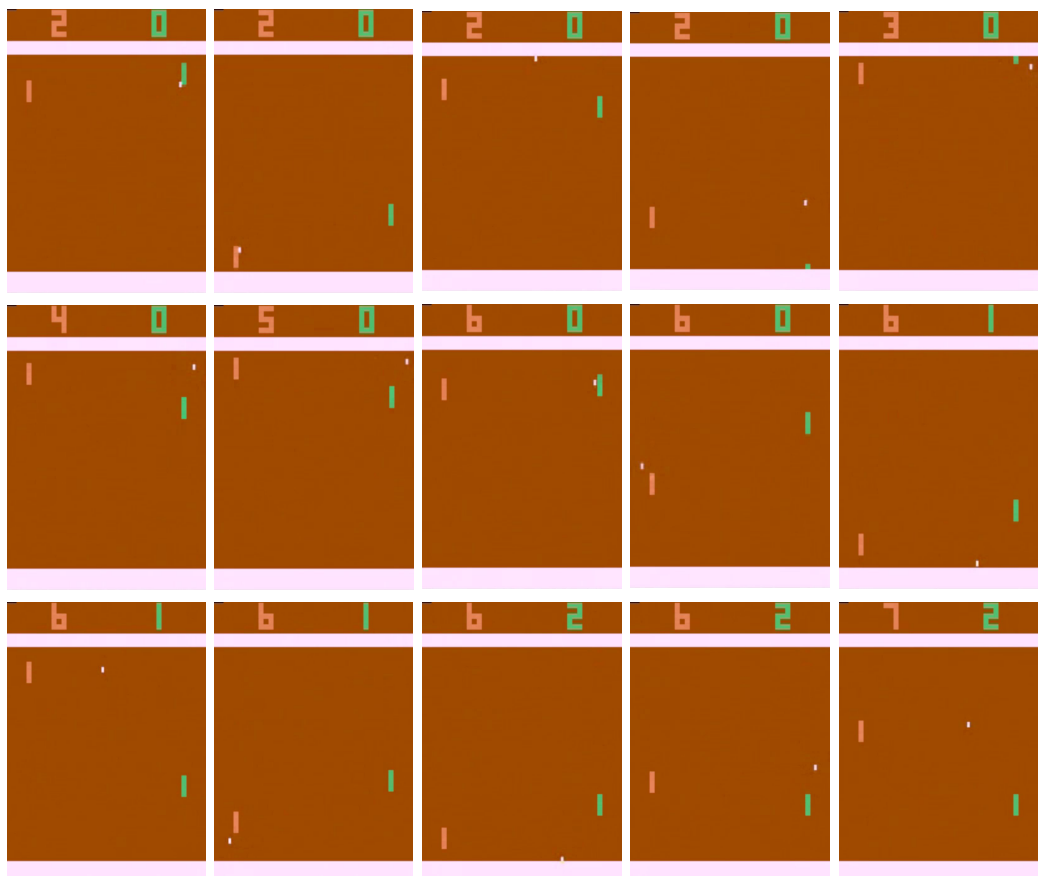}}
\caption{Salient frames extracted from 500 frames of Pong generated using our simulator
with actions chosen by a human player.}
\label{fig:pong-large}
\end{figure}

\begin{figure}[t]
\centering
\resizebox{\textwidth}{!}{\includegraphics[width=6.8cm,height=5.8cm]{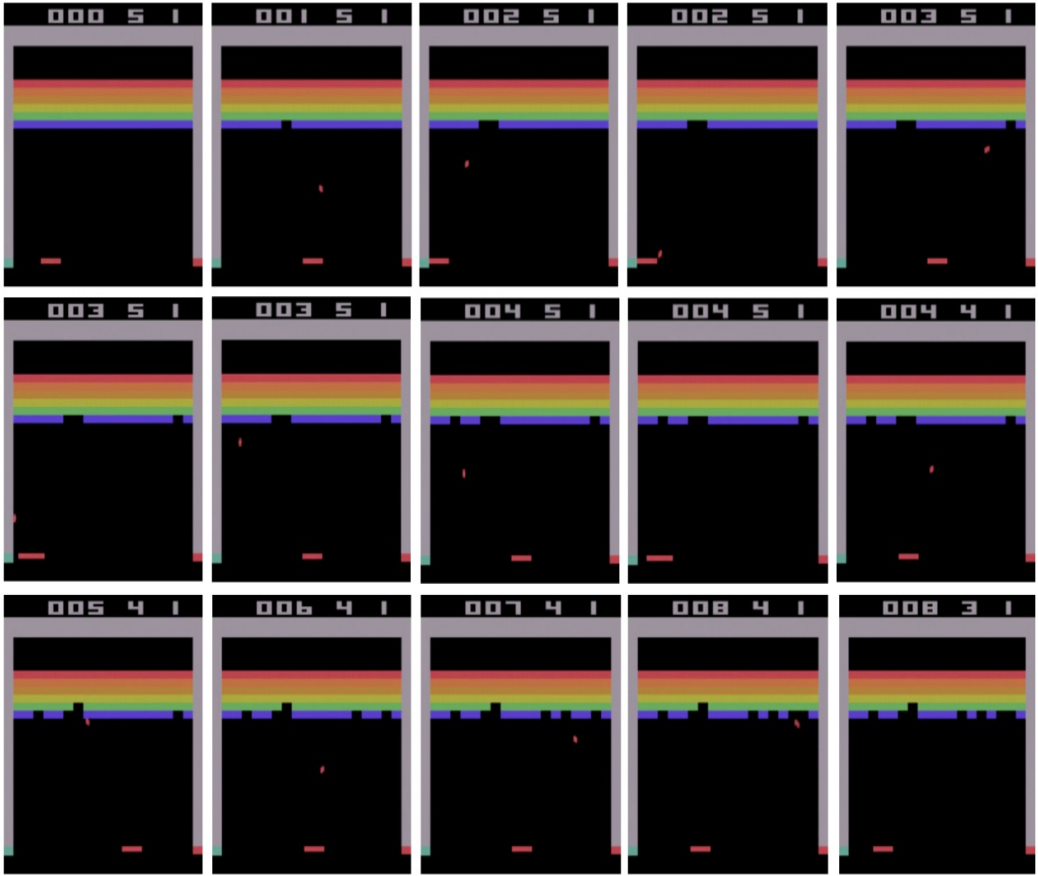}}
\caption{Salient frames extracted from 350 frames of Breakout generated using
our simulator with actions taken by a human player.}
\label{fig:breakout-large}
\end{figure}

Figs. \ref{fig:pong-large} and \ref{fig:breakout-large} are larger views of the same frames shown in \figref{fig:pong-breakout}.

% freeway_2000frames : 1, 0:40, 0:45, 2:19, 3:44, 3:51, 20:17, 20:37, 27:39, 29:01, 30:23, 30:45, 30:47, 30:48, 31:31

\subsection{3D Car Racing\label{sec:AppTorcs}}
We generated 10 and one million (180$\times$180) RGB images for training and testing respectively, with an agent trained with the asynchronous advantage actor critic algorithm (Fig. 2 in \citep{mnih16asynchronous}).
The agent could choose among the three actions accelerate straight, accelerate left, and accelerate right, according to an $\epsilon$-greedy policy, with $\epsilon$ selected at random between 0 and 0.5, independently for each episode. 
We added a 4th `do nothing' action when generating actions at random. Smaller $\epsilon$ lead to longer episodes ($\sim$1500 frames), while larger $\epsilon$ lead to shorter episodes ($\sim$200 frames). 

We could use the same number of convolutional layers, filters and kernel sizes as in Atari, with no padding.

\begin{figure}[t]
\centering
\resizebox{\textwidth}{!}{\includegraphics[]{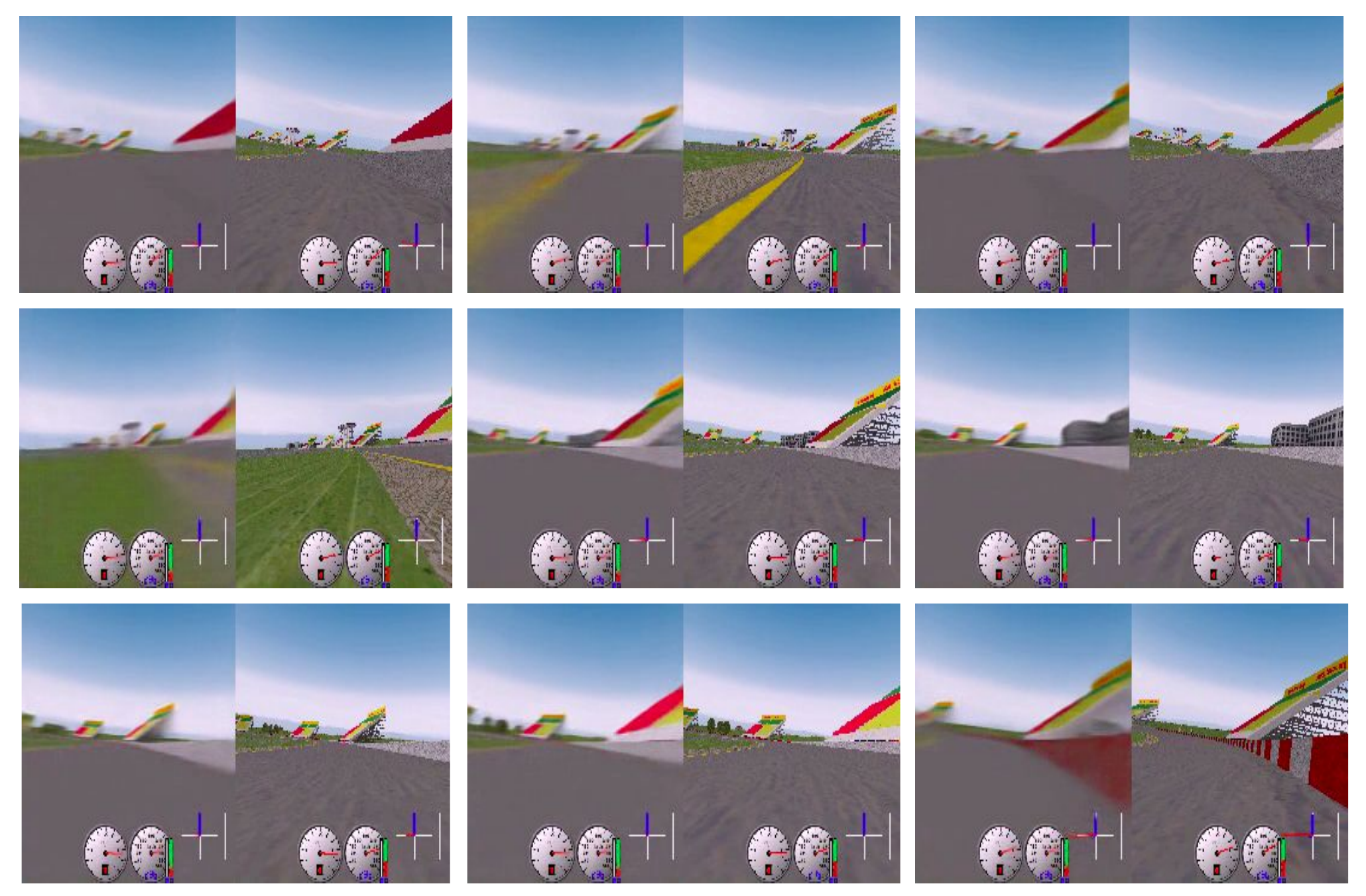}}
\caption{Salient frames, predicted (left) and real (right), for TORCS from a 200 time-steps video.}
\label{fig:torcstest}
\end{figure}

\figref{fig:torcstest} shows side by side predicted and real frames for up to $200$ actions. We found that this quality of predictions was very common.

When using our model as an interactive simulator, we observed that the car would slightly slow down when selecting no action, but fail
to stop. % unless the car was at the very beginning of the race, in which case it would not start moving as expected. 
Since the model had never seen
occurrences of the agent completely releasing the accelerator for more than
a few consecutive actions, it makes sense it would fail to deal with this case appropriately.

\subsection{3D Mazes\label{sec:AppMazes}}
Unlike Atari and TORCS, we could rely on agents with random policies to generate interesting sequences. The agent could choose one of five actions: forward, backward, rotate left, rotate right or do nothing. 
During an episode, the agent alternated between a random walk for 15 steps, and spinning on itself for 15 steps (roughly, a complete $360^\circ$ spin). This encourages coherent learning of the predicted frames after a spin. 
The random walk was with dithering of 0.7, meaning that new actions were chosen with a probability of 0.7 at every time-step. The training and test datasets were made of 7,600 and
1,100 episodes, respectively.  All episodes were of length 900 frames, resulting in 6,840,000 and 990,000 (48$\times$48) RGB images for training and testing respectively.

We adapted the encoding by having only $3$ convolutions with 64 filters of size $6\times 6$, stride 2, and padding 0, 1, and 2. The decoding transformation was adapted accordingly. 

\subsection{Model-based Exploration\label{sec:AppExploration}}
We observed that increasing the number of Monte-Carlo simulations beyond
$100$ made little to
no difference, probably because with $n_a$ possible actions the number of
possible Monte-Carlo simulations $n_a^d$ is so large that we quickly get
diminishing returns with every new simulation.

Increasing significantly the sequence length of actions beyond $d = 6$ lead to a
large decrease in
performance. To explain this, we observed that after $6$ steps, our average
prediction error was less than half the average prediction error
after $30$ steps ($0.16$ and $0.37$ respectively). Since the average minimum and
maximum distances did not vary significantly (from $0.23$ to $0.36$, and from
$0.24$ to $0.4$ respectively), for deep simulations we ended up with more noise
than signal in our predictions and our decisions were no better than random.

\begin{figure}[t]
\centering
\resizebox{\textwidth}{!}{
\begin{tabular}{c c}
 \includegraphics[]{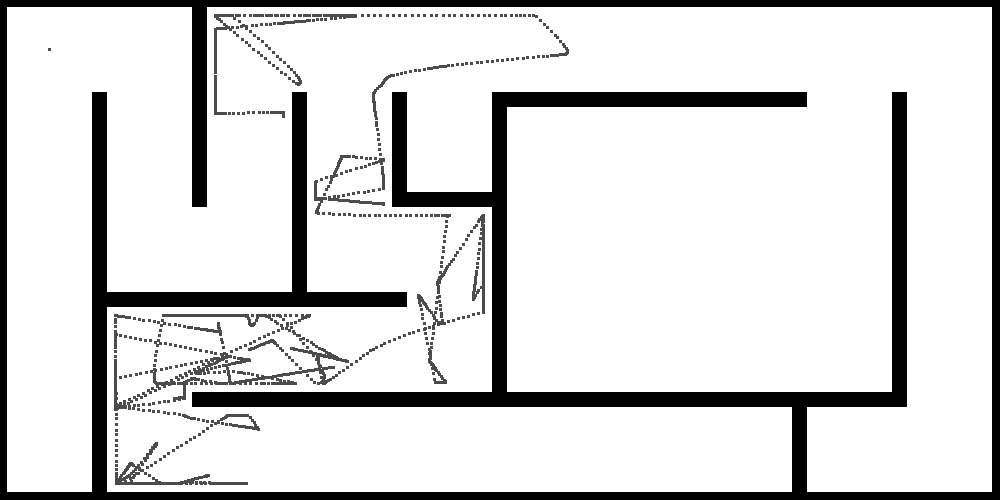} &
 \includegraphics[]{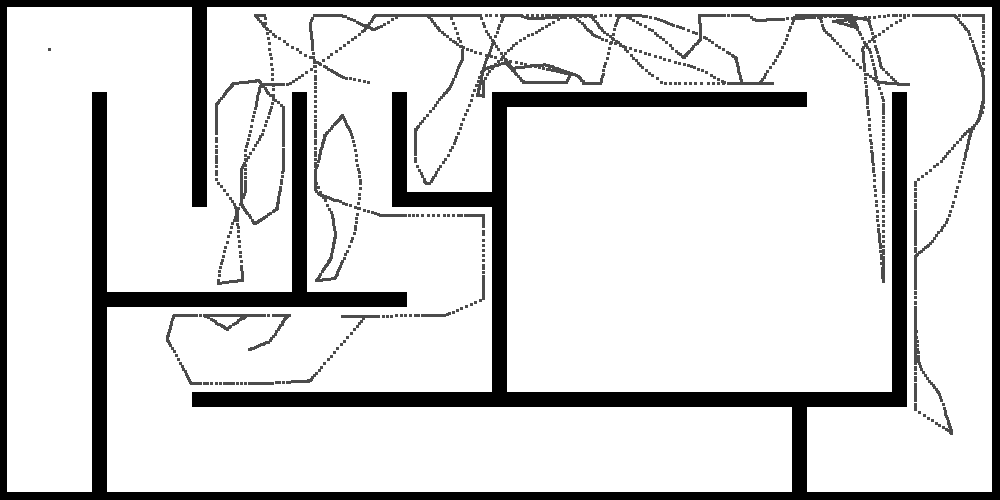} \\
 \includegraphics[]{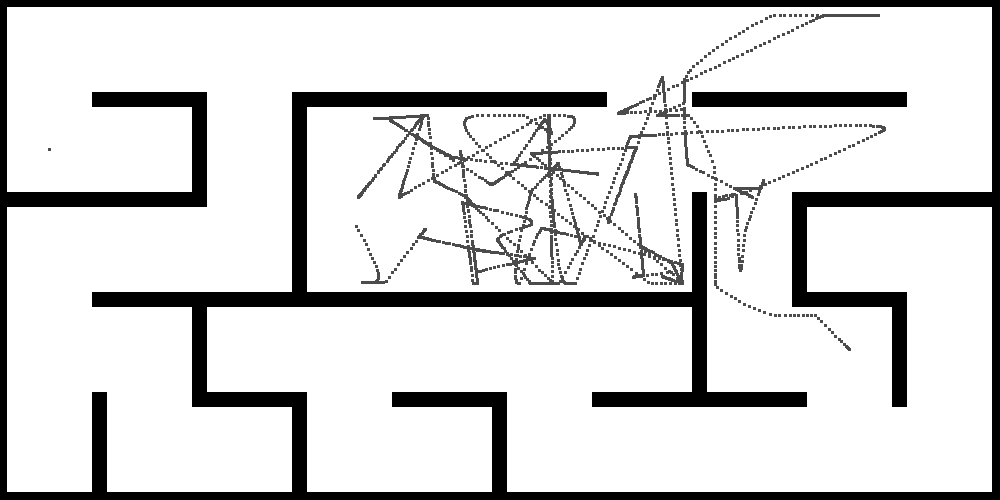} &
 \includegraphics[]{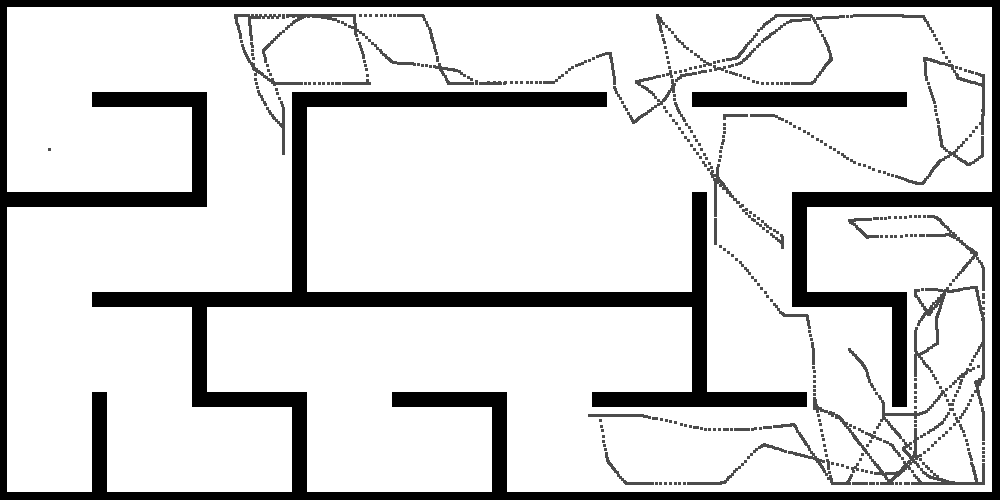} \\
 \includegraphics[]{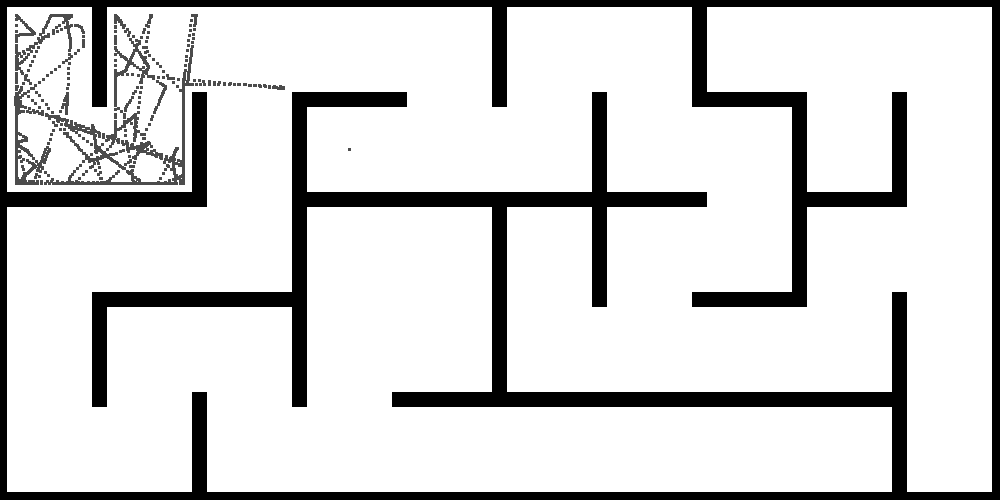} &
 \includegraphics[]{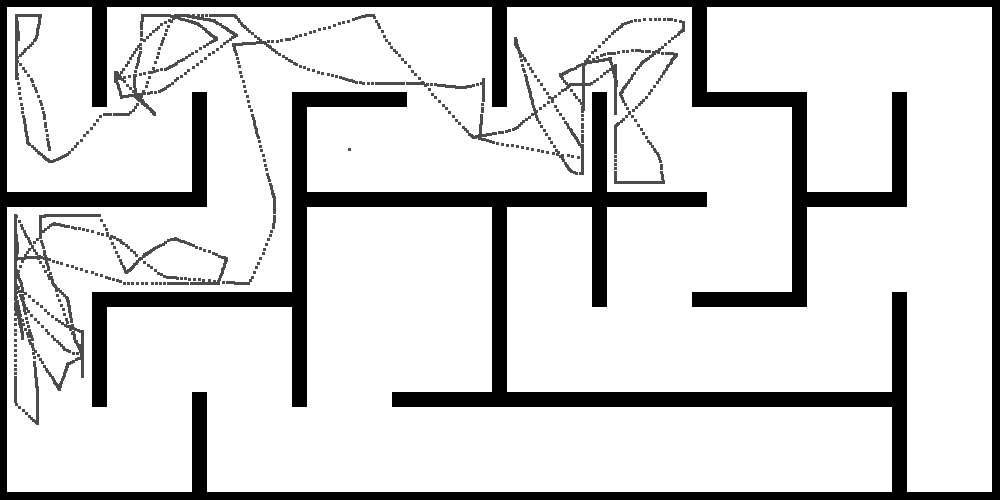} \\
 \includegraphics[]{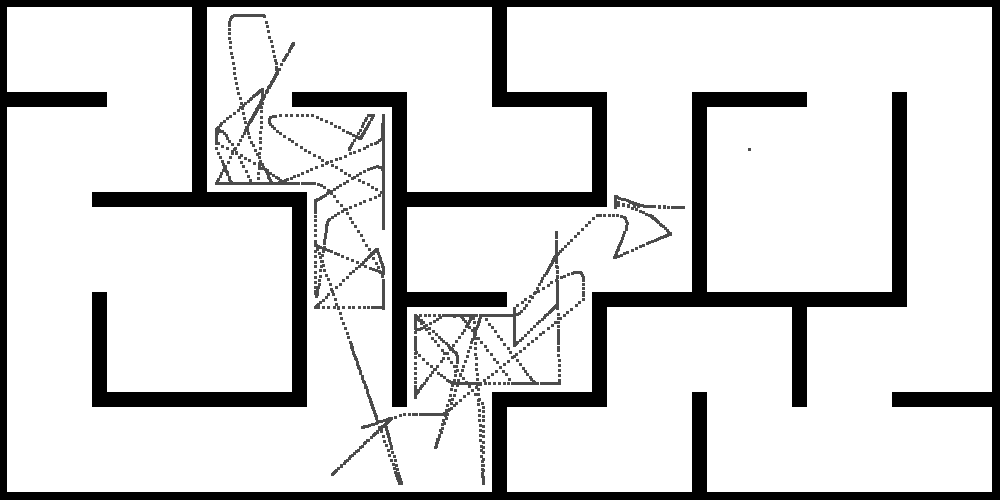} &
 \includegraphics[]{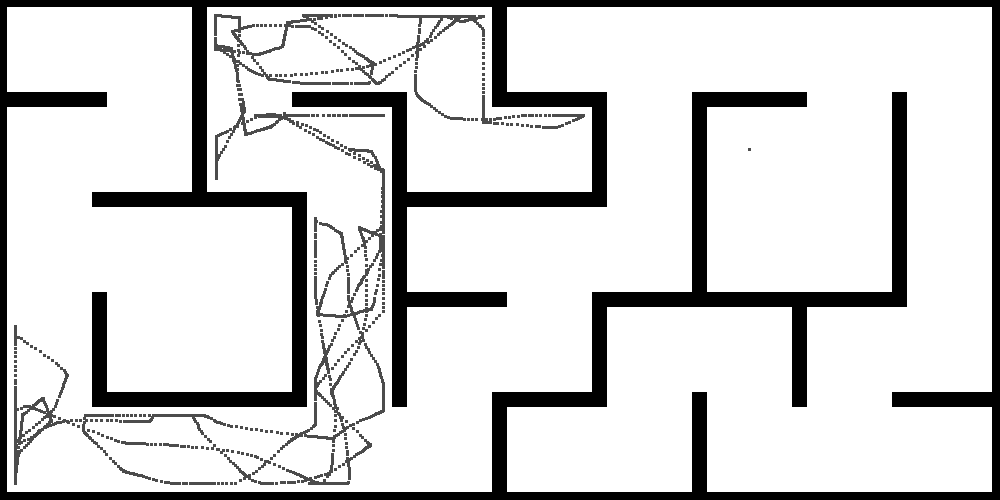}
\end{tabular}}
\caption{Examples of paths followed by random baseline (left), and 
explorers based on our simulator (right).}
\label{fig:exploration_trajectory_more_examples}
\end{figure}

\figref{fig:exploration_trajectory_more_examples} shows some examples of
trajectories chosen by our explorer. Note that all these trajectories are much
smoother than for our baseline agent.

\section{Prediction-Independent Simulators\label{sec:AppPISimulators}}
\vspace{-2mm}
In this section we compare different action-dependent state transitions and prediction lengths $T$ for the prediction-independent simulator.

\begin{figure}[t] % Figures obtained with predErrNoFBStruct
\vskip-0.7cm
\scalebox{0.79}{\includegraphics[]{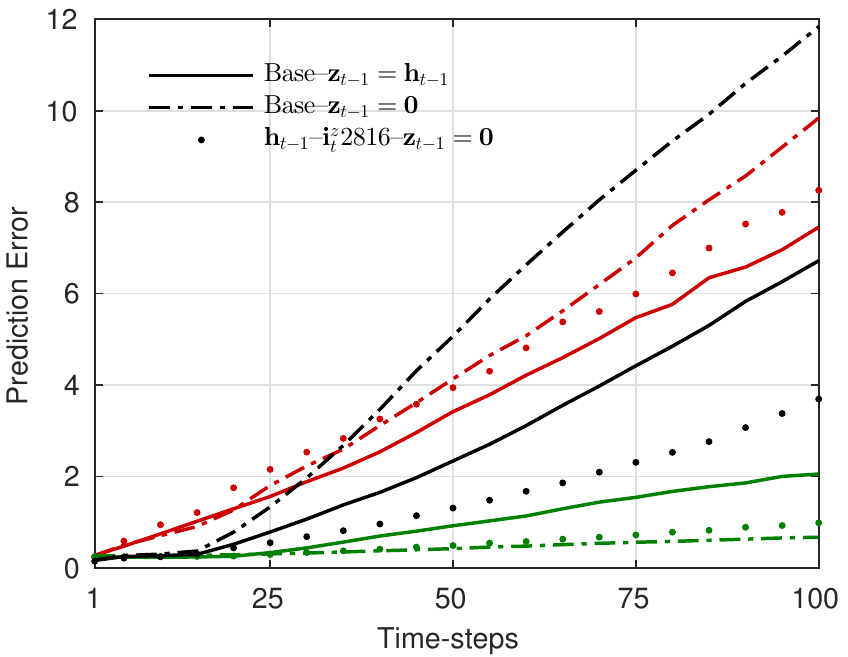}}
%\hskip0.1cm
\scalebox{0.79}{\includegraphics[]{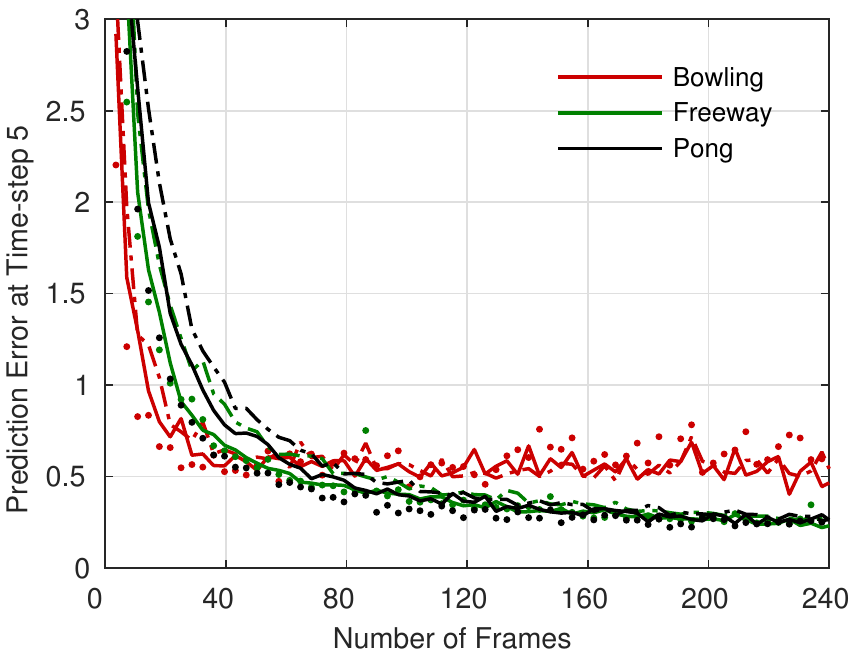}}
\subfigure[]{
\scalebox{0.79}{\includegraphics[]{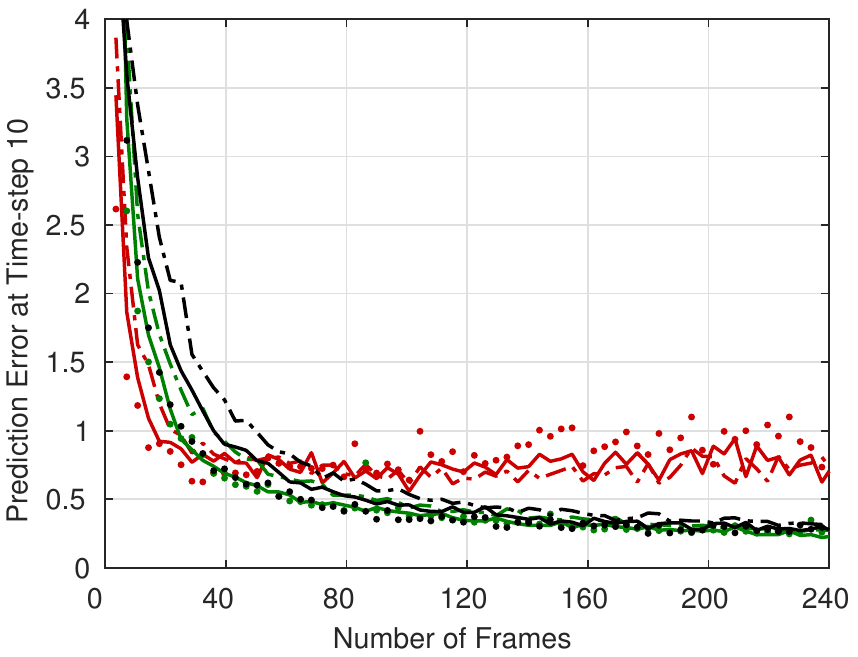}}
%\hskip0.1cm
\scalebox{0.79}{\includegraphics[]{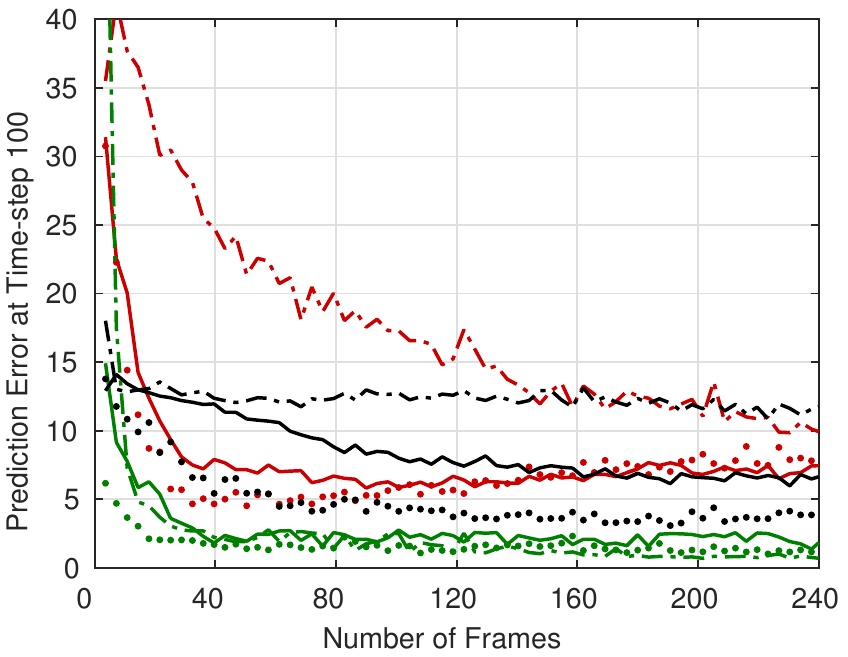}}}
\scalebox{0.79}{\includegraphics[]{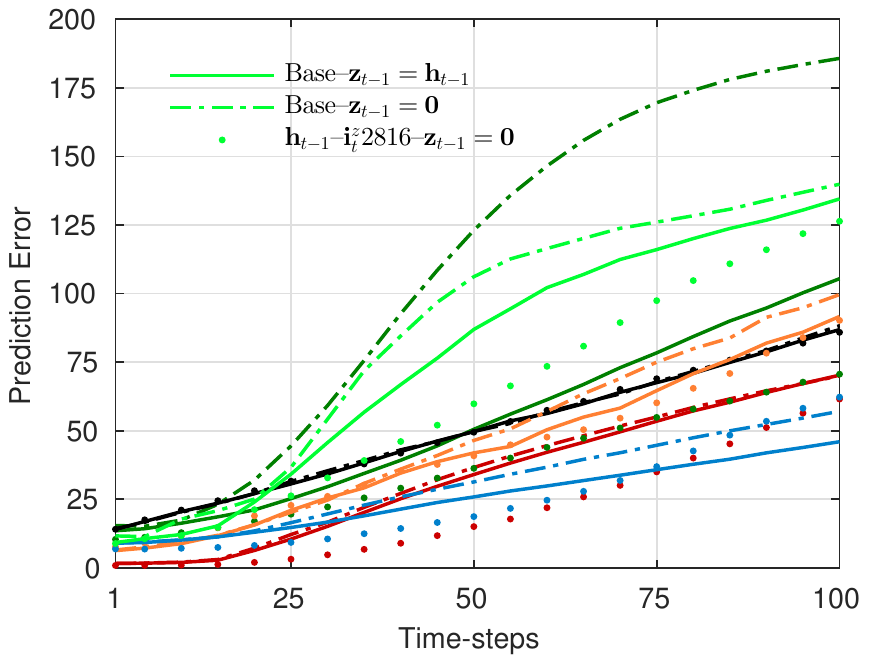}}
%\hskip0.1cm
\scalebox{0.79}{\includegraphics[]{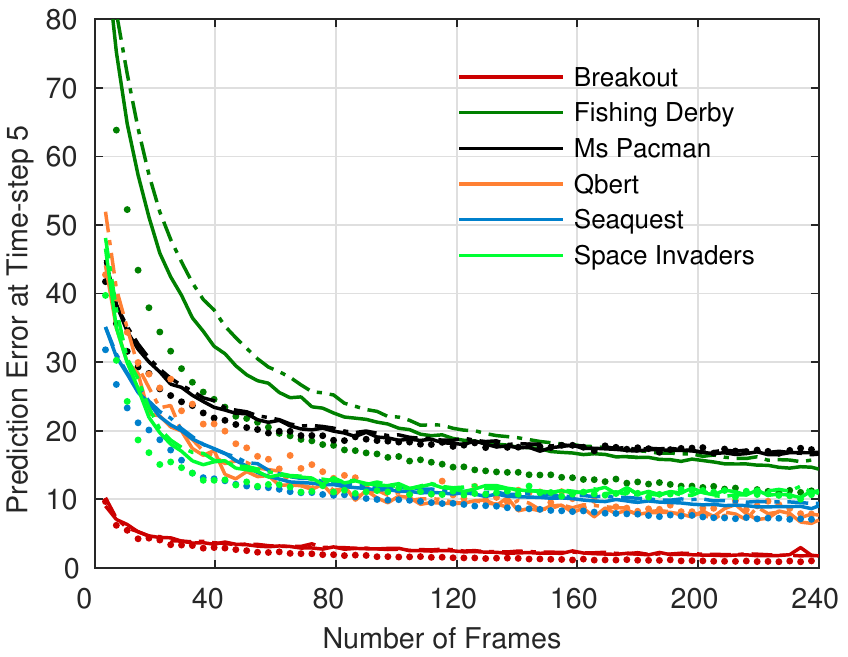}}
\subfigure[]{
\scalebox{0.79}{\includegraphics[]{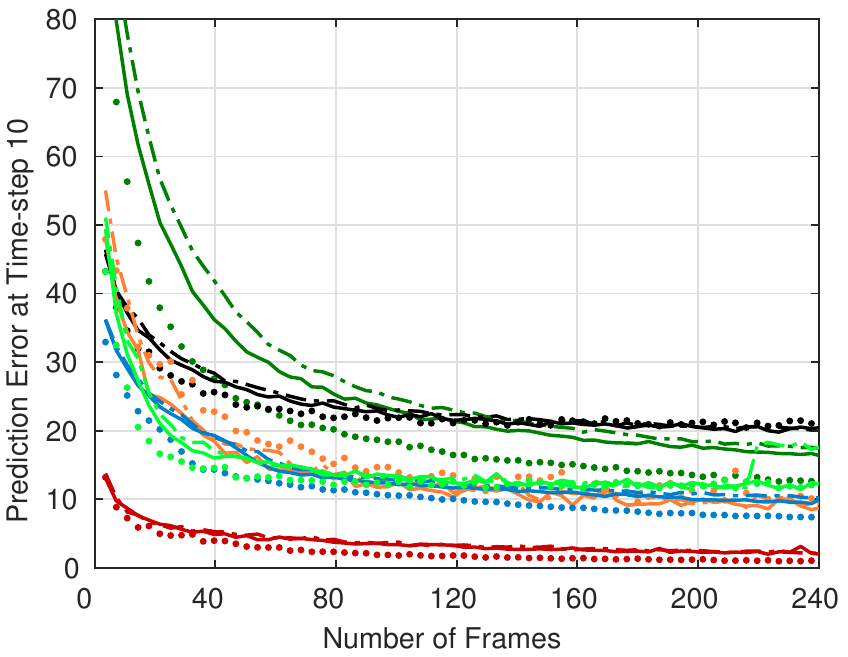}}
%\hskip0.1cm
\scalebox{0.79}{\includegraphics[]{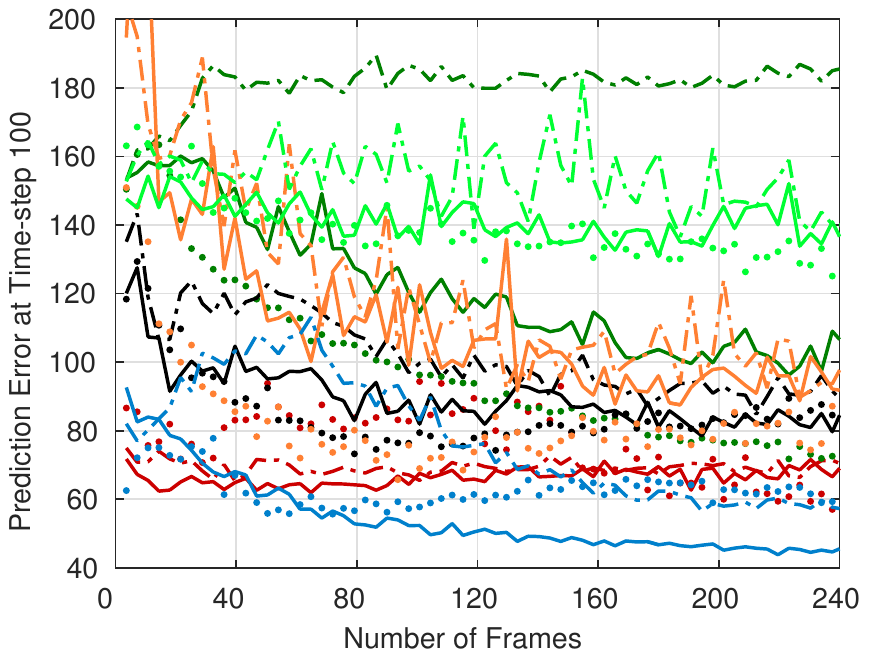}}}
\caption{Prediction error (average over 10,000 sequences) for the prediction-independent simulator with different action-dependent state transitions for (a) Bowling, Freeway, Pong, and (b)
Breakout, Fishing Derby, Ms Pacman, Qbert, Seaquest, Space Invaders. Number of frames is in million and excludes warm-up frames.}
\label{fig:predErrJumpyStruct}
\end{figure}

More specifically, in \figref{fig:predErrJumpyStruct} we compare (with $T=15$) the state transition 
\begin{align*}
\textrm{Encoding: } & \vz_{t-1}=
\begin{cases} 
{\cal C}(\vx_{t-1}) & \text{Up to $t-1 = \tau-1$}\,,\\
\vh_{t-1}  &  \text{From $t-1 = \tau$}\,,\\
\end{cases}\\
\textrm{Action fusion: } & \ha_{t}=\vW^h \vh_{t-1}\otimes \vW^{a}\va_{t-1}\,,\\
\textrm{Gate update: } & \vi_t = {\sigma}(\vW^{iv}\ha_{t} +\vW^{iz}\vz_{t-1})\,, \hskip0.3cm \vf_t = {\sigma}(\vW^{fv}\ha_t +\vW^{fz}\vz_{t-1})\,,\\ 
& \vo_t = {\sigma}(\vW^{ov}\ha_t + \vW^{oz}\vz_{t-1})\,,\\
\textrm{Cell update: } & \vc_t = \vf_t\otimes \vc_{t-1}+\vi_t\otimes \textrm{tanh}(\vW^{cv}\ha_{t}+\vW^{cz}\vz_{t-1})\,,
\end{align*}
where the vectors $\vh_{t-1}$ and $\ha_{t}$ have dimension 1024 and 2048 respectively and with different matrices $\vW$ for the warm-up and the prediction phases 
(the resulting model has around 40M parameters -- we refer to this structure as 'Base--$\vz_{t-1}=\vh_{t-1}$' in the figures), 
with the following alternatives:
\begin{description}
\item[$\bullet$ Base--$\vz_{t-1}={\mathbf 0}$:] Remove the action-independent transformation of $\vh_{t-1}$, \ie 
\begin{align*}
& \vz_{t-1}=
\begin{cases} 
{\cal C}(\vx_{t-1}) & \text{Up to $t-1 = \tau-1$}\,,\\
{\mathbf 0}  &  \text{From $t-1 = \tau$}\,,\\
\end{cases}\\
\end{align*}
where ${\mathbf 0}$ represents a zero-vector and with different matrices $\vW$ for the warm-up and the prediction phases. This model has around 40M parameters.
\item[$\bullet$ $\vh_{t-1}$--$\vi^z_t$2816--$\vz_{t-1}={\mathbf 0}$: ]
Substitute $\vz_{t-1}$ with $\vh_{t-1}$ in the gate updates and have a separate gating for the encoded frame, \ie %ATA5e6Tr1e6Tse0.2MPfalsebS16nH11nHid2816nLHid2816nF2816mask0-15-1.0-15-1SEnF64lR1e-5MNoXHGSX
\begin{align*}
& \vz_{t-1}=
\begin{cases} 
{\cal C}(\vx_{t-1}) & \text{Up to $t-1 = \tau-1$}\,,\\
{\mathbf 0}  &  \text{From $t-1 = \tau$}\,,\\
\end{cases}\\
& \ha_{t}=\vW^h \vh_{t-1}\otimes \vW^{a}\va_{t-1}\,,\\
& \vi_t = {\sigma}(\vW^{iv}\ha_{t} +\vW^{ih}\vh_{t-1})\,, \hskip0.3cm \vf_t = {\sigma}(\vW^{fv}\ha_t +\vW^{fh}\vh_{t-1})\,,\\ 
& \vo_t = {\sigma}(\vW^{ov}\ha_t + \vW^{oh}\vh_{t-1})\,,\\
& \vc_t = \vf_t\otimes \vc_{t-1}+\vi_t\otimes \textrm{tanh}(\vW^{cv}\ha_{t}+\vW^{cs}\vh_{t-1})+\vi^z_t\otimes \textrm{tanh}(\vz_{t-1})\,,
\end{align*}
with shared $\vW$ matrices for the warm-up and the prediction phases, without RReLU after the last convolution of the encoding, and 
with vectors $\vh_{t-1}$ and $\ha_t$ of dimensionality 2816. This model has around 95M parameters. 
\end{description}

As we can see, the 'Base--$\vz_{t-1}={\mathbf 0}$' state transition performs quite poorly for long-term prediction compared to the other transitions.
With this transition, the prediction-independent simulator performs much worse than the prediction-dependent simulator with the baseline state transition (\appref{sec:AppPDT}).
The best performance is obtained with the '$\vh_{t-1}$--$\vi^z_t$2816--$\vz_{t-1}={\mathbf 0}$' structure, which however has a large number of parameters.

\begin{figure}[t] % Figures obtained with predErrNoFBSeqLength
\vskip-0.5cm
\scalebox{0.79}{\includegraphics[]{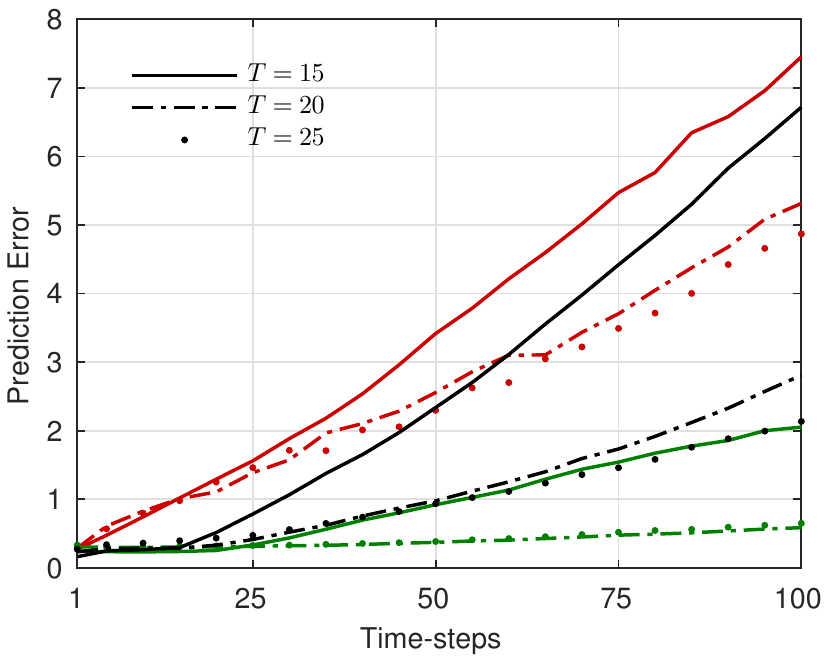}}
%\hskip0.1cm
\scalebox{0.79}{\includegraphics[]{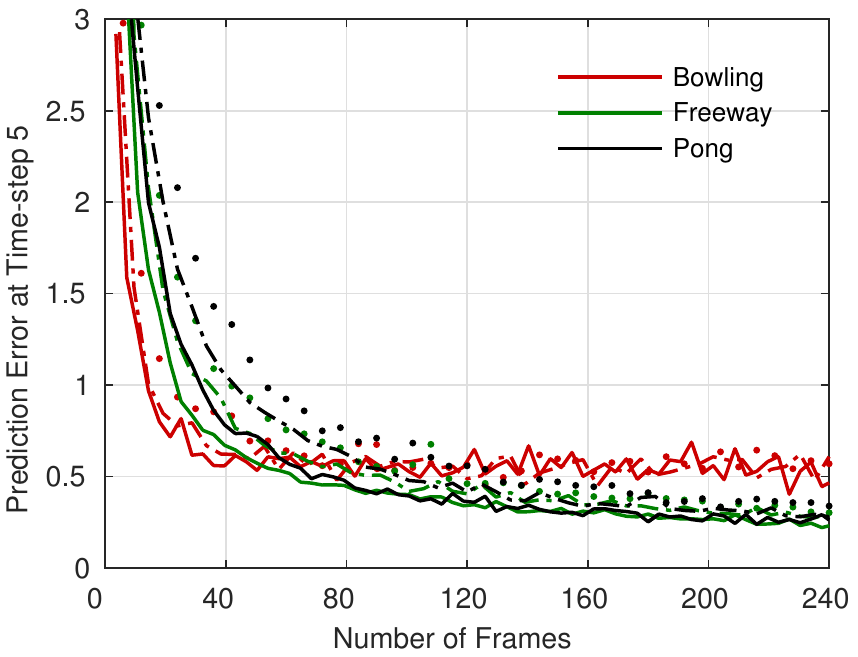}}
\subfigure[]{
\scalebox{0.79}{\includegraphics[]{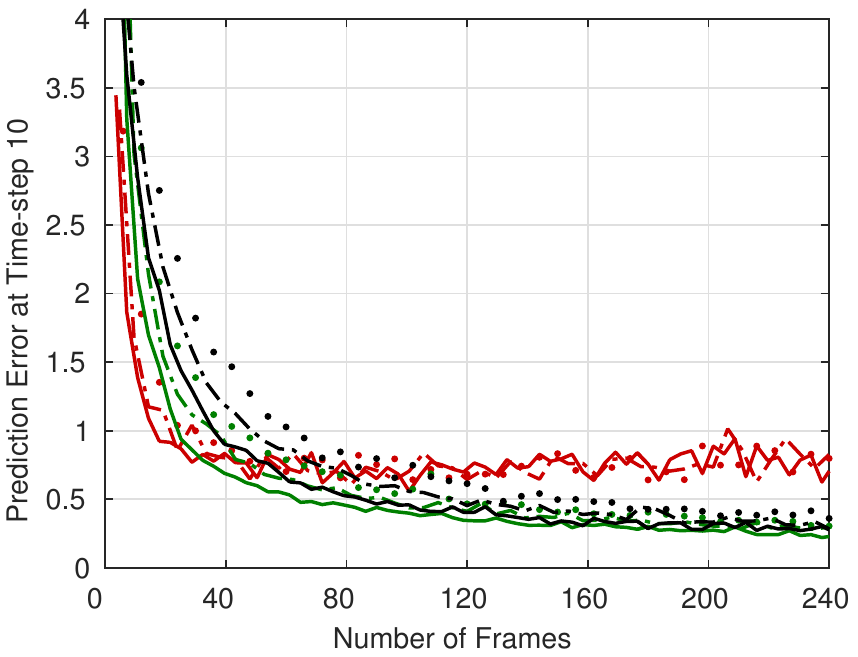}}
%\hskip0.1cm
\scalebox{0.79}{\includegraphics[]{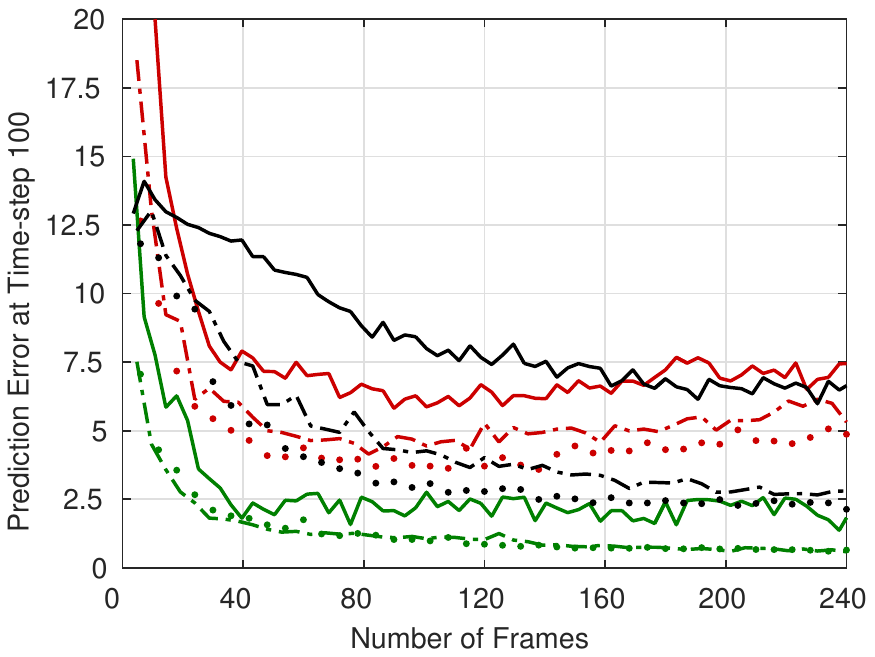}}}
\scalebox{0.79}{\includegraphics[]{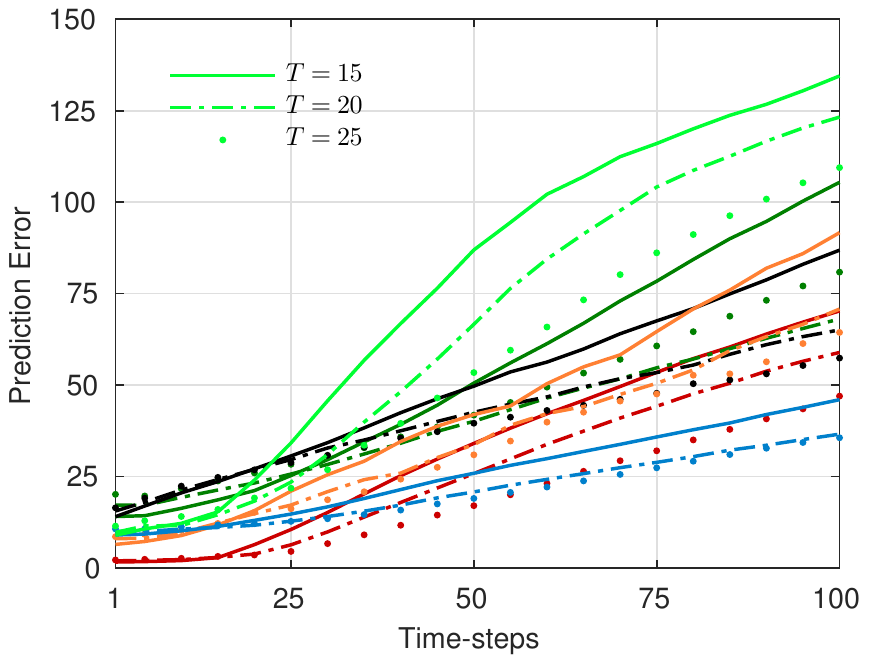}}
%\hskip0.1cm
\scalebox{0.79}{\includegraphics[]{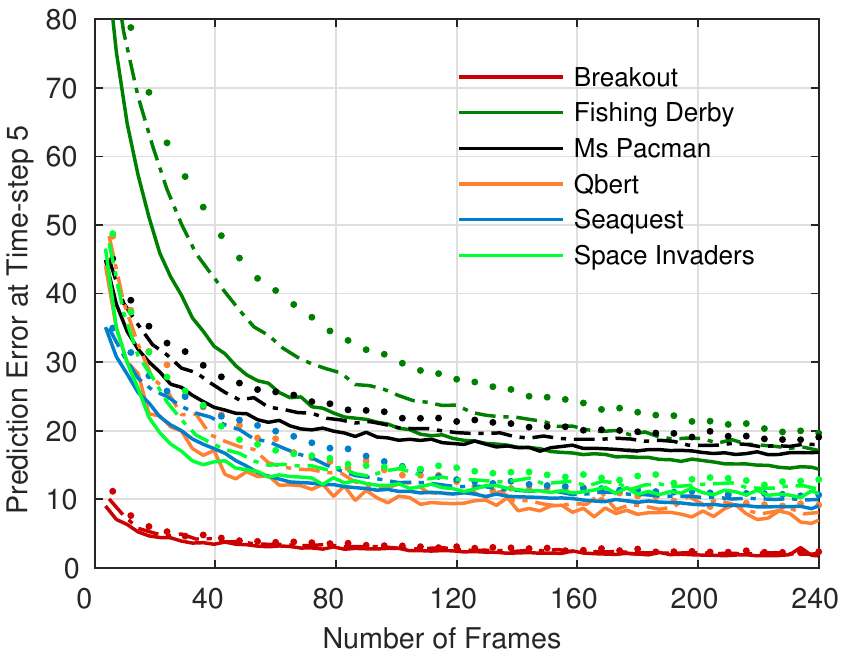}}
\subfigure[]{
\scalebox{0.79}{\includegraphics[]{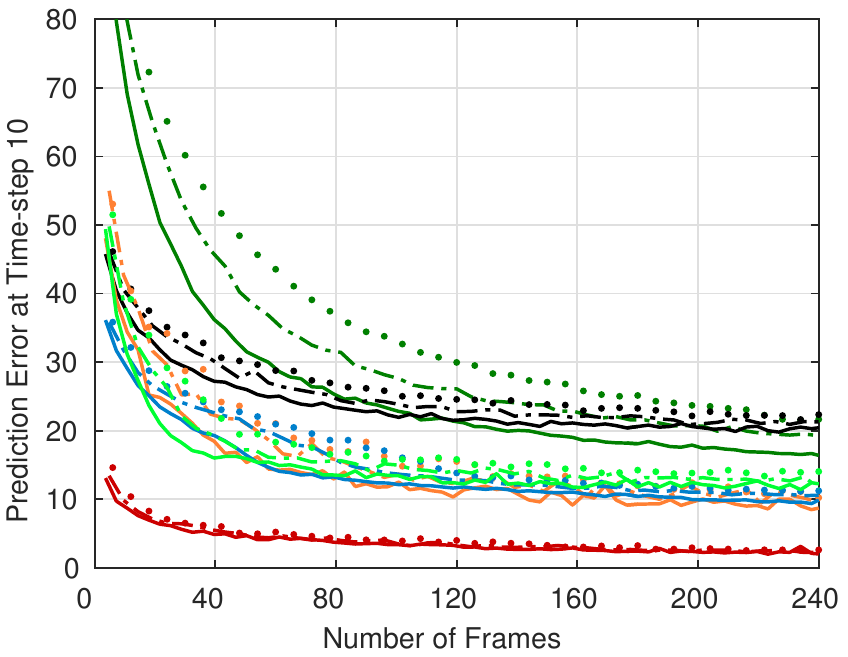}}
%\hskip0.1cm
\scalebox{0.79}{\includegraphics[]{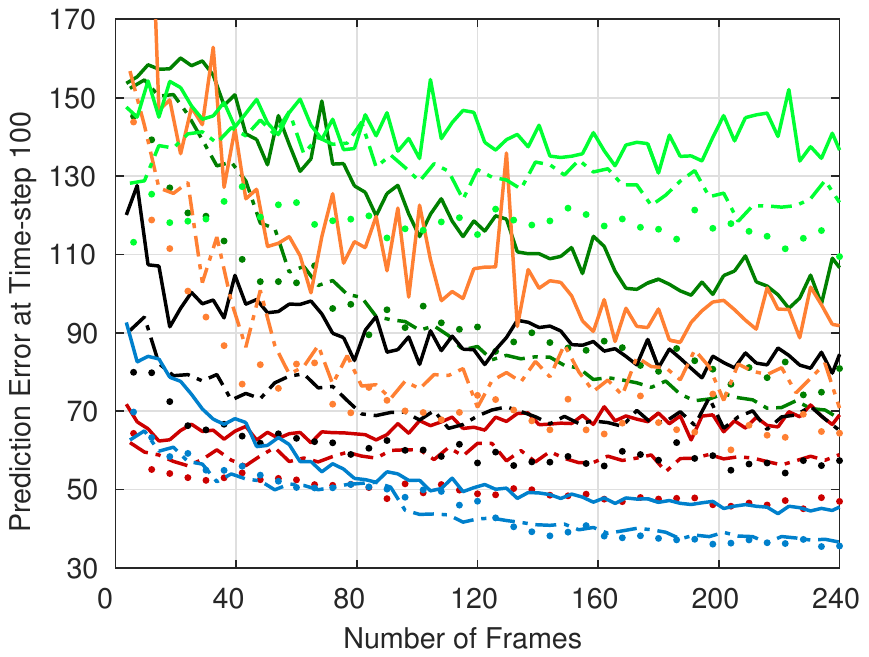}}}
\caption{Prediction error for the prediction-independent simulator with different prediction lengths $T\leq 25$ for (a) Bowling, Freeway, Pong, and (b)
Breakout, Fishing Derby, Ms Pacman, Qbert, Seaquest, Space Invaders.}
\label{fig:predErrJumpySeqLength}
\end{figure}

\begin{figure}[t] % Figures obtained with predErrNoFBSeqNum
\vskip-0.5cm
\scalebox{0.79}{\includegraphics[]{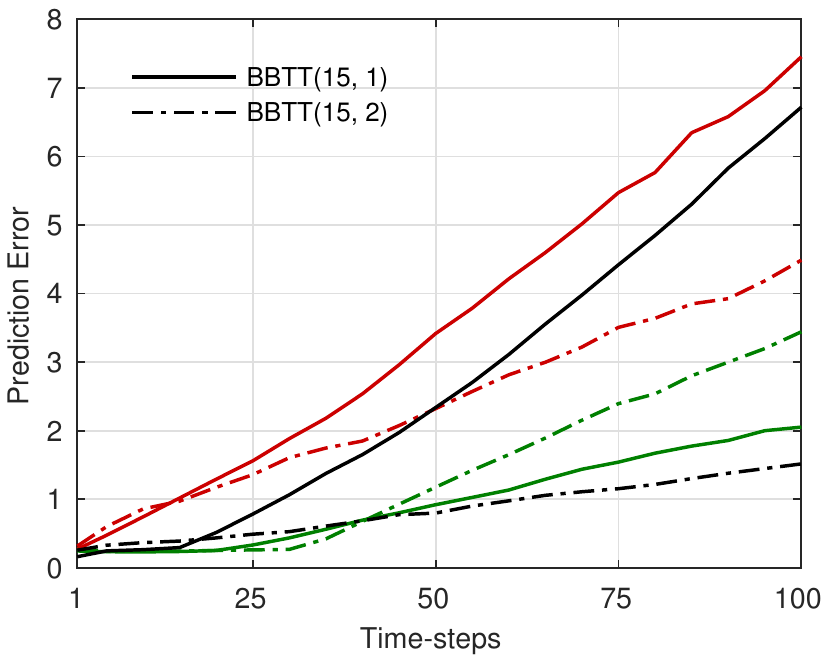}}
%\hskip0.1cm
\scalebox{0.79}{\includegraphics[]{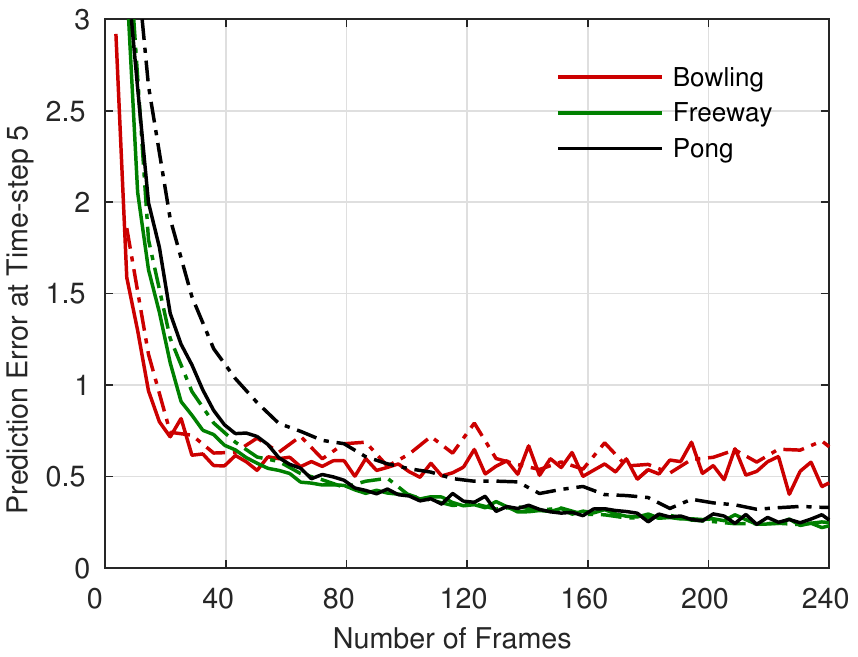}}
\subfigure[]{
\scalebox{0.79}{\includegraphics[]{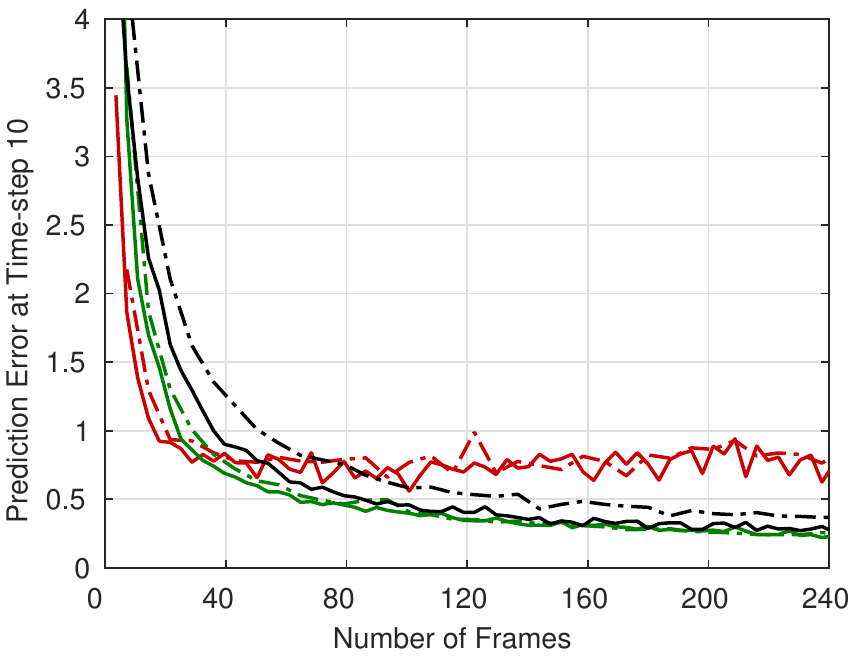}}
%\hskip0.1cm
\scalebox{0.79}{\includegraphics[]{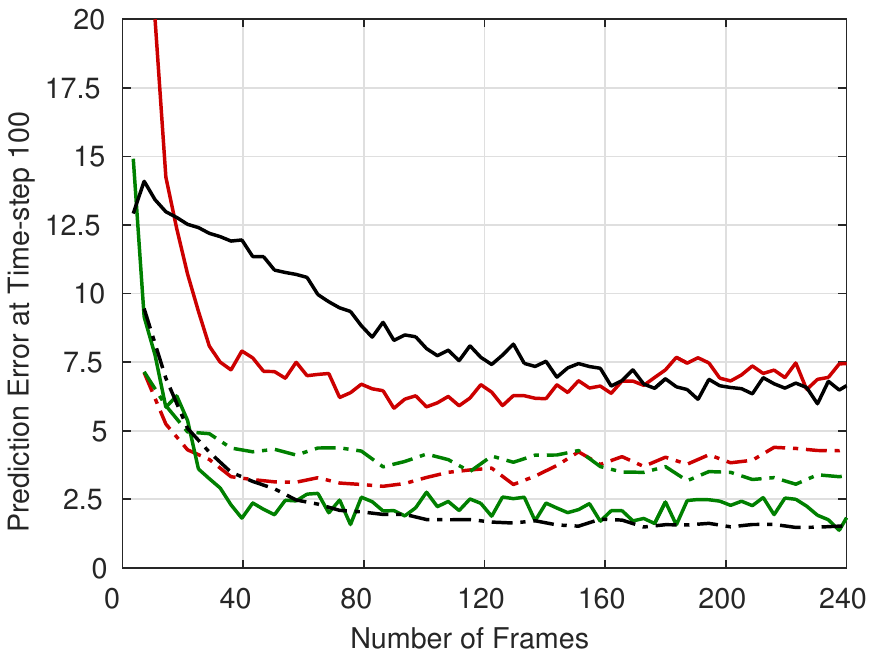}}}
\scalebox{0.79}{\includegraphics[]{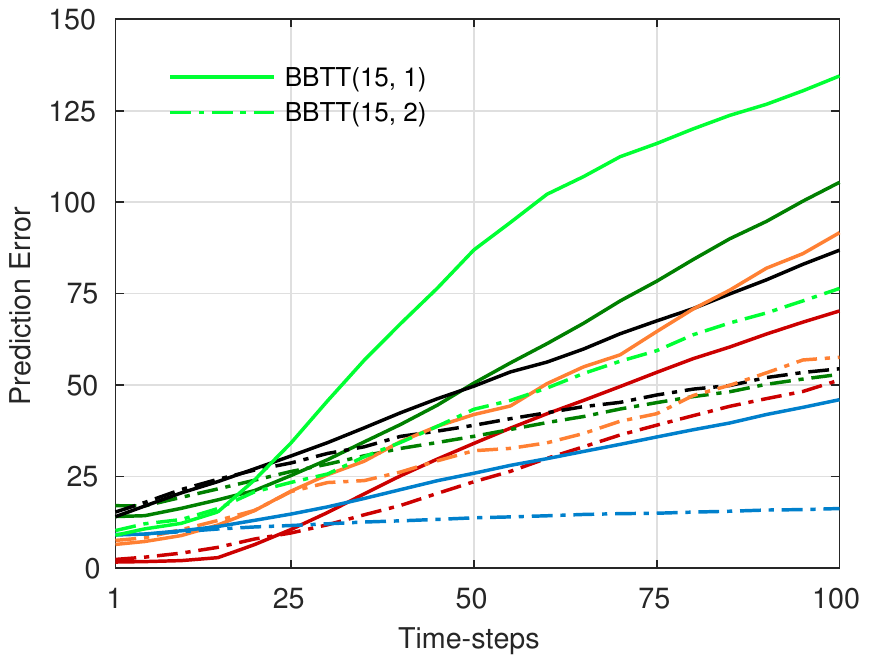}}
%\hskip0.1cm
\scalebox{0.79}{\includegraphics[]{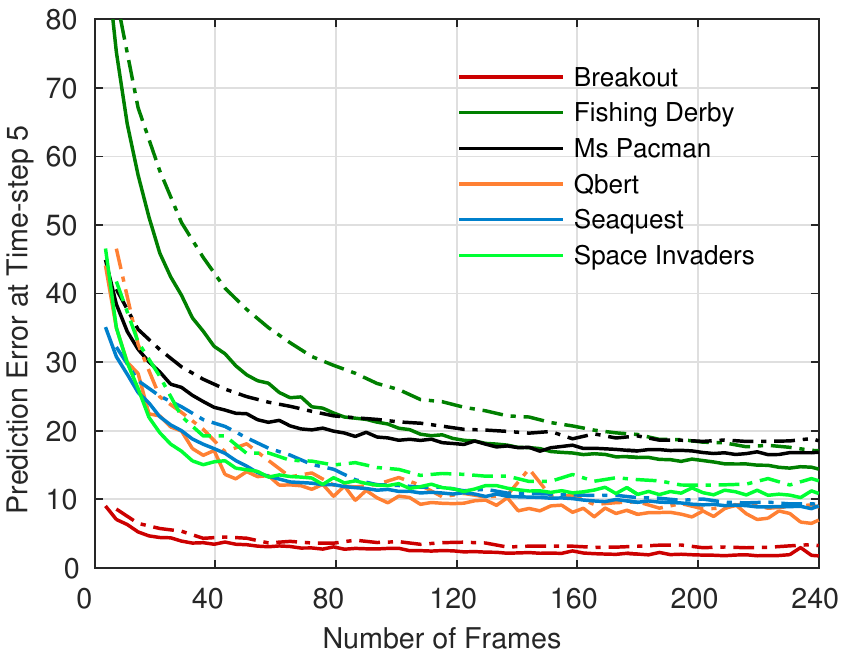}}
\subfigure[]{
\scalebox{0.79}{\includegraphics[]{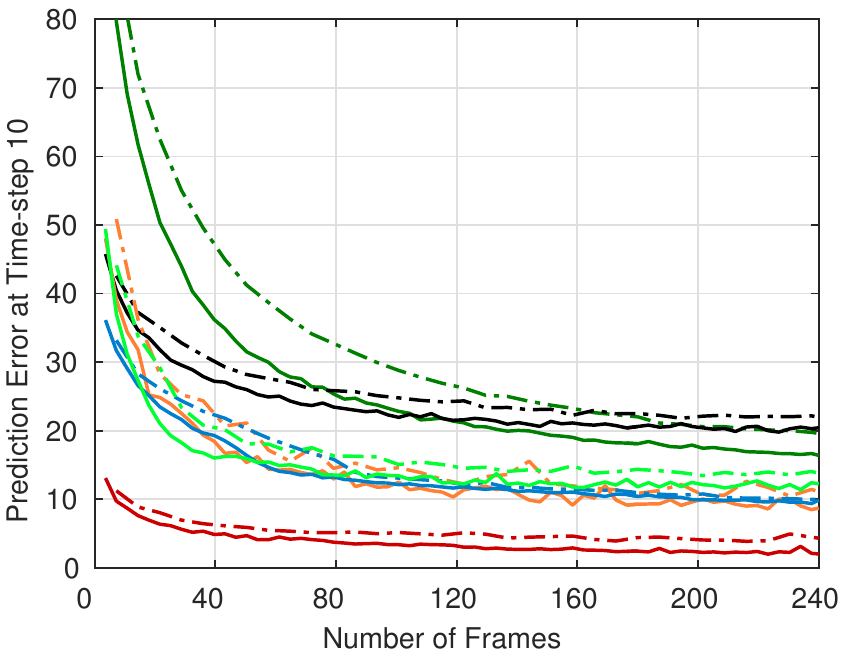}}
%\hskip0.1cm
\scalebox{0.79}{\includegraphics[]{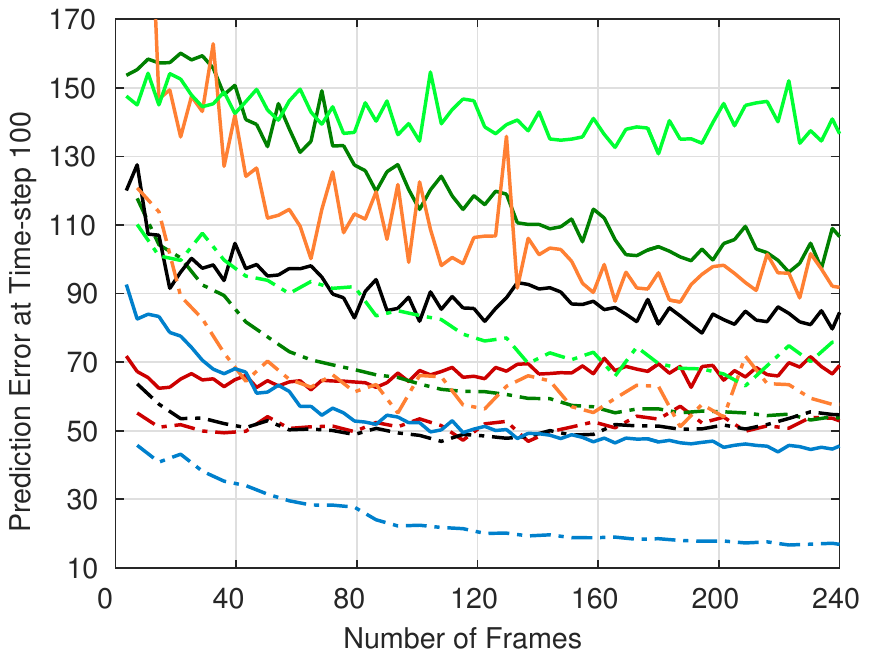}}}
\caption{Prediction error for the prediction-independent simulator with BPTT(15, 1) and BPTT(15, 2) for (a) Bowling, Freeway, Pong, and (b)
Breakout, Fishing Derby, Ms Pacman, Qbert, Seaquest, Space Invaders.}
\label{fig:predErrJumpySeqNum}
\end{figure}

In Figs. \ref{fig:predErrJumpySeqLength} and \ref{fig:predErrJumpySeqNum}, we show the effect of using different prediction lengths $T$ on the structure 'Base--$\vz_{t-1}=\vh_{t-1}$'. 
As we can see, using longer prediction lengths dramatically improves long-term. Overall, the best performance is obtained using two subsequences of length $T=15$.


\end{document}